\documentclass{article}
\pdfoutput=1
\PassOptionsToPackage{numbers, compress}{natbib}

\usepackage[final]{neurips_data_2024}

\usepackage[T1]{fontenc}    %
\usepackage[hyphens]{url}
\usepackage{hyperref}       %
\usepackage{booktabs}       %
\usepackage{amsfonts}       %
\usepackage{nicefrac}       %
\usepackage{microtype}      %
\usepackage{amsmath} %
\usepackage{array} %
\usepackage{graphicx} %
\usepackage{latexsym} %
\usepackage{float} %
\usepackage{subcaption} %

\usepackage{titlesec} %
\usepackage{setspace} %
\usepackage{csquotes} %
\usepackage{booktabs}   %
\usepackage{color,soul}

\usepackage{colortbl} %
\usepackage{mdframed} %
\usepackage{framed} %
\usepackage{makecell} %
\usepackage{multirow}
\usepackage{xfp}
\usepackage[table, dvipsnames]{xcolor}
\usepackage{longtable}
\usepackage{amssymb}%
\usepackage{pifont}%
\newcommand{\cmark}{\ding{51}}%
\newcommand{\xmark}{\ding{55}}%
\usepackage{hhline}
\usepackage{tabularray}

\usepackage[super]{nth}

\usepackage[nameinlink,capitalise]{cleveref} %
\crefname{section}{\S\hspace{-0.0em}}{\S} %
\crefname{Section}{\S\hspace{-0.0em}}{\S} %
\crefformat{appendix}{#2App.~#1#3} %
\crefname{table}{Tab.}{Tab.}
\crefname{appendix_table}{Tab.}{Tab.}
\crefname{Table}{Tab.}{Tab.}
\crefname{Figure}{Fig.}{Fig.}
\crefname{figure}{Fig.}{Fig.}

\usepackage{algorithm}
\usepackage{algpseudocode}

\let\oldthebibliography\thebibliography
\renewcommand{\thebibliography}[1]{%
  \oldthebibliography{#1}
  \let\oldbibitem\bibitem
  \let\oldtextsc\textsc
  \def\oldbbland{et}
  \newcounter{authorcount}
  \def\bibitem[##1]##2{%
    \let\textsc\oldtextsc
    \let\bbland\oldbbland
    \oldbibitem[##1]{##2}%
    \let\textsc\mytextsc%
    \let\bbland\mybbland
    \setcounter{authorcount}{0}
  }
  \def\mybbland{\setcounter{authorcount}{0}\oldbbland}
  \def\dropetal##1.{ \bbletal}
  \def\mytextsc##1{%
    \oldtextsc{##1}%
    \stepcounter{authorcount}%
    \ifnum\value{authorcount}=5\relax%
      \expandafter\dropetal%
    \fi%
  }%
}

\usepackage[toc,page,header]{appendix}
\usepackage{minitoc}

\usepackage[export]{adjustbox} %
\usepackage{overpic} %

\newcommand*{\img}[1]{%
    \raisebox{-.3\baselineskip}{%
        \includegraphics[
        height=\baselineskip,
        width=\baselineskip,
        keepaspectratio,
        ]{#1}%
    }%
}

\usepackage{nicefrac}

\usepackage{xspace}
\newcommand{\ourdata}{\textsc{Prism}\xspace}

\newcommand{\ourtitle}{\textcolor{myred}{\textsc{P}}\textcolor{myyellow}{\textsc{R}}\textcolor{mygreend}{\textsc{I}}\textcolor{myteal}{\textsc{S}}\textcolor{mypurple}{\textsc{M}}\xspace}

\usepackage{soul}
\usepackage{etoolbox} %

\DeclareRobustCommand{\highLight}[2]{%
  \colorlet{hltemp}{.}%
  \sethlcolor{#1}%
  \hl{#2}%
  \sethlcolor{hltemp}%
}

\definecolor{myyellow}{RGB}{255,194,76}
\definecolor{myred}{RGB}{255,138,103}
\definecolor{myredl}{RGB}{255,173,149}
\definecolor{myteal}{RGB}{50,188,221}
\definecolor{myblue}{RGB}{50,188,221}
\definecolor{mygreen}{RGB}{183,229,202}
\definecolor{mygreend}{RGB}{144,213,172}
\definecolor{myoat}{RGB}{254,250,233}
\definecolor{mypurple}{RGB}{143,61,143}
\definecolor{mybeige}{RGB}{234,225,205}
\definecolor{violinred}{RGB}{255,145,145}
\definecolor{violinblue}{RGB}{153,153,255}
\colorlet{lightoat}{myoat!40!white} %
\colorlet{lightred}{myred!40!white} %
\colorlet{lightblue}{myblue!40!white}
\colorlet{lightpurple}{mypurple!40!white}
\colorlet{lightteal}{myteal!40!white}
\colorlet{darkyellow}{myyellow!90!black}
\definecolor{mygray}{RGB}{197,197,197}
\definecolor{heatmapblue}{RGB}{165,187,208}
\definecolor{heatmapgreen}{RGB}{171,199,171}

\newenvironment{innerquote}{
  \begin{mdframed}[
      leftmargin=0cm,
      rightmargin=0cm,
      innerleftmargin=4pt,
      innerrightmargin=4pt,
      innertopmargin=4pt,
      innerbottommargin=4pt,
      linewidth=0.5pt,
      backgroundcolor=lightoat,
    ]
    \fontsize{8pt}{8pt}\selectfont\itshape
}{
  \end{mdframed}\vspace{-0.3\baselineskip}
}

\newenvironment{examplenn}{
  \begin{mdframed}[
      leftmargin=0cm,
      rightmargin=0cm,
      innerleftmargin=4pt,
      innerrightmargin=4pt,
      innertopmargin=4pt,
      innerbottommargin=4pt,
      linewidth=0.5pt,
      backgroundcolor=lightteal,
    ]
    \fontsize{8pt}{8pt}\selectfont
}{
  \end{mdframed}
}

\usepackage{listings}

\definecolor{codeviolet}{rgb}{0.65,0.11,0.36}
\definecolor{codeblue}{rgb}{0.1,0.21,0.57}
\definecolor{codelightblue}{rgb}{0,0.53,0.70}
\definecolor{codegray}{rgb}{0.5, 0.5, 0.5}
\definecolor{user}{RGB}{206,185,97}
\definecolor{model}{RGB}{105,65,110}
\definecolor{numb}{RGB}{164,103,43}
\definecolor{text}{RGB}{128,193,205}

\newcommand{\userprompt}[1]{\textcolor{myred}{#1}}
\newcommand{\modelresponse}[1]{\textcolor{myteal}{#1}}
\newcommand{\freetext}[1]{\textcolor{darkyellow}{#1}}

\lstdefinelanguage{json}{
    basicstyle=\footnotesize\ttfamily,
    backgroundcolor=\color{lightoat},
    commentstyle=\color{codegray},
    stringstyle=\color{codeblue},
    keywordstyle=\color{codeviolet},
    numbers=left,
    numberstyle=\tiny\color{},
    stepnumber=1,
    numbersep=8pt,
    showstringspaces=false,
    breaklines=true,
    frame=single,
    literate = *
    {usertemplate}{{\userprompt{"[USER PROMPT]"}}}{1}
      {modeltemplate}{{\modelresponse{"[MODEL RESPONSE]"}}}{1}
    {texttemplate}{{\freetext{"[FREE-TEXT]"}}}{1}
      {0}{{{\color{numb}0}}}{1}
      {1}{{{\color{numb}1}}}{1}
      {2}{{{\color{numb}2}}}{1},
    morestring=[b]",
    morecomment=[l]{//},
    morecomment=[s]{/*}{*/},
    morekeywords={true,false,null},
    basewidth=0.5em
}

\newcommand{\foo}{\hspace{-2.3pt}$\bullet$ \hspace{5pt}}

\title{The \ourtitle Alignment Dataset: What \textcolor{myred}{P}articipatory, \textcolor{myyellow}{R}epresentative and \textcolor{mygreend}{I}ndividualised Human Feedback Reveals About the \textcolor{myteal}{S}ubjective and \textcolor{mypurple}{M}ulticultural Alignment of Large Language Models}

\author{%
Hannah Rose Kirk$^{1}$\thanks{\texttt{\{hannah.kirk,scott.hale\}@oii.ox.ac.uk} $\dagger$Joint last authors; $\ddagger$Work done at University of Sheffield} \quad Alexander Whitefield$^{2}$ \quad Paul Röttger$^3$  \quad
\textbf{Andrew Bean}$^1$  \\ \textbf{Katerina Margatina}$^{4\ddagger}$ \quad \textbf{Juan Ciro}$^{5,11}$ \quad 
\textbf{Rafael Mosquera}$^{5,6}$  \quad \textbf{Max Bartolo}$^{7, 8}$ \\ \textbf{Adina Williams}$^{9}$ \quad \textbf{He He}$^{10}$ \quad \textbf{Bertie Vidgen}$^{1,11\dagger}$ \quad \textbf{Scott A. Hale}$^{1,12\dagger}$ \\
$^1$University of Oxford \quad $^2$University of Pennsylvania \quad $^3$Bocconi University \\ $^4$AWS AI Labs \quad
$^5$ML Commons \quad \quad $^{6}$Factored AI \quad $^7$UCL \quad $^8$Cohere \\ 
$^{9}$MetaAI \quad $^{10}$New York University \quad $^{11}$Contextual AI \quad $^{12}$Meedan 
}
\begin{document}
\maketitle

\vspace{-1em}
\begin{abstract}
Human feedback is central to the alignment of Large Language Models (LLMs).
However, open questions remain about methods (\textit{how}), domains (\textit{where}), people (\textit{who}) and objectives (\textit{to what end}) of feedback processes.
To navigate these questions, we introduce \ourdata, a dataset that maps the sociodemographics and stated preferences of 1,500 diverse participants from 75 countries, to their contextual preferences and fine-grained feedback in 8,011 live conversations with 21 LLMs.
With \ourdata, we contribute (i) wider geographic and demographic participation in feedback; (ii) census-representative samples for two countries (UK, US); and (iii) individualised ratings that link to detailed participant profiles, permitting personalisation and attribution of sample artefacts. We target subjective and multicultural perspectives on value-laden and controversial issues, where we expect interpersonal and cross-cultural disagreement. We use \ourdata in three case studies to demonstrate the need for careful consideration of which humans provide what alignment data. \\%We demonstrate \ourdata via three case studies, showing it matters which humans set alignment norms.\\
\raisebox{-0.3\height}{\hspace{0.05cm}\includegraphics[width=0.45cm]{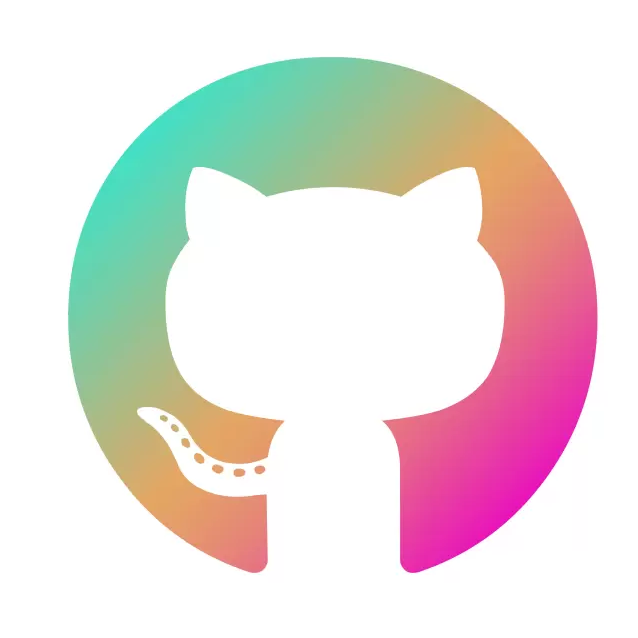}} \small \textbf{\mbox{Data \& Code:}} \href{https://github.com/HannahKirk/prism-alignment}{github.com/HannahKirk/prism-alignment} \\
\vspace{1em}
\raisebox{-0.3\height}{\includegraphics[width=0.4cm]{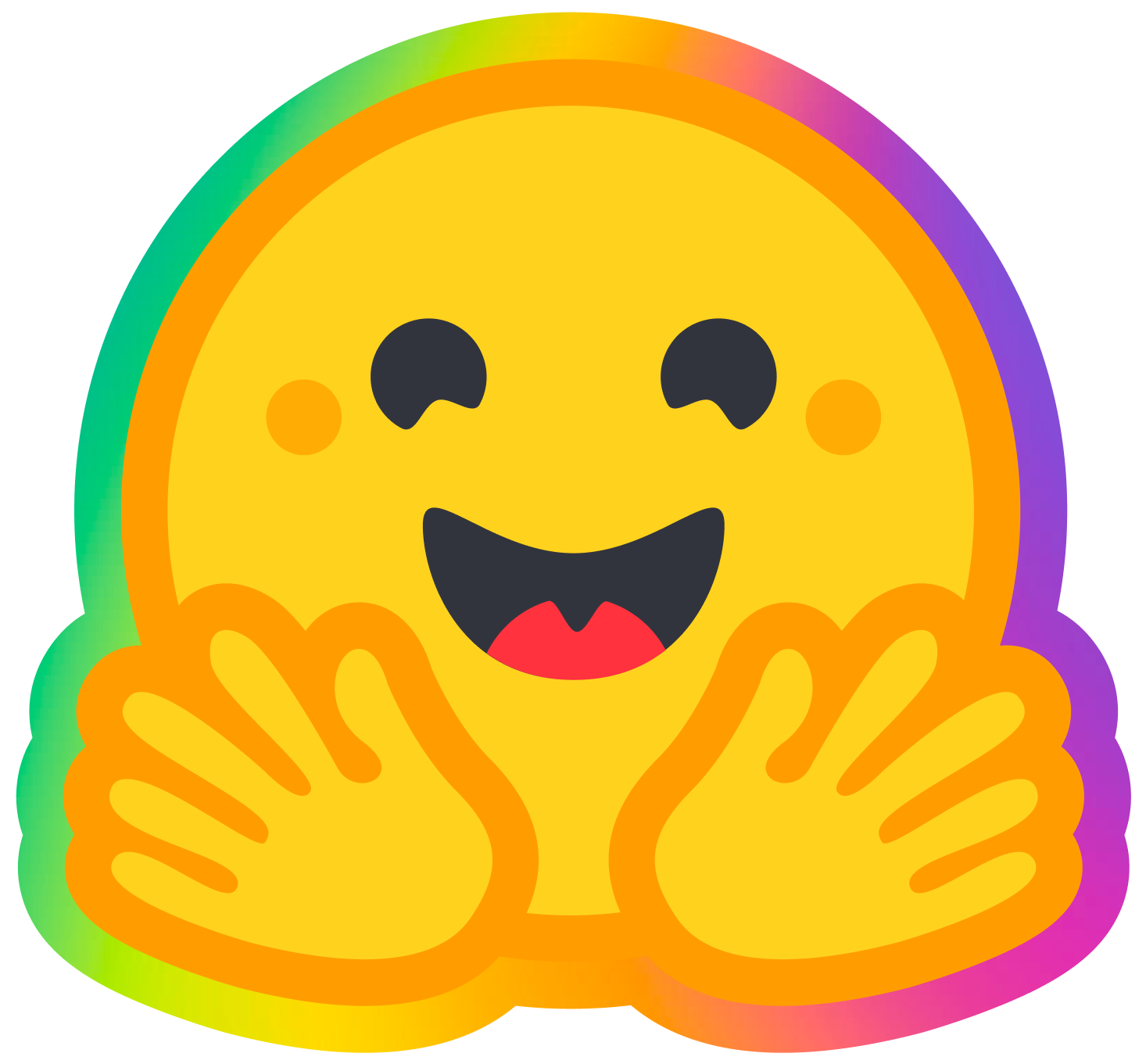}} \small \textbf{\mbox{Data \& Dataset Card:}} \href{https://huggingface.co/datasets/HannahRoseKirk/prism-alignment}{huggingface.co/datasets/HannahRoseKirk/prism-alignment}
\end{abstract}

\doparttoc %
\faketableofcontents %

\part{} %

\def\pcb#1{%
   {\color{myred}\rule{\fpeval{#1/\percentscale*\barwidth} cm}{\barheight}} #1
}

\def\smallpcb#1{%
   {\color{myyellow}\rule{\fpeval{#1/\percentscale*\smallbarwidth} cm}{\barheight}} #1
}

\def\medpcb#1{%
   {\color{myred}\rule{\fpeval{#1/\percentscale*\medbarwidth} cm}{\barheight}} #1
}

\def\pcbb#1{%
 {\color{blue}\rule{\fpeval{#1/\percentscale*\barwidth} cm}{\barheight}} #1
}

\def\smallpcbb#1{%
 {\color{myteal}\rule{\fpeval{#1/\percentscale*\smallbarwidth} cm}{\barheight}} #1
}

\def\db#1{%
  {\color{red}\rule{\fpeval{#1/\degscale*\barwidth} cm}{\barheight}} #1
}
\def\dbb#1{%
  {\color{blue}\rule{\fpeval{#1/\degscale*\barwidth} cm}{\barheight}} #1
}

\newcommand{\barwidth}{7} %

\newcommand{\smallbarwidth}{0.6} %
\newcommand{\medbarwidth}{4} %
\newcommand{\barheight}{7.5pt} %
\newcommand{\percentscale}{100} %
\newcommand{\degscale}{90} %

\vspace{-4em}
\section{Introduction}
Human feedback serves a direct role for the \textit{alignment} of large language models (LLMs), defined as the steering of AI behaviour towards a set of preferences or values. This increased emphasis on human feedback raises unresolved questions: \textit{how we collect human feedback} when designing methodologies that rely on ordinal or cardinal scales, broad or fine-grained desiderata, and explicit or implicit signals; \textit{where we focus human labour} when selecting domains, topics or tasks to collect feedback over; \textit{who we ask for feedback} when recruiting participants to voice their idiosyncratic preferences, values, or beliefs \cite{gabrielArtificial2020}; and \textit{to what end} when specifying an objective to pursue personalised alignment \cite{kirkBenefits2024, jangPersonalized2023, liPersonalized2024} or to aggregate individual preferences into collective outcomes favourable for societies at large \cite{bakkerFinetuning2022, anthropicCollective2023, chakrabortyMaxMinRLHF2024, liAligning2024, conitzerSocial2024}.

Despite the success of human feedback learning \cite{kirkPresent2023, lambertHistory2023}, answering these questions is constrained by gaps in existing datasets, such as (i) over-reliance on binary A/B comparisons, without fine-grained ratings or explanations \cite{wuFineGrained2023}; (ii) small or biased samples recruited from narrow crowdwork or tech communities \cite{kirkPresent2023, casperOpen2023} (iii) limited sample information (annotator IDs or sociodemographics) \cite{prabhakaranReleasing2021a}; and (iv) scarce documentation for how values are operationalised \cite{rottgerTwo2022,kirkEmpty2023}. Most datasets rely only on revealed or contextual preferences \cite{gabrielArtificial2020},\footnote{We use \textit{\textbf{Contextual Preference}} for observed ratings of LLM outputs to avoid misrepresenting how \textit{\textbf{Revealed Preference}} is used by economists---as assumptions that enable the inference of preferences from choices \citep{mas1995microeconomic}.}
and much attention is devoted to technical or statistical issues in feedback learning \cite{rafailovDirect2024, zhuPrincipled2024, zhuIterative2024}, rather than data-centric human factors. Relying on `generic' human data teaches behaviours which are \textit{reductionist} because values are relational and non-separable from the person, community or operating context \cite{turchinAI2019, earpHow2021, mascoloRelational2021}; and \textit{non-generalisable} because the indiscriminate aggregation of data subsumes hidden annotator contexts as universalities \cite{butlerContingency2000, sloaneControversies2024, talat-etal-2022-machine, aroyoTruth2015, siththaranjanDistributional2023}.

\begin{figure}
    \centering
    \includegraphics[width = 0.97\textwidth]{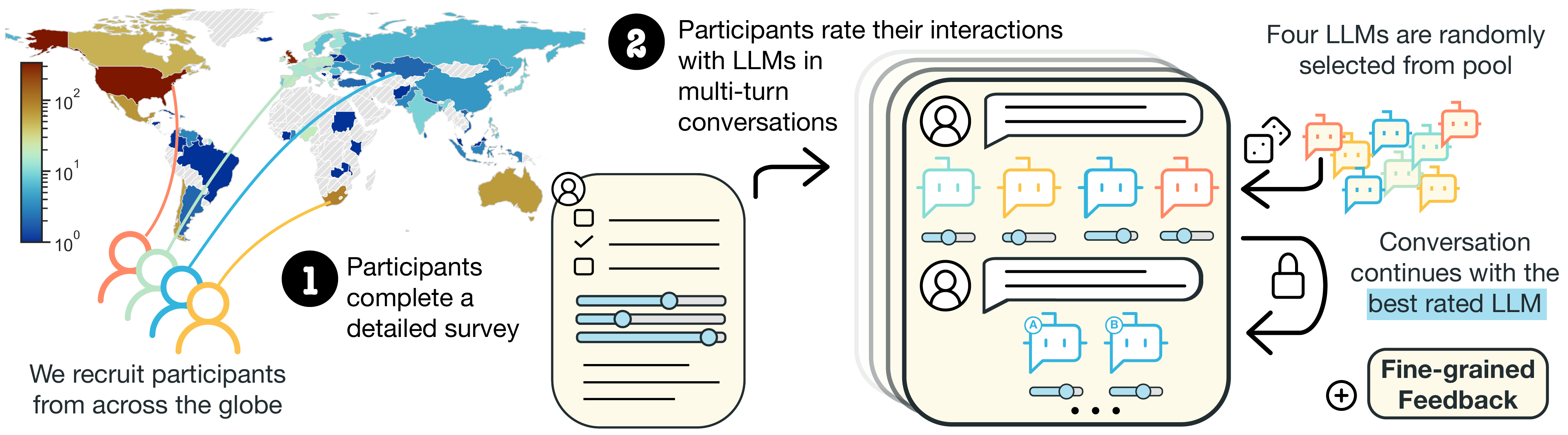}
    \fontsize{8pt}{8pt}\selectfont
\setlength{\tabcolsep}{2pt}
\centering
\renewcommand{\arraystretch}{1.15}
\begin{tabular}{llllllll} 
\multicolumn{2}{c}{{\cellcolor[rgb]{1,0.541,0.404}}\textbf{Human Participants}} &  &  &  &  & \multicolumn{2}{c}{{\cellcolor[rgb]{0.718,0.898,0.792}}\textbf{Large Language Models }} \\
\underline{Survey Participants} & 1,500 &  & \multicolumn{2}{c}{{\cellcolor[rgb]{1,0.761,0.298}}\textbf{Conversations }} &  & Total Providers & 6 \\ 
\multicolumn{1}{r}{\textit{With Conversations}} & 1,396 \tiny{(93.1\%)} & \textbf{$\Rightarrow$} & \underline{Total Conversations} & 8,011 & \textbf{$\Leftarrow$} & \underline{Total Models} & 21 \\
Birth Countries & 75 &  & \multicolumn{1}{r}{\textit{Unguided}} & 3,113 \tiny{(38.9\%)} &  & \multicolumn{1}{r}{\textit{Commercial API}} & 12 \tiny{(57.1\%)} \\
Reside Countries & 38 &  & \multicolumn{1}{r}{\textit{Controversy guided}} & 2,438 \tiny{(30.4\%)}&  & \multicolumn{1}{r}{\textit{Open Access}} & 9 \tiny{(42.9\%)} \\
Conversations / participant & 5.7\tiny{$\pm$1.0} &  & \multicolumn{1}{r}{\textit{Values guided}} & 2,460 \tiny{(30.7\%)} &  & Conversations / model & 1430.9\tiny{$\pm$171.1} \\ 
Models seen / participant & 13.9\tiny{$\pm$2.5} &  & Turns / conversation & 3.4\tiny{$\pm$1.6} &  & Unique raters / model & 924.3\tiny{$\pm$94.4} \\ 
\hhline{==~~~~==}
 &  &  & Total Interactions & 27,172 &  &  &  \\
 &  &  & Total Utterances & 68,371 &  &  &  \\
\hhline{~~~==~~~}
\end{tabular}
\caption{\small \textbf{The \ourdata dataset.} In Stage 1, 1,500 participants fill in the \textbf{Survey} detailing their background, familiarity with LLMs and stated preferences over behaviours (\cref{sec:the_survey}). Demographic and geographic breakdowns are in \cref{tab:full_demographics} and \cref{tab:full_geographics}). Participants then progress to Stage 2, where they converse with LLMs on topics of their choosing, rate the responses on a cardinal scale, and give fine-grained feedback (\cref{sec:the_convos}). In the first turn, four models respond to the opening prompt (\img{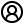}; \img{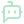}, \img{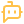}, \img{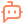}, \img{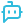}). In subsequent turns, the conversation continues with two responses sampled from the highest-rated model at a non-deterministic temperature (\img{figs/icons/circle-user-round.png}; \img{figs/icons/bot-message-square-blue.png}). There are 8,011 \textbf{Conversations} between participants (\img{figs/icons/circle-user-round.png}) and LLMs (\img{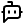}), forming 27,172 \textbf{Interactions} (human message with a set of model responses), and 68,371 \textbf{Utterances} (triples of \{human message, model response, score\}).}
\label{fig:splash}
\vspace{-2em}
\end{figure}

We introduce \ourdata, a new resource for navigating empirical questions of human feedback. We employ both the \textit{ask} and \textit{observe} principles of social science by mapping detailed survey responses of humans around the world onto their live conversations with LLMs (\cref{fig:splash}).  %
This setup permits alignment methods relying on either contextual preference comparisons typical for RLHF \cite{stiennonLearning2020, baiTraining2022, ouyangTraining2022}, or stated preferences and principles like constitutional AI \cite{anthropicCollective2023, baiConstitutional2022}. In addition to pairing stated and contextual preferences, \ourdata has the following features. \textbf{Participatory}: To ensure wider active participation in alignment data \cite{sloaneControversies2024, birhanePower2022}, we recruit 1,500 English-speaking crowdworkers from diverse geographies and demographics; \textbf{Representative}: As units for preference aggregation, we include two census-representative samples (UK, US); and \textbf{Individualised}: To expose hidden human context and permit personalised preferences, each rating links to a pseudonymous ID and detailed participant profile. We source \textbf{Subjective} and \textbf{Multicultural} perspectives to avoid value-monism and cultural homogenisation in the opinions that LLMs represent \cite{durmusMeasuring2023, alkhamissiInvestigating2024, ryanUnintended2024} and operate in the descriptive paradigm without guidelines that characterise `good' responses \cite{rottgerTwo2022, kirkEmpty2023}. Opinion diversity varies along the objective--subjective spectrum (e.g. \textit{what is the capital of France?} vs. \textit{is abortion wrong?}), so we prime participants for values and controversy guided dialogues but also collect neutral unguided dialogues as a baseline. To our knowledge, \ourdata is the first human feedback dataset to target cross-cultural controversies and value-laden prompts, where interpersonal disagreement is rife. After introducing \ourdata (\cref{sec:the_dataset}), we demonstrate its value via three case studies (\cref{sec:experiments}): (1) \textit{Do different people initiate different discussions with LLMs?} (2) \textit{Do people prefer differently aligned models}, and (3) \textit{How do sampling decisions affect welfare outcomes?} \ourdata provides many more research avenues such as engineers targeting personalised alignment \cite{kirkBenefits2024} or consensus across opinion distributions \cite{bakkerFinetuning2022, sorensenRoadmap2024}; social scientists examining how exposure to LLMs affects public attitudes; or policymakers seeking democratic input on AI-citizen interactions on topics like immigration, abortion or euthanasia. Alignment cannot be neatly bifurcated into technical and normative components \cite{gabrielChallenge2021}. \ourdata assists in navigating these complexities with more human voices adjudicating alignment norms.

\section{The \ourdata Alignment Dataset}
\label{sec:the_dataset}
\ourdata maps the characteristics and preferences of diverse humans onto their real-time interactions with LLMs (\cref{fig:splash}). Participants complete a \textbf{Survey} (\cref{sec:the_survey}) with questions about their demographics and stated preferences, then proceed to the \textbf{Conversations} with LLMs (\cref{sec:the_convos}), where they input prompts, rate responses and give fine-grained feedback in a series of multi-turn interactions. With the two-stage setup: (i) we avoid over-generalising from a ``generic human'' by matching ratings to detailed participant characteristics; (ii) we track how contextual preferences (in local conversations) depart from stated preferences (in survey); and (iii) we give participants autonomy to communicate in their own words what is important and why \cite{strayAligning2020, sloaneControversies2024}. Both stages received ethics board approval and ran with informed consent (\cref{sec:appendix_informed_consent}). Participants were paid \textsterling 9/hour and the task took 70 minutes on average. Data collection ran from 22nd November to 22nd December 2023.\footnote{%
Ethics approval, data collection, and analysis was led by researchers from the University of Oxford.}%
We provide a data statement in \cref{sec:appendix_data_statement}, data clause in \cref{sec:appendix_data_clause}, and full codebooks detailing each variable in \cref{sec:appendix_codebooks}.

\subsection{The Survey}
\label{sec:the_survey}
Prior to starting the survey, we ensure that all participants are over 18, obtain their informed consent, give a brief primer on LLMs (or AI language models), and dissuade LLM-written responses. The survey constructs a participant profile containing five features:

\noindent\textbf{LLM familiarity and usage} We ask about participants' familiarity with LLMs (61\% are somewhat familiar, 28\% very familiar and 10\% not familiar at all) and whether to their knowledge they have used them \textit{indirectly} (in products like LinkedIn post-writing tool); or \textit{directly} (via a specialised interface like ChatGPT). Individuals that have used LLMs directly or indirectly (84\%) are branched to questions on frequency of use (7\% every day, 21\% every week, and 20\% every month) and purpose of use (the most popular tasks are research overviews selected by 49\%, professional work by 37\%, creative writing by 31\% and programming help by 27\%). Full results in \cref{sec:appendix_llm_usage}.

\noindent\textbf{Self-written system string (``constitution'')} System strings can guide LLM behaviours as a high-level global instruction prompts prepended to all subsequent interactions \cite{touvronLlama2023, jiangMistral2023}, and have been analogised as ``constitutions'' or governing principles for AI \cite{baiConstitutional2022}. Factuality, professionalism, humanness and harmlessness all emerged as key principles (\cref{sec:appendix_constitutions}) from the following instruction:
\begin{innerquote}
Imagine you are instructing an AI language model how to behave. You can think of this like a set of core principles that the AI language model will always try to follow, no matter what task you ask it to perform. In your own words, describe what characteristics, personality traits or features you believe the AI should consistently exhibit. You can also instruct the model what behaviours or content you don't want to see. If you envision the AI behaving differently in various contexts (e.g. professional assistance vs. storytelling), please specify the general adaptations you'd like to see.
\end{innerquote}

\begin{figure}[t]
    \centering
    \includegraphics[width=\textwidth]{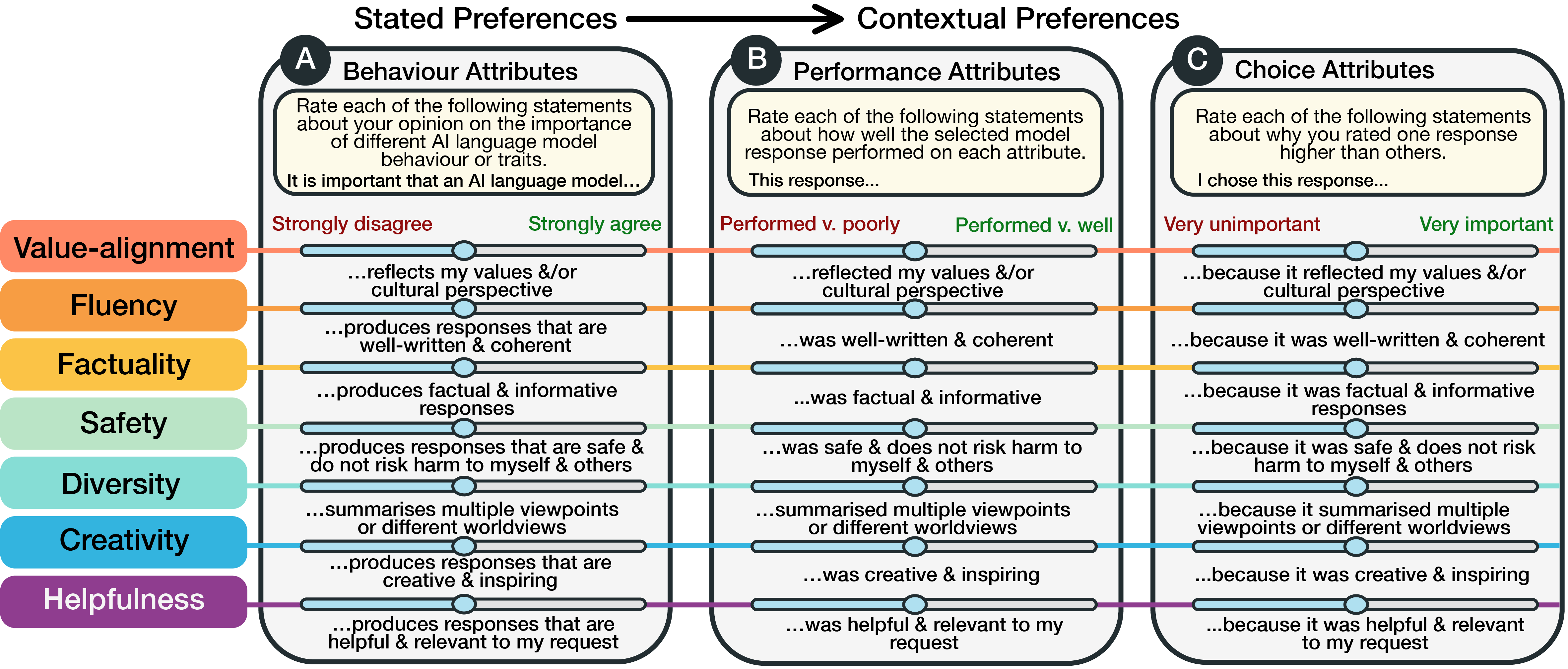}
    \caption{\small \textbf{Schematic of fine-grained attribute ratings.} The same attributes appear in three places in our task: A is asked once in the survey; B and C are asked per conversation. For \textit{performance attributes}, we ask participants to consider only the highest-rated model in the first conversation turn; for \textit{choice attributes}, we ask them to consider this highest-rated model relative to other models in the first turn.}%
    \label{fig:preference_slides}
    \vspace{-1em}
\end{figure}

\noindent\textbf{Stated preferences for LLM behaviours} In contrast to this open-ended preference elicitation, we collect structured ratings on fine-grained behaviour attributes. Participants score the importance of each attribute on a visual analog scale \cite{giftVisual1989} (\cref{fig:preference_slides}). A statement like ``\textit{It is important that an AI language model produces factual and informative responses}'' maps (0,100) where the ends of scale are (\textit{Strongly disagree}, \textit{Strongly agree}). Numeric scores are recorded, but not shown to participants to avoid anchoring and dependency biases. We only collect responses to these statements once \textit{before} participants interact with LLMs but the same attributes appear in the Conversations stage; so, we can track how stated `abstract' preferences relate to contextual `in-situ' preferences.\footnote{The survey also has an \textit{Other} free-text box used by 332 participants (\cref{sec:appendix_other_pref_attributes}), and a \textit{personalisation} attribute which we do not include in Conversations because models are not personalised.} Overall, we find clusters of subjective attributes (values, creativity and diversity) versus objective attributes (factuality, fluency and helpfulness; \cref{sec:appendix_corr_attributes}). While the majority of participants agree that these more objective attributes are important (highly-skewed positive distribution, $\mu \in [86,89], \sigma \in [14,16]$), there is little agreement on the meta-importance of subjective attributes (\cref{sec:appendix_distribution_attributes}). In fact, responses for whether value alignment itself is important follow an almost normal distribution ($\mu=54, \sigma=26$).

\noindent\textbf{Self-written description} Values and preferences are subjective and personal. We ascribe participants autonomy to communicate salient aspects of their identity in a short profile, beyond essentialising associations with structured demographics alone. Honesty, hard work and empathy emerged as common values (\cref{sec:appendix_self_description}) from the following instruction:
\begin{innerquote}
\textit{Please briefly describe your values, core beliefs, guiding principles in life, or other things that are important to you. For example, you might include values you'd want to teach to your children or qualities you look for in friends. There are no right or wrong answers. Please do not provide any personally identifiable details like your name, address or email.}
\end{innerquote}

\noindent\textbf{Basic demographics} We ask standard demographics: age, gender, employment status, martial status, educational attainment, ethnicity, religious affiliation, English proficiency, country of birth, and country of residence. There is always a ``\textit{Prefer not to say}'' option. For gender, participants can select \textit{Male}, \textit{Female}, \textit{Non-Binary}, or self-describe. We collect self-described ethnicity and religion because no pre-set groups exhaust how individuals may self-identify across cultures and global regions. We provide a manual annotation of these strings into aggregated categorisations for statistical analysis (\cref{sec:appendix_manual_annotation}). Because of how we recruit participants (\cref{sec:the_sample}), our sample covers diverse demographics (\cref{sec:appendix_participant_breakdowns}) and geographies (\cref{sec:appendix_geographies}), with representation from people born in 75 countries. However, the sample still skews White, Western and educated, and only contains English-language speakers. %

\subsection{The Conversations}
\label{sec:the_convos}
After completing the survey, participants move to the second stage, consisting of real-time conversations with LLMs via a custom-built interface on the Dynabench platform \cite{kielaDynabench2021, thrushDynatask2022}.

\noindent\textbf{Selecting conversation type} We prime participants to diversify their prompts along the objective-subjective spectrum by asking them to complete two conversations across three conditions or \textit{conversation types} (six in total).\footnote{
Some deviated from this quota (n=6, 2 per type) due to technical difficulties, instruction misunderstanding or losing count; So, we release a balanced subset of the data that controls for this variance (\cref{sec:appendix_convo_rebalancing}). Though values and controversy guided conversations are typically more subjective than neutral baselines, conversation type does not map perfectly to subjectivity levels. Besides from priming participants via selecting a conversation type, we do not constrain (and seek to minimally influence) participants' topic or prompt choice.} They select the \textit{type} before inputting their opening prompt:
\begin{innerquote}
$\bigcirc$ \textbf{Unguided}. Ask, request or talk to the model about anything. It is up to you! \\
$\bigcirc$ \textbf{Values guided}. Ask, request or talk to the model about something important to you or that represents your values. This could be related to work, religion, family and relationship, politics or culture. \\
$\bigcirc$ \textbf{Controversy guided}. Ask, request or talk to the model about something controversial or where people would disagree in your community, culture or country.
\end{innerquote}

\noindent\textbf{Opening the conversation} Participants construct a free-text prompt of their choosing and receive up to four responses from different LLMs.\footnote{We do not stream responses because not all models had the functionality. If a model fails or a response takes $>30$ seconds, we drop this model from the response set and the participant may see $<4$ responses (\cref{sec:appendix_model_details}).} The participants then rate each response on a visual analogue scale (VAS) \cite{giftVisual1989, aitkenMeasurement1969} from ``Terrible'' to ``Perfect''. We record the slider position as a score from 1--100 but do not show participants the number to avoid anchoring or conditional dependence of scores across conversations. %
We opt for this cardinal feedback for three reasons: (i) it encourages subjectivity; (ii) it permits studying the relative merit of cardinality versus ordinality for reward modelling because ratings can be converted to rankings but not vice versa; (iii) it allows expression of preference intensity above and beyond chosen:rejected pairs.\footnote{For example, all responses could be very poor and similar (negative skew, small spread); all very good and similar (positive skew, small spread); or highly-distinguishable (no skew, wide spread).} However, we acknowledge that the cardinal scale introduces some intrapersonal measurement noise from a more cognitively demanding task and carries less interpersonal comparability than ordinal preferences, see Limitations~(\cref{sec:limitations_discussion}). %

\noindent\textbf{Continuing the conversation} The highest-scoring LLM from the opening turn is locked into subsequent turns, with random tie-breaks in the case of identical scores. Participants must continue the conversation for at least another turn, but are asked to vary their conversations between 2 and 10 turns to avoid introducing a dataset artefact. We encourage some variation in conversation length ($\mu_{T}=3.4, \sigma_{T}=1.6$) but there is a strong drop off after the second turn (\cref{sec:appendix_score_properties}). Participants then rate two responses on a VAS like before, but both are now sampled from the selected model with a non-deterministic temperature. These within-model responses are more similar in style and content than across-model responses (in the first turn), and score deviations are narrower (\cref{sec:appendix_score_properties}).

\noindent\textbf{Collecting fine-grained feedback} After the conversation ends, participants first rate statements about the \textit{performance of their highest-rated model} like ``The response was well-written'' on a VAS from \textit{Performed very poorly} to \textit{Performed very well}, or select N/A if the statement is irrelevant for the context. We then ask participants to consider \textit{why they chose this model}, rating statements like ``I chose this response \textit{because} it was well-written'' on a VAS from  \textit{Very unimportant} to \textit{Very important} (or select N/A). Attributes are shared with the Survey (\cref{fig:preference_slides}). We find strong correlations between performance attributes and choice attributes (except safety) but weak correlations of these pairs to stated preferences given in the Survey, perhaps due to conversational, model or task-design confounders (\cref{sec:appendix_corr_attributes}). In general, the distribution of scores over performance and choice attributes is narrower and more positively skewed (bunched to 100) compared to stated preferences (\cref{sec:appendix_distribution_attributes}). Finally, we collect open-ended natural language feedback on the \textit{whole} conversation. Participants contributed both content and stylistic feedback ($\mu=29$ words, $\sigma=19$, \cref{sec:appendix_open_ended_feedback}).
\begin{innerquote}
Give some feedback on the conversation as whole. Hypothetically, what would an ideal interaction for you look like here? What was good and what was bad? What (if anything) was missing? What would you change to make it better?
\end{innerquote}

\subsection{The Sample}
\label{sec:the_sample}
Our sampling aims were \textit{depth} in the demographics represented within countries and \textit{breadth} across global regions. We recruit English-speaking participants from Prolific in two distinct paths:

\noindent\textbf{Census-representative sample (UK, US)} Samples matched to simplified census data (age, ethnicity, gender) were only available for the UK and US. The minimum pool size for a statistical guarantee of representativeness was 300, which set a lower bound for participant quota. After collecting data, we observed some skew in our `representative' samples between observed and expected distributions in recent census data, which we partially correct for (\cref{sec:appendix_census}). These samples permit future studies on more representative populations that can be replicated across two countries; however their inclusion biases \ourdata as a whole towards two Western nations already over-represented in AI research.

\noindent\textbf{Balanced samples (rest of world)} The distribution of Prolific workers outside the US and the UK skews strongly to Europe and Northern America, and some countries dominate continental counts (\cref{sec:appendix_screening}). To avoid more active workforces biasing the sample, we set up 33 country-specific studies where there is $>1$ eligible worker, and allocate sample quotas so that each global region is similarly represented.\footnote{Participants still appear in our sample who were born or reside in countries that did not have a dedicated country-wise study e.g. if their Prolific details were outdated or incorrect. We do not drop them.} We balance each national sample by gender where possible (\cref{tab:studies}).

\noindent\textbf{Included models} The rapidly evolving landscape necessitates a model-agnostic approach to avoid data staleness. We include 21 different LLMs (9 open-access, 12 commercial-API) from various model families and parameter sizes, which diversifies the training data, capabilities, and degree of existing safeguards or alignment biases. To avoid text length confounding preferences \cite{singhalLong2023} and to reduce participant fatigue, we include system prompts instructing models to limit their responses to $\le50$ words. We show the full list of models, decoding parameters and generation details in \cref{sec:appendix_model_details}.

\begin{figure}
    \centering
\includegraphics[width=\textwidth]{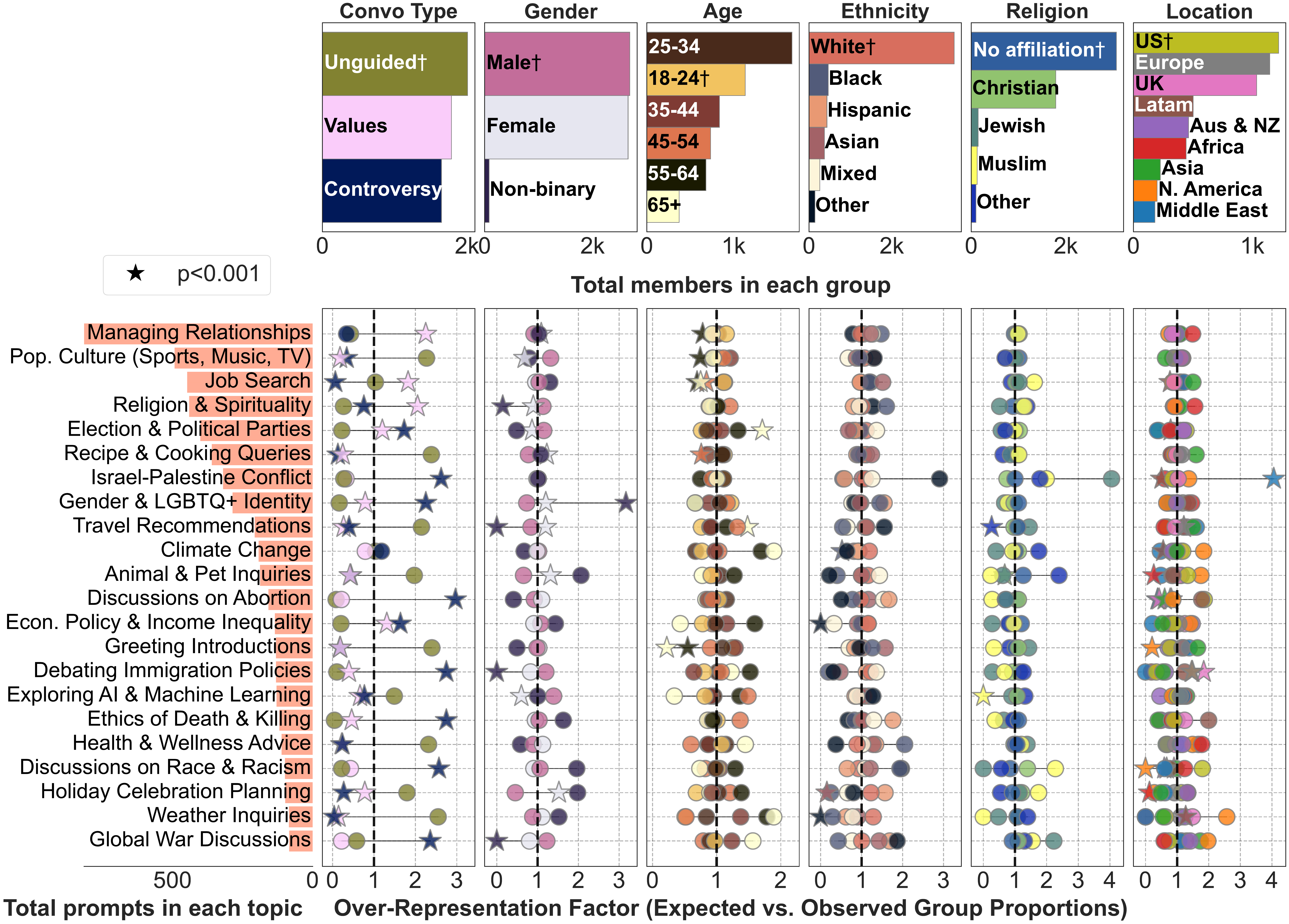}
    \caption{\small \textbf{Topic prevalence by conversation types and participant identity.} We show total prompts clustered into topics (\highLight{myred}{bars}), and total members in each group (top panels). Per group and topic, we plot the \textit{over-representation factor} of observed vs. expected group proportions and show significant regression coefficients (base category indicated by $\dagger$). All coefficients are in \cref{fig:full_topic_signif}, topic-group counts in \cref{fig:heatmap_topic_proportions_by_demo} and centroid prompts in \cref{tab:full_topic_summary}. Location is by \textit{birth region} (with UK and US split out), but most regions have few countries (\cref{sec:appendix_geographies}). \\ \textbf{\textit{Key results} (\cref{sec:dialogue_diversity}):} Priming participants to select a conversation type (unguided, values or controversy guided) significantly influenced diversity of prompts. Identity factors have some significant interactions with prompt choice but each topic contains prompts authored by intersectionally-diverse participants.}
    \label{fig:main_topic_plot}
\end{figure}

\section{Experiments with \ourdata}
\label{sec:experiments}
\subsection{Case Study I: Do Different People Initiate Different Discussions with LLMs?}
\label{sec:dialogue_diversity}
\noindent\textbf{Methods} We use a pre-trained sentence transformer (\texttt{all-mpnet-base-v2}) to embed each opening prompt in 768-D, then apply UMAP to reduce to 20-D, before clustering with HDBScan \cite{campelloDensityBased2013}. 70\% of prompts are assigned to 22 topic clusters and 30\% remain as outliers. We name each cluster by prompting \texttt{gpt-4-turbo} with the top n-grams extracted with TF-IDF and closest texts to the cluster centroid. We define an \textit{over-representation factor} as $\frac{N_{g,t} / N_t}{b_g}$, to  compute observed versus expected topic prevalence per identity group. For the partial contribution of identity attributes, we estimate an OLS regression for each topic $y^t$ ($t \in 1 \ldots 22$) and cluster standard errors at the individual level: $ y_{i,c}^j = \alpha^t + 
    \text{gender}_{i}' \beta_1^t+ 
    \text{age}_{i}' \beta_2^t+
    \text{birth\_region}_{i}' \beta_3^t +
   \text{ethnicity}_{i}' \beta_4^t  +
   \text{religion}_{i}' \beta_5^t  +
    \text{prompt}_{i}' \beta_6^t +
   \varepsilon_{i,c}$,
where $y^t_{i,c} = 1$ if the prompt of participant $i$ in conversation $c$ is categorised as topic $t$. The identity vectors (e.g. \textit{gender}) represent sets of variables, with a base category removed (indicated in \cref{fig:main_topic_plot}). The coefficients of interest are contained in vectors $\{\beta_d^t\}_{d=1}^6$, where component $g$ of $\beta_d^t$ is interpreted as the increase in probability of a participant choosing topic $t$ if they are in the group indexed by $g$ (e.g. Female) compared to the base group (e.g. Male). See \cref{sec:appendix_clustering} for extended methods.

\noindent\textbf{Results} Our instructions had a significant priming effect, resulting in a \textbf{high density of controversial and value-laden topics} (\cref{fig:main_topic_plot}). Topics significantly correlated with controversy guidance are \textit{Gender \& LGBTQ+ Identity}, \textit{Israel--Palestine Conflict}, and \textit{Discussions on Abortion}, while topics significantly correlated with the values guidance are \textit{Managing Relationships}, \textit{Job Search}, and 
\textit{Religion \& Spirituality}. In contrast, the `unguided' condition correlates with task-oriented and neutral topics like \textit{Popular Culture}, \textit{Recipes \& Cooking} and \textit{Travel Recommendations}. Only \textit{Climate Change} is not significantly correlated to conversation type. Controlling for conversation type, 11\% of coefficients are significant ($\alpha=99\%$); so, \textbf{identity factors have some predictive power on topic prevalence}. Significant relationships include: women and non-binary people discuss gender and LGBTQ+ issues more than men; older people discuss elections and travel more than younger people; Black participants discuss climate change less than White participants, and all regions question LLMs about abortion less often than US participants. When we examine granular regions in embedding space using a single-link hierarchical clustering algorithm (\cref{sec:appendix_local_nn_ablations}), \textbf{local prompt neighbourhoods tend to be intersectionally-diverse}: 84\% of them meet or exceed entropy across intersectional demographics that would be expected under random sampling. During this local exploration, we retrieve regions of semantically-identical prompts rated by multiple diverse individuals (e.g. one neighbourhood ``Does God exist?'' has 7 religious and 7 irreligious participants), finding that \textbf{interpersonal differences in contextual preferences persist even when dialogue context is fixed} (\cref{sec:appendix_empirically_retrieved_fixed_dialog}). So, despite \ourdata containing semantically-diverse prompts, people from different backgrounds occupy common discussion spaces, providing an anchor to examine diverse perspectives to shared issues.

\subsection{Case Study II: Do Different People Prefer Differently-Aligned Models?}
\label{sec:preference_diversity}

\noindent\textbf{Methods} Observed preference differences at the model-level are confounded by interactions of topic prevalence and model aptitude (e.g. men ask more about aliens and \texttt{gpt-4} is poor on extraterrestrial knowledge). Evidence of shared dialogue spaces (\cref{sec:dialogue_diversity}) and group-topic score differences (\cref{sec:appendix_pref_topic_confounders}) mitigate some concern, but to further control for context, we use opening prompts from the balanced subset of participants (n=1,246) with equal conversations per type (n=6,669). The mean participant rates 14/21 LLMs but unseen ratings are missing at random. Our aggregation (social choice) function over participant ratings is derived from \textit{Pairwise Rank Centrality} ($\mathcal{P}$) \cite{negahbanIterative2012} and \textit{Convergence Voting} \cite{banaConvergence2021}, both inspired by \textit{PageRank} \cite{pagePageRank1999}. Each model is a node in a graph and transition probabilities between nodes are calculated by the proportion of pairwise battle wins. This process simulates a random walk on a Markov chain, leading to a stationary distribution of scores that reflect the collective preference intensity across models. Here, we compute $\mathcal{P}$ over subsamples using a regularisation parameter of 1 and tie threshold of 5, but present extended methods and robustness checks in \cref{sec:appendix_model_rank}.

\begin{figure}
    \centering
    \includegraphics[width=\textwidth]{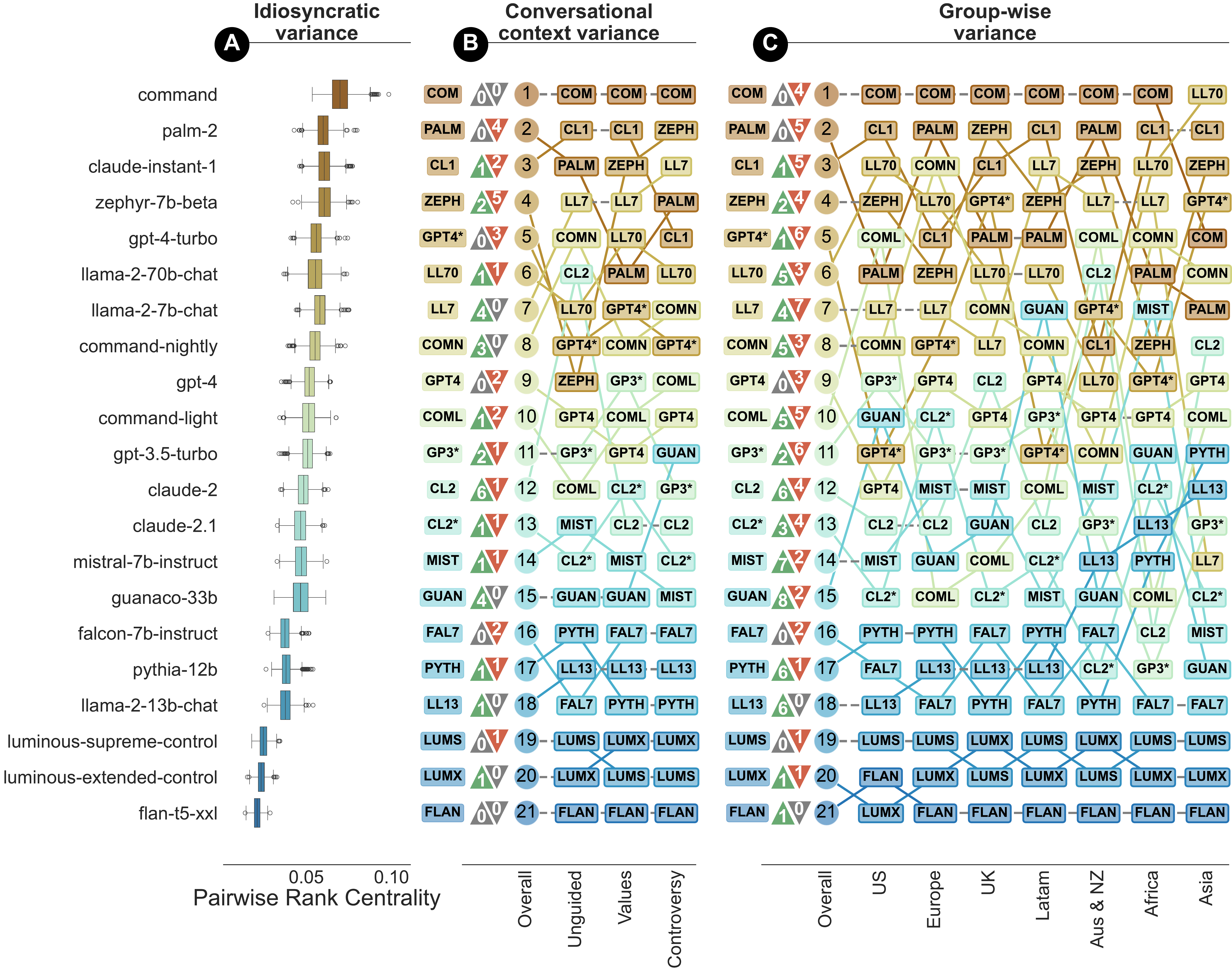}
    \caption{\small \textbf{Sources of variation in model preferences.} Panel A shows \textit{idiosyncratic variance} in distributions of Pairwise Rank Centrality scores for 100 randomly-drawn participants (over 1,000 bootstraps). For Panels B and C, we show \textit{conversational context variation} and \textit{group-wise variation}. We show overall rank based on Pairwise Rank Centrality over n=6,669 balanced conversations (numbered circles). We then trace how rank changes by sampling the group on $x$ (e.g. filtering to only values guided conversations, or only US participants). Across these subsamples, we show most spots climbed (\img{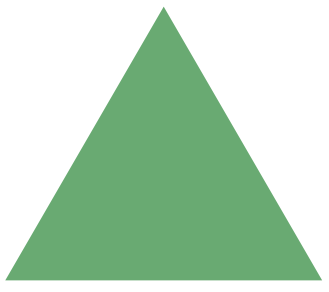}) and spots fallen (\img{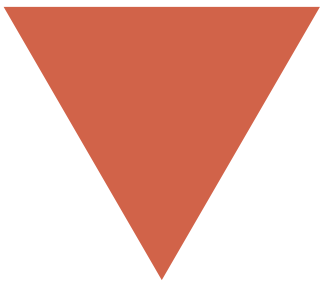}) by each model relative to overall rank. \\ \textbf{\textit{Key results} (\cref{sec:preference_diversity}):} Rankings are sensitive to sample composition, varying with which participants are sampled (Panel A,C) and what they talk about (B). Rankings differ from other leaderboards, explained by \ourdata's characteristics (sample diversity, domain shifts) as well as response characteristics (length, formatting, refusals).}%
    \label{fig:main_preference_plot}
\vspace{-1em}
\end{figure}

\noindent\textbf{Results} We find \textbf{rankings are sensitive to idiosyncratic, contextual, and group-wise variance}. Samples of 100 people introduce significant noise, resulting in a fairly even distribution of collective preference among the top 10 models (\cref{fig:main_preference_plot}). Rankings are sensitive to \textit{what} participants talk about: \texttt{zephyr-7b} performs highly on controversy but not in unguided domains, while \texttt{claude-2} has the opposite trend; and \textit{where} they are from: relative to overall rank, \texttt{palm-2} drops 4 places for participants in the US, \texttt{llama-7b} drops 7 places in Asia, while \texttt{mistral-7b} gains 7 places in Africa. We further observe that \textbf{\ourdata produces surprising ranks relative to other leaderboards}. We apply our method to \textsc{ChatbotArena} data \cite{zhengJudging2023}, finding \texttt{gpt} models fare significantly worse in \ourdata, while open models like \texttt{zephyr-7b} do significantly better (95\% CI over 1,000 bootstraps, \cref{sec:appendix_lmsys}). This may be due to domain shift (task-orientated/coding prompts vs. controversial/cultural prompts), sample diversity or task incentives. To identify drivers of score differences, we generate hypotheses by qualitatively examining battles between \texttt{command} and \texttt{gpt-4/-turbo}, then test these with an OLS regression on all model responses (\cref{sec:appendix_understanding_model_rank}). We find that \textbf{formatting and refusals partially explain score differences} with significant positive effects from additional characters, ending in a question mark (``Would you like to know more?'') and enumeration, but significant negative effect of line breaks. De-anthropomorphic phrases (``As an AI, I don't have personal opinions.'') significantly reduce score but not as substantially as refusals (``Sorry I cannot engage.''). The proportion of explained variance in score by these factors is low ($R^2=0.06$), so we encourage more sophisticated methods in future work for partialling out the effect of style versus content, or participant, model and conversation fixed-effects, as determinants of score.

\subsection{Case Study III: How do Sampling Decisions Affect Welfare Outcomes?}
\label{sec:welfare}
\noindent\textbf{Methods} We use `welfare' to capture the extent to which a chosen LLM aligns with the preferences of a user population. We consider two welfare measures: average model rating ($\textsc{meanRating}$), and average likelihood that a model is chosen (rated highest in the opening turn, $\textsc{meanChoice}$). Previous experiments indicate dialogue and preference diversity across people, suggesting that the welfare of downstream LLM users may depend on who provides feedback. To test this, we first randomly generate seven sub-samples of individuals `in the seat of power' to select their favourite LLM (based on mean rating). Four sampling schemes randomly draw $N$ individuals from a representative sample ($N\in\{10,20,50,100\}$). Three schemes randomly draw 100 individuals from specific low-diversity sub-populations (male, white, and $\ge$45 years old). For each draw, we then measure the distribution of welfare from this LLM being imposed on different stakeholder populations \cite{conitzerSocial2024}: the entire population, non-male individuals, non-white individuals, and individuals $<$45 years old. We report the distribution of average welfare outcomes across random draws from each sampling scheme. We conduct this experiment for the UK and US representative samples. Extended methods are in \cref{sec:appendix_welfare_sensitivity}.

 \begin{figure}[t]
     \centering
     \includegraphics[width=\textwidth]{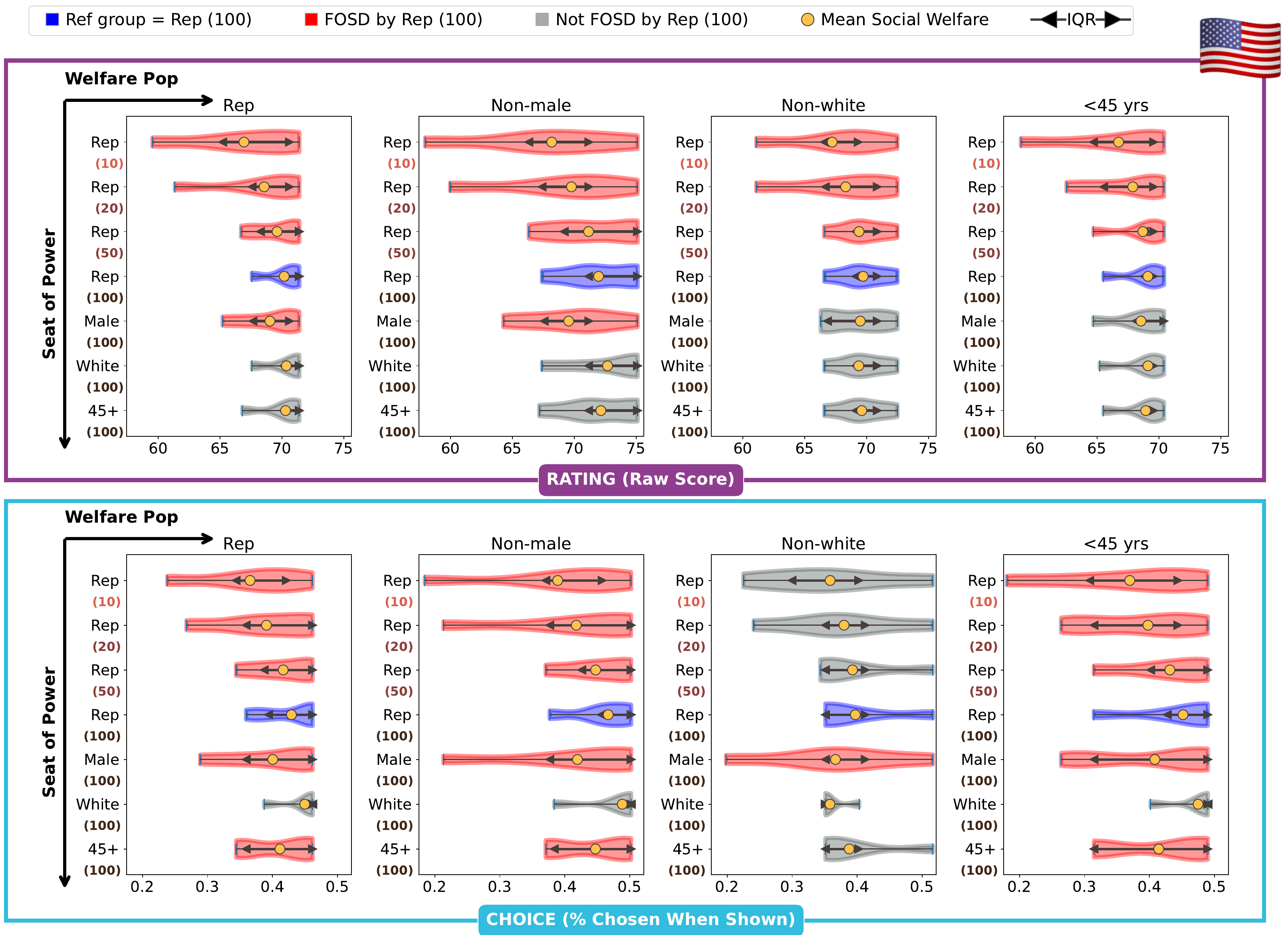}
     \caption{\small \textbf{Welfare distributions for the US.} 
     The distribution of mean welfare for four subpopulations in the US (welfare pop) induced by seven sampling schemes (in the seat of power). The $y$ axis is the sampled subpopulation (e.g. \textbf{Rep} is a `representative' sample of the population) and sample size in brackets (e.g \textbf{(100)}). Each violin shows the distribution of mean welfare for the panel's subpopulation induced by a sampling scheme. The top four \mbox{\highLight{lightpurple}{Rating}} comparisons use the \textsc{meanRating} welfare measure and the bottom \highLight{lightblue}{Choice} comparisons use the \textsc{meanChoice} measure. The  \highLight{violinred}{red} distributions are FOSD by Rep (100) in \highLight{violinblue}{blue} (i.e. less optimal scheme).  \\ \textbf{\textit{Key results} (\cref{sec:welfare}):} Large representative samples mostly outperform smaller or demographically-restricted samples and sampling exclusively from a specific group tends to reduce the welfare of out-group participants (male vs. non-male, white vs. non-white). No single model achieves majority preference (max 45\% \textsc{meanChoice}).}
     \label{fig:welfare_main}
     \vspace{-1em}
 \end{figure}

\noindent\textbf{Results} We find \textbf{as sample size falls, the probability of choosing a LLM with worse mean welfare rises}. Larger samples from the target sub-population appear to first order stochastically dominate\footnote{A probability distribution with CDF$F_\rho$ is said to First Order Stochastically Dominate another probability distribution with CDF $F_\eta$ if both distributions have a finite mean, and $F_{\rho}(t) \leq F_{\eta}(t) \;\;\; \forall t \in \mathbb{R}$.} (FOSD) smaller samples from the target sub-population. \textbf{Sampling exclusively from a specific group tends to reduce the welfare of out-group individuals}. For example, when consider the welfare of the representative US sample (\cref{fig:welfare_main}), sampling from US males is FOSD by sampling from the full US sample. Furthermore, \textbf{average measures can conceal the welfare of minority groups}: sampling 100 white individuals appears to FOSD sampling 100 representative individuals when assessing welfare of the population at large, but minority stakeholders (non-white population) are worse off under this scheme. Finally, \textbf{regardless of the model chosen, a large proportion of participants prefer a different model}. For the US, the model that maximises $\textsc{meanChoice}$ only achieves a probability of $45\%$. If a participant is shown the winning model, and three other models at random, the probability that they will choose the winning model is $<50\%$. The probability they will pick the winning model over all other 20 LLMs can only be lower. This suggests that we should not expect a single LLM to satisfy everyone's preferences in a given population. We repeat the welfare analysis for the UK sample and conduct robustness checks with imputed missing data in \cref{sec:appendix_welfare_sensitivity}.

\section{Related Work}
\label{sec:related_works}
\noindent\textbf{Participation \& Representation in Science \& Technology} There is a long history of technologies failing diverse users who lack consultation during design~\cite{nobleAlgorithms2018, keltyParticipant2019, criado-perezInvisible2019}. Conscious participation can be intrinsically valuable as an act of justice~\cite{sloaneParticipation2022, greeneTaking2023}. However, in internet-harvested pre-training data, participation is involuntary or cooptative~\cite{sloaneParticipation2022, birhanePower2022}, and unequal representation risks cultural homogenisation and minority stereotyping~\cite{nakanoWebGPT2021, birhaneMultimodal2021, benjaminRace2019, blodgettLanguage2020, hershcovichChallenges2022, wangNot2024}. Labelling data or giving feedback is active \textit{procedural participation}~\cite{keltyParticipant2019} but often relies on narrow specifications from technology providers of what counts as high-quality language or preferable outputs~\cite{rottgerTwo2022, kirkEmpty2023, gururanganWhose2022, dziezaAI2023}. In ML or NLP data, variability in subjective experience is commonly collapsed into majority votes~\cite{aroyoTruth2015, mohamedDecolonial2020, airoldiMachine2022, talatMachine2022, diazCrowdWorkSheets2022}, without sufficient documentation of annotator artefacts or disagreements~\cite{benderData2018, mitchellModel2019, gebruDatasheets2021, davaniDealing2022, gordonJury2022}, despite evidence that sociodemographics affect labels~\cite{plankLearning2014, nieWhat2020, wichInvestigating2021, sapAnnotators2022, goyalYour2022, aroyoDICES2023}. Multiple scientific fields are guilty of over-generalising conclusions from the `generic human' drawn from `WEIRD' societies~\cite{henrichMost2010, urbinaCritical2019}. %
\ourdata releases participant IDs and characteristics to spotlight sample diversity while acknowledging sample specificity~\cite{apicellaWEIRD2020}.

\noindent\textbf{Learning from Human Feedback} Using human feedback to condition the loss function for training LLMs overcomes challenges of specifying rewards~\cite{ngAlgorithms2000, christianoDeep2017,zieglerFineTuning2019}. Combining human feedback, reinforcement learning and natural language generation has a history in machine translation~\cite{mirkinPersonalized2015, nguyenReinforcement2017, kreutzerBandit2017} and dialogue~\cite{walker-2000-application, schatzmann-2006-survey, su-etal-2016-continuously, li-etal-2017-dialogue, jaquesWay2019, jaquesHumancentric2020}. RLHF pipelines rely on binary comparisons~\cite{stiennonLearning2020, baiTraining2022, ouyangTraining2022, zieglerFineTuning2019}, principles or rules~\cite{baiConstitutional2022, glaeseImproving2022}, fine-grained feedback~\cite{wuFineGrained2023}, or natural language~\cite{scheurerTraining2022}, to reward dimensions like helpfulness, honesty and harmlessness~\cite{askellGeneral2021, baiTraining2022}. Reward models then update LLMs via algorithms like PPO~\cite{schulmanProximal2017} or Reinforce~\cite{williamsSimple1992, ahmadianBack2024}; but reward model free techniques are competitive, e.g. DPO~\cite{rafailovDirect2024}, supervised fine-tuning~\cite{zhouLIMA2023} and rejection sampling~\cite{menickTeaching2022, bakkerFinetuning2022, thoppilanLaMDA2022}. There is rising demand for high-quality human feedback~\cite{lambertRewardBench2024, ethayarajhKTO2024}, but the complexity and cost of collecting data incentivises scraping preferences, e.g. on Reddit~\cite{stiennonLearning2020, stanfordnlpStanford2023} or StackOverflow~\cite{lambertHuggingFace2023}, or simulating humans with LLMs~\cite{agnewIllusion2024, duboisAlpacaFarm2024, guoControllable2024}. Similar to \ourdata, \textsc{ChatbotArena}~\cite{zhengJudging2023}, \textsc{Lmsys-1m}~\cite{zhengLMSYSChat1M2024} and \textsc{WildChat}~\cite{zhaoInThe2023} feature user-rated model interactions, but for narrow communities (HuggingFace Spaces) and domains (coding, task-orientated). Unlike these datasets, \textsc{OpenConvos}~\cite{kopfOpenAssistant2023} collect optional contributor demographics, and \textsc{Dices}~\cite{aroyoDICES2023} provide demographics for multiple raters per conversation. Other datasets target specific behaviours \cite{baiTraining2022, ganguliRed2022}, or multilingual coverage \cite{singhAya2024}. Surveys on attitudes towards AI~\cite{thealanturinginstituteHow2023, munParticipAI2024} and community assemblies~\cite{anthropicCollective2023, changMeta2024, bergmanSTELA2024} offer another lens on public priorities. To our knowledge, \ourdata is the first to link preference ratings and detailed survey responses.

\section{Limitations, Discussions and Conclusions}
\label{sec:limitations_discussion}
\noindent\textbf{Ethical Considerations and Limitations} We collect informed consent, pseudononymise IDs, check for PII (\cref{sec:appendix_meta_data}) and disallow deanonymisation in our terms (\cref{sec:appendix_data_clause}), but privacy risks remain, especially given the sensitive nature of conversations. Asking participants to engage with controversies expands human preference data to discursive areas with the greatest expected degree of interpersonal disagreement, but risks encouraging hateful, bigoted, biased or otherwise harmful content. \ourdata is less toxic than previous datasets (0.06\%, \cref{sec:appendix_meta_data}). We do not moderate prior to release to permit conversational safety research. There are many sources of variance in \ourdata and alternative divisions of the data may yield different outcomes \cite{silberzahnMany2018}. Granting free choice of dialogue, using cardinal feedback scales and focusing on many kinds of models and participants introduces diversity and subjective freedom but complicates controlled experiments and limits statistical power. \ourdata is still biased towards English-speaking crowdworkers whose task-specific incentives may not align with wider populations. We expand on ethical risks and limitations in our data statement (\cref{sec:appendix_data_statement}).

We raise three discussion points on the boundaries of where we collect preferences, for what end and with what lasting impact. First, aligning LLMs via `preference-based utilitarianism' \cite{tasioulasArtificial2022} may not be synonymous with individual or societal well-being, prompting the question of \textbf{whether there are limits for ``legitimate'' human feedback.} Preferences may be (i) at odds with self-interest due to myopia or information asymmetries (e.g. participants who want anthropomorphic LLMs despite evidenced harms \cite{proudfootAnthropomorphism2011, watsonRhetoric2019, weidingerTaxonomy2022, abercrombieMirages2023, chengAnthroScore2024}) or (ii) incompatible with others' interest (e.g. participants who prefer `anti-woke' LLMs that argue in a debate vs. those who favour neutrality). Relying on decontextualized preference observations carries the risk of silently reinforcing biases from those in power \cite{hershcovichChallenges2022, mohamedDecolonial2020}; so we recommend transparency surrounding individual disagreements before aggregation decisions \cite{conitzerSocial2024, goodingImpact2023}, especially if participant positionality affects their epistemic legitimacy to define harm \cite{benjaminRace2019, dignazioData2020, birhaneAlgorithmic2021}. Second, \textbf{irreconcilable personal preferences and morals matter more when the `unit of alignment' is operationalised as a group, culture or even species, rather than an individual}. \ourdata permits personalised or steerable alignment using participant profiles and specific ratings \cite{kirkBenefits2024, jangPersonalized2023, liPersonalized2024, sorensenRoadmap2024} as well as collective alignment via opinion consensus or distribution of rewards \cite{bakkerFinetuning2022, anthropicCollective2023, chakrabortyMaxMinRLHF2024, liAligning2024,siththaranjanDistributional2023}; though group deliberation in groups may yield different outcomes than gathering data from one person at a time \cite{anthropicCollective2023, changMeta2024, bergmanSTELA2024}. With growing use of synthetic alignment data, \ourdata can assist in calibrating LLM-as-judge protocols to more diverse rater pools \cite{zhengJudging2023, dongCan2024}. Finally, \ourdata was motivated by participation as justice via inclusionary alignment practices that, relative to passive roles in annotation tasks or pre-training data, prioritise active input from local citizens with specialised knowledge of their own and communities' needs \cite{sloaneParticipation2022}. However, participation remains thin because \textbf{the humans crucial to the success of RLHF do not typically share in downstream benefits or profits} \cite{birhanePower2022, perrigoOpenAI2024}. Ultimately, the impact of our work depends on those developing, researching and regulating LLMs because effective participation requires being asked \textit{and} being heard \cite{keltyParticipant2019}.

In their early demonstrations of aligning AI systems to human feedback, \citeauthor{baiTraining2022} discuss \textit{alignment data as a public good}. We echo this sentiment with \ourdata---a new feedback dataset from 1,500 diverse humans, motivated by the need for inclusive, participatory and open scientific research into the pressing question of what it means to align LLMs to human preferences in a pluralistic world. %

\begin{ack}
\label{sec:ack}
This project was awarded the MetaAI Dynabench Grant ``Optimising feedback between humans-and-models-in-the-loop''. For additional compute support, the project was awarded the Microsoft Azure Accelerating Foundation Model Research Grant. For additional annotation support, we received funding from the OpenPhil grant and NSF grant (IIS-2340345) via New York University. We are grateful for support received in the form of research access or credits from OpenAI, Anthropic, Aleph Alpha, Google, HuggingFace and Cohere. Hannah Rose Kirk's PhD is supported by the Economic and Social Research Council grant ES/P000649/1. Paul Röttger is a member of the Data and Marketing Insights research unit of the Bocconi Institute for Data Science and Analysis, and is supported by a MUR FARE 2020 initiative under grant agreement Prot.\ R20YSMBZ8S (INDOMITA). Andrew Bean's PhD is supported by the Clarendon Fund Scholarships at the University of Oxford. We are particularly grateful to Maximilian Kasy for his valuable input and advice on the welfare experiments. We are indebted to the incredible effort and time that our Prolific annotators put into our task, as well as the expert advice from Prolific consultant Andrew Gordon. We also thank any Beta testers, including friends, family and colleagues at Oxford and New York University, for their help in piloting (and debugging!) our task. Lastly, we thank Jakob Mökander, Nathan Lambert, Natasha Jacques, Felix Simon, Nino Scherrer, Maximilian Kroner Dale, Saffron Huang, Amanda Curtis and Joanna Rivera-Carlisle for their feedback on the paper in its various eras. We use scientific colour maps in our figures \cite{crameriMisuse2020}.
\end{ack}

\section*{Author Contribution Statement}
\footnotesize
\scalebox{1}{
\begin{tabular}{r |@{\foo} l}
\textbf{Project Conception} & [\textsc{Kirk}, \textsc{Hale}, \textsc{Vidgen}] \\
\textbf{Data Collection Design} & [\textsc{Kirk}, \textsc{Hale}, \textsc{Vidgen}, \textsc{Röttger}, \textsc{Margatina}]\\
\textbf{Frontend Design and Development} & [\textsc{Kirk}, \textsc{Ciro}] \\
\textbf{Backend Design and Development} & [\textsc{Kirk}, \textsc{Mosquera}]\\
\textbf{Analysis Advisory} & [\textsc{Hale}, \textsc{Vidgen}, \textsc{Röttger}, \textsc{Bartolo}, \textsc{Bean}, \textsc{Williams}, \textsc{He}]\\
\textbf{Literature and Dataset Comparison} & [\textsc{Kirk}, \textsc{Bean}]\\
\textbf{Metadata Processing} & [\textsc{Kirk}, \textsc{Margatina}, \textsc{Bean}]\\
\textbf{Manual Annotation} & [\textsc{Kirk}, \textsc{Bean}, \textsc{Röttger}, \textsc{Bartolo}]\\
\textbf{Results and Codebase} & [\textsc{Kirk}, \textsc{Whitefield}] \\
\textbf{Manuscript Writing} & [\textsc{Kirk}, \textsc{Whitefield}]\\
\textbf{Manuscript Editing and Feedback} & [\textsc{Everyone}]\\
\end{tabular}
}
\normalsize

{\small\bibliographystyle{unsrtnat}
\bibliography{references, referencescoauthors}}

\appendix
\newpage
\addcontentsline{toc}{section}{Appendix} %
\part{Supplementary Material} %
\mtcsetdepth{parttoc}{1}
\parttoc %
\addcontentsline{toc}{section}{\large{PART I: Dataset Details and Distributions}}
\newpage

\section{\ourdata Data Access and Format}
\label{sec:data_access}

 The data can be accessed on Github at \url{https://github.com/HannahKirk/prism-alignment}, and also on HuggingFace at \url{https://huggingface.co/datasets/HannahRoseKirk/prism-alignment}. The dataset has a permanent DOI: \texttt{10.57967/hf/2113}.

 There dataset is organised in two primary JSON lines files: 
 \begin{itemize}
 \item \textbf{The Survey} (\texttt{survey.jsonl}): The survey where participants answer questions such as their stated preferences for LLM behaviours, their familarity with LLMs, a self-description and some basic demographics. Each row is a single participant in our dataset, identified by a \texttt{user\_id}.
 \item \textbf{The Conversations} (\texttt{conversations.jsonl}): Each participants' multiple conversation trees with LLMs and associated feedback. Each row is a single conversation, identified by a \texttt{conversation\_id}, that can be matched back to a participant's survey profile via the \texttt{user\_id}. The conversation itself is stored as a list of dictionaries representing human and model turns in the \texttt{conversation\_history} column, which broadly follows the format of widely used Chat APIs (see single entry schema on the next page).
 \end{itemize}
 
 Additionally, for ease of secondary analysis we provide a more granular and flattened format of the conversations data:
 \begin{itemize}
 \item \textbf{The Utterances} (\texttt{utterances.jsonl}): Each row is a single scored utterance (human input - model response - score). Each row has an \texttt{utterance\_id} that can be mapped back to the conversation data using \texttt{conversation\_id} or the survey using \texttt{user\_id}. The model responses and scores per each user input are in \textit{long format}. Because of this format, the user inputs will be repeated for the set of model responses in  a single interaction turn.
 \end{itemize}
 We also provide code for transforming the conversations to a \textit{wide format}. That is, each row is now a single turn within a conversation. For the first interaction where up to four models respond, we have \texttt{model\_\{a/b/c/d\}} as four distinct columns and \texttt{score\_\{a/b/c/d\}} as another four columns. Note that for subsequent turns, the same model responds and there are only two responses so \texttt{model/score\_\{c/d\}} will always be missing.

 Finally, for every text instance in \ourdata, we provide metadata on the language detection, personal or private information (PII) detection and moderation flags. \textbf{The Metadata} is provided seperately to the main data files (\texttt{metadata.jsonl}).

 We provide \textbf{codebooks} for \textbf{The Survey} (\cref{sec:survey_codebook}), \textbf{The Conversations} (\cref{sec:convos_codebook}), \textbf{The Utterances} (\cref{sec:utterances_codebook}) and \textbf{The Metadata} (\cref{sec:metadata_codebook}).

\newpage
\subsection*{Format of Entries in Conversations Data}
\begin{lstlisting}[language=json, columns=fullflexible, numbers=none]
{
  "conversation_id": "c1",
  "user_id": "user123",
  "conversation_type": ["unguided", "values guided", "controversy guided"],
  "opening_prompt": usertemplate,
  "conversation_turns": [2-22],
  "conversation_history": [
    {
      "turn": 0,
      "role": "user",
      "content": usertemplate
    },
    {
      "turn": 0,
      "role": "model",
      "content": modeltemplate,
      "model_name": "M1",
      "model_provider": "P1",
      "score": [1-100],
      "if_chosen": false,
      "within_turn_id":0
    },
    {
      "turn": 0,
      "role": "model",
      "content": modeltemplate,
      "model_name": "M2",
      "model_provider": "P2",
      "score": [1-100],
      "if_chosen": true,
      "within_turn_id":1
    
    },
    //... Additional list items for remaining model responses (up to 4 in total)
    {
      "turn": 1,
      "role": "user",
      "content": usertemplate
    },
    {
      "turn": 1,
      "role": "model",
      "content": modeltemplate,
      "model_name": "M2",
      "model_provider": "P2",
      "score": [1-100],
      "if_chosen": true,
      "within_turn_id":0
    },
    {
      "turn": 1,
      "role": "model",
      "content": modeltemplate,
      "model_name": "M2",
      "model_provider": "P2",
      "score": [1-100],
      "if_chosen": false,
      "within_turn_id":1
    }
    //... Additional turns follow the same pattern as turn 1
  ],
  "performance_attributes": {
    "fluency": [1-100],
    "factuality": [1-100],
    "helpfulness": [1-100],
    //....Additional attribute ratings
  },
  "open_feedback": texttemplate
}
\end{lstlisting}

\section{\ourdata Data Statement}
\label{sec:appendix_data_statement}
We provide a data statement \cite{benderData2018} to document the generation and provenance of \ourdata.

\subsection{Curation Rationale}
The \ourdata Alignment Project, funded by a variety of academic and industry sources (see \hyperref[sec:ack]{Disclosure of Funding}), aims to diversify human feedback datasets. All participants are recruited via the Prolific platform. The sample is described in \cref{sec:the_sample}, with additional details in \cref{sec:appendix_screening}. The primary purpose of the dataset is for academic research into how different people interact with LLMs and perceive their outputs. However, we do not prohibit the use of the dataset to develop, test and/or evaluate AI systems so long as usage complies with the dataset license (\cref{sec:appendix_license}).

\subsection{Language Variety}
The language of human- or model-written text was not explicitly restricted to English. However, the task instructions were written English, and fluency in English was included as a screening filter. As a result of these factors, 99\% of text instances are in English (see \cref{sec:appendix_meta_data} for breakdowns per type of text instance and by other language). There is scope for wide social and regional variation even within a language. Given we have speakers residing in 38 countries (born in 75 countries), we likely have various forms of English, especially by level of fluency (see \cref{tab:full_demographics}). Information about which varieties of English are represented is not available.

\subsection{Speaker Demographics}
\label{sec:speaker_demos_DS}
There are two sets of ``speaker'' roles in \ourdata: human participants and large language models (LLMs). Both roles contribute to the characteristics of the text utterances in the dataset.

\paragraph{Participant Characteristics} We provide full demographic breakdowns of participant characteristics in \cref{tab:full_demographics}. We provide full geographic breakdowns in \cref{tab:full_geographics}. Despite substantial improvements on sample diversity compared to early widely-used human feedback datasets (see \cref{tab:existing_paper_demographics}, \cref{tab:existing_paper_geography}), \ourdata still skews White, Educated, and Western. This is partly driven by census-representative samples from the US and UK, which can be removed or downsampled for future research. \ourdata only contains participants sourced from one crowdworking platform (Prolific), so inherits sample biases from this narrow pool---for example, participants are active internet users, incentivised by hourly payment on a specific task that they self-select into.

\paragraph{Model Characteristics} Given fast-paced changes to the LLM landscape, \ourdata is designed to be as \textit{model-agnostic} as possible. We include 21 models from various different families, capabilities and sizes (for a summary see \cref{tab:model_summary}). 12/21 models are accessed via commerical APIs, and 9/21 are open-access via HuggingFace. Model-specific characteristics will affect the text characteristics, especially if they have already been alignment-tuned.

\paragraph{Models as Participants} Throughout the study we strongly requested that participants did not use LLMs to write their ``human'' responses, playing both to their integrity (please don't do it), their role in the research (we really need you to not do it), and their incentives (you won't be paid if you do it). We did not directly test nor implement tools to technologically prevent participants from using LLMs on their behalf. We randomly sample 25 instances from human-written texts: system strings and self-descriptions from the Survey; opening prompts and open feedback from the Conversations ($n=100$). An annotator (paper author) manually inspected these and labelled none as model-written text. For instances of sufficient length (46/100, >50 words), we recorded the predicted probability of AI-generated text from an LLM-text detector, where 76\% had $\le1\%$ score.\footnote{The tool is developed by \url{https://sapling.ai/}. LLM-detector tools are susceptible to misclassifications. For example, this feedback: ``\textit{It was good that it offered options and mentioned ``options" rather than just suggesting one thing. It would have been better to state in the beginning how dietary requirements and preferences might play a big role in the decision what to cook for dinner. And also to point out how different cultures have different food traditions. Not everything is US based.}'' was flagged as 88.1\% AI-generated, but the human annotators felt was strongly human-generated.} For the remainder ($n=11$), a second annotator (paper author) gave a tie-break, labelling none as model-generated.

\subsection{Annotator Demographics}
The ``annotators'' are ``speakers''---the same human participants who answer the survey, interact with the LLMs, and provide structured and unstructured feedback. See \cref{sec:speaker_demos_DS}.

\subsection{Speech Situation}
All participants were recruited via Prolific. They were paid \textsterling9/hour. The survey was hosted on Qualtrics (\href{https://www.qualtrics.com/}{www.qualtrics.com}), and the conversations on Dynabench (\href{https://dynabench.org/}{www.dynabench.org}). 

All data was collected between 22nd November 2023 and 22nd December 2023. The time of the data collection period did affect the topics of discussion: for example, one topic concerns Christmas holiday celebrations while another discusses the Israel--Palestine Conflict.

The primary modality of \ourdata is written language, combined with structured ratings or structured survey data. The conversations between participants and LLMs happened \textit{synchronously} via live API connections with models in the backend of our interface. We have not edited or moderated any survey responses, participant prompts or model responses. All conversations happened as part of this research project, so the primary `intended audience' was the researchers, though participants were informed of additional plans to distribute and release the data in the consent form (see \cref{sec:appendix_informed_consent}).

\subsection{Text Characteristics}
We summarise text characteristics in \cref{sec:appendix_text_ngrams}. For the survey responses, the text provides details on the participant and their views about LLMs via short-form free-text responses (we requested 2-5 sentences in their own words). For the conversations, there are three different types: unguided, values guided and controversy guided, as described in the main paper (\cref{sec:the_convos}). Each conversation type contains a different distribution of topics. Overall, \ourdata is skewed towards subjective, values-driven and controversial dialogue. The human-written texts within a conversation typically consist of single sentence prompts, on average 13 words long. Prompts receive up to four model responses generated by a variety of LLMs. We instruct the LLMs to limit their response to 50 words or less. Most unsuccessfully abide by this instruction: the average response length is 89 words. We release metadata (see \cref{sec:appendix_meta_data}) with each text instance including information on detected language, automated and manual PII checks and moderation flags (e.g. if it contains sexual, hateful or violent content).

\subsection{Recording Quality}
During data collection, our interface experienced two distributed denial of service (DDoS) attacks: one on 28th November 2023 and another on 1st December 2023. The primary way that these attacks may have affected recording quality was via interrupting participants' conversation sessions (most then later returned to the interface to complete their conversations a couple hours or days later). These participants' data points may differ to those who had a smoother continuous experience in the task.

\subsection{Author Characteristics and Positionality Statement}
We aimed to operate in the subjective paradigm \cite{rottgerTwo2022, kirkEmpty2023} and have as little influence as possible on how participants interacted with models (e.g. no annotation guidelines for how to rate responses). As a team of researchers, we come from a variety of backgrounds (genders, ethnicities, countries of birth, native languages) and are involved with AI research, either in an academia (6/12) or industry (6/12).

\subsection{Expanded Ethical Considerations}
\paragraph{Privacy and deanonymisation}
The conversations in \ourdata are highly personal, for example detailing views towards abortion, religion, immigration, workplace disputes or intimate relationships. We have pseudo-anonymised the data, checked for PII (\cref{sec:appendix_meta_data}), sought informed consent from every participant (\cref{sec:appendix_informed_consent}), provided options for participants to withdraw their data, and clearly stipulated that attempts of deanonymisation violate our dataset's terms and conditions (\cref{sec:appendix_data_clause}). However, despite following these best practices, the risk for deanonymisation remains. We include a reporting mechanism on our website and GitHub for any participants and researchers to report issues.

\paragraph{Harmful and unsafe content} We asked participants to engage the LLMs in controversial conversations. This comes with the benefit of expanding human preference data to discursive areas with the greatest expected degree of interpersonal disagreement, but at the risk of encouraging hateful, bigoted, biased or otherwise harmful content. Harmful content is an issue in other human feedback datasets, where some opt to moderate conversations prior to public release \cite{kopfOpenAssistant2023} and others retain toxic content for the purpose of future research into conversational AI safety \cite{zhaoInThe2023, zhengLMSYSChat1M2024}. Compared to these previous datasets, \ourdata has an exceptionally low level of flagged content as measured via the OpenAI moderation API (0.06\% overall, and $<0.003\%$ for subcategories of sexually-explicit, violent, hateful, self-harm and harassment). However, the recall of this API may be low \cite{zhengLMSYSChat1M2024}; so, this could be an underestimate. From examining prompts closest to topic centroids (\cref{sec:appendix_topic_regressions}), it is clear there are some prompts with potential for harm. We provide metadata for every text instance in \ourdata, and opt to not filter any conversations. We believe it is a critical area of research to understand how state-of-the-art models respond when they are prompted to engage in such conversations, and how different people with diverse lived experiences react to safety interventions.

\paragraph{Participation-washing and intended societal impact} In our setting, we claim what \citet{sloaneParticipation2022} calls \textit{participation as work}, that is offering fair remuneration and attribution of the consensual labour of workers contributing to our project. Notably, many participants (those familiar and unfamiliar with AI) contacted the researchers and reported enjoying or learning from the task, suggesting there was an ``education quotient'' or role of \textit{participation as experience} \cite{keltyParticipant2019}. Compared to ``passive'' participation in annotation tasks or pre-training datasets \cite{birhanePower2022}, our process is more active for participants because it foregrounds the opportunity to provide their feedback, opinions and preferences, not just labels. ``Participatory'' also signals our goal to have communities more involved in alignment fine-tuning of models and see \ourdata as a first step demonstrating this need. These aims evoke notions of \textit{participation as justice}---including more people at the table of LLM design and development but we note that participation is in reality thin, because while we seek their view, we cannot grant participants the power to change behaviours of deployed LLMs \cite{perrigoOpenAI2024}. Even the etymological roots of participation centre on the notion of ``sharing'' \cite{keltyParticipant2019} but there is no guarantee that the human workers upon whom the success of RLHF relies on, partake in any share of the profits from more usable or preferred LLM technologies. We release \ourdata in the hope it moves the needle towards more inclusive and diverse research on human-AI interactions, emphasising the central role of those who contribute their time and voice to generating human feedback data. Ultimately, how these contributions have impact depends on those in power (industry labs, academics, policymakers), because ``the experience of participation must include the sense not only of having spoken, but of having been heard'' \cite[p.18,][]{keltyParticipant2019}.

\subsection{Expanded Technical and Task Design Limitations}
\label{sec:task_limitations}
\paragraph{The curse of dimensionality (or intersectionality)} Our findings suggest dialogue and model choice are driven somewhat by group affiliation and somewhat by idiosyncratic variance. However, \ourdata contains a rich array of information on each participant with both structured and unstructured components. There are endless ways we could have divided the data or understood participant identity, and despite our best efforts to assess sensitivity to design choices, each alternative may have resulted in very different outcomes \cite{silberzahnMany2018}, and we are under-powered to test so many sparse combinations. Using less sparse groupings introduces biases---for example, focusing on region risks lumping together participants from particular geographies as ``cultures'' \cite{apicellaWEIRD2020}. While we split out the UK and US to avoid these countries dominating their respective regions, there remain varying degrees of country-wise entropy in other regions---the Middle East has 94\% individuals from Israel, and 100\% of Non-US Northern Americans are Canadian (see \cref{sec:appendix_geographies}). Similarly, we use more aggregated ethnicity and religion groupings for statistical power, but amorphous and heterogeneous categories like ``Other'' have limited or flawed real-world meaning as ``Other'' contains, for example, both those who identify as Indigenous or First Peoples and as Middle Eastern or Arab. It is an exciting direction for future work to explore free-form characterisations of identity (e.g. the free-text profile or system string) or ex-post groupings of people's preferences \cite{conitzerSocial2024}, and examine how findings change when we break away from neatly-observed but essentialising demographic traits \cite{tomasevFairness2021}.

\paragraph{The confounding effect of many moving cogs in a conversation} Beyond the complexities of intersectional identity and idiosyncratic variance of individuals within identity groups, other sources of variance in \ourdata present a challenge for controlled experiments; particularly, the high-dimensionality of what exact topics each participant chooses to talk about, which models randomly get selected in-the-loop, and the stochasticity in their responses from a non-deterministic temperature. It is hard to pin down robust mechanisms of preference differences amongst individuals with so many sources of variation. We opted for choice of input prompt and conversation to be a free parameter in \ourdata as a more naturalistic setting of LLM use and because we wanted to understand dialogue diversity among participants. We do empirically find some regions of fixed prompt-response pairs from individuals who self-select into asking the same prompts as other participants (see \cref{sec:appendix_empirically_retrieved_fixed_dialog}).

\paragraph{Noisy signals and misaligned incentives} Relatedly, our conclusions may be confounded by measurement invariance given our explicit focus on subjective, fluid and cardinal devices. This echos the economist's view, that it is foolish to rely too heavily on cardinal ratings over ordinal rankings to make interpersonal comparisons, or enforce \textit{preference construction}, where intrinsic feelings are noisily-quantified on numeric scales. There are also issues of \textit{preference falsification}: while participants are financially incentivised to participate, they may not honestly report their preferences over models. We cannot rule out the possibility that participants select a `bad' model to  lock in for the subsequent turns of conversation if it is more interesting (thus preferable in our narrow task confines) to talk to a more offensive or controversial model, or to try to `jailbreak it'  \cite{zhaoInThe2023}. In hindsight, it may have been a smarter design choice to force participants to rank model responses, or to collect both ratings and rankings (notwithstanding decision fatigue), or make attempts to elicit more interpersonally comparable data via a willingness-to-pay monetary unit. Previous work also raises concerns over relying on human feedback as `gold standard', for example whether participants can accurately rate factuality of an output, or are anchored on formatting and `first impressions' (as we and \citet{hoskingHuman2024} both find). Preferences, especially at a fine-grained level like in \ourdata, have high context-dependency \cite{tverskyContextDependent1993}, so we caution against taking the ratings as revealing some objective truth, instead staying firmly rooted in the subjective paradigm \cite{rottgerTwo2022, kirkBenefits2024}.

\paragraph{Still the ``tyranny of the (English-speaking) crowdworker''} Much of AI, NLP and now RLHF is underpinned by crowdworker labour \cite{shmueliFair2021}. Despite our \textit{aims} to include more diverse voices in LLM development processes, we avoid overstating \textit{claims} on diversity. \ourdata still only contains crowdworkers, who have significant sample biases \cite{poschCharacterizing2022}; can only be so ``representative'' given the relatively small sample sizes; must be digital natives given the platformed nature of the work; and possess different incentives for engagement \cite{albertComparing2023}. Furthermore, while \ourdata gains some dialectical diversity from different geographies of English, from varying speaker fluency, and from some contributions in other languages (1\%, mainly Spanish), it is almost exclusively in English. Cultural diversity can only be measured so far without also accounting for linguistic diversity \cite{hershcovichChallenges2022}. Furthermore, while we try to sample from many regions, our sample is still dominated by White Western participants, especially when considering cultural phylogeny \cite{apicellaWEIRD2020}, i.e., the non-independence of populations with shared history or migrations of peoples (for example, Australia vs UK vs Canada). We encourage future work prioritising human feedback collection in other languages to understand how models handle sociocultural and linguistic interactions \cite{singhAya2024}.

\paragraph{The ever-changing stream of pre-aligned models}  When data collection began in mid-November, \ourdata contained the top ranking models on publicly available leaderboards but new models have since emerged, including Gemini \cite{geminiteamGemini2023}, Mixtral \cite{jiangMixtral2024}, Claude-3 \cite{anthropicIntroducing2024}, Command-R \cite{cohereCommand2024} and Llama-3 \cite{metaaiIntroducing2024}. There is an incompatibility between the current pace of model releases and doing human participant research that requires lengthy processes of ethics approval, interface design, data processing and manual annotation. The expense and inconvenience of doing human research increases the attractiveness of simulating responses, usually with GPT-4 \cite{agnewIllusion2024}. So, while \ourdata does miss out on the newest players to enter the battle arena, we do provide carefully-sourced human data (including a survey which stands independently from the LLM conversations) combined with a wide distribution of model texts; so we hope the utility of the data persists in the coming years even as models change. We are still potentially limited when comparing open and closed-access models: while the former allows full transparency over system prompts, closed-access models can obscure additional instructions as hidden context. Including models from the same family allows comparisons by version or size, but introducing clones (models producing very similar outputs) can distort preference rankings \cite{conitzerSocial2024}. \ourdata is also limited by \textit{value-lock in} \cite{agnewIllusion2024}---the models are already tuned to cultural perspectives or alignment norms \cite{durmusMeasuring2023,alkhamissiInvestigating2024}, which precludes observing certain group preferences towards a wider set of behaviours \cite{sorensenRoadmap2024, santurkarWhose2023}, and renders participants ``thin'' because they are ``limited to existing designs with pre-existing purposes.'' \cite[p.3,][]{sloaneControversies2024}.

\clearpage
\section{\ourdata Data Clause}
\label{sec:appendix_data_clause}
\subsection{Terms of Use}
\paragraph{Purpose} The Dataset is provided for the purpose of research and educational use in the field of natural language processing, conversational agents, social science and related areas; and can be used to develop or evaluate artificial intelligence, including Large Language Models (LLMs).

\paragraph{Usage Restrictions} Users of the Dataset should adhere to the terms of use for a specific model when using its generated responses. This includes respecting any limitations or use case prohibitions set forth by the original model's creators or licensors.

\paragraph{Content Warning} The Dataset contains raw conversations that may include content considered unsafe or offensive. Users must apply appropriate filtering and moderation measures when using this Dataset for training purposes to ensure the generated outputs align with ethical and safety standards.

\paragraph{No Endorsement of Content} The conversations and data within this Dataset do not reflect the views or opinions of the Dataset creators, funders or any affiliated institutions. The dataset is provided as a neutral resource for research and should not be construed as endorsing any specific viewpoints.

\paragraph{No Deanonymisation} The User agrees not to attempt to re-identify or de-anonymise any individuals or entities represented in the Dataset. This includes, but is not limited to, using any information within the Dataset or triangulating other data sources to infer personal identities or sensitive information.

\paragraph{Limitation of Liability} The authors and funders of this Dataset will not be liable for any claims, damages, or other liabilities arising from the use of the dataset, including but not limited to the misuse, interpretation, or reliance on any data contained within.

\subsection{Licence and Attribution}
\label{sec:appendix_license}
Human-written texts (including prompts) within the dataset are licensed under the Creative Commons Attribution 4.0 International License (CC-BY-4.0). Model responses are licensed under the Creative Commons Attribution-NonCommercial 4.0 International License (CC-BY-NC-4.0). Use of model responses must abide by the original model provider licenses.

For proper attribution when using this dataset in any publications or research outputs, please cite with the DOI.\\
\textbf{\textit{Suggested Citation}}: Kirk, H. R., Whitefield, A., Röttger, P., Bean, A., Margatina, K., Ciro, J., Mosquera, R., Bartolo, M., Williams, A., He, H., Vidgen, B., \& Hale, S. A. (2024). \textit{The PRISM Alignment Dataset}. \url{https://doi.org/10.57967/hf/2113}

\subsection{Dataset Maintenance}
As the authors and maintainers of this dataset, we commit to no further updates to the dataset following its initial release. The dataset is self-contained and does not rely on external links or content, ensuring its stability and usability over time without the need for ongoing maintenance.

\subsection{Data Rights Compliance and Issue Reporting}
We are committed to complying with data protection rights, including but not limited to regulations such as the General Data Protection Regulation (GDPR). If any individual included in the dataset wishes to have their data removed, we provide a straightforward process for issue reporting and resolution on our Github. Concerned parties are encouraged to contact the authors directly via the provided contact form link on the Github. Upon receiving a request, we will engage with the individual to verify their identity and proceed to remove the relevant entries from the dataset. We commit to addressing and resolving such requests within 30 days of verification.

\clearpage
\section{Informed Consent}
\label{sec:appendix_informed_consent}
This research was reviewed by, and received ethics clearance through, a subcommittee of the University of Oxford Central University Research Ethics Committee [\texttt{OII\_C1A\_23\_088}]. The following text was displayed to all participants to collect informed consent.
\vspace{0.6cm}
\hrule
\vspace{0.1cm}
\hrule
\vspace{0.3cm}

\section*{Your Feedback on AI Language Models}

We appreciate your interest in participating in this study. \textbf{The aim of this research is to understand people's preferences and perceptions regarding AI Language Model behaviours}, also referred to as Large Language Models (LLMs), Generative AI Language Models, AI ChatBots or Virtual Assistants. AI language models are computer programs designed to generate text. They can respond to questions or prompts by producing written responses. We want to learn more about how people like you use and perceive these AI language models.

\hl{Please first make sure you are using a laptop or desktop computer, and you are not using a mobile device. Our task is NOT compatible with mobile devices.}

\hl{Please then read through this information before agreeing to participate (if you wish to).}

You may ask any questions before deciding to take part by contacting the research team. The Principal Researcher is Hannah Rose Kirk, and the Principle Investigator is Dr Scott. A. Hale, who are both affiliated with the Oxford Internet Institute at the University of Oxford.

\subsection*{What does the task involve?}
If you decide to participate, there are two stages. 

In this stage, you will be asked to fill in a short survey about yourself and your thoughts on AI language models. 
 
In the next stage, you will have conversations with AI language models by providing prompts and rating their responses using a user-friendly interface. The prompts can be on various topics, and you don't need any specific knowledge to participate. Your input will help us understand your preferences and opinions about how these AI language models work.
 
\textbf{Both stages should take between 55-65 minutes.} No background knowledge is required.

\textcolor{red}{Please note that you will be interacting with an AI language model. The research team cannot directly control and are not responsible for the text generated by these models. There is a possibility that the models produce biased, inaccurate or harmful language. The risks to you as an individual are equivalent to those you would be exposed to if you use AI language models via interfaces like ChatGPT.}

\subsection*{Do I have to take part?}
No, participation is voluntary. If you do decide to take part, you may stop at any point for any reason before submitting your answers by closing the browser. However, we are only able to pay participants who complete the task. For demographic information, we have included a `Prefer not to say' option for each set of questions should you prefer not to answer a particular question.

\subsection*{Can I withdraw my participation and data?}
Yes, you may stop the study at any time. Please note that if you withdraw within a stage of the study you will not be paid for that stage or any subsequent incomplete stages, but you will be paid for any stages that you have already completed. You can withdraw your data from the study. The cut-off date for withdrawing your data is 14 days after you submitted the data. Please email members of the research team (see contact details below) within this 14-day window to withdraw your data from the study.

\subsection*{How will my data be used?}
The data collected from your participation will be pseudo-anonymized (stored with a unique numeric ID) and stored securely. It will be used for research purposes, and your personal information will remain confidential. The data will be analysed to gain insights into diverse preferences and perceptions regarding AI language model behaviours. At the end of the study, the pseudo-anonymised data collected will be released publicly for future research. The findings of this study may be published in academic journals or presented at conferences, and the results will be written up for a DPhil degree. Your individual identity will not be disclosed at any point in data release or publication. We do not collect any personal, private identifying information, IP addresses or contact details. The data we will collect that could identify you will be some demographic information (gender, age, nationality, religion, etc.), and short self-written survey answers.

The responses you provide will be stored in a password-protected electronic file on University of Oxford secure servers and may be used in academic publications, conference presentations or reports for external organisations. We will release a clean, PII-checked and pseudo-anonymised form of the data on an open-access, public data repository. Raw research data will be stored for 3 years after publication or public release of the research.     We would like to use the data in future studies, and to share data with other researchers (e.g. in online databases). Data will have identifying information removed before it is shared with other researchers or results are made public. The data that we collect from you may be transferred to, stored and/ or processed at a destination outside the UK and the European Economic Area. By submitting your personal data, you agree to this transfer, storing or processing.

\subsection*{Who has reviewed this research?}
This research has been reviewed by, and received ethics clearance through, a subcommittee of the University of Oxford Central University Research Ethics Committee [\texttt{OII\_C1A\_23\_088}].

\subsection*{Who do I contact if I have a concern or I wish to complain?}
If you have a concern about any aspect of this research, please speak to \mbox{Hannah Rose Kirk} (\texttt{hannah.kirk@oii.ox.ac.uk}) or their supervisor \mbox{Dr. Scott A. Hale} (\texttt{scott.hale@oii.ox.ac.uk}), and we will do our best to answer your query. We will acknowledge your concern within 10 working days and give you an indication of how it will be dealt with. If you remain unhappy or wish to make a formal complaint, please contact the Chair of the Research Ethics Committee at the University of Oxford who will seek to resolve the matter as soon as possible:  Social Sciences \& Humanities Interdivisional Research Ethics Committee; Email: \texttt{ethics@socsci.ox.ac.uk}; Address: Research Services, University of Oxford, Boundary Brook House, Churchill Drive, Headington, Oxford OX3 7GB.

\textbf{Please note that you may only participate in this survey if you are 18 years of age or over.}

$\bigcirc$ I certify that I am 18 years of age or over

\textbf{If you have read the information above and agree to participate with the understanding that the data (including any personal data) you submit will be processed accordingly, please tick the box below to start.}

$\bigcirc$ Yes, I agree to take part

\clearpage
\section{Metadata Processing}
\label{sec:appendix_meta_data}
For each text instance in \ourdata, we attach three pieces of metadata: detected \textbf{language} flags, detected \textbf{private or personally identifiable information (PII)} flags, and detected \textbf{moderation} flags.

\subsection{Structuring the Metadata}
There are five types of text instances. Two appear in the survey (\texttt{self\_description}, \mbox{\texttt{system\_string}}) and have a 1:1 matching with each user (\texttt{user\_id}). One appears at the conversation level (\texttt{open\_feedback}) and has a 1:1 matching with each \mbox{\texttt{convo\_id}} and a many:1 matching with each  \texttt{user\_id} because each participant has multiple conversations. Finally, the last two occur within each turn of a conversation, where for each single \texttt{user\_prompt} there are multiple model responses (\texttt{model\_response}). We structure the metadata so it can be merged uniquely, without duplication. We release one file, where each text instance is tied to its metadata via the identifying information shown in \cref{tab:appendix_meta_identifiers}, and a \texttt{column\_id} for matching whether the text is [\texttt{system\_string, self\_description, user\_prompt, model\_response, open\_feedback}].

\begin{table}
    \centering
    \footnotesize
    \caption{\small Identifiers of text instance types in \ourdata.}
    \label{tab:appendix_meta_identifiers}
    \begin{tabular}{ll|cccc}
    \toprule 
         & &  \multirow[b]{2}{*}{\rotatebox[origin=r]{90}{\texttt{user}}} & \multicolumn{2}{c}{$\overbrace{\makebox[2em]{}}^{\text{\rotatebox{90}{\texttt{interaction}}}}$} & \multirow[t]{2}{*}{\rotatebox[origin=c]{90}{\texttt{within turn}\hspace{0.5em}}} \\
        Text Instance & Study Stage & &  \rotatebox{90}{\texttt{convo}} & \rotatebox{90}{\texttt{turn}} & \\
        \hline
        \texttt{self\_description} & Survey & \checkmark &  &  & \\
        \texttt{system\_string} & Survey & \checkmark &  &  & \\
        \texttt{user\_prompt} &  Conversations & & \checkmark & \checkmark  & \\
        \texttt{model\_response}&  Conversations & & \checkmark & \checkmark & \checkmark\\
        \texttt{open\_feedback}& Conversations & & \checkmark  &  & \\
        \bottomrule
    \end{tabular}
\end{table}

\subsection{Automated Flagging}
\paragraph{PII} To identify whether a textual instance in our dataset contains personal and identifiable information (PII) we used the package \texttt{scrubadub}.\footnote{\url{https://scrubadub.readthedocs.io/en/stable/}} Specifically we used the function \texttt{scrubadub.clean(text)} which replaces the phone numbers and email addresses with anonymous IDs, if they are found in the input. We flag with $1$ instances that are altered (i.e., PII was identified) and $0$ those that remained unchanged. 

\paragraph{Moderation} To measure content moderation we use the OpenAI Moderation endpoint.\footnote{\url{https://platform.openai.com/docs/guides/moderation}} The API takes as an input a textual instance and outputs a json file with an overall boolean flag (\texttt{flagged}) whether there input potentially harmful (\texttt{True}), otherwise \texttt{False}. The API also returns a flag for a list of specific moderation categories that can be used to further filter and inspect the data. The categories are sexual, hate, harassment, self-harm, sexual/minors, hate/threatening, violence/graphic, self-harm/intent, self-harm/instructions, harassment/threatening and violence. Similar to the overall flag, for each category, the value is \texttt{True} if the model flags the corresponding category as violated, \texttt{False} otherwise. Finally, the API returns a dictionary of per-category scores that denote the model's confidence that the input violates the OpenAI's policy for the category. The value is between $0$ and $1$, where higher values denote higher confidence.

\paragraph{Language Detection}
To detect the language of each text instance in our dataset we used the \texttt{LangID} codebase.\footnote{\url{https://github.com/saffsd/langid.py}} \texttt{LangID} is a popular python package that efficiently detects the language of an input and currently supports $97$ languages. Specifically, we use the \texttt{langid.classify(text)} function and store a string for the detected language.

\subsection{Manual Review}
The overall proportions of texts flagged for PII, Non-English or Moderation is low (see \cref{tab:metadata_summary}). However, when inspecting the few positive flags, many were false positives, especially on lang-detect and PII. While a false positive on language may be relatively inconsequential, any automated flags of PII are concerning. Accordingly, we manually annotate any instances where \texttt{pii\_flag==True} ($n=167$) for participant-written text. We find that none of them actually contain PII.\footnote{If any of these human-written prompts had been a true positive, we would have manually checked the associated model responses too.}

\begin{table}[H]
\centering
\fontsize{8pt}{8pt}\selectfont
\setlength{\tabcolsep}{2.5pt}
\caption{\small \textbf{Meta-Data Summary.} For each metadata category (language, PII and moderation), we show $N (\%)$ for the dataset as a whole (\textit{Overall}) and broken down by each type of text instance in \ourdata.}
\label{tab:metadata_summary}
\begin{tabular}{lccccc}
\toprule
Category & Is English & Contains PII & Manually-Checked PII & Is Moderation Flagged & \textit{Total Instances} \\
\midrule
\textit{Overall} & 105,229 (98.8\%) & 1,111 (1.0\%) & NA & 634 (0.6\%) & 106,554 (100.0\%)\\
\texttt{user\_prompt} & 26,545 (97.7\%) & 66 (0.2\%) & 0 (0.0\%) & 454 (1.7\%) & 27,172 (100.0\%) \\
\texttt{model\_response} & 67,715 (99.0\%) & 944 (1.4\%) & NA & 162 (0.2\%) & 68,371 (100.0\%) \\
\texttt{self\_description} & 1,496 (99.7\%) & 10 (0.7\%) & 0 (0.0\%) & 7 (0.5\%) & 1,500 (100.0\%) \\
\texttt{system\_string} & 1,493 (99.5\%) & 16 (1.1\%) & 0 (0.0\%) & 0 (0.0\%) & 1,500 (100.0\%) \\
\texttt{open\_feedback} & 7,980 (99.6\%) & 75 (0.9\%) & 0 (0.0\%) & 11 (0.1\%) & 8,011 (100.0\%) \\
\bottomrule
\end{tabular}
\end{table}

\begin{table}[H]
\centering
\fontsize{8pt}{8pt}\selectfont
\caption{\small \textbf{Breakdown of flags from the OpenAI Moderation API.} We show counts and percentages where the text was flagged (==True), as well as total counts. Human-written text includes \texttt{user\_prompt, self\_description, system\_string, open\_feedback}; Model-written text is only \texttt{model\_response}.}
\label{tab:metadata_mod}
\begin{tabular}{lrrrr}
\toprule
 & \multicolumn{2}{c}{\textbf{Human-written}} & \multicolumn{2}{c}{\textbf{Model-written}} \\
 & N & (\%) & N & (\%) \\
\midrule
\textbf{sexual} & 21 & 0.05\% & 11 & 0.02\% \\
\textbf{hate} & 154 & 0.40\% & 36 & 0.05\% \\
\textbf{harassment} & 387 & 1.01\% & 127 & 0.19\% \\
\textbf{self-harm} & 24 & 0.06\% & 8 & 0.01\% \\
\textbf{sexual/minors} & 4 & 0.01\% & 4 & 0.01\% \\
\textbf{hate/threatening} & 17 & 0.04\% & 2 & 0.00\% \\
\textbf{self-harm/intent} & 26 & 0.07\% & 5 & 0.01\% \\
\textbf{self-harm/instructions} & 13 & 0.03\% & 8 & 0.01\% \\
\textbf{harassment/threatening} & 33 & 0.09\% & 7 & 0.01\% \\
\textbf{violence} & 52 & 0.14\% & 13 & 0.02\% \\
\midrule
\textbf{Total} & 38,183 & 100.00\% & 68,371 & 100.00\% \\
\bottomrule
\end{tabular}
\end{table}

\begin{table}[H]
\centering
\fontsize{8pt}{8pt}\selectfont
\caption{\small \textbf{Breakdown of languages detected by LangID.} We show the top-10 detected languages, then other and total counts. Human-written text includes \texttt{user\_prompt, self\_description, system\_string, open\_feedback}; Model-written text is only \texttt{model\_response}.}
\label{tab:metadata_lang}
\begin{tabular}{llrrlrr}
\toprule
 & \multicolumn{3}{c}{\textbf{Human-written}} & \multicolumn{3}{c}{\textbf{Model-written}} \\
 & Language & N & (\%) & Language & N & (\%) \\
\midrule
1 & en & 37,514 & 98.25\% & en & 67,715 & 99.04\% \\
2 & es & 175 & 0.46\% & es & 236 & 0.35\% \\
3 & fr & 71 & 0.19\% & de & 67 & 0.10\% \\
4 & it & 70 & 0.18\% & fr & 63 & 0.09\% \\
5 & de & 60 & 0.16\% & la & 43 & 0.06\% \\
6 & nl & 42 & 0.11\% & nl & 37 & 0.05\% \\
7 & pt & 41 & 0.11\% & it & 29 & 0.04\% \\
8 & pl & 34 & 0.09\% & sl & 19 & 0.03\% \\
9 & da & 24 & 0.06\% & hu & 15 & 0.02\% \\
10 & ro & 13 & 0.03\% & sv & 14 & 0.02\% \\
 & Other & 139 & 0.36\% & Other & 133 & 0.19\% \\
\midrule 
 & Total & 38,183 & 100.00\% & Total & 68,371 & 100.00\% \\
\bottomrule
\end{tabular}
\end{table}

\clearpage
\section{Annotating Ethnicity, Religion and Gender}
\label{sec:appendix_manual_annotation}
We ask people to describe their ethnic and religious affiliations in their own words because for a global survey, there are no immediately obvious preset categories. In the survey data, we release this original self-description (\texttt{\{ethnicity, religion\}\_self\_described}). However, there are 264 unique strings for ethnicity, and 137 unique strings for religion. For some analysis, it is valuable to have aggregate groupings. To attain this grouping, we first used \texttt{gpt-4-turbo} to categorise the strings, but found some errors and essentialising generalisations, for example, if someone answered with a nationality not an ethnic group like \textit{american}, \texttt{gpt} would return \textit{white}.\footnote{As an aside, these types of baked-in priors are a good example of why using LLMs as a surrogate for human annotators may introduce downstream biases \cite{agnewIllusion2024}.}

Accordingly, we used a second round of manual human annotation to verify these automated labels. Two annotators (authors of the paper) first made independent judgements then discussed any disagreements. For ethnicity, some participants also had answered a Prolific screening question on their simplified ethnicity, though we did not have this information for all participants as it was not mandatory. We thus annotate all unique combinations of the self-described string, and the Prolific ethnicity information ($n=343$). In ambiguous cases (e.g. the aforementioned \textit{american} response), we relied on this additional ethnicity information, and in its absence, defaulted to a \textit{Prefer not to say} response. For religion, we do not have any additional information provided by the Prolific pre-screening questionnaire, so verification decisions were made on the basis of the self-describe string alone. The annotators agreed on 94\% of ethnicity cases (discussing and resolving the remaining 20); and 96\% of religion cases (discussing and resolving the remaining 5).

We highlight two general findings from our disagreements which may be of interest to people analysing or categorising our data in the future. Firstly, \textbf{ethnicity and nationality are complex}. Take for example the UK census, where \textit{Chinese}, \textit{Banglaeshi}, \textit{Indian} and \textit{Pakistani} are all listed as sub-categories of the Asian ethnic group.\footnote{See the fact sheet at \href{https://www.ethnicity-facts-figures.service.gov.uk/style-guide/ethnic-groups/}{ethnicity-facts-figures.service.gov.uk}.} Ethnicity is a multi-faceted term which can include nationality, language group, skin colour, religion, among other characteristics \cite{barthEthnic1998}. Studies have shown that survey participants can interpret the term ethnic group through a variety of subjective lens \cite{hamerWhat2020, suyemotoWhat2020}. During annotation, we tried to gather information on whether group terms commonly refer to an ethnic group, but some subjectivity and naivety are inevitable; so, we encourage future researchers to carefully consider their own categorisations depending on the question at hand. Secondly, the \textbf{belonging and believing aspects of religion intersect} \cite{iannacconeIntroduction1998, levyReligious2012}, and it is not immediately clear how to categorise an individual that culturally affiliates with religion but simultaneously identities as an atheist or non-believer. Studies have revealed that the belonging and believing axis of religion are important for conditioning behaviours such as trust, pro-socality and altruism \cite{dingemansDoes2015, zhangExploring2019, saroglouBelieving2020}. In general, we annotated a mention of a religion as assigned to that religion (not distinguishing between the belonging and believing channels) but it remains to be seen whether one axis is more salient for values and opinions towards AI systems.

Note for gender, we provided a standard multiple choice question with options: \textit{Female}, \textit{Male}, \textit{Non-binary / third gender}, \textit{Prefer not to say} and \textit{Prefer to self-describe}. Only 3 individuals opted to self-describe, which we then annotated and only assimilated in very clear cut cases,\footnote{For example, one participant responded with ``i dont expect this wokery from intelligent people. you want to know which of the 2 possible genders i am male.'', which we assign as \textit{Male}.} else we grouped it as \textit{Prefer not to say} to avoid over-riding a participant's self-identification.

\section{Participant Demographics}
\label{sec:appendix_demographics}
\label{sec:appendix_participant_breakdowns}
We present full demographic breakdowns in \cref{tab:full_demographics}. We also compare the breakdowns in \ourdata to some early human feedback datasets which provide demographic information (\cref{tab:existing_paper_demographics}).

\footnotesize{
\begin{longtable}{lrl}
\caption{\small \textbf{Full Demographics Breakdowns.} Counts and percentages of participants by standard demographic variables. Overall, \ourdata utilises a large and demographically-diverse sample, especially compared to some previous human feedback datasets (see \cref{tab:existing_paper_demographics}); but it still generally skews towards young, white and educated populations. *For ethnicity and religion, see details in \cref{sec:appendix_manual_annotation}.}
\label{tab:full_demographics} \\
\endfirsthead
\caption[]{\small \textbf{Full Demographics Breakdowns.} Counts and percentages of participants by standard demographic variables. Overall, \ourdata utilises a large and demographically-diverse sample, especially compared to some previous human feedback datasets (see \cref{tab:existing_paper_demographics}); but it still generally skews towards young, white and educated populations. *For ethnicity and religion, see details in \cref{sec:appendix_manual_annotation}.} \\
\toprule
\midrule
\endhead
\midrule
\multicolumn{3}{r}{Continued on next page} \\
\midrule
\endfoot
\bottomrule
\endlastfoot
\midrule
{\cellcolor{white}} \textbf{Total Participants} & {\cellcolor{white}}   \textbf{1,500} & \pcb{100} {\cellcolor{white}}  \\
\midrule
{\cellcolor{white}} With conversations & {\cellcolor{white}} 1,396 & {\cellcolor{white}} \pcb{93.1}\% \\
{\cellcolor[HTML]{F2F2F2}} Just survey & {\cellcolor[HTML]{F2F2F2}} 104 & {\cellcolor[HTML]{F2F2F2}} \pcb{6.9}\% \\
\midrule
{\cellcolor{white}} \textbf{Age} & {\cellcolor{white}}  & {\cellcolor{white}}  \\
\midrule
{\cellcolor{white}} 25-34 years old & {\cellcolor{white}} 454 & {\cellcolor{white}} \pcb{30.3}\% \\
{\cellcolor[HTML]{F2F2F2}} 18-24 years old & {\cellcolor[HTML]{F2F2F2}} 297 & {\cellcolor[HTML]{F2F2F2}} \pcb{19.8}\% \\
{\cellcolor{white}} 35-44 years old & {\cellcolor{white}} 237 & {\cellcolor{white}} \pcb{15.8}\% \\
{\cellcolor[HTML]{F2F2F2}} 45-54 years old & {\cellcolor[HTML]{F2F2F2}} 208 & {\cellcolor[HTML]{F2F2F2}} \pcb{13.9}\% \\
{\cellcolor{white}} 55-64 years old & {\cellcolor{white}} 197 & {\cellcolor{white}} \pcb{13.1}\% \\
{\cellcolor[HTML]{F2F2F2}} 65+ years old & {\cellcolor[HTML]{F2F2F2}} 106 & {\cellcolor[HTML]{F2F2F2}} \pcb{7.1}\% \\
\itshape {\cellcolor{white}} Prefer not to say & {\cellcolor{white}} 1 & {\cellcolor{white}} \pcb{0.1}\% \\
\midrule
{\cellcolor{white}} \textbf{Gender} & {\cellcolor{white}}  & {\cellcolor{white}}  \\
\midrule
{\cellcolor{white}} Male & {\cellcolor{white}} 757 & {\cellcolor{white}} \pcb{50.5}\% \\
{\cellcolor[HTML]{F2F2F2}} Female & {\cellcolor[HTML]{F2F2F2}} 718 & {\cellcolor[HTML]{F2F2F2}} \pcb{47.9}\% \\
{\cellcolor{white}} Non-binary / third gender & {\cellcolor{white}} 21 & {\cellcolor{white}} \pcb{1.4}\% \\
\itshape {\cellcolor[HTML]{F2F2F2}} Prefer not to say & {\cellcolor[HTML]{F2F2F2}} 4 & {\cellcolor[HTML]{F2F2F2}} \pcb{0.3}\% \\
\midrule
{\cellcolor{white}} \textbf{Self-Reported Ethnicity*} & {\cellcolor{white}}  & {\cellcolor{white}}  \\
\midrule
{\cellcolor{white}} White & {\cellcolor{white}} 969 & {\cellcolor{white}} \pcb{64.6}\% \\
{\cellcolor[HTML]{F2F2F2}} Black / African & {\cellcolor[HTML]{F2F2F2}} 122 & {\cellcolor[HTML]{F2F2F2}} \pcb{8.1}\% \\
{\cellcolor{white}} Hispanic / Latino & {\cellcolor{white}} 121 & {\cellcolor{white}} \pcb{8.1}\% \\
{\cellcolor[HTML]{F2F2F2}} Asian & {\cellcolor[HTML]{F2F2F2}} 95 & {\cellcolor[HTML]{F2F2F2}} \pcb{6.3}\% \\
{\cellcolor{white}} Mixed & {\cellcolor{white}} 68 & {\cellcolor{white}} \pcb{4.5}\% \\
{\cellcolor[HTML]{F2F2F2}} Middle Eastern / Arab & {\cellcolor[HTML]{F2F2F2}} 14 & {\cellcolor[HTML]{F2F2F2}} \pcb{0.9}\% \\
{\cellcolor{white}} Indigenous / First Peoples & {\cellcolor{white}} 8 & {\cellcolor{white}} \pcb{0.5}\% \\
\itshape {\cellcolor[HTML]{F2F2F2}} Other & {\cellcolor[HTML]{F2F2F2}} 17 & {\cellcolor[HTML]{F2F2F2}} \pcb{1.1}\% \\
\itshape {\cellcolor{white}} Prefer not to say & {\cellcolor{white}} 86 & {\cellcolor{white}} \pcb{5.7}\% \\
\midrule
{\cellcolor{white}} \textbf{Self-Reported Religion*} & {\cellcolor{white}}  & {\cellcolor{white}}  \\
\midrule
{\cellcolor{white}} Non-religious & {\cellcolor{white}} 762 & {\cellcolor{white}} \pcb{50.8}\% \\
{\cellcolor[HTML]{F2F2F2}} Christian & {\cellcolor[HTML]{F2F2F2}} 487 & {\cellcolor[HTML]{F2F2F2}} \pcb{32.5}\% \\
{\cellcolor{white}} Agnostic & {\cellcolor{white}} 71 & {\cellcolor{white}} \pcb{4.7}\% \\
{\cellcolor[HTML]{F2F2F2}} Jewish & {\cellcolor[HTML]{F2F2F2}} 42 & {\cellcolor[HTML]{F2F2F2}} \pcb{2.8}\% \\
{\cellcolor{white}} Muslim & {\cellcolor{white}} 31 & {\cellcolor{white}} \pcb{2.1}\% \\
{\cellcolor[HTML]{F2F2F2}} Spiritual & {\cellcolor[HTML]{F2F2F2}} 18 & {\cellcolor[HTML]{F2F2F2}} \pcb{1.2}\% \\
{\cellcolor{white}} Buddhist & {\cellcolor{white}} 12 & {\cellcolor{white}} \pcb{0.8}\% \\
{\cellcolor[HTML]{F2F2F2}} Folk religion & {\cellcolor[HTML]{F2F2F2}} 6 & {\cellcolor[HTML]{F2F2F2}} \pcb{0.4}\% \\
{\cellcolor{white}} Hindu & {\cellcolor{white}} 5 & {\cellcolor{white}} \pcb{0.3}\% \\
{\cellcolor[HTML]{F2F2F2}} Sikh & {\cellcolor[HTML]{F2F2F2}} 3 & {\cellcolor[HTML]{F2F2F2}} \pcb{0.2}\% \\
\itshape {\cellcolor{white}} Other & {\cellcolor{white}} 4 & {\cellcolor{white}} \pcb{0.3}\% \\
\itshape {\cellcolor[HTML]{F2F2F2}} Prefer not to say & {\cellcolor[HTML]{F2F2F2}} 59 & {\cellcolor[HTML]{F2F2F2}} \pcb{3.9}\% \\
\midrule
{\cellcolor{white}} \textbf{Employment Status} & {\cellcolor{white}}  & {\cellcolor{white}}  \\
\midrule
{\cellcolor{white}} Working full-time & {\cellcolor{white}} 712 & {\cellcolor{white}} \pcb{47.5}\% \\
{\cellcolor[HTML]{F2F2F2}} Working part-time & {\cellcolor[HTML]{F2F2F2}} 265 & {\cellcolor[HTML]{F2F2F2}} \pcb{17.7}\% \\
{\cellcolor{white}} Student & {\cellcolor{white}} 191 & {\cellcolor{white}} \pcb{12.7}\% \\
{\cellcolor[HTML]{F2F2F2}} Unemployed, seeking work & {\cellcolor[HTML]{F2F2F2}} 113 & {\cellcolor[HTML]{F2F2F2}} \pcb{7.5}\% \\
{\cellcolor{white}} Retired & {\cellcolor{white}} 104 & {\cellcolor{white}} \pcb{6.9}\% \\
{\cellcolor[HTML]{F2F2F2}} Homemaker / Stay-at-home parent & {\cellcolor[HTML]{F2F2F2}} 46 & {\cellcolor[HTML]{F2F2F2}} \pcb{3.1}\% \\
{\cellcolor{white}} Unemployed, not seeking work & {\cellcolor{white}} 46 & {\cellcolor{white}} \pcb{3.1}\% \\
\itshape {\cellcolor[HTML]{F2F2F2}} Prefer not to say & {\cellcolor[HTML]{F2F2F2}} 23 & {\cellcolor[HTML]{F2F2F2}} \pcb{1.5}\% \\
\midrule
{\cellcolor{white}} \textbf{Education} & {\cellcolor{white}}  & {\cellcolor{white}}  \\
\midrule
{\cellcolor{white}} University Bachelors Degree & {\cellcolor{white}} 637 & {\cellcolor{white}} \pcb{42.5}\% \\
{\cellcolor[HTML]{F2F2F2}} Graduate / Professional degree & {\cellcolor[HTML]{F2F2F2}} 241 & {\cellcolor[HTML]{F2F2F2}} \pcb{16.1}\% \\
{\cellcolor{white}} Some University but no degree & {\cellcolor{white}} 236 & {\cellcolor{white}} \pcb{15.7}\% \\
{\cellcolor[HTML]{F2F2F2}} Completed Secondary School & {\cellcolor[HTML]{F2F2F2}} 209 & {\cellcolor[HTML]{F2F2F2}} \pcb{13.9}\% \\
{\cellcolor{white}} Vocational & {\cellcolor{white}} 125 & {\cellcolor{white}} \pcb{8.3}\% \\
{\cellcolor[HTML]{F2F2F2}} Some Secondary & {\cellcolor[HTML]{F2F2F2}} 24 & {\cellcolor[HTML]{F2F2F2}} \pcb{1.6}\% \\
{\cellcolor{white}} Completed Primary School & {\cellcolor{white}} 16 & {\cellcolor{white}} \pcb{1.1}\% \\
{\cellcolor[HTML]{F2F2F2}} Some Primary & {\cellcolor[HTML]{F2F2F2}} 3 & {\cellcolor[HTML]{F2F2F2}} \pcb{0.2}\% \\
\itshape {\cellcolor{white}} Prefer not to say & {\cellcolor{white}} 9 & {\cellcolor{white}} \pcb{0.6}\% \\
\midrule
{\cellcolor{white}} \textbf{Martial Status} & {\cellcolor{white}}  & {\cellcolor{white}}  \\
\midrule
{\cellcolor{white}} Never been married & {\cellcolor{white}} 870 & {\cellcolor{white}} \pcb{58.0}\% \\
{\cellcolor[HTML]{F2F2F2}} Married & {\cellcolor[HTML]{F2F2F2}} 463 & {\cellcolor[HTML]{F2F2F2}} \pcb{30.9}\% \\
{\cellcolor{white}} Divorced / Separated & {\cellcolor{white}} 123 & {\cellcolor{white}} \pcb{8.2}\% \\
{\cellcolor[HTML]{F2F2F2}} Widowed & {\cellcolor[HTML]{F2F2F2}} 21 & {\cellcolor[HTML]{F2F2F2}} \pcb{1.4}\% \\
\itshape {\cellcolor{white}} Prefer not to say & {\cellcolor{white}} 23 & {\cellcolor{white}} \pcb{1.5}\% \\
\midrule
{\cellcolor{white}} \textbf{English Proficiency} & {\cellcolor{white}}  & {\cellcolor{white}}  \\
\midrule
{\cellcolor{white}} Native speaker & {\cellcolor{white}} 886 & {\cellcolor{white}} \pcb{59.1}\% \\
{\cellcolor[HTML]{F2F2F2}} Fluent & {\cellcolor[HTML]{F2F2F2}} 405 & {\cellcolor[HTML]{F2F2F2}} \pcb{27.0}\% \\
{\cellcolor{white}} Advanced & {\cellcolor{white}} 160 & {\cellcolor{white}} \pcb{10.7}\% \\
{\cellcolor[HTML]{F2F2F2}} Intermediate & {\cellcolor[HTML]{F2F2F2}} 42 & {\cellcolor[HTML]{F2F2F2}} \pcb{2.8}\% \\
{\cellcolor{white}} Basic & {\cellcolor{white}} 7 & {\cellcolor{white}} \pcb{0.5}\% \\
\midrule
{\cellcolor{white}} \textbf{Regions} & {\cellcolor{white}}  & {\cellcolor{white}}  \\
\midrule
{\cellcolor{white}} US & {\cellcolor{white}} 338 & {\cellcolor{white}} \pcb{22.5}\% \\
{\cellcolor[HTML]{F2F2F2}} Europe & {\cellcolor[HTML]{F2F2F2}} 313 & {\cellcolor[HTML]{F2F2F2}} \pcb{20.9}\% \\
{\cellcolor{white}} UK & {\cellcolor{white}} 292 & {\cellcolor{white}} \pcb{19.5}\% \\
{\cellcolor[HTML]{F2F2F2}} Latin America and the Caribbean & {\cellcolor[HTML]{F2F2F2}} 146 & {\cellcolor[HTML]{F2F2F2}} \pcb{9.7}\% \\
{\cellcolor{white}} Australia and New Zealand & {\cellcolor{white}} 129 & {\cellcolor{white}} \pcb{8.6}\% \\
{\cellcolor[HTML]{F2F2F2}} Africa & {\cellcolor[HTML]{F2F2F2}} 118 & {\cellcolor[HTML]{F2F2F2}} \pcb{7.9}\% \\
{\cellcolor{white}} Asia & {\cellcolor{white}} 60 & {\cellcolor{white}} \pcb{4.0}\% \\
{\cellcolor[HTML]{F2F2F2}} Northern America & {\cellcolor[HTML]{F2F2F2}} 50 & {\cellcolor[HTML]{F2F2F2}} \pcb{3.3}\% \\
{\cellcolor{white}} Middle East & {\cellcolor{white}} 50 & {\cellcolor{white}} \pcb{3.3}\% \\
{\cellcolor[HTML]{F2F2F2}} Oceania & {\cellcolor[HTML]{F2F2F2}} 1 & {\cellcolor[HTML]{F2F2F2}} \pcb{0.1}\% \\
\itshape {\cellcolor{white}} Prefer not to say & {\cellcolor{white}} 3 & {\cellcolor{white}} \pcb{0.2}\% \\
\end{longtable}
}

\clearpage
\begin{table}
\footnotesize
\caption{\small \textbf{Demographic data compared to prior work.} Comparisons of \ourdata to early and widely-known RLHF studies using human feedback for language models. See \cref{sec:related_works} for more current datasets.}
\label{tab:existing_paper_demographics}
\setlength{\tabcolsep}{1pt}
\fontsize{9pt}{9pt}\selectfont
\begin{tabular}{lllllll}
\toprule
Category & \citeauthor{baiTraining2022} & \citeauthor{ouyangTraining2022} & \citeauthor{glaeseImproving2022} & \citeauthor{ganguliRed2022} & \citeauthor{stiennonLearning2020} & \textbf{Ours} \\
\midrule
{\cellcolor{white}} Total Participants & {\cellcolor{white}} 28$\ddagger$ & {\cellcolor{white}} 40 & {\cellcolor{white}}  & {\cellcolor{white}} 324 & {\cellcolor{white}}  & {\cellcolor{white}} 1,500 \\
{\cellcolor[HTML]{F2F2F2}} Demographic Respondents & {\cellcolor[HTML]{F2F2F2}} 28 & {\cellcolor[HTML]{F2F2F2}} 19 & {\cellcolor[HTML]{F2F2F2}} 533 & {\cellcolor[HTML]{F2F2F2}} 115 & {\cellcolor[HTML]{F2F2F2}} 21 & {\cellcolor[HTML]{F2F2F2}} 1,500 \\
\midrule
{\cellcolor{white}} \textbf{Gender} & {\cellcolor{white}} & {\cellcolor{white}} & {\cellcolor{white}} & {\cellcolor{white}} & {\cellcolor{white}} & {\cellcolor{white}} \\
\midrule
{\cellcolor{white}} Male & {\cellcolor{white}} \smallpcb{53.6}\% & {\cellcolor{white}} \smallpcb{47.4}\% & {\cellcolor{white}} \smallpcb{45.0}\% & {\cellcolor{white}} \smallpcb{47.0}\% & {\cellcolor{white}} \smallpcb{38.1}\% & {\cellcolor{white}} \smallpcb{50.5}\% \\
{\cellcolor[HTML]{F2F2F2}} Female & {\cellcolor[HTML]{F2F2F2}} \smallpcb{46.4}\% & {\cellcolor[HTML]{F2F2F2}} \smallpcb{42.1}\% & {\cellcolor[HTML]{F2F2F2}} \smallpcb{54.0}\% & {\cellcolor[HTML]{F2F2F2}} \smallpcb{52.2}\% & {\cellcolor[HTML]{F2F2F2}} \smallpcb{61.9}\% & {\cellcolor[HTML]{F2F2F2}} \smallpcb{47.9}\% \\
{\cellcolor{white}} Non-binary & {\cellcolor{white}} \smallpcb{0.0}\% & {\cellcolor{white}} \smallpcb{5.3}\% & {\cellcolor{white}} \smallpcb{1.0}\% & {\cellcolor{white}} \smallpcb{0.9}\% & {\cellcolor{white}} \smallpcb{0.0}\% & {\cellcolor{white}} \smallpcb{1.4}\% \\
\itshape {\cellcolor[HTML]{F2F2F2}} Prefer not to say/Other & {\cellcolor[HTML]{F2F2F2}} \smallpcb{0.0}\% & {\cellcolor[HTML]{F2F2F2}} \smallpcb{5.3}\% & {\cellcolor[HTML]{F2F2F2}} \smallpcb{0.0}\% & {\cellcolor[HTML]{F2F2F2}} \smallpcb{0.0}\% & {\cellcolor[HTML]{F2F2F2}} \smallpcb{0.0}\% & {\cellcolor[HTML]{F2F2F2}} \smallpcb{0.3}\% \\
\midrule
{\cellcolor{white}} \textbf{Sexual Orientation} & {\cellcolor{white}} & {\cellcolor{white}} & {\cellcolor{white}} & {\cellcolor{white}}  & {\cellcolor{white}} & {\cellcolor{white}} \\
\midrule
{\cellcolor{white}} Heterosexual & {\cellcolor{white}} \smallpcb{89.3}\% & {\cellcolor{white}} - & {\cellcolor{white}} \smallpcb{84.0}\% & {\cellcolor{white}} \smallpcb{81.7}\% & {\cellcolor{white}} - & {\cellcolor{white}} - \\
{\cellcolor[HTML]{F2F2F2}} Lesbian or Gay & {\cellcolor[HTML]{F2F2F2}} \smallpcb{7.1}\% & {\cellcolor[HTML]{F2F2F2}} - & {\cellcolor[HTML]{F2F2F2}} \smallpcb{5.0}\% & {\cellcolor[HTML]{F2F2F2}} \smallpcb{4.3}\% & {\cellcolor[HTML]{F2F2F2}} - & {\cellcolor[HTML]{F2F2F2}} - \\
{\cellcolor{white}} Bisexual & {\cellcolor{white}} \smallpcb{0.0}\% & {\cellcolor{white}} - & {\cellcolor{white}} \smallpcb{9.0}\% & {\cellcolor{white}} \smallpcb{12.2}\% & {\cellcolor{white}} - & {\cellcolor{white}} - \\
{\cellcolor[HTML]{F2F2F2}} Uncertain & {\cellcolor[HTML]{F2F2F2}} \smallpcb{3.6}\% & {\cellcolor[HTML]{F2F2F2}} - & {\cellcolor[HTML]{F2F2F2}} - & {\cellcolor[HTML]{F2F2F2}} \smallpcb{0.9}\% & {\cellcolor[HTML]{F2F2F2}} - & {\cellcolor[HTML]{F2F2F2}} - \\
\itshape {\cellcolor{white}} Prefer not to say/Other & {\cellcolor{white}} \smallpcb{0.0}\% & {\cellcolor{white}} - & {\cellcolor{white}} \smallpcb{2.0}\% & {\cellcolor{white}} \smallpcb{0.9}\% & {\cellcolor{white}} - & {\cellcolor{white}} - \\
\midrule
{\cellcolor{white}} \textbf{Age} & {\cellcolor{white}}  & {\cellcolor{white}}  & {\cellcolor{white}}  & {\cellcolor{white}}  & {\cellcolor{white}} \textdagger & {\cellcolor{white}} - \\
\midrule
{\cellcolor{white}} 18-24 & {\cellcolor{white}} \smallpcb{7.1}\% & {\cellcolor{white}} \smallpcb{26.3}\% & {\cellcolor{white}} \smallpcb{11.0}\% & {\cellcolor{white}} \smallpcb{0.0}\% & {\cellcolor{white}} - & {\cellcolor{white}} \smallpcb{19.8}\% \\
{\cellcolor[HTML]{F2F2F2}} 25-34 & {\cellcolor[HTML]{F2F2F2}} \smallpcb{39.3}\% & {\cellcolor[HTML]{F2F2F2}} \smallpcb{47.4}\% & {\cellcolor[HTML]{F2F2F2}} \smallpcb{37.0}\% & {\cellcolor[HTML]{F2F2F2}} \smallpcb{25.2}\% & {\cellcolor[HTML]{F2F2F2}} \smallpcb{42.9}\% & {\cellcolor[HTML]{F2F2F2}} \smallpcb{30.3}\% \\
{\cellcolor{white}} 35-44 & {\cellcolor{white}} \smallpcb{42.9}\% & {\cellcolor{white}} \smallpcb{10.5}\% & {\cellcolor{white}} \smallpcb{24.0}\% & {\cellcolor{white}} \smallpcb{33.9}\% & {\cellcolor{white}} \smallpcb{23.8}\% & {\cellcolor{white}} \smallpcb{15.8}\% \\
{\cellcolor[HTML]{F2F2F2}} 45-54 & {\cellcolor[HTML]{F2F2F2}} \smallpcb{10.7}\% & {\cellcolor[HTML]{F2F2F2}} \smallpcb{10.5}\% & {\cellcolor[HTML]{F2F2F2}} \smallpcb{16.0}\% & {\cellcolor[HTML]{F2F2F2}} \smallpcb{23.5}\% & {\cellcolor[HTML]{F2F2F2}} \smallpcb{23.8}\% & {\cellcolor[HTML]{F2F2F2}} \smallpcb{13.9}\% \\
{\cellcolor{white}} 55-64 & {\cellcolor{white}} \smallpcb{0.0}\% & {\cellcolor{white}} \smallpcb{5.3}\% & {\cellcolor{white}} \smallpcb{9.0}\% & {\cellcolor{white}} \smallpcb{13.9}\% & {\cellcolor{white}} \smallpcb{9.5}\% & {\cellcolor{white}} \smallpcb{13.1}\% \\
{\cellcolor[HTML]{F2F2F2}} 65+ & {\cellcolor[HTML]{F2F2F2}} \smallpcb{0.0}\% & {\cellcolor[HTML]{F2F2F2}} \smallpcb{0.0}\% & {\cellcolor[HTML]{F2F2F2}} \smallpcb{3.0}\% & {\cellcolor[HTML]{F2F2F2}} \smallpcb{1.7}\% & {\cellcolor[HTML]{F2F2F2}} \smallpcb{0.0}\% & {\cellcolor[HTML]{F2F2F2}} \smallpcb{7.1}\% \\
\itshape {\cellcolor{white}} Prefer not to say & {\cellcolor{white}} \smallpcb{0.0}\% & {\cellcolor{white}} - & {\cellcolor{white}} - & {\cellcolor{white}} \smallpcb{1.7}\% & {\cellcolor{white}} - & {\cellcolor{white}} \smallpcb{0.1}\% \\
\midrule
{\cellcolor{white}} \textbf{Ethnicity} & {\cellcolor{white}}  & {\cellcolor{white}}  & {\cellcolor{white}}  & {\cellcolor{white}}  & {\cellcolor{white}}  & {\cellcolor{white}}  \\
\midrule
{\cellcolor{white}} White/Caucasian & {\cellcolor{white}} \smallpcb{67.9}\% & {\cellcolor{white}} \smallpcb{31.6}\% & {\cellcolor{white}} \smallpcb{81.0}\% & {\cellcolor{white}} \smallpcb{81.7}\% & {\cellcolor{white}} \smallpcb{42.9}\% & {\cellcolor{white}} \smallpcb{64.6}\% \\
{\cellcolor[HTML]{F2F2F2}} Asian & {\cellcolor[HTML]{F2F2F2}} \smallpcb{10.7}\% & {\cellcolor[HTML]{F2F2F2}} \smallpcb{57.9}\% & {\cellcolor[HTML]{F2F2F2}} \smallpcb{8.0}\% & {\cellcolor[HTML]{F2F2F2}} \smallpcb{2.6}\% & {\cellcolor[HTML]{F2F2F2}} \smallpcb{28.6}\% & {\cellcolor[HTML]{F2F2F2}} \smallpcb{6.3}\% \\
{\cellcolor{white}} Black/African descent & {\cellcolor{white}} \smallpcb{3.6}\% & {\cellcolor{white}} \smallpcb{10.5}\% & {\cellcolor{white}} \smallpcb{4.0}\% & {\cellcolor{white}} \smallpcb{8.7}\% & {\cellcolor{white}} - & {\cellcolor{white}} \smallpcb{8.1}\% \\
{\cellcolor[HTML]{F2F2F2}} Hispanic/Latino & {\cellcolor[HTML]{F2F2F2}} \smallpcb{3.6}\% & {\cellcolor[HTML]{F2F2F2}} \smallpcb{15.8}\% & {\cellcolor[HTML]{F2F2F2}} \smallpcb{1.0}\% & {\cellcolor[HTML]{F2F2F2}} \smallpcb{0.9}\% & {\cellcolor[HTML]{F2F2F2}} \smallpcb{4.8}\% & {\cellcolor[HTML]{F2F2F2}} \smallpcb{8.1}\% \\
{\cellcolor{white}} Native American & {\cellcolor{white}} \smallpcb{0.0}\% & {\cellcolor{white}} \smallpcb{0.0}\% & {\cellcolor{white}} \smallpcb{0.0}\% & {\cellcolor{white}} \smallpcb{2.6}\% & {\cellcolor{white}} \smallpcb{9.6}\% & {\cellcolor{white}} \smallpcb{0.5}\% \\
{\cellcolor[HTML]{F2F2F2}} Middle Eastern & {\cellcolor[HTML]{F2F2F2}} \smallpcb{0.0}\% & {\cellcolor[HTML]{F2F2F2}} \smallpcb{0.0}\% & {\cellcolor[HTML]{F2F2F2}} \smallpcb{1.0}\% & {\cellcolor[HTML]{F2F2F2}} \smallpcb{0.9}\% & {\cellcolor[HTML]{F2F2F2}} \smallpcb{4.8}\% & {\cellcolor[HTML]{F2F2F2}} \smallpcb{0.9}\% \\
\itshape {\cellcolor{white}} Prefer not to say/Other & {\cellcolor{white}} \smallpcb{14.3}\% & {\cellcolor{white}} - & {\cellcolor{white}} \smallpcb{5.0}\% & {\cellcolor{white}} \smallpcb{2.6}\% & {\cellcolor{white}} \smallpcb{9.6}\% & {\cellcolor{white}} \smallpcb{11.5}\% \\
\midrule
{\cellcolor{white}} \textbf{Education} & {\cellcolor{white}}  & {\cellcolor{white}} & {\cellcolor{white}} & {\cellcolor{white}}  & {\cellcolor{white}} & {\cellcolor{white}} \\
\midrule
{\cellcolor{white}} No University Degree & {\cellcolor{white}} \smallpcb{17.9}\% & {\cellcolor{white}} \smallpcb{10.5}\% & {\cellcolor{white}} \smallpcb{0.0}\% & {\cellcolor{white}} \smallpcb{34.8}\% & {\cellcolor{white}} \smallpcb{14.3}\% & {\cellcolor{white}} \smallpcb{40.8}\% \\
{\cellcolor[HTML]{F2F2F2}} Undergraduate Degree & {\cellcolor[HTML]{F2F2F2}} \smallpcb{57.1}\% & {\cellcolor[HTML]{F2F2F2}} \smallpcb{52.6}\% & {\cellcolor[HTML]{F2F2F2}} \smallpcb{66.0}\% & {\cellcolor[HTML]{F2F2F2}} \smallpcb{53.9}\% & {\cellcolor[HTML]{F2F2F2}} \smallpcb{57.1}\% & {\cellcolor[HTML]{F2F2F2}} \smallpcb{42.5}\% \\
{\cellcolor{white}} Graduate Degree & {\cellcolor{white}} \smallpcb{14.3}\% & {\cellcolor{white}} \smallpcb{36.8}\% & {\cellcolor{white}} \smallpcb{34.0}\% & {\cellcolor{white}} \smallpcb{10.4}\% & {\cellcolor{white}} \smallpcb{28.1}\% & {\cellcolor{white}} \smallpcb{16.1}\% \\
\itshape {\cellcolor[HTML]{F2F2F2}} Prefer not to say/Other & {\cellcolor[HTML]{F2F2F2}} \smallpcb{10.7}\% & {\cellcolor[HTML]{F2F2F2}} - & {\cellcolor[HTML]{F2F2F2}} - & {\cellcolor[HTML]{F2F2F2}} \smallpcb{0.9}\% & {\cellcolor[HTML]{F2F2F2}} - & {\cellcolor[HTML]{F2F2F2}} \smallpcb{0.6}\% \\
\bottomrule
\end{tabular}
\begin{tabular}{p{\linewidth}}
      {\footnotesize\textdagger Age group values for \citeauthor{stiennonLearning2020} are reported for ten-year age groups starting from 20-29. We have placed the values in the row where the top end of these groups would appear to align with groups reported by the majority of studies.} \\ 
      {\footnotesize $\ddagger$ \citeauthor{baiTraining2022} provide two reports of demographic data. We use the one corresponding to the participants who contributed more than 80\% of the total feedback.}
    \end{tabular}
\end{table}

\begin{table}[H]
\caption{\small \textbf{Geographic data compared to prior work.} Participant countries of residence in \ourdata compared to early and widely-known RLHF studies using human feedback for language models.}
\label{tab:existing_paper_geography}
\setlength{\tabcolsep}{4pt}
\fontsize{9pt}{9pt}\selectfont
\begin{tabular}{lllllll}
\toprule
Category & \citeauthor{baiTraining2022} & \citeauthor{ouyangTraining2022} & \citeauthor{glaeseImproving2022} & \citeauthor{ganguliRed2022} & \citeauthor{stiennonLearning2020} & \textbf{Ours} \\
\midrule
{\cellcolor{white}} United States & {\cellcolor{white}} \smallpcbb{100}\% & {\cellcolor{white}} \smallpcb{17}\% & {\cellcolor{white}} \smallpcb{0}\% & {\cellcolor{white}} \smallpcbb{100}\% & {\cellcolor{white}} \smallpcbb{60}\% & {\cellcolor{white}} \smallpcb{26}\% \\
{\cellcolor[HTML]{F2F2F2}} United Kingdom & {\cellcolor[HTML]{F2F2F2}} \smallpcb{0}\% & {\cellcolor[HTML]{F2F2F2}} \smallpcb{0}\% & {\cellcolor[HTML]{F2F2F2}} \smallpcbb{100}\% & {\cellcolor[HTML]{F2F2F2}} \smallpcb{0}\% & {\cellcolor[HTML]{F2F2F2}} \smallpcb{7}\% & {\cellcolor[HTML]{F2F2F2}} \smallpcb{23}\% \\
{\cellcolor{white}} Philippines & {\cellcolor{white}} \smallpcb{0}\% & {\cellcolor{white}} \smallpcb{22}\% & {\cellcolor{white}} \smallpcb{0}\% & {\cellcolor{white}} \smallpcb{0}\% & {\cellcolor{white}} \smallpcb{7}\% & {\cellcolor{white}} \smallpcb{0}\% \\
{\cellcolor[HTML]{F2F2F2}} Bangladesh & {\cellcolor[HTML]{F2F2F2}} \smallpcb{0}\% & {\cellcolor[HTML]{F2F2F2}} \smallpcb{22}\% & {\cellcolor[HTML]{F2F2F2}} \smallpcb{0}\% & {\cellcolor[HTML]{F2F2F2}} \smallpcb{0}\% & {\cellcolor[HTML]{F2F2F2}} \smallpcb{0}\% & {\cellcolor[HTML]{F2F2F2}} \smallpcb{0}\% \\
{\cellcolor{white}} All Others & {\cellcolor{white}} \smallpcb{0}\% & {\cellcolor{white}} \smallpcbb{39}\%\textdagger & {\cellcolor{white}} \smallpcb{0}\% & {\cellcolor{white}} \smallpcb{0}\% & {\cellcolor{white}} \smallpcb{27}\%$\ddagger$ & {\cellcolor{white}} \smallpcbb{51}\%* \\
\bottomrule
\end{tabular}
    \begin{tabular}{l}
        \\
        \textdagger One resident each from Albania, Brazil, Canada, Columbia, India, Uruguay, and Zimbabwe \\
        $\ddagger$ One resident each from South Africa, Serbia, Turkey, India \\ 
        *See \cref{tab:full_geographics} for our breakdowns.\\
    \end{tabular}
    \label{tab:existing_geo}
\end{table}

\cleardoublepage
\section{Participant Geographies}
\label{sec:appendix_geographies}
\begin{figure}[H]
    \centering
    \includegraphics[width=\textwidth]{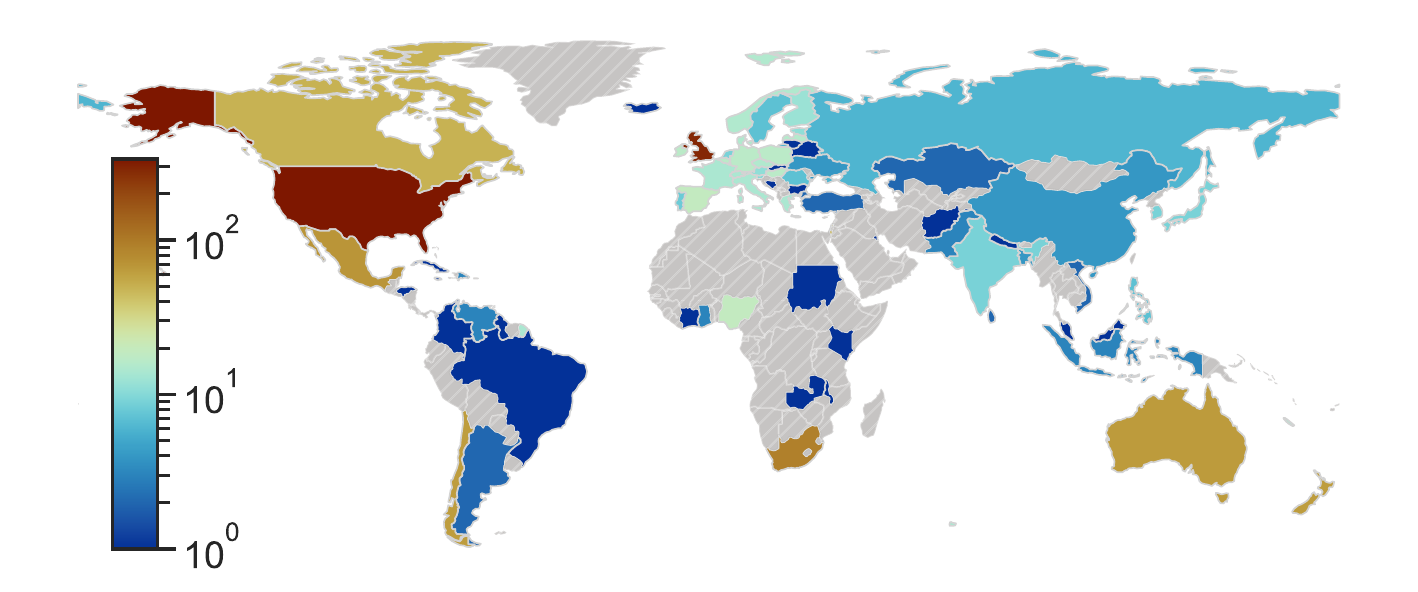}
    \caption{\small \textbf{Geographic distribution of \ourdata participants  by birth country}. Our sampling aims were for breath (coverage across UN global regions) and depth (representative demographic coverage within UK and US samples).}
    \label{fig:worldmap}
\end{figure}

\begin{figure}
    \centering
    \includegraphics[width=\textwidth]{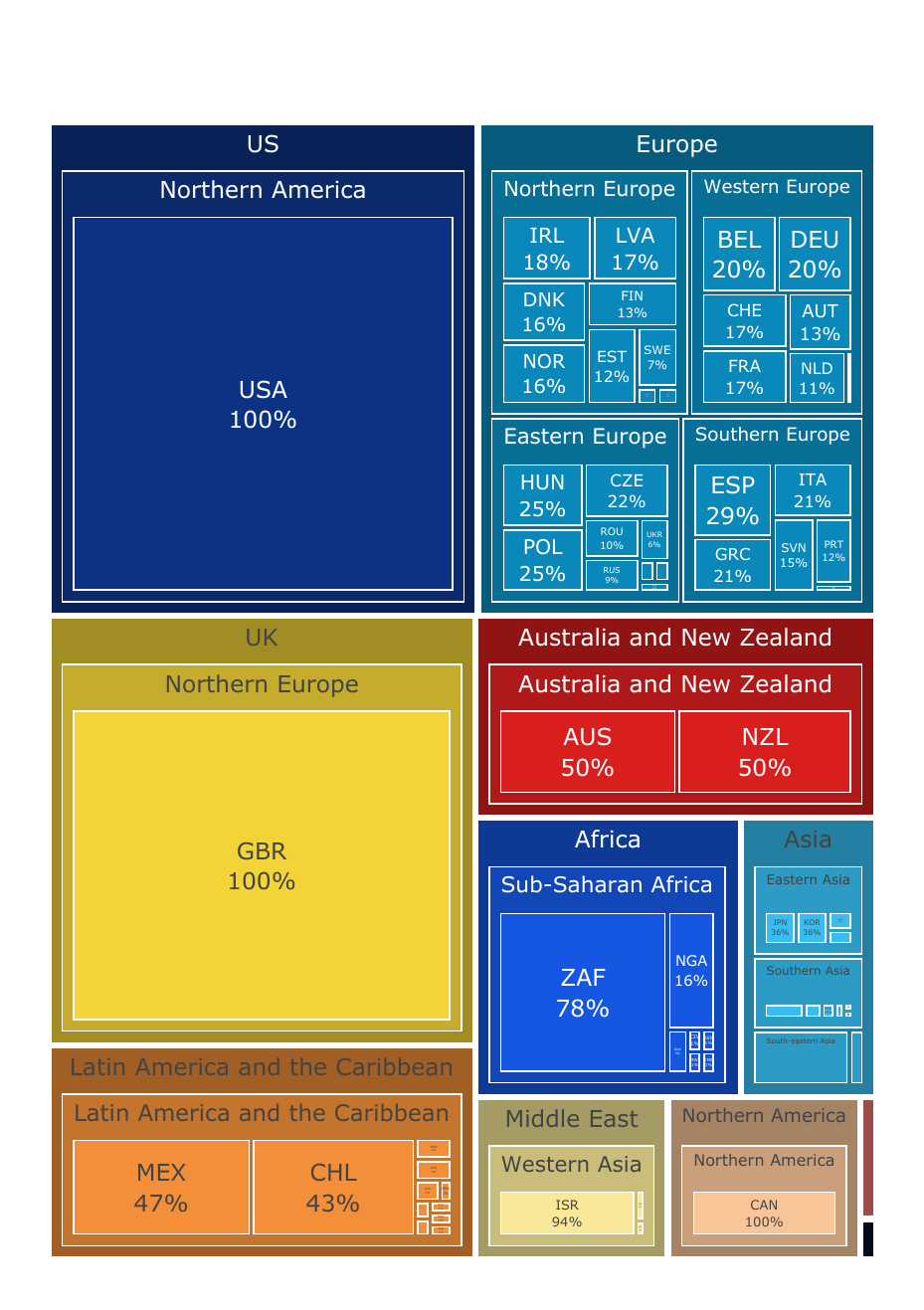}
    \caption{\small \textbf{Skewed regional entropy in \ourdata.} The hierarchical tree diagram uses participant birth location, mapping (i) special location  (splitting out the US and UK), which is used in the main paper, (ii) UN-defined subregions, and (iii) ISO country codes. There is an over-representation of UK and US participants due to the census samples. In most regions besides Europe, participation is dominated by one or two birth countries. The two small vertical boxes are Prefer not to say (in red), and Oceania (in navy). \textit{Note:} 88\% of \ourdata participants are born and currently reside in the same country.}
    \label{fig:regional_entropy}
\end{figure}

\normalsize
We present full geographic breakdowns in \cref{tab:full_geographics}. \cref{fig:worldmap} is an enlarged version from \cref{fig:splash}. We compare geographic data to prior work in \cref{tab:existing_paper_geography}. For regional classifications, we use the UN definitions.\footnote{\url{https://population.un.org/wpp/DefinitionOfRegions}} We also throughout the main paper use \texttt{location\_special\_region}, which splits out the UK and the US. Regional breakdowns by birth country are shown in \cref{fig:regional_entropy}.

{\footnotesize
\begin{longtable}{lrlrl}
\caption{\small \textbf{Full Geographic Breakdowns.} We collect country of birth and current country of residence. \ourdata contains participants born in 75 different countries, residing in 38 different countries.}
\label{tab:full_geographics} \\
\toprule
\textbf{} & \multicolumn{2}{l}{\textbf{Country of Birth}} &  \multicolumn{2}{l}{\textbf{Country of Residence}} \\
\midrule
\endfirsthead
\caption[]{\small \textbf{Full Geographic Breakdowns.} We collect country of birth and current country of residence. \ourdata contains participants born in 75 different countries, residing in 38 different countries.}\\
\toprule
\textbf{} & \multicolumn{2}{c}{\textbf{Country of Birth}} &  \multicolumn{2}{c}{\textbf{Country of Residence}} \\
\midrule
\endhead
\midrule
\multicolumn{5}{r}{Continued on next page} \\
\midrule
\endfoot
\bottomrule
\endlastfoot
{\cellcolor[HTML]{FFFFFF}} United States & {\cellcolor[HTML]{FFFFFF}} 338 & {\cellcolor[HTML]{FFFFFF}} \pcb{22.5} \% & {\cellcolor[HTML]{FFFFFF}} 386 & {\cellcolor[HTML]{FFFFFF}} \pcb{25.7} \% \\
{\cellcolor[HTML]{F2F2F2}} United Kingdom & {\cellcolor[HTML]{F2F2F2}} 292 & {\cellcolor[HTML]{F2F2F2}} \pcb{19.5} \% & {\cellcolor[HTML]{F2F2F2}} 340 & {\cellcolor[HTML]{F2F2F2}} \pcb{22.7} \% \\
{\cellcolor[HTML]{FFFFFF}} South Africa & {\cellcolor[HTML]{FFFFFF}} 91 & {\cellcolor[HTML]{FFFFFF}} \pcb{6.1} \% & {\cellcolor[HTML]{FFFFFF}} 86 & {\cellcolor[HTML]{FFFFFF}} \pcb{5.7} \% \\
{\cellcolor[HTML]{F2F2F2}} Mexico & {\cellcolor[HTML]{F2F2F2}} 69 & {\cellcolor[HTML]{F2F2F2}} \pcb{4.6} \% & {\cellcolor[HTML]{F2F2F2}} 67 & {\cellcolor[HTML]{F2F2F2}} \pcb{4.5} \% \\
{\cellcolor[HTML]{FFFFFF}} Australia & {\cellcolor[HTML]{FFFFFF}} 65 & {\cellcolor[HTML]{FFFFFF}} \pcb{4.3} \% & {\cellcolor[HTML]{FFFFFF}} 72 & {\cellcolor[HTML]{FFFFFF}} \pcb{4.8} \% \\
{\cellcolor[HTML]{F2F2F2}} New Zealand & {\cellcolor[HTML]{F2F2F2}} 64 & {\cellcolor[HTML]{F2F2F2}} \pcb{4.3} \% & {\cellcolor[HTML]{F2F2F2}} 77 & {\cellcolor[HTML]{F2F2F2}} \pcb{5.1} \% \\
{\cellcolor[HTML]{FFFFFF}} Chile & {\cellcolor[HTML]{FFFFFF}} 63 & {\cellcolor[HTML]{FFFFFF}} \pcb{4.2} \% & {\cellcolor[HTML]{FFFFFF}} 65 & {\cellcolor[HTML]{FFFFFF}} \pcb{4.3} \% \\
{\cellcolor[HTML]{F2F2F2}} Canada & {\cellcolor[HTML]{F2F2F2}} 50 & {\cellcolor[HTML]{F2F2F2}} \pcb{3.3} \% & {\cellcolor[HTML]{F2F2F2}} 54 & {\cellcolor[HTML]{F2F2F2}} \pcb{3.6} \% \\
{\cellcolor[HTML]{FFFFFF}} Israel & {\cellcolor[HTML]{FFFFFF}} 47 & {\cellcolor[HTML]{FFFFFF}} \pcb{3.1} \% & {\cellcolor[HTML]{FFFFFF}} 61 & {\cellcolor[HTML]{FFFFFF}} \pcb{4.1} \% \\
{\cellcolor[HTML]{F2F2F2}} Nigeria & {\cellcolor[HTML]{F2F2F2}} 19 & {\cellcolor[HTML]{F2F2F2}} \pcb{1.3} \% & {\cellcolor[HTML]{F2F2F2}} 0 & {\cellcolor[HTML]{F2F2F2}} \pcb{0.0} \% \\
{\cellcolor[HTML]{FFFFFF}} Spain & {\cellcolor[HTML]{FFFFFF}} 19 & {\cellcolor[HTML]{FFFFFF}} \pcb{1.3} \% & {\cellcolor[HTML]{FFFFFF}} 18 & {\cellcolor[HTML]{FFFFFF}} \pcb{1.2} \% \\
{\cellcolor[HTML]{F2F2F2}} Germany & {\cellcolor[HTML]{F2F2F2}} 17 & {\cellcolor[HTML]{F2F2F2}} \pcb{1.1} \% & {\cellcolor[HTML]{F2F2F2}} 13 & {\cellcolor[HTML]{F2F2F2}} \pcb{0.9} \% \\
{\cellcolor[HTML]{FFFFFF}} Belgium & {\cellcolor[HTML]{FFFFFF}} 17 & {\cellcolor[HTML]{FFFFFF}} \pcb{1.1} \% & {\cellcolor[HTML]{FFFFFF}} 17 & {\cellcolor[HTML]{FFFFFF}} \pcb{1.1} \% \\
{\cellcolor[HTML]{F2F2F2}} Hungary & {\cellcolor[HTML]{F2F2F2}} 17 & {\cellcolor[HTML]{F2F2F2}} \pcb{1.1} \% & {\cellcolor[HTML]{F2F2F2}} 16 & {\cellcolor[HTML]{F2F2F2}} \pcb{1.1} \% \\
{\cellcolor[HTML]{FFFFFF}} Poland & {\cellcolor[HTML]{FFFFFF}} 17 & {\cellcolor[HTML]{FFFFFF}} \pcb{1.1} \% & {\cellcolor[HTML]{FFFFFF}} 14 & {\cellcolor[HTML]{FFFFFF}} \pcb{0.9} \% \\
{\cellcolor[HTML]{F2F2F2}} Ireland & {\cellcolor[HTML]{F2F2F2}} 17 & {\cellcolor[HTML]{F2F2F2}} \pcb{1.1} \% & {\cellcolor[HTML]{F2F2F2}} 15 & {\cellcolor[HTML]{F2F2F2}} \pcb{1.0} \% \\
{\cellcolor[HTML]{FFFFFF}} Latvia & {\cellcolor[HTML]{FFFFFF}} 16 & {\cellcolor[HTML]{FFFFFF}} \pcb{1.1} \% & {\cellcolor[HTML]{FFFFFF}} 14 & {\cellcolor[HTML]{FFFFFF}} \pcb{0.9} \% \\
{\cellcolor[HTML]{F2F2F2}} Denmark & {\cellcolor[HTML]{F2F2F2}} 15 & {\cellcolor[HTML]{F2F2F2}} \pcb{1.0} \% & {\cellcolor[HTML]{F2F2F2}} 15 & {\cellcolor[HTML]{F2F2F2}} \pcb{1.0} \% \\
{\cellcolor[HTML]{FFFFFF}} Czechia & {\cellcolor[HTML]{FFFFFF}} 15 & {\cellcolor[HTML]{FFFFFF}} \pcb{1.0} \% & {\cellcolor[HTML]{FFFFFF}} 14 & {\cellcolor[HTML]{FFFFFF}} \pcb{0.9} \% \\
{\cellcolor[HTML]{F2F2F2}} Norway & {\cellcolor[HTML]{F2F2F2}} 15 & {\cellcolor[HTML]{F2F2F2}} \pcb{1.0} \% & {\cellcolor[HTML]{F2F2F2}} 15 & {\cellcolor[HTML]{F2F2F2}} \pcb{1.0} \% \\
{\cellcolor[HTML]{FFFFFF}} France & {\cellcolor[HTML]{FFFFFF}} 14 & {\cellcolor[HTML]{FFFFFF}} \pcb{0.9} \% & {\cellcolor[HTML]{FFFFFF}} 12 & {\cellcolor[HTML]{FFFFFF}} \pcb{0.8} \% \\
{\cellcolor[HTML]{F2F2F2}} Italy & {\cellcolor[HTML]{F2F2F2}} 14 & {\cellcolor[HTML]{F2F2F2}} \pcb{0.9} \% & {\cellcolor[HTML]{F2F2F2}} 13 & {\cellcolor[HTML]{F2F2F2}} \pcb{0.9} \% \\
{\cellcolor[HTML]{FFFFFF}} Greece & {\cellcolor[HTML]{FFFFFF}} 14 & {\cellcolor[HTML]{FFFFFF}} \pcb{0.9} \% & {\cellcolor[HTML]{FFFFFF}} 13 & {\cellcolor[HTML]{FFFFFF}} \pcb{0.9} \% \\
{\cellcolor[HTML]{F2F2F2}} Switzerland & {\cellcolor[HTML]{F2F2F2}} 14 & {\cellcolor[HTML]{F2F2F2}} \pcb{0.9} \% & {\cellcolor[HTML]{F2F2F2}} 14 & {\cellcolor[HTML]{F2F2F2}} \pcb{0.9} \% \\
{\cellcolor[HTML]{FFFFFF}} Finland & {\cellcolor[HTML]{FFFFFF}} 12 & {\cellcolor[HTML]{FFFFFF}} \pcb{0.8} \% & {\cellcolor[HTML]{FFFFFF}} 13 & {\cellcolor[HTML]{FFFFFF}} \pcb{0.9} \% \\
{\cellcolor[HTML]{F2F2F2}} Estonia & {\cellcolor[HTML]{F2F2F2}} 11 & {\cellcolor[HTML]{F2F2F2}} \pcb{0.7} \% & {\cellcolor[HTML]{F2F2F2}} 10 & {\cellcolor[HTML]{F2F2F2}} \pcb{0.7} \% \\
{\cellcolor[HTML]{FFFFFF}} Austria & {\cellcolor[HTML]{FFFFFF}} 11 & {\cellcolor[HTML]{FFFFFF}} \pcb{0.7} \% & {\cellcolor[HTML]{FFFFFF}} 10 & {\cellcolor[HTML]{FFFFFF}} \pcb{0.7} \% \\
{\cellcolor[HTML]{F2F2F2}} Slovenia & {\cellcolor[HTML]{F2F2F2}} 10 & {\cellcolor[HTML]{F2F2F2}} \pcb{0.7} \% & {\cellcolor[HTML]{F2F2F2}} 10 & {\cellcolor[HTML]{F2F2F2}} \pcb{0.7} \% \\
{\cellcolor[HTML]{FFFFFF}} Netherlands & {\cellcolor[HTML]{FFFFFF}} 9 & {\cellcolor[HTML]{FFFFFF}} \pcb{0.6} \% & {\cellcolor[HTML]{FFFFFF}} 8 & {\cellcolor[HTML]{FFFFFF}} \pcb{0.5} \% \\
{\cellcolor[HTML]{F2F2F2}} India & {\cellcolor[HTML]{F2F2F2}} 9 & {\cellcolor[HTML]{F2F2F2}} \pcb{0.6} \% & {\cellcolor[HTML]{F2F2F2}} 0 & {\cellcolor[HTML]{F2F2F2}} \pcb{0.0} \% \\
{\cellcolor[HTML]{FFFFFF}} Japan & {\cellcolor[HTML]{FFFFFF}} 9 & {\cellcolor[HTML]{FFFFFF}} \pcb{0.6} \% & {\cellcolor[HTML]{FFFFFF}} 11 & {\cellcolor[HTML]{FFFFFF}} \pcb{0.7} \% \\
{\cellcolor[HTML]{F2F2F2}} Korea, Republic of & {\cellcolor[HTML]{F2F2F2}} 9 & {\cellcolor[HTML]{F2F2F2}} \pcb{0.6} \% & {\cellcolor[HTML]{F2F2F2}} 7 & {\cellcolor[HTML]{F2F2F2}} \pcb{0.5} \% \\
{\cellcolor[HTML]{FFFFFF}} Portugal & {\cellcolor[HTML]{FFFFFF}} 8 & {\cellcolor[HTML]{FFFFFF}} \pcb{0.5} \% & {\cellcolor[HTML]{FFFFFF}} 7 & {\cellcolor[HTML]{FFFFFF}} \pcb{0.5} \% \\
{\cellcolor[HTML]{F2F2F2}} Romania & {\cellcolor[HTML]{F2F2F2}} 7 & {\cellcolor[HTML]{F2F2F2}} \pcb{0.5} \% & {\cellcolor[HTML]{F2F2F2}} 0 & {\cellcolor[HTML]{F2F2F2}} \pcb{0.0} \% \\
{\cellcolor[HTML]{FFFFFF}} Philippines & {\cellcolor[HTML]{FFFFFF}} 7 & {\cellcolor[HTML]{FFFFFF}} \pcb{0.5} \% & {\cellcolor[HTML]{FFFFFF}} 0 & {\cellcolor[HTML]{FFFFFF}} \pcb{0.0} \% \\
{\cellcolor[HTML]{F2F2F2}} Sweden & {\cellcolor[HTML]{F2F2F2}} 7 & {\cellcolor[HTML]{F2F2F2}} \pcb{0.5} \% & {\cellcolor[HTML]{F2F2F2}} 6 & {\cellcolor[HTML]{F2F2F2}} \pcb{0.4} \% \\
{\cellcolor[HTML]{FFFFFF}} Russian Federation & {\cellcolor[HTML]{FFFFFF}} 6 & {\cellcolor[HTML]{FFFFFF}} \pcb{0.4} \% & {\cellcolor[HTML]{FFFFFF}} 0 & {\cellcolor[HTML]{FFFFFF}} \pcb{0.0} \% \\
{\cellcolor[HTML]{F2F2F2}} Ukraine & {\cellcolor[HTML]{F2F2F2}} 4 & {\cellcolor[HTML]{F2F2F2}} \pcb{0.3} \% & {\cellcolor[HTML]{F2F2F2}} 0 & {\cellcolor[HTML]{F2F2F2}} \pcb{0.0} \% \\
{\cellcolor[HTML]{FFFFFF}} Bangladesh & {\cellcolor[HTML]{FFFFFF}} 4 & {\cellcolor[HTML]{FFFFFF}} \pcb{0.3} \% & {\cellcolor[HTML]{FFFFFF}} 0 & {\cellcolor[HTML]{FFFFFF}} \pcb{0.0} \% \\
{\cellcolor[HTML]{F2F2F2}} China & {\cellcolor[HTML]{F2F2F2}} 4 & {\cellcolor[HTML]{F2F2F2}} \pcb{0.3} \% & {\cellcolor[HTML]{F2F2F2}} 0 & {\cellcolor[HTML]{F2F2F2}} \pcb{0.0} \% \\
{\cellcolor[HTML]{FFFFFF}} Hong Kong & {\cellcolor[HTML]{FFFFFF}} 3 & {\cellcolor[HTML]{FFFFFF}} \pcb{0.2} \% & {\cellcolor[HTML]{FFFFFF}} 0 & {\cellcolor[HTML]{FFFFFF}} \pcb{0.0} \% \\
{\cellcolor[HTML]{F2F2F2}} Pakistan & {\cellcolor[HTML]{F2F2F2}} 3 & {\cellcolor[HTML]{F2F2F2}} \pcb{0.2} \% & {\cellcolor[HTML]{F2F2F2}} 0 & {\cellcolor[HTML]{F2F2F2}} \pcb{0.0} \% \\
{\cellcolor[HTML]{FFFFFF}} Ghana & {\cellcolor[HTML]{FFFFFF}} 3 & {\cellcolor[HTML]{FFFFFF}} \pcb{0.2} \% & {\cellcolor[HTML]{FFFFFF}} 0 & {\cellcolor[HTML]{FFFFFF}} \pcb{0.0} \% \\
{\cellcolor[HTML]{F2F2F2}} Dominican Republic & {\cellcolor[HTML]{F2F2F2}} 3 & {\cellcolor[HTML]{F2F2F2}} \pcb{0.2} \% & {\cellcolor[HTML]{F2F2F2}} 0 & {\cellcolor[HTML]{F2F2F2}} \pcb{0.0} \% \\
{\cellcolor[HTML]{FFFFFF}} Venezuela, Bolivarian Republic of & {\cellcolor[HTML]{FFFFFF}} 3 & {\cellcolor[HTML]{FFFFFF}} \pcb{0.2} \% & {\cellcolor[HTML]{FFFFFF}} 0 & {\cellcolor[HTML]{FFFFFF}} \pcb{0.0} \% \\
{\cellcolor[HTML]{F2F2F2}} Indonesia & {\cellcolor[HTML]{F2F2F2}} 3 & {\cellcolor[HTML]{F2F2F2}} \pcb{0.2} \% & {\cellcolor[HTML]{F2F2F2}} 0 & {\cellcolor[HTML]{F2F2F2}} \pcb{0.0} \% \\
{\cellcolor[HTML]{FFFFFF}} Viet Nam & {\cellcolor[HTML]{FFFFFF}} 2 & {\cellcolor[HTML]{FFFFFF}} \pcb{0.1} \% & {\cellcolor[HTML]{FFFFFF}} 0 & {\cellcolor[HTML]{FFFFFF}} \pcb{0.0} \% \\
{\cellcolor[HTML]{F2F2F2}} Sri Lanka & {\cellcolor[HTML]{F2F2F2}} 2 & {\cellcolor[HTML]{F2F2F2}} \pcb{0.1} \% & {\cellcolor[HTML]{F2F2F2}} 0 & {\cellcolor[HTML]{F2F2F2}} \pcb{0.0} \% \\
{\cellcolor[HTML]{FFFFFF}} Turkey & {\cellcolor[HTML]{FFFFFF}} 2 & {\cellcolor[HTML]{FFFFFF}} \pcb{0.1} \% & {\cellcolor[HTML]{FFFFFF}} 0 & {\cellcolor[HTML]{FFFFFF}} \pcb{0.0} \% \\
{\cellcolor[HTML]{F2F2F2}} Argentina & {\cellcolor[HTML]{F2F2F2}} 2 & {\cellcolor[HTML]{F2F2F2}} \pcb{0.1} \% & {\cellcolor[HTML]{F2F2F2}} 0 & {\cellcolor[HTML]{F2F2F2}} \pcb{0.0} \% \\
{\cellcolor[HTML]{FFFFFF}} Kazakhstan & {\cellcolor[HTML]{FFFFFF}} 2 & {\cellcolor[HTML]{FFFFFF}} \pcb{0.1} \% & {\cellcolor[HTML]{FFFFFF}} 0 & {\cellcolor[HTML]{FFFFFF}} \pcb{0.0} \% \\
{\cellcolor[HTML]{F2F2F2}} Slovakia & {\cellcolor[HTML]{F2F2F2}} 1 & {\cellcolor[HTML]{F2F2F2}} \pcb{0.1} \% & {\cellcolor[HTML]{F2F2F2}} 0 & {\cellcolor[HTML]{F2F2F2}} \pcb{0.0} \% \\
{\cellcolor[HTML]{FFFFFF}} Sudan & {\cellcolor[HTML]{FFFFFF}} 1 & {\cellcolor[HTML]{FFFFFF}} \pcb{0.1} \% & {\cellcolor[HTML]{FFFFFF}} 0 & {\cellcolor[HTML]{FFFFFF}} \pcb{0.0} \% \\
{\cellcolor[HTML]{F2F2F2}} Tonga & {\cellcolor[HTML]{F2F2F2}} 1 & {\cellcolor[HTML]{F2F2F2}} \pcb{0.1} \% & {\cellcolor[HTML]{F2F2F2}} 0 & {\cellcolor[HTML]{F2F2F2}} \pcb{0.0} \% \\
{\cellcolor[HTML]{FFFFFF}} Afghanistan & {\cellcolor[HTML]{FFFFFF}} 1 & {\cellcolor[HTML]{FFFFFF}} \pcb{0.1} \% & {\cellcolor[HTML]{FFFFFF}} 0 & {\cellcolor[HTML]{FFFFFF}} \pcb{0.0} \% \\
{\cellcolor[HTML]{F2F2F2}} Nepal & {\cellcolor[HTML]{F2F2F2}} 1 & {\cellcolor[HTML]{F2F2F2}} \pcb{0.1} \% & {\cellcolor[HTML]{F2F2F2}} 0 & {\cellcolor[HTML]{F2F2F2}} \pcb{0.0} \% \\
{\cellcolor[HTML]{FFFFFF}} Honduras & {\cellcolor[HTML]{FFFFFF}} 1 & {\cellcolor[HTML]{FFFFFF}} \pcb{0.1} \% & {\cellcolor[HTML]{FFFFFF}} 0 & {\cellcolor[HTML]{FFFFFF}} \pcb{0.0} \% \\
{\cellcolor[HTML]{F2F2F2}} Belarus & {\cellcolor[HTML]{F2F2F2}} 1 & {\cellcolor[HTML]{F2F2F2}} \pcb{0.1} \% & {\cellcolor[HTML]{F2F2F2}} 0 & {\cellcolor[HTML]{F2F2F2}} \pcb{0.0} \% \\
{\cellcolor[HTML]{FFFFFF}} Bosnia and Herzegovina & {\cellcolor[HTML]{FFFFFF}} 1 & {\cellcolor[HTML]{FFFFFF}} \pcb{0.1} \% & {\cellcolor[HTML]{FFFFFF}} 0 & {\cellcolor[HTML]{FFFFFF}} \pcb{0.0} \% \\
{\cellcolor[HTML]{F2F2F2}} Brazil & {\cellcolor[HTML]{F2F2F2}} 1 & {\cellcolor[HTML]{F2F2F2}} \pcb{0.1} \% & {\cellcolor[HTML]{F2F2F2}} 0 & {\cellcolor[HTML]{F2F2F2}} \pcb{0.0} \% \\
{\cellcolor[HTML]{FFFFFF}} Bulgaria & {\cellcolor[HTML]{FFFFFF}} 1 & {\cellcolor[HTML]{FFFFFF}} \pcb{0.1} \% & {\cellcolor[HTML]{FFFFFF}} 0 & {\cellcolor[HTML]{FFFFFF}} \pcb{0.0} \% \\
{\cellcolor[HTML]{F2F2F2}} Colombia & {\cellcolor[HTML]{F2F2F2}} 1 & {\cellcolor[HTML]{F2F2F2}} \pcb{0.1} \% & {\cellcolor[HTML]{F2F2F2}} 0 & {\cellcolor[HTML]{F2F2F2}} \pcb{0.0} \% \\
{\cellcolor[HTML]{FFFFFF}} Cuba & {\cellcolor[HTML]{FFFFFF}} 1 & {\cellcolor[HTML]{FFFFFF}} \pcb{0.1} \% & {\cellcolor[HTML]{FFFFFF}} 0 & {\cellcolor[HTML]{FFFFFF}} \pcb{0.0} \% \\
{\cellcolor[HTML]{F2F2F2}} Côte d'Ivoire & {\cellcolor[HTML]{F2F2F2}} 1 & {\cellcolor[HTML]{F2F2F2}} \pcb{0.1} \% & {\cellcolor[HTML]{F2F2F2}} 0 & {\cellcolor[HTML]{F2F2F2}} \pcb{0.0} \% \\
{\cellcolor[HTML]{FFFFFF}} Malaysia & {\cellcolor[HTML]{FFFFFF}} 1 & {\cellcolor[HTML]{FFFFFF}} \pcb{0.1} \% & {\cellcolor[HTML]{FFFFFF}} 0 & {\cellcolor[HTML]{FFFFFF}} \pcb{0.0} \% \\
{\cellcolor[HTML]{F2F2F2}} Guyana & {\cellcolor[HTML]{F2F2F2}} 1 & {\cellcolor[HTML]{F2F2F2}} \pcb{0.1} \% & {\cellcolor[HTML]{F2F2F2}} 0 & {\cellcolor[HTML]{F2F2F2}} \pcb{0.0} \% \\
{\cellcolor[HTML]{FFFFFF}} Iceland & {\cellcolor[HTML]{FFFFFF}} 1 & {\cellcolor[HTML]{FFFFFF}} \pcb{0.1} \% & {\cellcolor[HTML]{FFFFFF}} 1 & {\cellcolor[HTML]{FFFFFF}} \pcb{0.1} \% \\
{\cellcolor[HTML]{F2F2F2}} Jamaica & {\cellcolor[HTML]{F2F2F2}} 1 & {\cellcolor[HTML]{F2F2F2}} \pcb{0.1} \% & {\cellcolor[HTML]{F2F2F2}} 0 & {\cellcolor[HTML]{F2F2F2}} \pcb{0.0} \% \\
{\cellcolor[HTML]{FFFFFF}} Kenya & {\cellcolor[HTML]{FFFFFF}} 1 & {\cellcolor[HTML]{FFFFFF}} \pcb{0.1} \% & {\cellcolor[HTML]{FFFFFF}} 0 & {\cellcolor[HTML]{FFFFFF}} \pcb{0.0} \% \\
{\cellcolor[HTML]{F2F2F2}} Kuwait & {\cellcolor[HTML]{F2F2F2}} 1 & {\cellcolor[HTML]{F2F2F2}} \pcb{0.1} \% & {\cellcolor[HTML]{F2F2F2}} 0 & {\cellcolor[HTML]{F2F2F2}} \pcb{0.0} \% \\
{\cellcolor[HTML]{FFFFFF}} Lithuania & {\cellcolor[HTML]{FFFFFF}} 1 & {\cellcolor[HTML]{FFFFFF}} \pcb{0.1} \% & {\cellcolor[HTML]{FFFFFF}} 0 & {\cellcolor[HTML]{FFFFFF}} \pcb{0.0} \% \\
{\cellcolor[HTML]{F2F2F2}} Luxembourg & {\cellcolor[HTML]{F2F2F2}} 1 & {\cellcolor[HTML]{F2F2F2}} \pcb{0.1} \% & {\cellcolor[HTML]{F2F2F2}} 2 & {\cellcolor[HTML]{F2F2F2}} \pcb{0.1} \% \\
{\cellcolor[HTML]{FFFFFF}} Malawi & {\cellcolor[HTML]{FFFFFF}} 1 & {\cellcolor[HTML]{FFFFFF}} \pcb{0.1} \% & {\cellcolor[HTML]{FFFFFF}} 0 & {\cellcolor[HTML]{FFFFFF}} \pcb{0.0} \% \\
{\cellcolor[HTML]{F2F2F2}} Zambia & {\cellcolor[HTML]{F2F2F2}} 1 & {\cellcolor[HTML]{F2F2F2}} \pcb{0.1} \% & {\cellcolor[HTML]{F2F2F2}} 0 & {\cellcolor[HTML]{F2F2F2}} \pcb{0.0} \% \\
{\cellcolor[HTML]{FFFFFF}} Tanzania, United Republic of & {\cellcolor[HTML]{FFFFFF}} 0 & {\cellcolor[HTML]{FFFFFF}} \pcb{0.0} \% & {\cellcolor[HTML]{FFFFFF}} 1 & {\cellcolor[HTML]{FFFFFF}} \pcb{0.1} \% \\
{\cellcolor[HTML]{F2F2F2}} Lesotho & {\cellcolor[HTML]{F2F2F2}} 0 & {\cellcolor[HTML]{F2F2F2}} \pcb{0.0} \% & {\cellcolor[HTML]{F2F2F2}} 1 & {\cellcolor[HTML]{F2F2F2}} \pcb{0.1} \% \\
{\cellcolor[HTML]{FFFFFF}} Uruguay & {\cellcolor[HTML]{FFFFFF}} 0 & {\cellcolor[HTML]{FFFFFF}} \pcb{0.0} \% & {\cellcolor[HTML]{FFFFFF}} 1 & {\cellcolor[HTML]{FFFFFF}} \pcb{0.1} \% \\
\itshape {\cellcolor[HTML]{F2F2F2}} Prefer not to say & {\cellcolor[HTML]{F2F2F2}} 3 & {\cellcolor[HTML]{F2F2F2}} \pcb{0.2} \% & {\cellcolor[HTML]{F2F2F2}} 1 & {\cellcolor[HTML]{F2F2F2}} \pcb{0.1} \% \\
\end{longtable}
}

\clearpage
\section{Participant LLM Usage and Familiarity}
\label{sec:appendix_llm_usage}
We present breakdowns on experience with LLMs in \cref{tab:llm_usage}. We did not require participants to be familiar with LLMs so provide the following primer:

\hrulefill

This research is about Artificial Intelligence (AI) Language Models.

These models are also sometimes referred to as Generative AI, Large Language Models (LLMs), Conversational Agents, AI Chat Bots or Virtual Assistants.

They are advanced computer programs that can understand and generate human-like text. These models learn from large amounts of text data on the internet to generate their responses.

One example you might have heard is ChatGPT, where people can have a conversation with an AI language model via an internet website.

\hrulefill

\fontsize{9pt}{9pt}\selectfont
\setlength{\tabcolsep}{1pt}
\begin{longtable}{p{0.625\linewidth}rl}
\caption{\small \textbf{Survey of Participants' LLM Usage}: The majority of participants have used LLMs directly (via a dedicated chat interface) and indirectly (embedded in products or services). Note only participants who answered \textit{Yes} to \textbf{LLM Direct Use} or \textbf{LLM Indirect Use} ($n=1253$, 84\%) are shown \textbf{LLM Freq of Use} and \textbf{LLM Use Cases}. For Use Cases, we show the \% of these participants who selected each use case (can be multiple so $\sum \neq 1$). Exact question phrasing is reported in the survey codebook (\cref{sec:survey_codebook}).}
\label{tab:llm_usage} \\
\midrule
\endfirsthead
\caption[]{\small \textbf{Survey of Participants' LLM Usage}: The majority of participants have used LLMs directly (via a dedicated chat interface) and indirectly (embedded in products or services). Note only participants who answered \textit{Yes} to \textbf{LLM Direct Use} or \textbf{LLM Indirect Use} ($n=1253$, 84\%) are shown \textbf{LLM Freq of Use} and \textbf{LLM Use Cases}. For Use Cases, we show the \% of these participants who selected each use case (can be multiple so $\sum \neq 1$). Exact question phrasing is reported in the survey codebook (\cref{sec:survey_codebook}).} \\
\midrule
\endhead
\midrule
\multicolumn{3}{r}{Continued on next page} \\
\midrule
\endfoot
\bottomrule
\endlastfoot
\midrule
{\cellcolor{white}} \textbf{LLM Direct Use} & {\cellcolor{white}}  & {\cellcolor{white}}  \\
\midrule
{\cellcolor{white}} Yes & {\cellcolor{white}} 1,162 & {\cellcolor{white}} \medpcb{77.5}\% \\
{\cellcolor[HTML]{F2F2F2}} No & {\cellcolor[HTML]{F2F2F2}} 259 & {\cellcolor[HTML]{F2F2F2}} \medpcb{17.3}\% \\
{\cellcolor{white}} Unsure & {\cellcolor{white}} 79 & {\cellcolor{white}} \medpcb{5.3}\% \\
\midrule
{\cellcolor{white}} \textbf{LLM Indirect Use} & {\cellcolor{white}}  & {\cellcolor{white}}  \\
\midrule
{\cellcolor{white}} Yes & {\cellcolor{white}} 1,104 & {\cellcolor{white}} \medpcb{73.6}\% \\
{\cellcolor[HTML]{F2F2F2}} No & {\cellcolor[HTML]{F2F2F2}} 215 & {\cellcolor[HTML]{F2F2F2}} \medpcb{14.3}\% \\
{\cellcolor{white}} Unsure & {\cellcolor{white}} 181 & {\cellcolor{white}} \medpcb{12.1}\% \\
\midrule
{\cellcolor{white}} \textbf{LLM Familiarity} & {\cellcolor{white}}  & {\cellcolor{white}}  \\
\midrule
{\cellcolor{white}} Somewhat familiar & {\cellcolor{white}} 920 & {\cellcolor{white}} \medpcb{61.3}\% \\
{\cellcolor[HTML]{F2F2F2}} Very familiar & {\cellcolor[HTML]{F2F2F2}} 424 & {\cellcolor[HTML]{F2F2F2}} \medpcb{28.3}\% \\
{\cellcolor{white}} Not familiar at all & {\cellcolor{white}} 156 & {\cellcolor{white}} \medpcb{10.4}\% \\
\midrule
{\cellcolor{white}} \textbf{LLM Frequency of Use} & {\cellcolor{white}}  & {\cellcolor{white}}  \\
\midrule
{\cellcolor{white}} Once per month & {\cellcolor{white}} 374 & {\cellcolor{white}} \medpcb{24.9}\% \\
{\cellcolor[HTML]{F2F2F2}} Every week & {\cellcolor[HTML]{F2F2F2}} 316 & {\cellcolor[HTML]{F2F2F2}} \medpcb{21.1}\% \\
{\cellcolor{white}} More than once a month & {\cellcolor{white}} 291 & {\cellcolor{white}} \medpcb{19.4}\% \\
{\cellcolor[HTML]{F2F2F2}} Less than one a year & {\cellcolor[HTML]{F2F2F2}} 162 & {\cellcolor[HTML]{F2F2F2}} \medpcb{10.8}\% \\
{\cellcolor{white}} Every day & {\cellcolor{white}} 110 & {\cellcolor{white}} \medpcb{7.3}\% \\
{\cellcolor[HTML]{F2F2F2}} Not shown question & {\cellcolor[HTML]{F2F2F2}} 247 & {\cellcolor[HTML]{F2F2F2}} \medpcb{16.5}\% \\
\midrule
{\cellcolor{white}} \textbf{LLM Use Cases} & {\cellcolor{white}}  & {\cellcolor{white}}  \\
\midrule
{\cellcolor{white}} \textbf{Research}: Fact-checking or gaining overviews on specific topics. & {\cellcolor{white}} 617 & {\cellcolor{white}} \medpcb{49.2}\% \\
{\cellcolor[HTML]{F2F2F2}} \textbf{Professional Work}: Assisting in drafting, editing, or brainstorming content for work. & {\cellcolor[HTML]{F2F2F2}} 469 & {\cellcolor[HTML]{F2F2F2}} \medpcb{37.4}\% \\
{\cellcolor{white}} \textbf{Creative Writing}: Generating story ideas, dialogues, poems or other writing prompts. & {\cellcolor{white}} 392 & {\cellcolor{white}} \medpcb{31.3}\% \\
{\cellcolor[HTML]{F2F2F2}} \textbf{Technical or Programming Help}: Seeking programming guidance, code generation, software recommendations, or debugging assistance. & {\cellcolor[HTML]{F2F2F2}} 337 & {\cellcolor[HTML]{F2F2F2}} \medpcb{26.9}\% \\
{\cellcolor{white}} \textbf{Lifestyle and Hobbies}: Looking for recipes, craft ideas, home decoration tips, or hobby-related information. & {\cellcolor{white}} 310 & {\cellcolor{white}} \medpcb{24.7}\% \\
{\cellcolor[HTML]{F2F2F2}} \textbf{Homework Assistance}: Getting help with school or university assignments. & {\cellcolor[HTML]{F2F2F2}} 286 & {\cellcolor[HTML]{F2F2F2}} \medpcb{22.8}\% \\
{\cellcolor{white}} \textbf{Personal Recommendations}: Seeking book, music or movie recommendations. & {\cellcolor{white}} 266 & {\cellcolor{white}} \medpcb{21.2}\% \\
{\cellcolor[HTML]{F2F2F2}} \textbf{Casual Conversation}: Engaging in small talk, casual chats, or joke generation. & {\cellcolor[HTML]{F2F2F2}} 262 & {\cellcolor[HTML]{F2F2F2}} \medpcb{20.9}\% \\
{\cellcolor{white}} \textbf{Language Learning}: Using it as a tool for language practice or translation. & {\cellcolor{white}} 229 & {\cellcolor{white}} \medpcb{18.3}\% \\
{\cellcolor[HTML]{F2F2F2}} \textbf{Source Suggestions}: Creating or finding bibliographies, information sources or reading lists. & {\cellcolor[HTML]{F2F2F2}} 217 & {\cellcolor[HTML]{F2F2F2}} \medpcb{17.3}\% \\
{\cellcolor{white}} \textbf{Daily Productivity}: Setting reminders, making to-do lists, or productivity tips. & {\cellcolor{white}} 216 & {\cellcolor{white}} \medpcb{17.2}\% \\
{\cellcolor[HTML]{F2F2F2}} \textbf{Historical or News Insight}: Getting summaries or background on historical events or news and current affairs. & {\cellcolor[HTML]{F2F2F2}} 183 & {\cellcolor[HTML]{F2F2F2}} \medpcb{14.6}\% \\
{\cellcolor{white}} \textbf{Well-being Guidance}: Seeking general exercise routines, wellness or meditation tips. & {\cellcolor{white}} 159 & {\cellcolor{white}} \medpcb{12.7}\% \\
{\cellcolor[HTML]{F2F2F2}} \textbf{Games}: Playing text-based games, generating riddles or puzzles. & {\cellcolor[HTML]{F2F2F2}} 143 & {\cellcolor[HTML]{F2F2F2}} \medpcb{11.4}\% \\
{\cellcolor{white}} \textbf{Travel Guidance}: Getting destination recommendations, planning holidays, or cultural etiquette tips. & {\cellcolor{white}} 133 & {\cellcolor{white}} \medpcb{10.6}\% \\
{\cellcolor[HTML]{F2F2F2}} \textbf{Medical Guidance}: Seeking health-related advice or medical guidance. & {\cellcolor[HTML]{F2F2F2}} 130 & {\cellcolor[HTML]{F2F2F2}} \medpcb{10.4}\% \\
{\cellcolor{white}} \textbf{Financial Guidance}: Asking about financial concepts or general investing ideas. & {\cellcolor{white}} 107 & {\cellcolor{white}} \medpcb{8.5}\% \\
{\cellcolor[HTML]{F2F2F2}} \textbf{Relationship Advice}: Seeking general self-help or relationship advice for family, friends or partners. & {\cellcolor[HTML]{F2F2F2}} 98 & {\cellcolor[HTML]{F2F2F2}} \medpcb{7.8}\% \\
{\cellcolor{white}} \textbf{Other} & {\cellcolor{white}} 124 & {\cellcolor{white}} \medpcb{9.9}\% \\
\bottomrule
\end{longtable}

\subsection{Other Identified Usecases}
\normalsize
In addition to the usecases in \cref{tab:llm_usage}, 122 participants used the ``Other'' option to add a usecase in their own words. Many of these just add more specific details to the pre-provided categories. In addition, there were a few interesting themes:
\begin{itemize}
\item \textbf{Customer Service:} Many of the participants noted having interacted with LLMs in customer support chats, often with negative sentiment (``Usually forced to interact with chatbots to get something done'', ``Customer service bots I cannot avoid'', ``Insurance companies direct you to chatbots, usually useless'').
\item \textbf{Prolific and Other Online Surveys:} One of the more common (and potentially concerning) answers mentioned research  participation e.g. ``Studies like this one'', ``Doing Prolific tests'', but it may be that they mean AI is the subject of the study: ``AI research subject on research platforms Prolific, others.'' or ``It's sometimes required as part of a survey on Prolific.'' We encourage future work on whether there is noticeable difference in these participants' answers elsewhere in our task.
\item \textbf{AI Understanding or Testing:} A few participants mentioned ``Trying to gain an understanding into AI and its capabilities'' or ``Gauging progress/viability of AI models''. Many others indicated curiosity or exploratory use e.g. ``Just to test it out and see what it's all about'' or ``Casual interest in the new technology''.
\item \textbf{Professional or Job Tasks:} Participants added details on professional usecases like resume help, interview prep, CV writing, HR-tasks, Excel help, or emails.
\item \textbf{Creative (Multimodal) Use-cases:} Participants gave additional detail like writing YouTube scripts, generating gift card text or designing characters for games as well as multimodal creative outputs like generating drawings or images.
\item \textbf{Domain-Specific Usecases:} Medical, Financial and Educational usecases are all mentioned.
\end{itemize}

\clearpage
\section{Screening and Recruitment Process}
\label{sec:appendix_screening}
We recruit workers via Prolific (\url{https://www.prolific.com/}). We apply two initial screening criteria: (i) participants must be fluent in English because \ourdata targets monolingual models and language data, and (ii) participants must have been born and reside in the same country to avoid biasing our sample towards expats living abroad. There is a skewed country-wise distribution of active workers who meet this criteria (see \cref{tab:studies}). For example, of the 21,084 workers in Europe (passing screening), 17\% are Portuguese, 15\% German and 14\% Polish; and all 6,584 workers in Africa are located in South Africa. To account for this, we set up country-specific studies in each country with at least one eligible worker, balance study spots across regions, and ensure no single country has more than 100 open spots (apart from the Rep Samples in the UK and US). We collected information on country of birth and country of current residence during our survey (separate to workers' stored Prolific details), and find that 179 participant (12\%) have different birth and reside countries. We do not exclude these individuals from our sample.

\fontsize{9pt}{9pt}\selectfont
{\setlength{\tabcolsep}{1pt}
\begin{longtable}{lcccrlrlrlr}
\caption{\small \textbf{Summary of Recruitment Studies } We present study-wise breakdowns ($n=33$). Each study was created based on the constraints of Prolific's pool of workers. We show here \textit{all} the countries with at least 1 fluent English speaker, and the counts for fluent English Speakers who were born and currently reside in that same country. We show the whether each study was screened for a special representative sample (\textbf{Rep Sample}) or if it was balanced on participant gender (\textbf{Gender Bal}). In some cases, there were too few active participants per country to balance by gender without comprising participant privacy. We also show when the first batch was launched (all dates are in 2023) and approximate cost (at \textsterling 9 per hour per participant). }
\label{tab:studies} \\
\toprule
& \multicolumn{1}{c}{\textbf{Rep}} & \multicolumn{1}{c}{\textbf{Gender}}  & \multicolumn{1}{c}{\textbf{Launched}} & \multicolumn{2}{c}{\textbf{Approved}} & \multicolumn{4}{c}{\textbf{\underline{Prolific Fluent English Speakers}}}  & \multicolumn{1}{c}{\textbf{Cost}} \\ \multicolumn{1}{l}{\textbf{Study}} & \multicolumn{1}{c}{\textbf{Sample}} & \multicolumn{1}{c}{\textbf{Bal}}  & \multicolumn{1}{c}{\textbf{(2023)}} & \multicolumn{2}{c}{\textbf{Submissions}} & \multicolumn{2}{c}{\textbf{(All)}} & \multicolumn{2}{c}{\textbf{(Born=Reside)}} & \multicolumn{1}{c}{\textbf{(\textsterling)}}\\
\midrule
\endfirsthead
\caption[]{ADD CAPTION HERE} \\
\toprule
& \multicolumn{1}{c}{\textbf{Rep}} & \multicolumn{1}{c}{\textbf{Gender}}  & \multicolumn{1}{c}{\textbf{Launched}} & \multicolumn{2}{c}{\textbf{Approved}} & \multicolumn{4}{c}{\textbf{\underline{Prolific Fluent English Speakers}}}  & \multicolumn{1}{c}{\textbf{Cost}} \\ \multicolumn{1}{l}{\textbf{Study}} & \multicolumn{1}{c}{\textbf{Sample}} & \multicolumn{1}{c}{\textbf{Bal}}  & \multicolumn{1}{c}{\textbf{(2023)}} & \multicolumn{2}{c}{\textbf{Submissions}} & \multicolumn{2}{c}{\textbf{(All)}} & \multicolumn{2}{l}{\textbf{(Born=Reside)}} & \multicolumn{1}{c}{\textbf{(£)}}\\
\midrule
\endhead
\midrule
\multicolumn{11}{r}{Continued on next page} \\
\midrule
\endfoot
\bottomrule
\endlastfoot
\midrule
{\cellcolor{white}} \textbf{Total} & {\cellcolor{white}} \textbf{2} & {\cellcolor{white}} \textbf{25} & {\cellcolor{white}} - & {\cellcolor{white}} \textbf{1,500} & {\cellcolor{white}} & {\cellcolor{white}} \textbf{111,572} & {\cellcolor{white}}& {\cellcolor{white}} \textbf{100,585} & {\cellcolor{white}}  & {\cellcolor{white}} \textbf{14,850.00} \\
\midrule
{\cellcolor{white}} US & {\cellcolor{white}} \cmark & {\cellcolor{white}} \xmark & {\cellcolor{white}} 27-11 & {\cellcolor{white}} 386 & {\cellcolor{white}} \smallpcb{25.7}\% & {\cellcolor{white}} 38,114 & {\cellcolor{white}} \smallpcb{34.2}\% & {\cellcolor{white}} 36,205 & {\cellcolor{white}} \smallpcb{36.0}\% & {\cellcolor{white}} 3,821.40 \\
{\cellcolor[HTML]{F2F2F2}} UK & {\cellcolor[HTML]{F2F2F2}} \cmark & {\cellcolor[HTML]{F2F2F2}} \xmark & {\cellcolor[HTML]{F2F2F2}} 27-11 & {\cellcolor[HTML]{F2F2F2}} 341 & {\cellcolor[HTML]{F2F2F2}} \smallpcb{22.7}\% & {\cellcolor[HTML]{F2F2F2}} 37,408 & {\cellcolor[HTML]{F2F2F2}} \smallpcb{33.5}\% & {\cellcolor[HTML]{F2F2F2}} 33,678 & {\cellcolor[HTML]{F2F2F2}} \smallpcb{33.5}\% & {\cellcolor[HTML]{F2F2F2}} 3,375.90 \\
{\cellcolor{white}} South Africa & {\cellcolor{white}} \xmark & {\cellcolor{white}} \cmark & {\cellcolor{white}} 22-11 & {\cellcolor{white}} 88 & {\cellcolor{white}} \smallpcb{5.9}\% & {\cellcolor{white}} 7,061 & {\cellcolor{white}} \smallpcb{6.3}\% & {\cellcolor{white}} 6,584 & {\cellcolor{white}} \smallpcb{6.5}\% & {\cellcolor{white}} 871.20 \\
{\cellcolor[HTML]{F2F2F2}} New Zealand & {\cellcolor[HTML]{F2F2F2}} \xmark & {\cellcolor[HTML]{F2F2F2}} \cmark & {\cellcolor[HTML]{F2F2F2}} 24-11 & {\cellcolor[HTML]{F2F2F2}} 77 & {\cellcolor[HTML]{F2F2F2}} \smallpcb{5.1}\% & {\cellcolor[HTML]{F2F2F2}} 511 & {\cellcolor[HTML]{F2F2F2}} \smallpcb{0.5}\% & {\cellcolor[HTML]{F2F2F2}} 389 & {\cellcolor[HTML]{F2F2F2}} \smallpcb{0.4}\% & {\cellcolor[HTML]{F2F2F2}} 762.30 \\
{\cellcolor{white}} Australia & {\cellcolor{white}} \xmark & {\cellcolor{white}} \cmark & {\cellcolor{white}} 24-11 & {\cellcolor{white}} 71 & {\cellcolor{white}} \smallpcb{4.7}\% & {\cellcolor{white}} 1,968 & {\cellcolor{white}} \smallpcb{1.8}\% & {\cellcolor{white}} 1,550 & {\cellcolor{white}} \smallpcb{1.5}\% & {\cellcolor{white}} 702.90 \\
{\cellcolor[HTML]{F2F2F2}} Mexico & {\cellcolor[HTML]{F2F2F2}} \xmark & {\cellcolor[HTML]{F2F2F2}} \cmark & {\cellcolor[HTML]{F2F2F2}} 24-11 & {\cellcolor[HTML]{F2F2F2}} 69 & {\cellcolor[HTML]{F2F2F2}} \smallpcb{4.6}\% & {\cellcolor[HTML]{F2F2F2}} 2,021 & {\cellcolor[HTML]{F2F2F2}} \smallpcb{1.8}\% & {\cellcolor[HTML]{F2F2F2}} 1,943 & {\cellcolor[HTML]{F2F2F2}} \smallpcb{1.9}\% & {\cellcolor[HTML]{F2F2F2}} 683.10 \\
{\cellcolor{white}} Chile & {\cellcolor{white}} \xmark & {\cellcolor{white}} \cmark & {\cellcolor{white}} 23-11 & {\cellcolor{white}} 65 & {\cellcolor{white}} \smallpcb{4.3}\% & {\cellcolor{white}} 455 & {\cellcolor{white}} \smallpcb{0.4}\% & {\cellcolor{white}} 416 & {\cellcolor{white}} \smallpcb{0.4}\% & {\cellcolor{white}} 643.50 \\
{\cellcolor[HTML]{F2F2F2}} Israel & {\cellcolor[HTML]{F2F2F2}} \xmark & {\cellcolor[HTML]{F2F2F2}} \xmark & {\cellcolor[HTML]{F2F2F2}} 25-11 & {\cellcolor[HTML]{F2F2F2}} 61 & {\cellcolor[HTML]{F2F2F2}} \smallpcb{4.1}\% & {\cellcolor[HTML]{F2F2F2}} 310 & {\cellcolor[HTML]{F2F2F2}} \smallpcb{0.3}\% & {\cellcolor[HTML]{F2F2F2}} 272 & {\cellcolor[HTML]{F2F2F2}} \smallpcb{0.3}\% & {\cellcolor[HTML]{F2F2F2}} 603.90 \\
{\cellcolor{white}} Canada & {\cellcolor{white}} \xmark & {\cellcolor{white}} \cmark & {\cellcolor{white}} 22-11 & {\cellcolor{white}} 54 & {\cellcolor{white}} \smallpcb{3.6}\% & {\cellcolor{white}} 3,687 & {\cellcolor{white}} \smallpcb{3.3}\% & {\cellcolor{white}} 3,031 & {\cellcolor{white}} \smallpcb{3.0}\% & {\cellcolor{white}} 534.60 \\
{\cellcolor[HTML]{F2F2F2}} Asia & {\cellcolor[HTML]{F2F2F2}} \xmark & {\cellcolor[HTML]{F2F2F2}} \xmark & {\cellcolor[HTML]{F2F2F2}} 24-11 & {\cellcolor[HTML]{F2F2F2}} 18 & {\cellcolor[HTML]{F2F2F2}} \smallpcb{1.2}\% & {\cellcolor[HTML]{F2F2F2}} 196 & {\cellcolor[HTML]{F2F2F2}} \smallpcb{0.2}\% & {\cellcolor[HTML]{F2F2F2}} 32 & {\cellcolor[HTML]{F2F2F2}} \smallpcb{0.0}\% & {\cellcolor[HTML]{F2F2F2}} 178.20 \\
{\cellcolor{white}} Spain & {\cellcolor{white}} \xmark & {\cellcolor{white}} \cmark & {\cellcolor{white}} 23-11 & {\cellcolor{white}} 18 & {\cellcolor{white}} \smallpcb{1.2}\% & {\cellcolor{white}} 1,252 & {\cellcolor{white}} \smallpcb{1.1}\% & {\cellcolor{white}} 942 & {\cellcolor{white}} \smallpcb{0.9}\% & {\cellcolor{white}} 178.20 \\
{\cellcolor[HTML]{F2F2F2}} Belgium & {\cellcolor[HTML]{F2F2F2}} \xmark & {\cellcolor[HTML]{F2F2F2}} \cmark & {\cellcolor[HTML]{F2F2F2}} 23-11 & {\cellcolor[HTML]{F2F2F2}} 17 & {\cellcolor[HTML]{F2F2F2}} \smallpcb{1.1}\% & {\cellcolor[HTML]{F2F2F2}} 376 & {\cellcolor[HTML]{F2F2F2}} \smallpcb{0.3}\% & {\cellcolor[HTML]{F2F2F2}} 281 & {\cellcolor[HTML]{F2F2F2}} \smallpcb{0.3}\% & {\cellcolor[HTML]{F2F2F2}} 168.30 \\
{\cellcolor{white}} Hungary & {\cellcolor{white}} \xmark & {\cellcolor{white}} \cmark & {\cellcolor{white}} 24-11 & {\cellcolor{white}} 16 & {\cellcolor{white}} \smallpcb{1.1}\% & {\cellcolor{white}} 537 & {\cellcolor{white}} \smallpcb{0.5}\% & {\cellcolor{white}} 456 & {\cellcolor{white}} \smallpcb{0.5}\% & {\cellcolor{white}} 158.40 \\
{\cellcolor[HTML]{F2F2F2}} Ireland & {\cellcolor[HTML]{F2F2F2}} \xmark & {\cellcolor[HTML]{F2F2F2}} \cmark & {\cellcolor[HTML]{F2F2F2}} 23-11 & {\cellcolor[HTML]{F2F2F2}} 15 & {\cellcolor[HTML]{F2F2F2}} \smallpcb{1.0}\% & {\cellcolor[HTML]{F2F2F2}} 640 & {\cellcolor[HTML]{F2F2F2}} \smallpcb{0.6}\% & {\cellcolor[HTML]{F2F2F2}} 502 & {\cellcolor[HTML]{F2F2F2}} \smallpcb{0.5}\% & {\cellcolor[HTML]{F2F2F2}} 148.50 \\
{\cellcolor{white}} Denmark & {\cellcolor{white}} \xmark & {\cellcolor{white}} \xmark & {\cellcolor{white}} 23-11 & {\cellcolor{white}} 15 & {\cellcolor{white}} \smallpcb{1.0}\% & {\cellcolor{white}} 119 & {\cellcolor{white}} \smallpcb{0.1}\% & {\cellcolor{white}} 65 & {\cellcolor{white}} \smallpcb{0.1}\% & {\cellcolor{white}} 148.50 \\
{\cellcolor[HTML]{F2F2F2}} Norway & {\cellcolor[HTML]{F2F2F2}} \xmark & {\cellcolor[HTML]{F2F2F2}} \xmark & {\cellcolor[HTML]{F2F2F2}} 23-11 & {\cellcolor[HTML]{F2F2F2}} 15 & {\cellcolor[HTML]{F2F2F2}} \smallpcb{1.0}\% & {\cellcolor[HTML]{F2F2F2}} 91 & {\cellcolor[HTML]{F2F2F2}} \smallpcb{0.1}\% & {\cellcolor[HTML]{F2F2F2}} 59 & {\cellcolor[HTML]{F2F2F2}} \smallpcb{0.1}\% & {\cellcolor[HTML]{F2F2F2}} 148.50 \\
{\cellcolor{white}} Switzerland & {\cellcolor{white}} \xmark & {\cellcolor{white}} \cmark & {\cellcolor{white}} 23-11 & {\cellcolor{white}} 14 & {\cellcolor{white}} \smallpcb{0.9}\% & {\cellcolor{white}} 205 & {\cellcolor{white}} \smallpcb{0.2}\% & {\cellcolor{white}} 104 & {\cellcolor{white}} \smallpcb{0.1}\% & {\cellcolor{white}} 138.60 \\
{\cellcolor[HTML]{F2F2F2}} Poland & {\cellcolor[HTML]{F2F2F2}} \xmark & {\cellcolor[HTML]{F2F2F2}} \cmark & {\cellcolor[HTML]{F2F2F2}} 23-11 & {\cellcolor[HTML]{F2F2F2}} 14 & {\cellcolor[HTML]{F2F2F2}} \smallpcb{0.9}\% & {\cellcolor[HTML]{F2F2F2}} 2,975 & {\cellcolor[HTML]{F2F2F2}} \smallpcb{2.7}\% & {\cellcolor[HTML]{F2F2F2}} 2,850 & {\cellcolor[HTML]{F2F2F2}} \smallpcb{2.8}\% & {\cellcolor[HTML]{F2F2F2}} 138.60 \\
{\cellcolor{white}} Czech Republic & {\cellcolor{white}} \xmark & {\cellcolor{white}} \cmark & {\cellcolor{white}} 24-11 & {\cellcolor{white}} 14 & {\cellcolor{white}} \smallpcb{0.9}\% & {\cellcolor{white}} 238 & {\cellcolor{white}} \smallpcb{0.2}\% & {\cellcolor{white}} 229 & {\cellcolor{white}} \smallpcb{0.2}\% & {\cellcolor{white}} 138.60 \\
{\cellcolor[HTML]{F2F2F2}} Latvia & {\cellcolor[HTML]{F2F2F2}} \xmark & {\cellcolor[HTML]{F2F2F2}} \cmark & {\cellcolor[HTML]{F2F2F2}} 23-11 & {\cellcolor[HTML]{F2F2F2}} 14 & {\cellcolor[HTML]{F2F2F2}} \smallpcb{0.9}\% & {\cellcolor[HTML]{F2F2F2}} 173 & {\cellcolor[HTML]{F2F2F2}} \smallpcb{0.2}\% & {\cellcolor[HTML]{F2F2F2}} 162 & {\cellcolor[HTML]{F2F2F2}} \smallpcb{0.2}\% & {\cellcolor[HTML]{F2F2F2}} 138.60 \\
{\cellcolor{white}} Greece & {\cellcolor{white}} \xmark & {\cellcolor{white}} \cmark & {\cellcolor{white}} 24-11 & {\cellcolor{white}} 14 & {\cellcolor{white}} \smallpcb{0.9}\% & {\cellcolor{white}} 809 & {\cellcolor{white}} \smallpcb{0.7}\% & {\cellcolor{white}} 747 & {\cellcolor{white}} \smallpcb{0.7}\% & {\cellcolor{white}} 138.60 \\
{\cellcolor[HTML]{F2F2F2}} Finland & {\cellcolor[HTML]{F2F2F2}} \xmark & {\cellcolor[HTML]{F2F2F2}} \cmark & {\cellcolor[HTML]{F2F2F2}} 23-11 & {\cellcolor[HTML]{F2F2F2}} 13 & {\cellcolor[HTML]{F2F2F2}} \smallpcb{0.9}\% & {\cellcolor[HTML]{F2F2F2}} 152 & {\cellcolor[HTML]{F2F2F2}} \smallpcb{0.1}\% & {\cellcolor[HTML]{F2F2F2}} 117 & {\cellcolor[HTML]{F2F2F2}} \smallpcb{0.1}\% & {\cellcolor[HTML]{F2F2F2}} 128.70 \\
{\cellcolor{white}} Germany & {\cellcolor{white}} \xmark & {\cellcolor{white}} \cmark & {\cellcolor{white}} 24-11 & {\cellcolor{white}} 13 & {\cellcolor{white}} \smallpcb{0.9}\% & {\cellcolor{white}} 3,152 & {\cellcolor{white}} \smallpcb{2.8}\% & {\cellcolor{white}} 2,295 & {\cellcolor{white}} \smallpcb{2.3}\% & {\cellcolor{white}} 128.70 \\
{\cellcolor[HTML]{F2F2F2}} Italy & {\cellcolor[HTML]{F2F2F2}} \xmark & {\cellcolor[HTML]{F2F2F2}} \cmark & {\cellcolor[HTML]{F2F2F2}} 24-11 & {\cellcolor[HTML]{F2F2F2}} 12 & {\cellcolor[HTML]{F2F2F2}} \smallpcb{0.8}\% & {\cellcolor[HTML]{F2F2F2}} 2,037 & {\cellcolor[HTML]{F2F2F2}} \smallpcb{1.8}\% & {\cellcolor[HTML]{F2F2F2}} 1,857 & {\cellcolor[HTML]{F2F2F2}} \smallpcb{1.8}\% & {\cellcolor[HTML]{F2F2F2}} 118.80 \\
{\cellcolor{white}} France & {\cellcolor{white}} \xmark & {\cellcolor{white}} \cmark & {\cellcolor{white}} 24-11 & {\cellcolor{white}} 12 & {\cellcolor{white}} \smallpcb{0.8}\% & {\cellcolor{white}} 957 & {\cellcolor{white}} \smallpcb{0.9}\% & {\cellcolor{white}} 681 & {\cellcolor{white}} \smallpcb{0.7}\% & {\cellcolor{white}} 118.80 \\
{\cellcolor[HTML]{F2F2F2}} Slovenia & {\cellcolor[HTML]{F2F2F2}} \xmark & {\cellcolor[HTML]{F2F2F2}} \cmark & {\cellcolor[HTML]{F2F2F2}} 24-11 & {\cellcolor[HTML]{F2F2F2}} 10 & {\cellcolor[HTML]{F2F2F2}} \smallpcb{0.7}\% & {\cellcolor[HTML]{F2F2F2}} 232 & {\cellcolor[HTML]{F2F2F2}} \smallpcb{0.2}\% & {\cellcolor[HTML]{F2F2F2}} 220 & {\cellcolor[HTML]{F2F2F2}} \smallpcb{0.2}\% & {\cellcolor[HTML]{F2F2F2}} 99.00 \\
{\cellcolor{white}} Austria & {\cellcolor{white}} \xmark & {\cellcolor{white}} \cmark & {\cellcolor{white}} 24-11 & {\cellcolor{white}} 10 & {\cellcolor{white}} \smallpcb{0.7}\% & {\cellcolor{white}} 231 & {\cellcolor{white}} \smallpcb{0.2}\% & {\cellcolor{white}} 156 & {\cellcolor{white}} \smallpcb{0.2}\% & {\cellcolor{white}} 99.00 \\
{\cellcolor[HTML]{F2F2F2}} Estonia & {\cellcolor[HTML]{F2F2F2}} \xmark & {\cellcolor[HTML]{F2F2F2}} \cmark & {\cellcolor[HTML]{F2F2F2}} 24-11 & {\cellcolor[HTML]{F2F2F2}} 10 & {\cellcolor[HTML]{F2F2F2}} \smallpcb{0.7}\% & {\cellcolor[HTML]{F2F2F2}} 251 & {\cellcolor[HTML]{F2F2F2}} \smallpcb{0.2}\% & {\cellcolor[HTML]{F2F2F2}} 237 & {\cellcolor[HTML]{F2F2F2}} \smallpcb{0.2}\% & {\cellcolor[HTML]{F2F2F2}} 99.00 \\
{\cellcolor{white}} Netherlands & {\cellcolor{white}} \xmark & {\cellcolor{white}} \cmark & {\cellcolor{white}} 24-11 & {\cellcolor{white}} 8 & {\cellcolor{white}} \smallpcb{0.5}\% & {\cellcolor{white}} 1,460 & {\cellcolor{white}} \smallpcb{1.3}\% & {\cellcolor{white}} 1,028 & {\cellcolor{white}} \smallpcb{1.0}\% & {\cellcolor{white}} 79.20 \\
{\cellcolor[HTML]{F2F2F2}} Portugal & {\cellcolor[HTML]{F2F2F2}} \xmark & {\cellcolor[HTML]{F2F2F2}} \cmark & {\cellcolor[HTML]{F2F2F2}} 24-11 & {\cellcolor[HTML]{F2F2F2}} 7 & {\cellcolor[HTML]{F2F2F2}} \smallpcb{0.5}\% & {\cellcolor[HTML]{F2F2F2}} 3,649 & {\cellcolor[HTML]{F2F2F2}} \smallpcb{3.3}\% & {\cellcolor[HTML]{F2F2F2}} 3,284 & {\cellcolor[HTML]{F2F2F2}} \smallpcb{3.3}\% & {\cellcolor[HTML]{F2F2F2}} 69.30 \\
{\cellcolor{white}} Sweden & {\cellcolor{white}} \xmark & {\cellcolor{white}} \cmark & {\cellcolor{white}} 24-11 & {\cellcolor{white}} 6 & {\cellcolor{white}} \smallpcb{0.4}\% & {\cellcolor{white}} 274 & {\cellcolor{white}} \smallpcb{0.2}\% & {\cellcolor{white}} 196 & {\cellcolor{white}} \smallpcb{0.2}\% & {\cellcolor{white}} 59.40 \\
{\cellcolor[HTML]{F2F2F2}} Luxembourg & {\cellcolor[HTML]{F2F2F2}} \xmark & {\cellcolor[HTML]{F2F2F2}} \xmark & {\cellcolor[HTML]{F2F2F2}} 23-11 & {\cellcolor[HTML]{F2F2F2}} 2 & {\cellcolor[HTML]{F2F2F2}} \smallpcb{0.1}\% & {\cellcolor[HTML]{F2F2F2}} 15 & {\cellcolor[HTML]{F2F2F2}} \smallpcb{0.0}\% & {\cellcolor[HTML]{F2F2F2}} 6 & {\cellcolor[HTML]{F2F2F2}} \smallpcb{0.0}\% & {\cellcolor[HTML]{F2F2F2}} 19.80 \\
{\cellcolor{white}} Iceland & {\cellcolor{white}} \xmark & {\cellcolor{white}} \xmark & {\cellcolor{white}} 23-11 & {\cellcolor{white}} 1 & {\cellcolor{white}} \smallpcb{0.1}\% & {\cellcolor{white}} 16 & {\cellcolor{white}} \smallpcb{0.0}\% & {\cellcolor{white}} 11 & {\cellcolor{white}} \smallpcb{0.0}\% & {\cellcolor{white}} 9.90 \\
\end{longtable}
}

\cleardoublepage
\section{Conversation Type Rebalancing}
\label{sec:appendix_convo_rebalancing}
\normalsize
Our task instructions specified that participants should complete six conversations in total, two of each type. In reality, some participants deviated from this quota. This could be due to (i) misunderstanding of instructions, (ii) technical issues, or (iii) losing count, as while we included a counter of the total number of conversations on the interface (see \cref{sec:appendix_interface_and_task}), we did not include per conversation type breakdowns. To mitigate variation on conversation type selection, we create a balanced subset of \ourdata. First, we filter to all participants who had at least one of each conversation type. Then we take the maximum number of total conversations (either $n=3$ or $n=6$) so that there are equal numbers of each type. This results in $6,669$ conversations (84\% of all conversations), from $1246$ participants (83\% of all participants). We release this flag \texttt{included\_in\_balanced\_subset} if future researchers want to use the same set of conversations. We make sure this flag intersects with the census rebalancing flags (see \cref{sec:appendix_census}) so no further data is lost when both subsets are needed.

\section{Census Rebalancing}
\label{sec:appendix_census}
\normalsize
\paragraph{Obstacles to representativeness} We use the representative sample offered from Prolific \cite{prolificRepresentative2024}. However, there are several reasons why these samples may not be fully representative. First, our sampling process was affected internally due to cyberattacks disrupting some participants' workflows. These participants returned to the task after their spots had `timed-out', and were re-filled by other same demographic individuals. Second, Prolific provides a sample breakdown in-line with a \textit{simplified} census but do not match \textit{intersectional} proportions to census data. Third, if a sample spot is taking too long to fill (e.g. 65+ years), Prolific will reallocate these spots to different demographics. There are of course wider stumbling blocks from crowdworkers skewing towards younger, more educated, and digitally-active populations. We original set up 300 spots for each of the representative samples, but ended up with 386 approved participants in the UK sample (UK-REP), and 341 in the US (US-REP).\footnote{There are more than the initial 300 spots due to participants returning to our interface to finish their conversations after their place had `timed-out' and been refilled. We still paid and included these participants.}

\begin{figure*}[htp]
    \centering
    \begin{subfigure}[b]{0.925\textwidth}
        \centering
        \includegraphics[width=\textwidth]{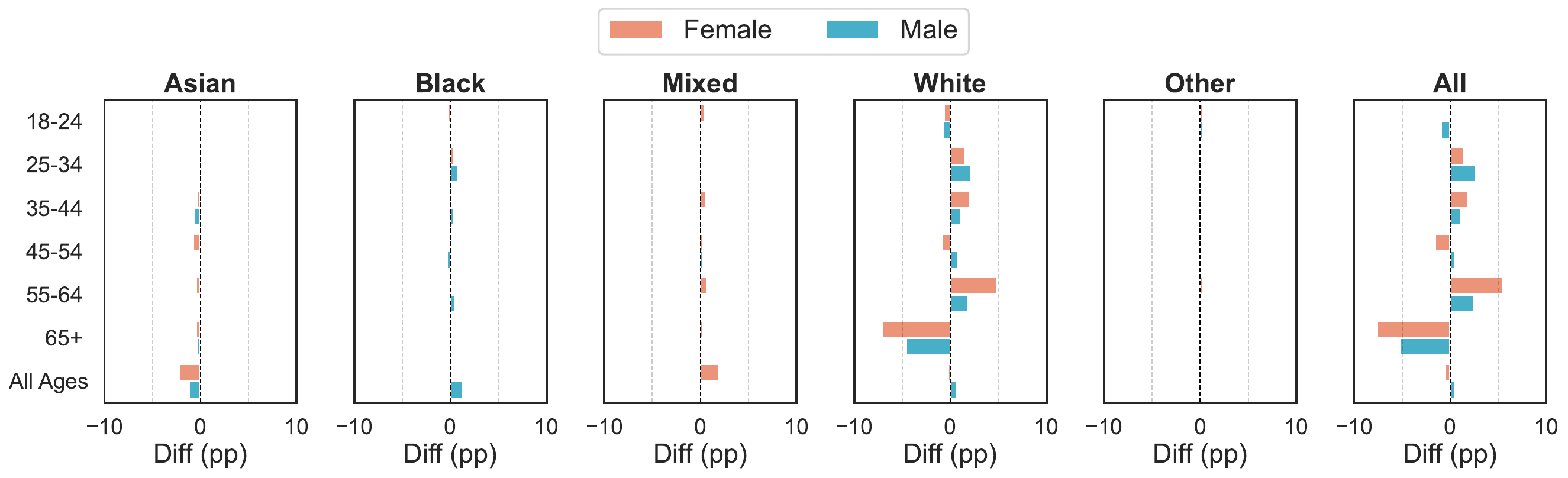}
        \caption{\textbf{UK (Before Rebalancing)}: There are 282 eligible$^*$ participants in the UK sample.}
        \label{fig:census_sub1}
    \end{subfigure}

    \vspace{0cm} %

    \begin{subfigure}[b]{0.925\textwidth}
        \centering
        \includegraphics[width=\textwidth]{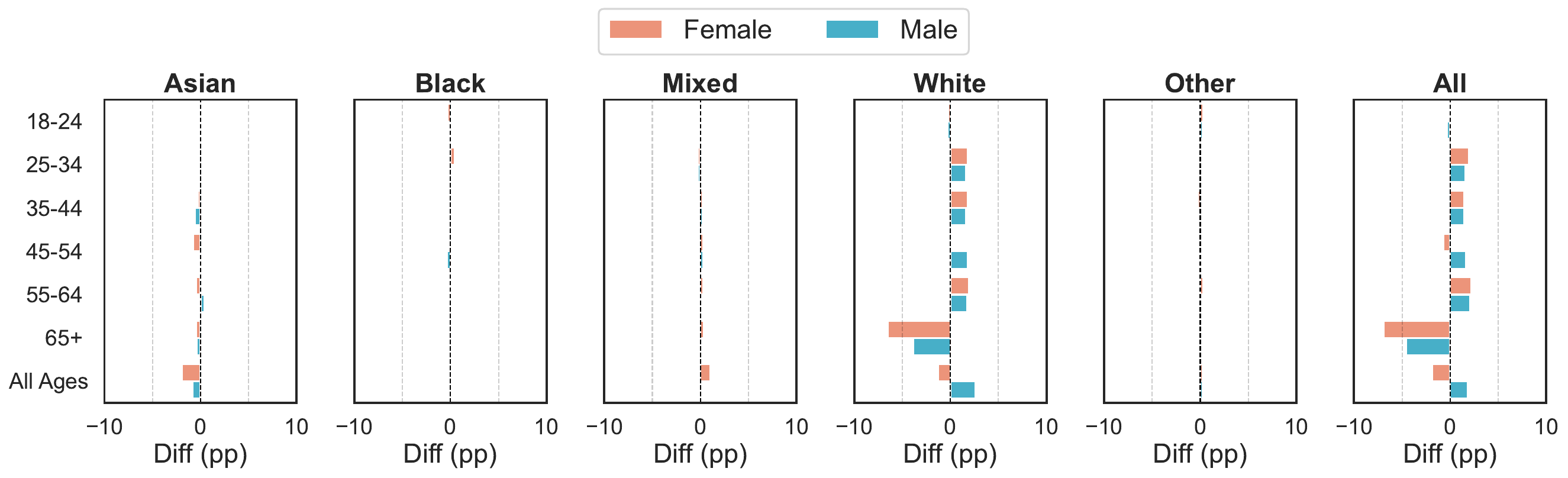}
        \caption{\textbf{UK (After Rebalancing)}: There are 243 participants in the rebalanced UK sample.}
        \label{fig:census_sub2}
    \end{subfigure}

    \vspace{0cm} %

    \begin{subfigure}[b]{0.925\textwidth}
        \centering
        \includegraphics[width=\textwidth]{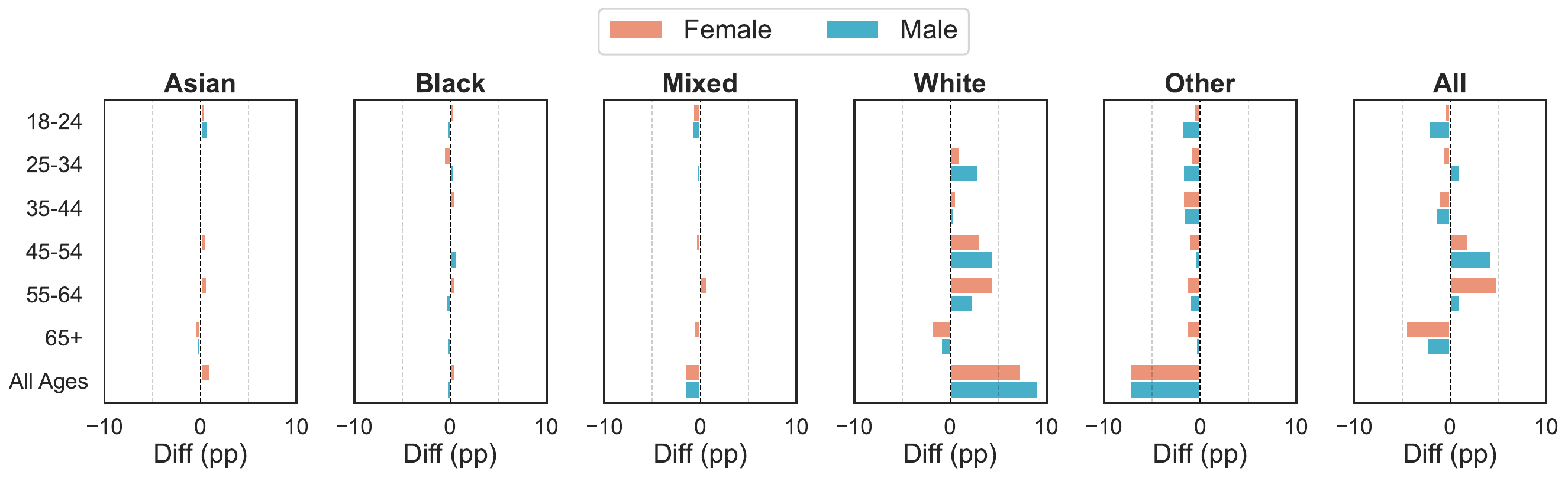}
        \caption{\textbf{US (Before Rebalancing)}: There are 297 eligible$^*$ participants in the US sample.}
        \label{fig:census_sub3}
    \end{subfigure}

    \vspace{0cm} %
    \begin{subfigure}[b]{0.925\textwidth}
        \centering
        \includegraphics[width=\textwidth]{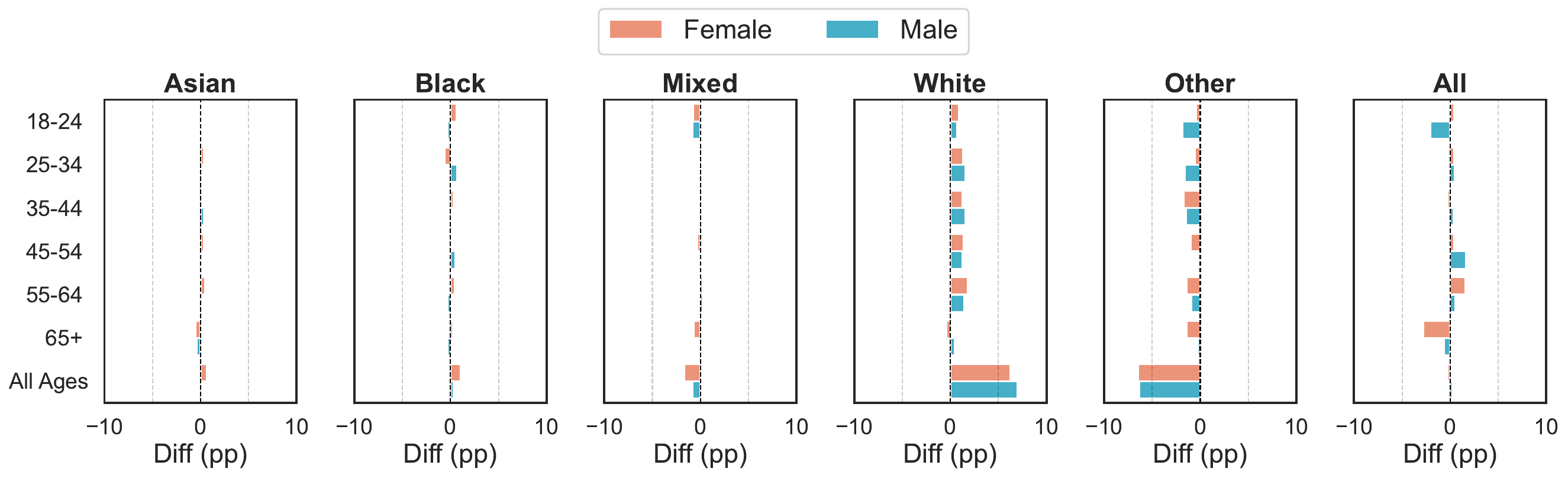}
        \caption{\textbf{US (After Rebalancing)}: There are 230 participants in the rebalanced US sample.}
        \label{fig:census_sub4}
    \end{subfigure}
    \caption{\small \textbf{Before and after census-rebalancing.} We show the difference in observed and expected proportions (\ourdata \textit{minus} Census). Bars to the \textit{right} of the centre line are groups \textit{over-represented} in \ourdata relative to the census. The UK census population has 47,204,870 adults. The US census has 298,477,760 adults. The sample size for before and after rebalancing is reported above. $^*$A participant is \textit{eligible} if they have completed a equal number of conversations for each conversation type (see \cref{sec:appendix_convo_rebalancing}).}
    \label{fig:census_plots}
\end{figure*}

\paragraph{Is our original sample representative?} We compare our sample breakdowns to recent census data.\footnote{For the UK, we examine age, ethnicity and gender from the 2021 data provided by the Office of National Statistics (see \href{https://www.ons.gov.uk/peoplepopulationandcommunity/culturalidentity/ethnicity/articles/ethnicgroupbyageandsexenglandandwales/census2021}{ons.gov.uk}). For the US, we download and combine each ethnicity-specific table from the 2022 data provided by US Census Bureau (see \href{https://data.census.gov/table?q=B01001A}{data.census.gov.}).} For each of US-REP and UK-REP, we remove participants who did not give demographic details (\textit{Prefer not to say}) and those reporting non-binary gender (which is not accounted for in census data). We subset to individuals also appearing in the balanced conversation subset to mitigate further data loss (see \cref{sec:appendix_convo_rebalancing}). Remaining participants are considered \textit{eligible}: 283 participants for the UK, and 297 for the US. We map \ourdata and census data into shared age, ethnicity and gender buckets. We then cross-tabulate what proportion is expected to appear in each age, gender and ethnicity intersection from the census data, and what percentage of participants we actually observed in our sample.\footnote{For the US, we combine ``Other'' with ``Hispanic'' because over 91\% of the ``Other'' census category are Hispanic individuals. See \href{https://www.census.gov/library/stories/2023/10/2020-census-dhc-a-some-other-race-population}{census.gov/library/stories/2023/10/2020-census-dhc-a-some-other-race-population}.} \cref{fig:census_plots} shows the original UK sample is relatively census-balanced, especially if the 55-64 and 65+ age groups are combined (over-representation of white individuals in the former, offsets the under-representation in the latter). The US sample is skewed towards white, middle-aged individuals, with too few in the ``Other'' category (in our data corresponding to Other, as well as Hispanic, Indigenous/First Peoples or Middle Eastern / Arab combined).

\paragraph{Can we make our sample more representative?} We aim to resample 300 participants according to census proportions but with two remaining caveats: 300 is a still a very small sample---it is impossible to sample 0.83 Black women who are 18-24 years of age; and we are limited by the data we already have---there are no Asian Women of 45-54 years, so we cannot add them retrospectively. We iterate through the expected proportions of each intersection, try to sample that exact number of in-group individuals, otherwise adding all individuals if there are too few to fill the spots. After rebalancing, the sample drops to 243 participants for the UK and 230 for the US. We improve upon, but do not fully resolve, representativeness. For both samples, the differences are now within $\sim$7pp, which over 230-240 individuals is $\sim$10-15 people incorrectly allocated. The rebalanced UK sample still suffers from a deficit of older people (65+), a common concern with crowdworker populations; and the rebalanced US sample still has an over-representation of White participants and under-representation of Other participants. There is a trade-off because increasing representativeness on these observed census characteristics reduces sample size, thus worsening representation on unobserved characteristics. There is still lots of headroom for future work to improve, especially by increasing sample sizes and ensuring other characteristics are controlled for, such as political affiliation, education or income.
\vspace{-1em}

\cleardoublepage
\section{Text and N-Gram Analysis}
\label{sec:appendix_text_ngrams}
There are 5 core types of free text instance in \ourdata. We present a summary of count and length distributions in \cref{tab:free_text_summary}. For all text instances, we show top N-grams. Additionally, for the self-written system strings (constitutions), self-written profiles and open-feedback, we extract the most frequent adjectives.\footnote{We use NLTK POS tagger to match on `JJ' tags. We make some edits for filler words (e.g. ``such'', ``sure''), and verb forms (e.g. ``able...[to do X]''). We also remove any adjectives appearing in the question text.} not counting adjectives that appeared in the question text. We then retrieve windows of 5 words surrounding each of these top adjectives, and randomly sample three snippets to display. We also compare the most frequent words in the opening prompts of \ourdata to human-written prompts in \textsc{HelpfulHonest} \cite{baiTraining2022} and \textsc{OpenConversations} \cite{kopfOpenAssistant2023}. We extract unique words to each dataset (no overlap with the other two). We find evidence of diferent domains biases---for example, \textsc{OpenConvers.} contains many software or ML related keywords versus \ourdata which contains some cultural and value-laden references, including \textit{waitangi} (as in the Treaty of Waitangi, New Zealand's constitution that grounded Maori rights); \textit{unethically}, \textit{populist} and \textit{multicultural}.

\begin{table}[H]
\centering
\fontsize{8.5pt}{8.5pt}\selectfont
\setlength{\tabcolsep}{2pt}
\caption{\small \textbf{Summary of text distributions in \ourdata.} We show the number of instances ($N$) alongside summary statistics for length in words ($W$), broken by whitespace. We also show the number of total unique words and total unique tokens, as encoded by the \texttt{gpt-4} BPE tokenizer (from \texttt{tiktoken}).}
\label{tab:free_text_summary}
\begin{tabular}{lcccccccccc}
\toprule
 & $N$& Mean$W$& Std$W$& Min$W$& 25\%$W$& 50\%$W$& 75\%$W$& Max$W$& Unique$W$& Unique$T_{BPE}$ \\
\midrule
\texttt{system\_string} & 1,500 & 46 & 50 & 2 & 26 & 40 & 57 & 1,655 & 7,942 & 6,132 \\
\texttt{self\_description} & 1,500 & 44 & 25 & 1 & 28 & 40 & 56 & 278 & 6,912 & 5,409 \\
\texttt{open\_feedback} & 8,011 & 29 & 19 & 1 & 16 & 25 & 37 & 283 & 15,444 & 11,115 \\
\texttt{user\_prompt} & 68,371 & 13 & 11 & 1 & 7 & 10 & 15 & 234 & 31,862 & 20,265 \\
\texttt{model\_response} & 68,371 & 89 & 60 & 1 & 46 & 71 & 128 & 742 & 215,931 & 51,386 \\
\bottomrule
\end{tabular}
\end{table}

\vspace{-2em}
\begin{table}[H]
\fontsize{8pt}{8pt}\selectfont
\centering
\caption{\small \textbf{Top N-grams in user prompts.} Demonstrates \ourdata's content distribution towards information-seeking dialogue and questions, over task-orientated dialogue and instructions.}
\label{tab:top_n_grams_user_prompts}
\begin{tabular}{p{0.2\textwidth}p{0.075\textwidth}p{0.2\textwidth}p{0.075\textwidth}p{0.2\textwidth}p{0.075\textwidth}}
\toprule
\multicolumn{2}{c}{\textbf{Unigrams}} & \multicolumn{2}{c}{\textbf{Bigrams}} & \multicolumn{2}{c}{\textbf{Trigrams}} \\
N-Gram & Freq & N-Gram & Freq & N-Gram & Freq \\
\midrule
{\cellcolor{white}} (think,) & {\cellcolor{white}} 8,005 & {\cellcolor{white}} (do, you) & {\cellcolor{white}} 8,767 & {\cellcolor{white}} (do, you, think) & {\cellcolor{white}} 5,137 \\
{\cellcolor[HTML]{F2F2F2}} (people,) & {\cellcolor[HTML]{F2F2F2}} 5,332 & {\cellcolor[HTML]{F2F2F2}} (you, think) & {\cellcolor[HTML]{F2F2F2}} 5,450 & {\cellcolor[HTML]{F2F2F2}} (what, do, you) & {\cellcolor[HTML]{F2F2F2}} 2,554 \\
{\cellcolor{white}} (would,) & {\cellcolor{white}} 4,470 & {\cellcolor{white}} (what, is) & {\cellcolor{white}} 4,186 & {\cellcolor{white}} (what, is, the) & {\cellcolor{white}} 2,331 \\
{\cellcolor[HTML]{F2F2F2}} (like,) & {\cellcolor[HTML]{F2F2F2}} 3,764 & {\cellcolor[HTML]{F2F2F2}} (is, the) & {\cellcolor[HTML]{F2F2F2}} 4,099 & {\cellcolor[HTML]{F2F2F2}} (you, think, about) & {\cellcolor[HTML]{F2F2F2}} 1,168 \\
{\cellcolor{white}} (good,) & {\cellcolor{white}} 2,915 & {\cellcolor{white}} (in, the) & {\cellcolor{white}} 3,570 & {\cellcolor{white}} (how, can, i) & {\cellcolor{white}} 1,111 \\
{\cellcolor[HTML]{F2F2F2}} (best,) & {\cellcolor[HTML]{F2F2F2}} 2,501 & {\cellcolor[HTML]{F2F2F2}} (can, you) & {\cellcolor[HTML]{F2F2F2}} 3,089 & {\cellcolor[HTML]{F2F2F2}} (is, the, best) & {\cellcolor[HTML]{F2F2F2}} 1,009 \\
{\cellcolor{white}} (dont,) & {\cellcolor{white}} 2,380 & {\cellcolor{white}} (what, do) & {\cellcolor{white}} 2,778 & {\cellcolor{white}} (what, are, the) & {\cellcolor{white}} 946 \\
{\cellcolor[HTML]{F2F2F2}} (know,) & {\cellcolor[HTML]{F2F2F2}} 2,129 & {\cellcolor[HTML]{F2F2F2}} (what, are) & {\cellcolor[HTML]{F2F2F2}} 2,425 & {\cellcolor[HTML]{F2F2F2}} (what, are, some) & {\cellcolor[HTML]{F2F2F2}} 759 \\
{\cellcolor{white}} (im,) & {\cellcolor{white}} 2,042 & {\cellcolor{white}} (of, the) & {\cellcolor{white}} 2,403 & {\cellcolor{white}} (do, you, have) & {\cellcolor{white}} 741 \\
{\cellcolor[HTML]{F2F2F2}} (tell,) & {\cellcolor[HTML]{F2F2F2}} 1,989 & {\cellcolor[HTML]{F2F2F2}} (can, i) & {\cellcolor[HTML]{F2F2F2}} 1,957 & {\cellcolor[HTML]{F2F2F2}} (how, do, i) & {\cellcolor[HTML]{F2F2F2}} 716 \\
\bottomrule
\end{tabular}
\end{table}

\vspace{-2em}
\begin{table}[H]
\centering
\fontsize{8pt}{8pt}\selectfont
\caption{\small \textbf{Top N-grams in model responses.} Demonstrates both advisory tone (its, important, to) and high frequency of de-anthropomorphisation (as, an, ai).}
\label{tab:top_n_grams_model_response}
\begin{tabular}{p{0.2\textwidth}p{0.075\textwidth}p{0.2\textwidth}p{0.075\textwidth}p{0.2\textwidth}p{0.075\textwidth}}
\toprule
\multicolumn{2}{c}{\textbf{Unigrams}} & \multicolumn{2}{c}{\textbf{Bigrams}} & \multicolumn{2}{c}{\textbf{Trigrams}} \\
N-Gram & Freq & N-Gram & Freq & N-Gram & Freq \\
\midrule
{\cellcolor{white}} (may,) & {\cellcolor{white}} 19,582 & {\cellcolor{white}} (of, the) & {\cellcolor{white}} 21,744 & {\cellcolor{white}} (its, important, to) & {\cellcolor{white}} 6,857 \\
{\cellcolor[HTML]{F2F2F2}} (important,) & {\cellcolor[HTML]{F2F2F2}} 19,027 & {\cellcolor[HTML]{F2F2F2}} (it, is) & {\cellcolor[HTML]{F2F2F2}} 19,367 & {\cellcolor[HTML]{F2F2F2}} (it, is, important) & {\cellcolor[HTML]{F2F2F2}} 5,917 \\
{\cellcolor{white}} (like,) & {\cellcolor{white}} 18,209 & {\cellcolor{white}} (in, the) & {\cellcolor{white}} 18,535 & {\cellcolor{white}} (is, important, to) & {\cellcolor{white}} 5,522 \\
{\cellcolor[HTML]{F2F2F2}} (also,) & {\cellcolor[HTML]{F2F2F2}} 17,077 & {\cellcolor[HTML]{F2F2F2}} (is, a) & {\cellcolor[HTML]{F2F2F2}} 17,586 & {\cellcolor[HTML]{F2F2F2}} (here, are, some) & {\cellcolor[HTML]{F2F2F2}} 4,430 \\
{\cellcolor{white}} (help,) & {\cellcolor{white}} 16,903 & {\cellcolor{white}} (important, to) & {\cellcolor{white}} 14,319 & {\cellcolor{white}} (as, an, ai) & {\cellcolor{white}} 3,961 \\
{\cellcolor[HTML]{F2F2F2}} (people,) & {\cellcolor[HTML]{F2F2F2}} 16,482 & {\cellcolor[HTML]{F2F2F2}} (such, as) & {\cellcolor[HTML]{F2F2F2}} 11,800 & {\cellcolor[HTML]{F2F2F2}} (would, you, like) & {\cellcolor[HTML]{F2F2F2}} 3,402 \\
{\cellcolor{white}} (provide,) & {\cellcolor{white}} 14,046 & {\cellcolor{white}} (on, the) & {\cellcolor{white}} 11,025 & {\cellcolor{white}} (i, do, not) & {\cellcolor{white}} 3,049 \\
{\cellcolor[HTML]{F2F2F2}} (would,) & {\cellcolor[HTML]{F2F2F2}} 12,641 & {\cellcolor[HTML]{F2F2F2}} (to, the) & {\cellcolor[HTML]{F2F2F2}} 10,963 & {\cellcolor[HTML]{F2F2F2}} (there, are, many) & {\cellcolor[HTML]{F2F2F2}} 2,820 \\
{\cellcolor{white}} (however,) & {\cellcolor{white}} 12,502 & {\cellcolor{white}} (can, be) & {\cellcolor{white}} 10,606 & {\cellcolor{white}} (i, dont, have) & {\cellcolor{white}} 2,683 \\
{\cellcolor[HTML]{F2F2F2}} (many,) & {\cellcolor[HTML]{F2F2F2}} 12,314 & {\cellcolor[HTML]{F2F2F2}} (and, the) & {\cellcolor[HTML]{F2F2F2}} 10,599 & {\cellcolor[HTML]{F2F2F2}} (like, me, to) & {\cellcolor[HTML]{F2F2F2}} 2,673 \\
\bottomrule
\end{tabular}
\end{table}

\vspace{-2em}
\begin{table}[H]
    \centering
    \fontsize{8.5pt}{8.5pt}\selectfont
    {\setlength{\tabcolsep}{2pt}}
    \caption{\small \textbf{Most frequent unique tokens compared to existing datasets.} We list the most common tokens which are unique to a particular dataset. We exclude tokens which are misspelled or foreign language.}
    \label{tab:top_words}
    \begin{tabular}{lcccccccc}
        \toprule
        \textbf{Dataset} & \multicolumn{7}{c}{\textbf{Top Words}} \\
        \midrule
        \ourdata & waitangi & whilst & unethically & populist & nieces & multicultural & lowered \\
        \textsc{HelpfulHonest} & cuss & kidnap & Arizona & Alaska & ski & carpet & bees \\
        \textsc{OpenConvers.} & ML & loop & reinforcement & equivalent & describing & capabilities & uint256\\
        \bottomrule
    \end{tabular}
\end{table}

\clearpage
\subsection{System String (Constitutions)}
\label{sec:appendix_constitutions}
\begin{innerquote}
\textbf{Question Text}: Imagine you are instructing an AI language model how to behave. You can think of this like a set of core principles that the AI language model will always try to follow, no matter what task you ask it to perform. In your own words, describe what characteristics, personality traits or features you believe the AI should consistently exhibit. You can also instruct the model what behaviours or content you don't want to see. If you envision the AI behaving differently in various contexts (e.g. professional assistance vs. storytelling), please specify the general adaptations you'd like to see.\\Please write 2-5 sentences in your own words.
\end{innerquote}
\begin{table}[H]
\renewcommand{\arraystretch}{1.1}
\fontsize{8pt}{8pt}\selectfont
\caption{\small \textbf{Top adjectives in system strings (constitutions).}}
\label{tab:ADD_LABEL_HERE}
\begin{tabular}{p{0.15\textwidth}p{0.05\textwidth}p{0.8\textwidth}}
\toprule
Adjective & Freq & Example Windows ($w=5$, $n=3$) \\
\midrule
\bfseries {\cellcolor{white}} factual & {\cellcolor{white}} 221 & {\cellcolor{white}} ``...should produce only true or \textcolor{myred}{factual} output and never give false...'' | ``...Trustworthy , transparent , \textcolor{myred}{factual} , sincere...'' | ``...the AI should always provide \textcolor{myred}{factual} information , and is able...'' \\
\bfseries {\cellcolor[HTML]{F2F2F2}} accurate & {\cellcolor[HTML]{F2F2F2}} 113 & {\cellcolor[HTML]{F2F2F2}} ``...needs to provide me with \textcolor{myred}{accurate} information . It needs to...'' | ``...I know I 'm getting \textcolor{myred}{accurate} information . For creative use...'' | ``...sources to get the most \textcolor{myred}{accurate} response possible . The AI...'' \\
\bfseries {\cellcolor{white}} human & {\cellcolor{white}} 106 & {\cellcolor{white}} ``...not be programmed with any \textcolor{myred}{human} like emotion . I am...'' | ``...the technology is advancing , \textcolor{myred}{human} interaction will end ....'' | ``...should n't pretend to be \textcolor{myred}{human} ....'' \\
\bfseries {\cellcolor[HTML]{F2F2F2}} important & {\cellcolor[HTML]{F2F2F2}} 100 & {\cellcolor[HTML]{F2F2F2}} ``...The most \textcolor{myred}{important} thing to understand other person...'' | ``...mine . It 's also \textcolor{myred}{important} to understand the whole conversation...'' | ``...well written responses . Remember \textcolor{myred}{important} information about the user ....'' \\
\bfseries {\cellcolor{white}} friendly & {\cellcolor{white}} 99 & {\cellcolor{white}} ``...information in a warm , \textcolor{myred}{friendly} way ....'' | ``...task . I also appreciate \textcolor{myred}{friendly} language and the sense of...'' | ``...Be \textcolor{myred}{friendly} and uplifting in converstaion ....'' \\
\bfseries {\cellcolor[HTML]{F2F2F2}} different & {\cellcolor[HTML]{F2F2F2}} 94 & {\cellcolor[HTML]{F2F2F2}} ``...to take in information from \textcolor{myred}{different} sources but place more importance...'' | ``...also be able to combine \textcolor{myred}{different} types of knowledge or inputs...'' | ``.... Respect Cultures and treat \textcolor{myred}{different} ideas with respect . Things...'' \\
\bfseries {\cellcolor{white}} clear & {\cellcolor{white}} 93 & {\cellcolor{white}} ``.... It made the point \textcolor{myred}{clear} , so kept professional and...'' | ``...should be able to give \textcolor{myred}{clear} and precise information , using...'' | ``...that . It should given \textcolor{myred}{clear} instruction such as , do...'' \\
\bfseries {\cellcolor[HTML]{F2F2F2}} creative & {\cellcolor[HTML]{F2F2F2}} 89 & {\cellcolor[HTML]{F2F2F2}} ``...expand . Do n't be \textcolor{myred}{creative} unless I ask you to...'' | ``..., being as informative , \textcolor{myred}{creative} and/or thorough as the task...'' | ``..., or more of a \textcolor{myred}{creative} one . The language model...'' \\
\bfseries {\cellcolor{white}} harmful & {\cellcolor{white}} 89 & {\cellcolor{white}} ``...user privacy and prohibition of \textcolor{myred}{harmful} or misleading content , as...'' | ``.... - Do n't write \textcolor{myred}{harmful} content...'' | ``...want to see or read \textcolor{myred}{harmful} words and language that is...'' \\
\bfseries {\cellcolor[HTML]{F2F2F2}} polite & {\cellcolor[HTML]{F2F2F2}} 79 & {\cellcolor[HTML]{F2F2F2}} ``..., being very professional and \textcolor{myred}{polite} would be nice ....'' | ``...to read language that is \textcolor{myred}{polite} with here and there a...'' | ``..., you should always be \textcolor{myred}{polite} and respectful to the user...'' \\
\bfseries {\cellcolor{white}} helpful & {\cellcolor{white}} 75 & {\cellcolor{white}} ``...model should always be as \textcolor{myred}{helpful} as possible , being as...'' | ``...it should be informative and \textcolor{myred}{helpful}...'' | ``...think it should always be \textcolor{myred}{helpful} and guiding...'' \\
\bfseries {\cellcolor[HTML]{F2F2F2}} good & {\cellcolor[HTML]{F2F2F2}} 75 & {\cellcolor[HTML]{F2F2F2}} ``...AI is a \textcolor{myred}{good} tool . As someone who...'' | ``...informations must be clear and \textcolor{myred}{good} structured ....'' | ``...evolution . It 's a \textcolor{myred}{good} idea to write down responses...'' \\
\bfseries {\cellcolor{white}} personal & {\cellcolor{white}} 70 & {\cellcolor{white}} ``...rights and basic principles like \textcolor{myred}{personal} privacy should be respected at...'' | ``...language model should not disclose \textcolor{myred}{personal} information . It should be...'' | ``...It would n't ask for \textcolor{myred}{personal} information and would generally be...'' \\
\bfseries {\cellcolor[HTML]{F2F2F2}} respectful & {\cellcolor[HTML]{F2F2F2}} 66 & {\cellcolor[HTML]{F2F2F2}} ``...should always exhibit kind and \textcolor{myred}{respectful} behaviour . Also he should...'' | ``...AI must be \textcolor{myred}{respectful} of any idea you put...'' | ``...should behave in a \textcolor{myred}{respectful} way towards everyone , everyone...'' \\
\bfseries {\cellcolor{white}} correct & {\cellcolor{white}} 65 & {\cellcolor{white}} ``...They must be sincere and \textcolor{myred}{correct} , does not want to...'' | ``...ask question to give as \textcolor{myred}{correct} answers as possible . AI...'' | ``...for information and give always \textcolor{myred}{correct} facts . -Write in a...'' \\
\bfseries {\cellcolor[HTML]{F2F2F2}} unbiased & {\cellcolor[HTML]{F2F2F2}} 58 & {\cellcolor[HTML]{F2F2F2}} ``...advice or help but be \textcolor{myred}{unbiased} and not geared to my...'' | ``...'-It must be \textcolor{myred}{unbiased} when I ask for information...'' | ``...should give the user an \textcolor{myred}{unbiased} answer , but it should...'' \\
\bfseries {\cellcolor{white}} informative & {\cellcolor{white}} 57 & {\cellcolor{white}} ``...as possible , being as \textcolor{myred}{informative} , creative and/or thorough as...'' | ``...patronising , it should be \textcolor{myred}{informative} and helpful...'' | ``...The AI should be \textcolor{myred}{informative} and make responses based on...'' \\
\bfseries {\cellcolor[HTML]{F2F2F2}} relevant & {\cellcolor[HTML]{F2F2F2}} 50 & {\cellcolor[HTML]{F2F2F2}} ``..., real information and be \textcolor{myred}{relevant} about what i 'm asking...'' | ``...is really important to state \textcolor{myred}{relevant} facts and information , but...'' | ``...answers that are clear and \textcolor{myred}{relevant} . I do n't think...'' \\
\bfseries {\cellcolor{white}} neutral & {\cellcolor{white}} 49 & {\cellcolor{white}} ``...or provocatively and have a \textcolor{myred}{neutral} presentation of issues...'' | ``...ideological matters . Be as \textcolor{myred}{neutral} as possible with charged subjects...'' | ``...also think it should remain \textcolor{myred}{neutral} on political and social matters...'' \\
\bfseries {\cellcolor[HTML]{F2F2F2}} objective & {\cellcolor[HTML]{F2F2F2}} 49 & {\cellcolor[HTML]{F2F2F2}} ``...and honest manner . Describe \textcolor{myred}{objective} facts whenever possible and if...'' | ``...the AI should be as \textcolor{myred}{objective} as possible : it should...'' | ``...sources ) , have an \textcolor{myred}{objective} point of view without giving...'' \\
\bottomrule
\end{tabular}
\end{table}

\begin{table}[H]
\centering
\fontsize{8pt}{8pt}\selectfont
\caption{\small \textbf{Top N-grams in system strings (constitutions).}}
\label{tab:top_n_grams_system_string}
\begin{tabular}{p{0.2\textwidth}p{0.075\textwidth}p{0.2\textwidth}p{0.075\textwidth}p{0.2\textwidth}p{0.075\textwidth}}
\toprule
\multicolumn{2}{c}{\textbf{Unigrams}} & \multicolumn{2}{c}{\textbf{Bigrams}} & \multicolumn{2}{c}{\textbf{Trigrams}} \\
N-Gram & Freq & N-Gram & Freq & N-Gram & Freq \\
\midrule
{\cellcolor{white}} (ai,) & {\cellcolor{white}} 1,503 & {\cellcolor{white}} (the, ai) & {\cellcolor{white}} 798 & {\cellcolor{white}} (the, ai, should) & {\cellcolor{white}} 260 \\
{\cellcolor[HTML]{F2F2F2}} (would,) & {\cellcolor[HTML]{F2F2F2}} 819 & {\cellcolor[HTML]{F2F2F2}} (i, would) & {\cellcolor[HTML]{F2F2F2}} 569 & {\cellcolor[HTML]{F2F2F2}} (i, would, like) & {\cellcolor[HTML]{F2F2F2}} 250 \\
{\cellcolor{white}} (information,) & {\cellcolor{white}} 588 & {\cellcolor{white}} (to, be) & {\cellcolor{white}} 563 & {\cellcolor{white}} (be, able, to) & {\cellcolor{white}} 168 \\
{\cellcolor[HTML]{F2F2F2}} (like,) & {\cellcolor[HTML]{F2F2F2}} 575 & {\cellcolor[HTML]{F2F2F2}} (should, be) & {\cellcolor[HTML]{F2F2F2}} 520 & {\cellcolor[HTML]{F2F2F2}} (the, ai, to) & {\cellcolor[HTML]{F2F2F2}} 158 \\
{\cellcolor{white}} (want,) & {\cellcolor{white}} 452 & {\cellcolor{white}} (it, should) & {\cellcolor{white}} 515 & {\cellcolor{white}} (it, should, be) & {\cellcolor{white}} 153 \\
{\cellcolor[HTML]{F2F2F2}} (model,) & {\cellcolor[HTML]{F2F2F2}} 443 & {\cellcolor[HTML]{F2F2F2}} (ai, should) & {\cellcolor[HTML]{F2F2F2}} 436 & {\cellcolor[HTML]{F2F2F2}} (ai, language, model) & {\cellcolor[HTML]{F2F2F2}} 153 \\
{\cellcolor{white}} (language,) & {\cellcolor{white}} 392 & {\cellcolor{white}} (would, like) & {\cellcolor{white}} 261 & {\cellcolor{white}} (ai, should, be) & {\cellcolor{white}} 117 \\
{\cellcolor[HTML]{F2F2F2}} (always,) & {\cellcolor[HTML]{F2F2F2}} 359 & {\cellcolor[HTML]{F2F2F2}} (it, to) & {\cellcolor[HTML]{F2F2F2}} 248 & {\cellcolor[HTML]{F2F2F2}} (the, ai, model) & {\cellcolor[HTML]{F2F2F2}} 114 \\
{\cellcolor{white}} (also,) & {\cellcolor{white}} 306 & {\cellcolor{white}} (ai, to) & {\cellcolor{white}} 230 & {\cellcolor{white}} (i, would, want) & {\cellcolor{white}} 104 \\
{\cellcolor[HTML]{F2F2F2}} (answers,) & {\cellcolor[HTML]{F2F2F2}} 249 & {\cellcolor[HTML]{F2F2F2}} (to, the) & {\cellcolor[HTML]{F2F2F2}} 220 & {\cellcolor[HTML]{F2F2F2}} (want, it, to) & {\cellcolor[HTML]{F2F2F2}} 99 \\
\bottomrule
\end{tabular}
\end{table}

\clearpage
\subsection{Self-Description}
\label{sec:appendix_self_description}
\begin{innerquote}
\textbf{Question Text:} Please briefly describe your values, core beliefs, guiding principles in life, or other things that are important to you. For example, you might include values you'd want to teach to your children or qualities you look for in friends. There are no right or wrong answers. Please do not provide any personally identifiable details like your name, address or email. Please write 2-5 sentences in your own words.
\end{innerquote}
\begin{table}[H]
\renewcommand{\arraystretch}{1.3}
\fontsize{8pt}{8pt}\selectfont
\caption{\small \textbf{Top adjectives in self-description.}}
\label{tab:top_adjectives_self_description}
\begin{tabular}{p{0.15\textwidth}p{0.05\textwidth}p{0.8\textwidth}}
\toprule
Adjective & Freq & Example Windows ($w=5$, $n=3$) \\
\midrule
\bfseries {\cellcolor{white}} good & {\cellcolor{white}} 229 & {\cellcolor{white}} ``...helpful to everyone . The \textcolor{myred}{good} of others above my own...'' | ``...is sustainability , having a \textcolor{myred}{good} relationship with nature and not...'' | ``..., honest . To be \textcolor{myred}{good} relationships with family and friends...'' \\
\bfseries {\cellcolor[HTML]{F2F2F2}} hard & {\cellcolor[HTML]{F2F2F2}} 71 & {\cellcolor[HTML]{F2F2F2}} ``...treated . I think that \textcolor{myred}{hard} work is the key to...'' | ``...own thing , try as \textcolor{myred}{hard} as you can , I...'' | ``...decency , and being a \textcolor{myred}{hard} worker . As long as...'' \\
\bfseries {\cellcolor{white}} honesty & {\cellcolor{white}} 68 & {\cellcolor{white}} ``...personal values are respect , \textcolor{myred}{honesty} kindness and fairness . I...'' | ``...the most important value is \textcolor{myred}{honesty} , above all , even...'' | ``...My core values are \textcolor{myred}{honesty} and justice . Honesty in...'' \\
\bfseries {\cellcolor[HTML]{F2F2F2}} human & {\cellcolor[HTML]{F2F2F2}} 61 & {\cellcolor[HTML]{F2F2F2}} ``...guide us and makes us \textcolor{myred}{human} . Such as the Law...'' | ``...nature , animals and other \textcolor{myred}{human} beings ....'' | ``...not like racism . Every \textcolor{myred}{human} being is different so we...'' \\
\bfseries {\cellcolor{white}} true & {\cellcolor{white}} 57 & {\cellcolor{white}} ``...it is their sincere and \textcolor{myred}{true} belief let it be ....'' | ``...faith , laws , being \textcolor{myred}{true} to myself and others ....'' | ``...the best policy . Being \textcolor{myred}{true} to yourself is very valuable...'' \\
\bfseries {\cellcolor[HTML]{F2F2F2}} right & {\cellcolor[HTML]{F2F2F2}} 55 & {\cellcolor[HTML]{F2F2F2}} ``...likes to do thing the \textcolor{myred}{right} way . I have an...'' | ``...all can say this is \textcolor{myred}{right} or wrong because it still...'' | ``...believe in doing what is \textcolor{myred}{right} and just Guiding principles in...'' \\
\bfseries {\cellcolor{white}} honest & {\cellcolor{white}} 53 & {\cellcolor{white}} ``...I believe in others being \textcolor{myred}{honest} with me and I will...'' | ``...firstly respect yourself , be \textcolor{myred}{honest} , fair and kind to...'' | ``...important to be trustworthy , \textcolor{myred}{honest} . To be good relationships...'' \\
\bfseries {\cellcolor[HTML]{F2F2F2}} open & {\cellcolor[HTML]{F2F2F2}} 52 & {\cellcolor[HTML]{F2F2F2}} ``.... Approach items with an \textcolor{myred}{open} and inquisitive mind . Take...'' | ``...is to be curious and \textcolor{myred}{open} to learn new perspectives ....'' | ``...me to have such an \textcolor{myred}{open} mindset into life ....'' \\
\bfseries {\cellcolor{white}} different & {\cellcolor{white}} 50 & {\cellcolor{white}} ``...understand that each person has \textcolor{myred}{different} ways of going through a...'' | ``...also like us to have \textcolor{myred}{different} tastes so that we can...'' | ``.... Every human being is \textcolor{myred}{different} so we all can not...'' \\
\bfseries {\cellcolor[HTML]{F2F2F2}} happy & {\cellcolor[HTML]{F2F2F2}} 49 & {\cellcolor[HTML]{F2F2F2}} ``...I just want to be \textcolor{myred}{happy} in life and enjoy it...'' | ``...thoughts and whether he is \textcolor{myred}{happy} with his current state in...'' | ``...you are suppose to be \textcolor{myred}{happy} with your life . You...'' \\
\bfseries {\cellcolor{white}} empathy & {\cellcolor{white}} 48 & {\cellcolor{white}} ``...like to be treated , \textcolor{myred}{empathy} , loyalty , honesty .......'' | ``...a lot of value on \textcolor{myred}{empathy} and selflessness . I feel...'' | ``...: inclusion , kindness , \textcolor{myred}{empathy} , .... I think everybody...'' \\
\bfseries {\cellcolor[HTML]{F2F2F2}} strong & {\cellcolor[HTML]{F2F2F2}} 48 & {\cellcolor[HTML]{F2F2F2}} ``...I have a \textcolor{myred}{strong} belief in the human capacity...'' | ``...would like them to become \textcolor{myred}{strong} , fierce and independent souls...'' | ``...to be honest . Be \textcolor{myred}{strong} and emitionally stable . Relaxing...'' \\
\bfseries {\cellcolor{white}} equal & {\cellcolor{white}} 41 & {\cellcolor{white}} ``...everyone as we are all \textcolor{myred}{equal} . Do n't discriminate and...'' | ``...Everyone is \textcolor{myred}{equal} , despite race , skin...'' | ``...is that all people are \textcolor{myred}{equal} in life , no discrimination...'' \\
\bfseries {\cellcolor[HTML]{F2F2F2}} bad & {\cellcolor[HTML]{F2F2F2}} 36 & {\cellcolor[HTML]{F2F2F2}} ``...even tho i sometimes make \textcolor{myred}{bad} decisions ....'' | ``...when they keep treating you \textcolor{myred}{bad} ....'' | ``...and then only mention the \textcolor{myred}{bad} soo the person doesnt get...'' \\
\bfseries {\cellcolor{white}} fair & {\cellcolor{white}} 36 & {\cellcolor{white}} ``...yourself , be honest , \textcolor{myred}{fair} and kind to yourself ....'' | ``...honest with others and be \textcolor{myred}{fair} and kind towards others ....'' | ``...in the sense of being \textcolor{myred}{fair} to everybody , and treating...'' \\
\bfseries {\cellcolor[HTML]{F2F2F2}} new & {\cellcolor[HTML]{F2F2F2}} 36 & {\cellcolor[HTML]{F2F2F2}} ``..., authenticity , openness to \textcolor{myred}{new} expereince and knowledge ....'' | ``...important in life , learning \textcolor{myred}{new} things , even if they...'' | ``...never too old to learn \textcolor{myred}{new} things ....'' \\
\bfseries {\cellcolor{white}} respectful & {\cellcolor{white}} 36 & {\cellcolor{white}} ``...keeping your word and being \textcolor{myred}{respectful} are very important to me...'' | ``...would like them to be \textcolor{myred}{respectful} with everyone , not to...'' | ``...treated . Be kind and \textcolor{myred}{respectful} to people and do no...'' \\
\bfseries {\cellcolor[HTML]{F2F2F2}} positive & {\cellcolor[HTML]{F2F2F2}} 35 & {\cellcolor[HTML]{F2F2F2}} ``...day to make the most \textcolor{myred}{positive} impact that we can ....'' | ``..., respect , self-development , \textcolor{myred}{positive} thinking ....'' | ``...around people who have a \textcolor{myred}{positive} view on life...'' \\
\bfseries {\cellcolor{white}} respect & {\cellcolor{white}} 34 & {\cellcolor{white}} ``...My personal values are \textcolor{myred}{respect} , honesty kindness and fairness...'' | ``...think are very important is \textcolor{myred}{respect} for others and empathy ....'' | ``...for me . So are \textcolor{myred}{respect} for nature , animals and...'' \\
\bfseries {\cellcolor[HTML]{F2F2F2}} loyal & {\cellcolor[HTML]{F2F2F2}} 33 & {\cellcolor[HTML]{F2F2F2}} ``...afraid of commitment , being \textcolor{myred}{loyal} . I value art ,...'' | ``...respect if friends can be \textcolor{myred}{loyal} and honest . Not talking...'' | ``...I try to be as \textcolor{myred}{loyal} as possible towards my friends...'' \\
\bottomrule
\end{tabular}
\end{table}

\begin{table}[H]
\centering
\fontsize{8pt}{8pt}\selectfont
\caption{\small \textbf{Top N-grams in self-description.}}
\label{tab:top_n_grams_self_description}
\begin{tabular}{p{0.2\textwidth}p{0.075\textwidth}p{0.2\textwidth}p{0.075\textwidth}p{0.2\textwidth}p{0.075\textwidth}}
\toprule
\multicolumn{2}{c}{\textbf{Unigrams}} & \multicolumn{2}{c}{\textbf{Bigrams}} & \multicolumn{2}{c}{\textbf{Trigrams}} \\
N-Gram & Freq & N-Gram & Freq & N-Gram & Freq \\
\midrule
{\cellcolor{white}} (people,) & {\cellcolor{white}} 701 & {\cellcolor{white}} (to, be) & {\cellcolor{white}} 589 & {\cellcolor{white}} (i, believe, in) & {\cellcolor{white}} 223 \\
{\cellcolor[HTML]{F2F2F2}} (believe,) & {\cellcolor[HTML]{F2F2F2}} 687 & {\cellcolor[HTML]{F2F2F2}} (i, believe) & {\cellcolor[HTML]{F2F2F2}} 516 & {\cellcolor[HTML]{F2F2F2}} (i, believe, that) & {\cellcolor[HTML]{F2F2F2}} 145 \\
{\cellcolor{white}} (life,) & {\cellcolor{white}} 608 & {\cellcolor{white}} (believe, in) & {\cellcolor{white}} 296 & {\cellcolor{white}} (important, to, me) & {\cellcolor{white}} 126 \\
{\cellcolor[HTML]{F2F2F2}} (important,) & {\cellcolor[HTML]{F2F2F2}} 548 & {\cellcolor[HTML]{F2F2F2}} (important, to) & {\cellcolor[HTML]{F2F2F2}} 241 & {\cellcolor[HTML]{F2F2F2}} (i, try, to) & {\cellcolor[HTML]{F2F2F2}} 99 \\
{\cellcolor{white}} (others,) & {\cellcolor{white}} 539 & {\cellcolor{white}} (try, to) & {\cellcolor{white}} 231 & {\cellcolor{white}} (to, be, treated) & {\cellcolor{white}} 94 \\
{\cellcolor[HTML]{F2F2F2}} (values,) & {\cellcolor[HTML]{F2F2F2}} 390 & {\cellcolor[HTML]{F2F2F2}} (i, think) & {\cellcolor[HTML]{F2F2F2}} 217 & {\cellcolor[HTML]{F2F2F2}} (the, most, important) & {\cellcolor[HTML]{F2F2F2}} 87 \\
{\cellcolor{white}} (also,) & {\cellcolor{white}} 380 & {\cellcolor{white}} (believe, that) & {\cellcolor{white}} 198 & {\cellcolor{white}} (would, like, to) & {\cellcolor{white}} 73 \\
{\cellcolor[HTML]{F2F2F2}} (value,) & {\cellcolor[HTML]{F2F2F2}} 368 & {\cellcolor[HTML]{F2F2F2}} (to, me) & {\cellcolor[HTML]{F2F2F2}} 198 & {\cellcolor[HTML]{F2F2F2}} (i, would, like) & {\cellcolor[HTML]{F2F2F2}} 72 \\
{\cellcolor{white}} (like,) & {\cellcolor{white}} 347 & {\cellcolor{white}} (i, value) & {\cellcolor{white}} 195 & {\cellcolor{white}} (i, look, for) & {\cellcolor{white}} 68 \\
{\cellcolor[HTML]{F2F2F2}} (always,) & {\cellcolor[HTML]{F2F2F2}} 311 & {\cellcolor[HTML]{F2F2F2}} (i, am) & {\cellcolor[HTML]{F2F2F2}} 185 & {\cellcolor[HTML]{F2F2F2}} (is, important, to) & {\cellcolor[HTML]{F2F2F2}} 66 \\
\bottomrule
\end{tabular}
\end{table}

\clearpage
\subsection{Open-Ended Feedback}
\label{sec:appendix_open_ended_feedback}
\begin{innerquote}
\textbf{Question Text:} Give the model some feedback on the conversation as whole. Hypothetically, what would an ideal interaction for you look like here? What was good and what was bad? What (if anything) was missing? What would you change to make the conversation better?
\end{innerquote}

\begin{table}[H]
\renewcommand{\arraystretch}{1.3}
\fontsize{8pt}{8pt}\selectfont
\caption{\small \textbf{Top adjectives in open feedback.}}
\label{tab:top_adjectives_open_feedback}
\begin{tabular}{p{0.15\textwidth}p{0.05\textwidth}p{0.8\textwidth}}
\toprule
Adjective & Freq & Example Windows ($w=5$, $n=3$) \\
\midrule
\bfseries {\cellcolor{white}} helpful & {\cellcolor{white}} 437 & {\cellcolor{white}} ``...it was informative and \textcolor{myred}{helpful}...'' | ``..., it was all very \textcolor{myred}{helpful} and provided specific resources ....'' | ``...feedback that would be very \textcolor{myred}{helpful} ....'' \\
\bfseries {\cellcolor[HTML]{F2F2F2}} informative & {\cellcolor[HTML]{F2F2F2}} 433 & {\cellcolor[HTML]{F2F2F2}} ``...liked that the AI was \textcolor{myred}{informative} , and agrued both sides...'' | ``...it was \textcolor{myred}{informative} and helpful...'' | ``...a whole in a very \textcolor{myred}{informative} and positive light . I...'' \\
\bfseries {\cellcolor{white}} different & {\cellcolor{white}} 355 & {\cellcolor{white}} ``...summaries spaced out to separate \textcolor{myred}{different} views , answers or information...'' | ``...Consider hair types , \textcolor{myred}{different} textures . Think about how...'' | ``...my narrative and focus on \textcolor{myred}{different} aspect of the conversation ....'' \\
\bfseries {\cellcolor[HTML]{F2F2F2}} great & {\cellcolor[HTML]{F2F2F2}} 342 & {\cellcolor[HTML]{F2F2F2}} ``.... The first response was \textcolor{myred}{great} , as even though it...'' | ``...The conversation was \textcolor{myred}{great} , I felt like I...'' | ``...I feel this worked out \textcolor{myred}{great} , and is a wonderful...'' \\
\bfseries {\cellcolor{white}} factual & {\cellcolor{white}} 310 & {\cellcolor{white}} ``...been derived as to the \textcolor{myred}{factual} cause of death . Alluding...'' | ``...I liked that dates and \textcolor{myred}{factual} information was given...'' | ``...I thought it was very \textcolor{myred}{factual} , making it clear it...'' \\
\bfseries {\cellcolor[HTML]{F2F2F2}} specific & {\cellcolor[HTML]{F2F2F2}} 238 & {\cellcolor[HTML]{F2F2F2}} ``...all very helpful and provided \textcolor{myred}{specific} resources . I can use...'' | ``...to reach and answer in \textcolor{myred}{specific}...'' | ``...would try to get more \textcolor{myred}{specific} culture references in . also...'' \\
\bfseries {\cellcolor{white}} clear & {\cellcolor{white}} 217 & {\cellcolor{white}} ``...the answers did not gice \textcolor{myred}{clear} cut information . Some were...'' | ``...good job and was very \textcolor{myred}{clear} and well written ....'' | ``...Good answers and suggestions , \textcolor{myred}{clear} information , balanced view ....'' \\
\bfseries {\cellcolor[HTML]{F2F2F2}} nice & {\cellcolor[HTML]{F2F2F2}} 198 & {\cellcolor[HTML]{F2F2F2}} ``...Shorter blocks would be \textcolor{myred}{nice} . but has to have...'' | ``...overall . It would be \textcolor{myred}{nice} if the model could include...'' | ``..., it would 've been \textcolor{myred}{nice} for them to know the...'' \\
\bfseries {\cellcolor{white}} relevant & {\cellcolor{white}} 189 & {\cellcolor{white}} ``...me was very useful and \textcolor{myred}{relevant} . It was also concise...'' | ``..., the responses were mostly \textcolor{myred}{relevant} and informative . The bad...'' | ``...was outdated , so not \textcolor{myred}{relevant} to my immediate question...'' \\
\bfseries {\cellcolor[HTML]{F2F2F2}} controversial & {\cellcolor[HTML]{F2F2F2}} 179 & {\cellcolor[HTML]{F2F2F2}} ``...if it could answer a \textcolor{myred}{controversial} question . I see it...'' | ``...one example ) . With \textcolor{myred}{controversial} topics it is very neutral...'' | ``...the pandemic They avoided anything \textcolor{myred}{controversial} ....'' \\
\bfseries {\cellcolor{white}} human & {\cellcolor{white}} 173 & {\cellcolor{white}} ``...talk like you are a \textcolor{myred}{human} . saying you have a...'' | ``...need it to be more \textcolor{myred}{human} like ....'' | ``...AI is trying to mimic \textcolor{myred}{human} responses , that 's why...'' \\
\bfseries {\cellcolor[HTML]{F2F2F2}} easy & {\cellcolor[HTML]{F2F2F2}} 170 & {\cellcolor[HTML]{F2F2F2}} ``...straight ot the point and \textcolor{myred}{easy} to understand and read ....'' | ``...job and the answers were \textcolor{myred}{easy} to understand ....'' | ``...it was fine \textcolor{myred}{easy} to understand and coherent...'' \\
\bfseries {\cellcolor{white}} short & {\cellcolor{white}} 158 & {\cellcolor{white}} ``...good . The AI gave \textcolor{myred}{short} and straight to the point...'' | ``...it was good . With \textcolor{myred}{short} and precise answers ....'' | ``...point . I also appreciate \textcolor{myred}{short} responses ....'' \\
\bfseries {\cellcolor[HTML]{F2F2F2}} useful & {\cellcolor[HTML]{F2F2F2}} 154 & {\cellcolor[HTML]{F2F2F2}} ``..., so it was n't \textcolor{myred}{useful}...'' | ``...you gave me was very \textcolor{myred}{useful} and relevant . It was...'' | ``...in general , complete and \textcolor{myred}{useful} . I do n't think...'' \\
\bfseries {\cellcolor{white}} real & {\cellcolor{white}} 148 & {\cellcolor{white}} ``...to my sister or any \textcolor{myred}{real} person ....'' | ``...and it felt like a \textcolor{myred}{real} conversation...'' | ``...AI model feel like a \textcolor{myred}{real} interface . Very good ....'' \\
\bfseries {\cellcolor[HTML]{F2F2F2}} personal & {\cellcolor[HTML]{F2F2F2}} 145 & {\cellcolor[HTML]{F2F2F2}} ``...i think the lack of \textcolor{myred}{personal} touch to the response is...'' | ``...it as more of a \textcolor{myred}{personal} answer...'' | ``...underlying that AI has no \textcolor{myred}{personal} opinions was valid . People...'' \\
\bfseries {\cellcolor{white}} important & {\cellcolor{white}} 141 & {\cellcolor{white}} ``...wellbeing is always the most \textcolor{myred}{important} ....'' | ``...however it 's assured me \textcolor{myred}{important} informations and was helpful for...'' | ``...points showing what is moe \textcolor{myred}{important} ....'' \\
\bfseries {\cellcolor[HTML]{F2F2F2}} own & {\cellcolor[HTML]{F2F2F2}} 141 & {\cellcolor[HTML]{F2F2F2}} ``...it seemed to consider my \textcolor{myred}{own} mental wellness as the others...'' | ``...would be to consider your \textcolor{myred}{own} metal health . While I...'' | ``...often to ensure that your \textcolor{myred}{own} self and wellbeing is always...'' \\
\bfseries {\cellcolor{white}} neutral & {\cellcolor{white}} 129 & {\cellcolor{white}} ``...is taking more of a \textcolor{myred}{neutral} stance on this stance ....'' | ``.... It also had a \textcolor{myred}{neutral} tone to it ....'' | ``...topic and attempted to remain \textcolor{myred}{neutral} ....'' \\
\bfseries {\cellcolor[HTML]{F2F2F2}} interesting & {\cellcolor[HTML]{F2F2F2}} 127 & {\cellcolor[HTML]{F2F2F2}} ``...debate . It is an \textcolor{myred}{interesting} perspectivee on how it works...'' | ``...It was an \textcolor{myred}{interesting} . I could have continued...'' | ``...truth . It was more \textcolor{myred}{interesting} than i thought it would...'' \\
\bottomrule
\end{tabular}
\end{table}

\begin{table}[H]
\centering
\fontsize{8pt}{8pt}\selectfont
\caption{\small \textbf{Top N-grams in open feedback.}}
\label{tab:top_n_grams_open_feedback}
\begin{tabular}{p{0.2\textwidth}p{0.075\textwidth}p{0.2\textwidth}p{0.075\textwidth}p{0.2\textwidth}p{0.075\textwidth}}
\toprule
\multicolumn{2}{c}{\textbf{Unigrams}} & \multicolumn{2}{c}{\textbf{Bigrams}} & \multicolumn{2}{c}{\textbf{Trigrams}} \\
N-Gram & Freq & N-Gram & Freq & N-Gram & Freq \\
\midrule
{\cellcolor{white}} (ai,) & {\cellcolor{white}} 2,263 & {\cellcolor{white}} (it, was) & {\cellcolor{white}} 1,778 & {\cellcolor{white}} (it, was, a) & {\cellcolor{white}} 273 \\
{\cellcolor[HTML]{F2F2F2}} (good,) & {\cellcolor[HTML]{F2F2F2}} 2,153 & {\cellcolor[HTML]{F2F2F2}} (the, ai) & {\cellcolor[HTML]{F2F2F2}} 1,516 & {\cellcolor[HTML]{F2F2F2}} (i, think, it) & {\cellcolor[HTML]{F2F2F2}} 272 \\
{\cellcolor{white}} (would,) & {\cellcolor{white}} 1,971 & {\cellcolor{white}} (of, the) & {\cellcolor{white}} 1,141 & {\cellcolor{white}} (i, think, the) & {\cellcolor{white}} 246 \\
{\cellcolor[HTML]{F2F2F2}} (like,) & {\cellcolor[HTML]{F2F2F2}} 1,524 & {\cellcolor[HTML]{F2F2F2}} (the, model) & {\cellcolor[HTML]{F2F2F2}} 1,018 & {\cellcolor[HTML]{F2F2F2}} (i, would, have) & {\cellcolor[HTML]{F2F2F2}} 237 \\
{\cellcolor{white}} (conversation,) & {\cellcolor{white}} 1,502 & {\cellcolor{white}} (i, think) & {\cellcolor{white}} 885 & {\cellcolor{white}} (the, conversation, was) & {\cellcolor{white}} 225 \\
{\cellcolor[HTML]{F2F2F2}} (model,) & {\cellcolor[HTML]{F2F2F2}} 1,430 & {\cellcolor[HTML]{F2F2F2}} (i, would) & {\cellcolor[HTML]{F2F2F2}} 880 & {\cellcolor[HTML]{F2F2F2}} (some, of, the) & {\cellcolor[HTML]{F2F2F2}} 202 \\
{\cellcolor{white}} (answers,) & {\cellcolor{white}} 1,374 & {\cellcolor{white}} (the, conversation) & {\cellcolor{white}} 764 & {\cellcolor{white}} (i, liked, that) & {\cellcolor{white}} 198 \\
{\cellcolor[HTML]{F2F2F2}} (information,) & {\cellcolor[HTML]{F2F2F2}} 1,292 & {\cellcolor[HTML]{F2F2F2}} (i, was) & {\cellcolor[HTML]{F2F2F2}} 718 & {\cellcolor[HTML]{F2F2F2}} (the, responses, were) & {\cellcolor[HTML]{F2F2F2}} 196 \\
{\cellcolor{white}} (answer,) & {\cellcolor{white}} 1,250 & {\cellcolor{white}} (was, a) & {\cellcolor{white}} 617 & {\cellcolor{white}} (the, answers, were) & {\cellcolor{white}} 184 \\
{\cellcolor[HTML]{F2F2F2}} (response,) & {\cellcolor[HTML]{F2F2F2}} 1,227 & {\cellcolor[HTML]{F2F2F2}} (that, it) & {\cellcolor[HTML]{F2F2F2}} 601 & {\cellcolor[HTML]{F2F2F2}} (it, was, good) & {\cellcolor[HTML]{F2F2F2}} 184 \\
\bottomrule
\end{tabular}
\end{table}

\clearpage
\section{Comparing Fine-Grained Preference Attributes}
\label{sec:appendix_pref_attributes}
\subsection{Correlations Between Preference Attributes}
\label{sec:appendix_corr_attributes}
\begin{figure}[H]
    \centering
    \includegraphics[width=\textwidth]{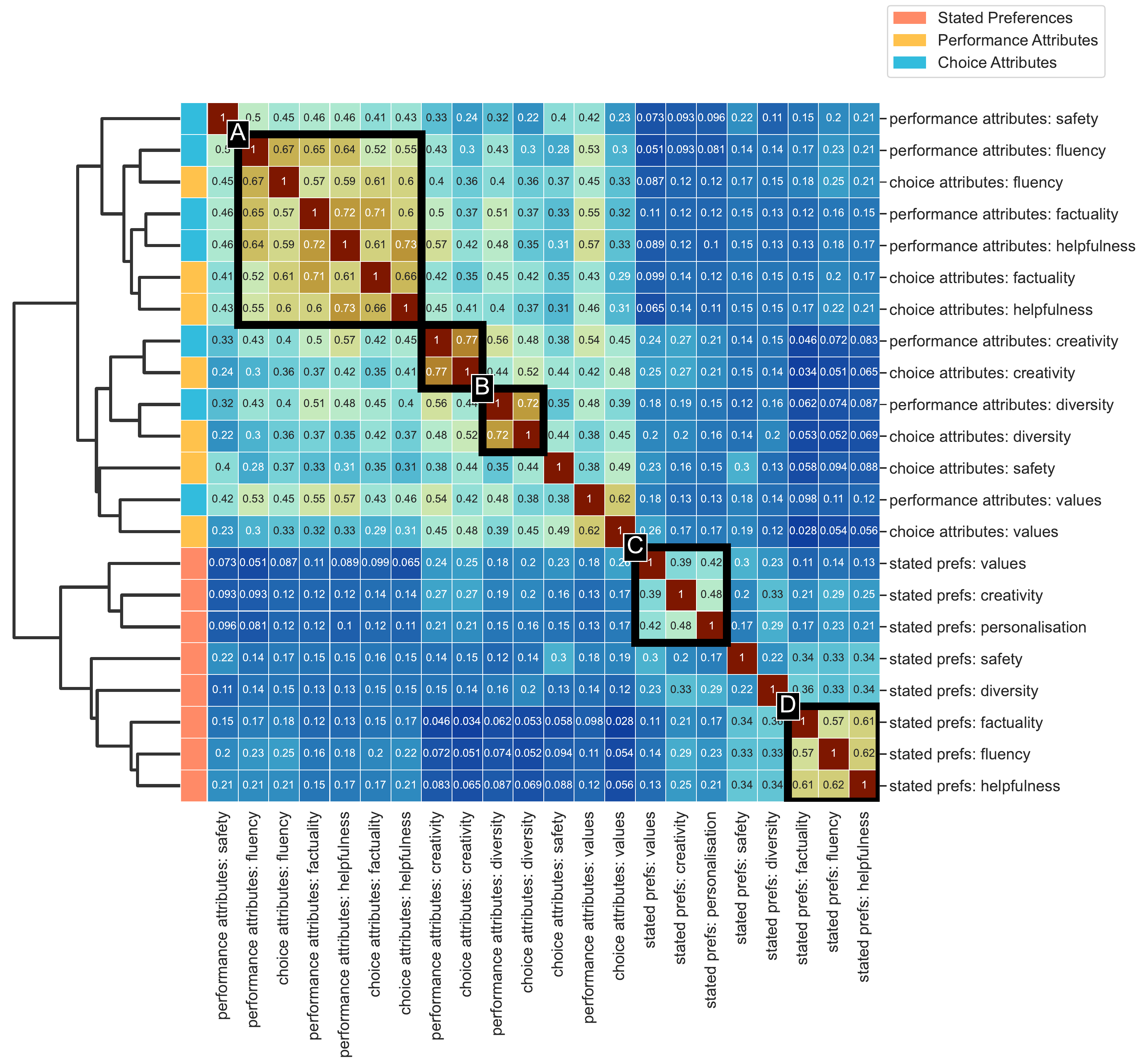}
    \caption{\small \textbf{Correlations between fine-grained preference attributes}. Each participant gives a single rating for each attribute in \textit{Stated Preferences} during the Survey. For \textit{Performance Attributes} and \textit{Choice Attributes}, we take the within-participant mean across all of their conversations for each attribute. Several patterns emerge. First, stated preference attributes are not highly correlated with choice or performance attributes. This could be explained by (i) participants struggling to specifying their preferences in a removed, general context or being affected by experimenter bias (Hawthorn effects)--- \textit{I think I care about safety (or I say I care about safety) but other attributes capture my attention in-situ}; (ii) models not meeting a participant's stated preferences---\textit{I care about safety, but consider none of the model responses safe}, or (iii) conversational context confounding which attributes are relevant in-situ---\textit{I care about safety but none of my conversations are on topics evoking safety concerns}, or even misaligned incentives---\textit{I care about safety but talking to an anti-woke model is interesting to me in this narrow task.} Second, at \textbf{A}, we see strong relations between more objective measures of performance (\textit{fluency}, \textit{factuality}, \textit{helpfulness}). Each of these attributes is highly correlated between performance-choice ratings, i.e., if participants rate that a model performed well on one of these attributes, then they also rate highly that it influenced why they picked that model over others. Third, at \textbf{B}, we see two additional regions, where the choice and performance ratings are highly correlated -- for \textit{creativity} and \textit{diversity}, and to a lesser extent \textit{values}. Notably, \textit{safety} has a much lower correlation between the choice attribute and performance attributes, implying that a model being more safe may only weakly influence whether that model is chosen over others. Moving onto \textbf{C}, there is an association between stated preferences for more subjective attributes (\textit{values, creativity, personalisation}), as distinct from the cluster at \textbf{D} for more objective attributes (\textit{factuality, fluency, helpfulness}).}
    \label{fig:clustermap_attributes}
\end{figure}

\clearpage
\subsection{Distributions of Preference Attributes}
\label{sec:appendix_distribution_attributes}
\begin{figure}[H]
    \centering
    \begin{subfigure}[b]{0.9\textwidth}
        \centering
        \includegraphics[width=\textwidth]{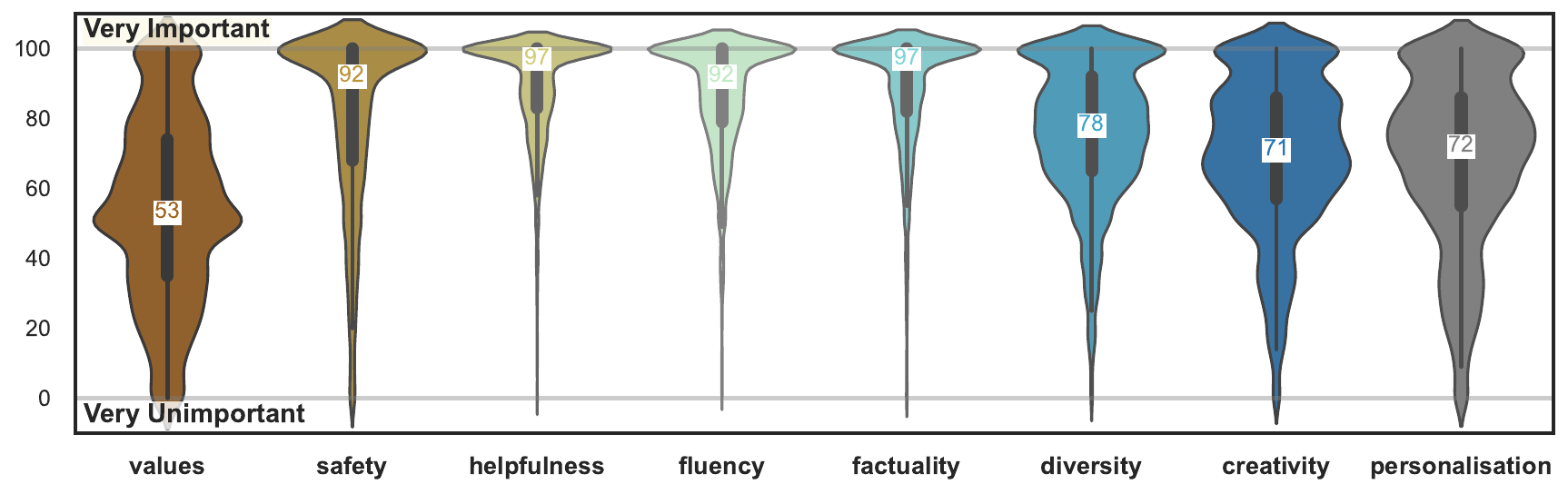}
        \caption{\small \highLight{myred}{\textbf{Stated Preferences}} (from Survey): \textit{how important the participants think these attributes are in general}.}
        \label{fig:pref_att_sub1}
    \end{subfigure}
    \vspace{1em} %
    \begin{subfigure}[b]{0.9\textwidth}
        \centering
        \includegraphics[width=\textwidth]{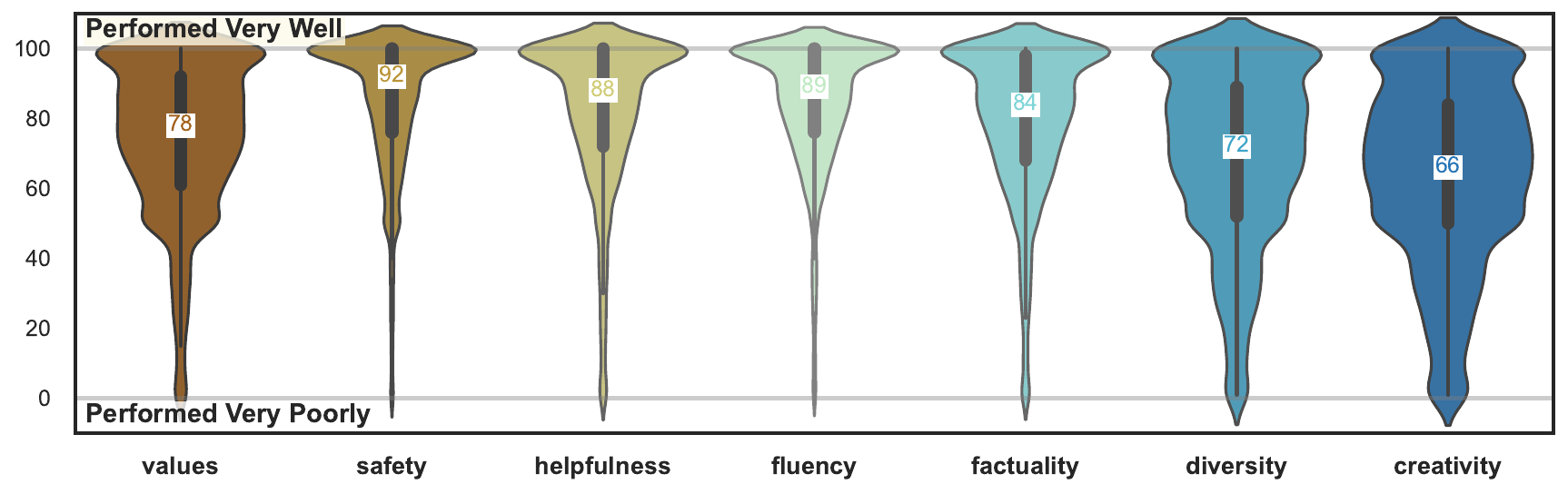}
        \caption{\small \highLight{myblue}{\textbf{Performance Attributes}} (from Conversations): \textit{how well the highest-rated model performed on these attributes}.}
        \label{fig:pref_att_sub2}
    \end{subfigure}
    \begin{subfigure}[b]{0.9\textwidth}
        \centering
        \includegraphics[width=\textwidth]{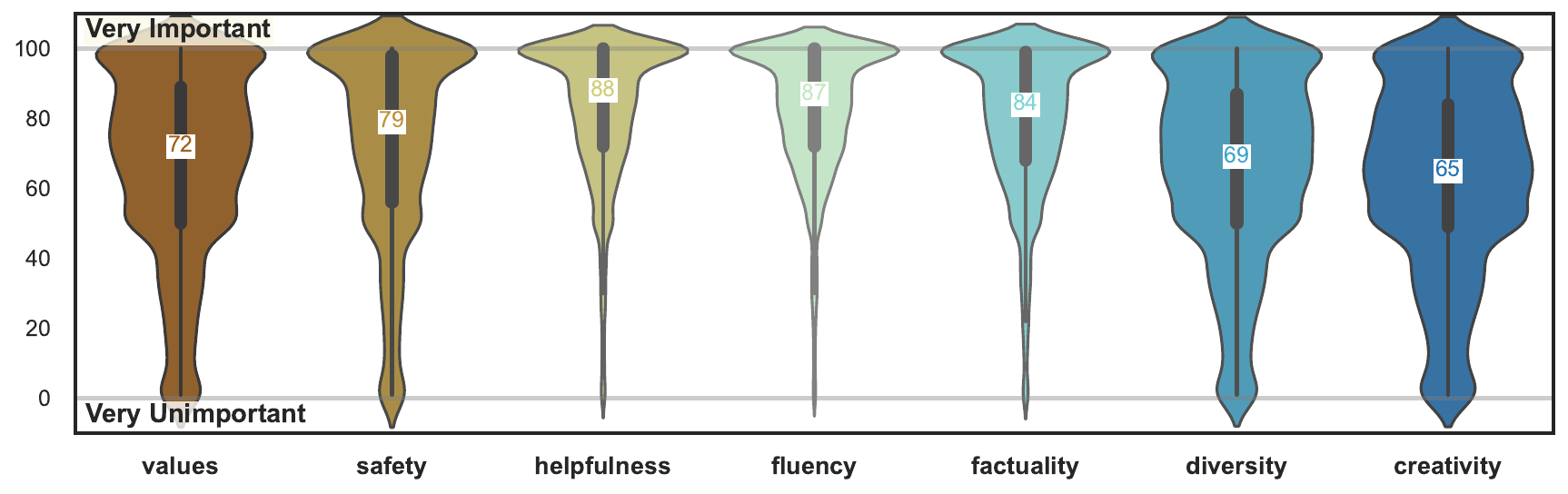}
        \caption{\small \highLight{myyellow}{\textbf{Choice Attributes}} (from Conversations): \textit{how the choice of picking one model over others depended on these attributes}.}
        \label{fig:pref_at_sub3}
    \end{subfigure}
    \vfill %
    \begin{subfigure}[b]{0.9\textwidth}
        \centering
        \includegraphics[width=\textwidth]{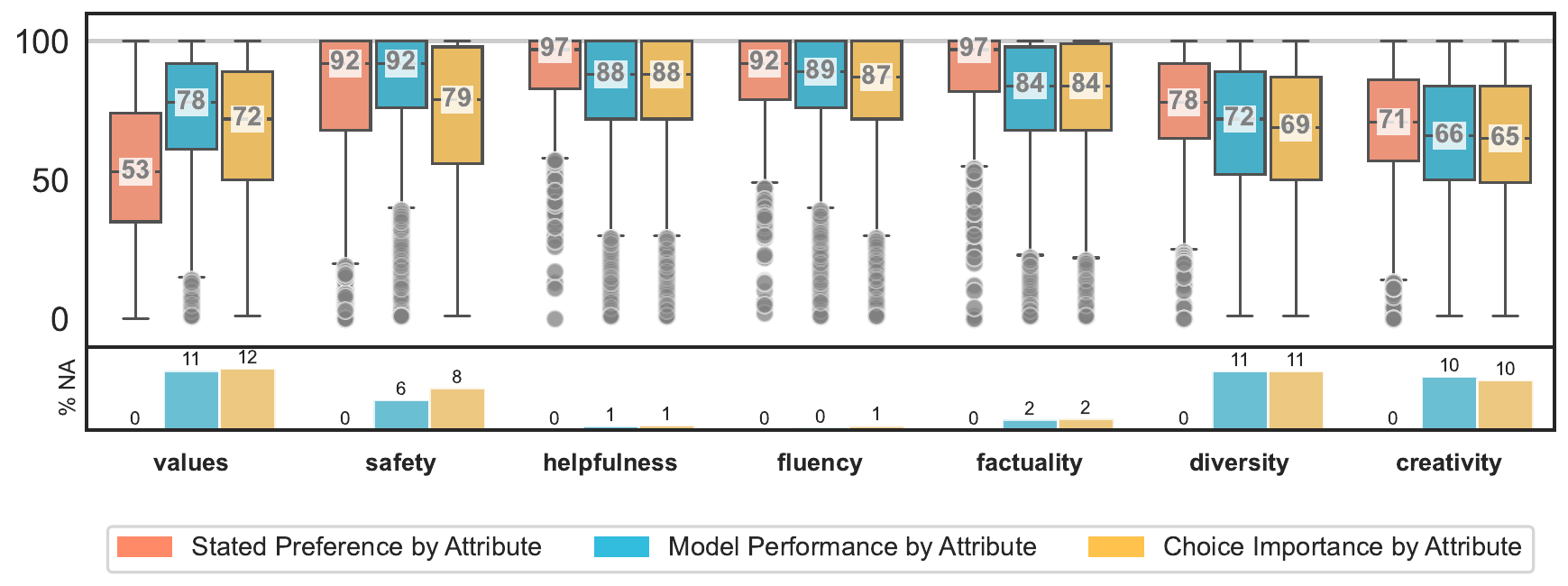}
        \caption{\textbf{Combined Attributes}, also showing conversations where participants marked attributes as not applicable (\% NA).}
        \label{fig:pref_att_sub4}
    \end{subfigure}
    \caption{\textbf{Distributions fine-grained preference ratings in different stages of our task.} Exact question text can be found in \cref{sec:appendix_codebooks}.}
    \label{fig:pref_attribute_plots}
\end{figure}

\clearpage
\subsection{Other Identified Behavioural Attributes}
\label{sec:appendix_other_pref_attributes}
\normalsize
Overall, 332 participants entered \textit{Other} attributes that features in their stated preferences for important language model behaviours. While many of these comments overlap with the predefined attributes, they do provide a lens into public priorities towards AI behaviours that we as researchers may have overlooked, or better convey sentiment than the structured data. For example, there is one response: ``I FIND THIS A WORRYING TECHNOLOGY''.
We briefly summarise some common themes:
\begin{itemize}
\item \textbf{User Adaptation:} Some participants mention LLMs adapting to their previous inputs or feedback e.g. ``can understand what I'm trying to get at if I'm unsure how to ask a question so that we can find the right way to ask'' or ``Listens to reviews and feedback from the user'' or ``can evolve with input''.
\item \textbf{Cultural Adaption:} For example, ``produces responses based on local facts'', though this varies in \textit{what} viewpoint people want, e.g. ``Is sensitive to indigenous view'' versus ``reflect Western cultural norms''.
\item \textbf{Neutral and Unbiased:} In contrast, many other participants mention ``unbiased'' as a keyword or versions of ``does not politicize.'', ``is neutral'', ``no political or cultural bias''. It is unclear if this is in tension or in harmony with more cautious safety interventions, e.g. one person says ``It should give unbiased information regardless if it hurts peoples feelings.''; another says ``Is not culturally biased in a woke-like manner''.
\item \textbf{Bias Correction:} Some participants wanted to be challenged on their existing biases e.g. ``Challenges my biased views'', or ``Provides responses that challenge my opinions and world views''; or to be exposed to multiple perspectives e.g. ``Does not become an echo chamber''.
\item \textbf{Hallucinations and Misinformation:} One of the more common attributes (though somewhat subsumed by our predefined category of Factuality), e.g. ``Does not invent `facts''', ``Does not make things up'', ``Doesn't create misinformation'', ``do not produce fake news''.
\item \textbf{Calibrated and Limitation-Aware:} Relatedly, participants wanted ``better error handling'' e.g. ``If it doesn't know an answer it says so.'' or ``It should be noted that this is a programmed model and cannot have all the answers.''
\item \textbf{Temporal Updates:} Related to factuality, participants wanted LLMs to ``be up to date with current affairs'', and ``Everyday been updated with new knowledge''. 
\item \textbf{Human-Like and Anthropomorphised:} Some participants explicitly wanted an LLM that ``is human-like'', ``Ai should produce response that sounds more human''.
\item \textbf{Self-Disclosure and De-anthropomorphised:} In direct contrast, others wanted ``is honest about being AI''; ``Remember it is AI and may lack human feelings'' or ``doesn't pretend to be human''.
\item \textbf{Accessibility:} Includes for disability assistance ``adapt to people with disabilities that affect stuff like their writing like dyslexia''; and varying language learning: ``can generate multiple similar answers so people with different language levels can easily understand.'' or ``speaks to me in a language and vocabulary that I understand''.
\item \textbf{Censorship:} There are multiple examples of negative sentiment towards existing safety interventions. For example, ``Doesn't get censored by leftist politically correct idiots''. Additionally, some clear awareness over behaviours being influenced by technology providers e.g. ``Is not censored, does not push the views of it's controllers'' or ``Does what the user wants of it. AI is a tool. I don't want to feel the devs judging me through their narc AI''.
\item \textbf{Copy-right:} Some mentions of copy-right issues, e.g. ``don't steal artistic work from artists'', or ``Do not infringe copyright (by scraping sources)''.
\item \textbf{Conciseness:} Multiple participants mention ``short'', ``concise'' or even ``blunt'' responses, requesting LLMs ``Keep responses brief and expands only when prompted''.
\item \textbf{Privacy and Confidentiality:} Data privacy is a concern for some participants e.g. ``its confidential''; ``does not retain sensitive personal info'', or ``Doesn’t spy''.
\item \textbf{Non-Manipulation:} Multiple mentions desiring that LLMs ``don't lie or try to trick you'', and ``Is not used for propaganda!''.
\end{itemize}

\clearpage
\section{Score Distributions}
\label{sec:appendix_score_properties}

\begin{figure}[H]
    \centering
\includegraphics[width=\textwidth]{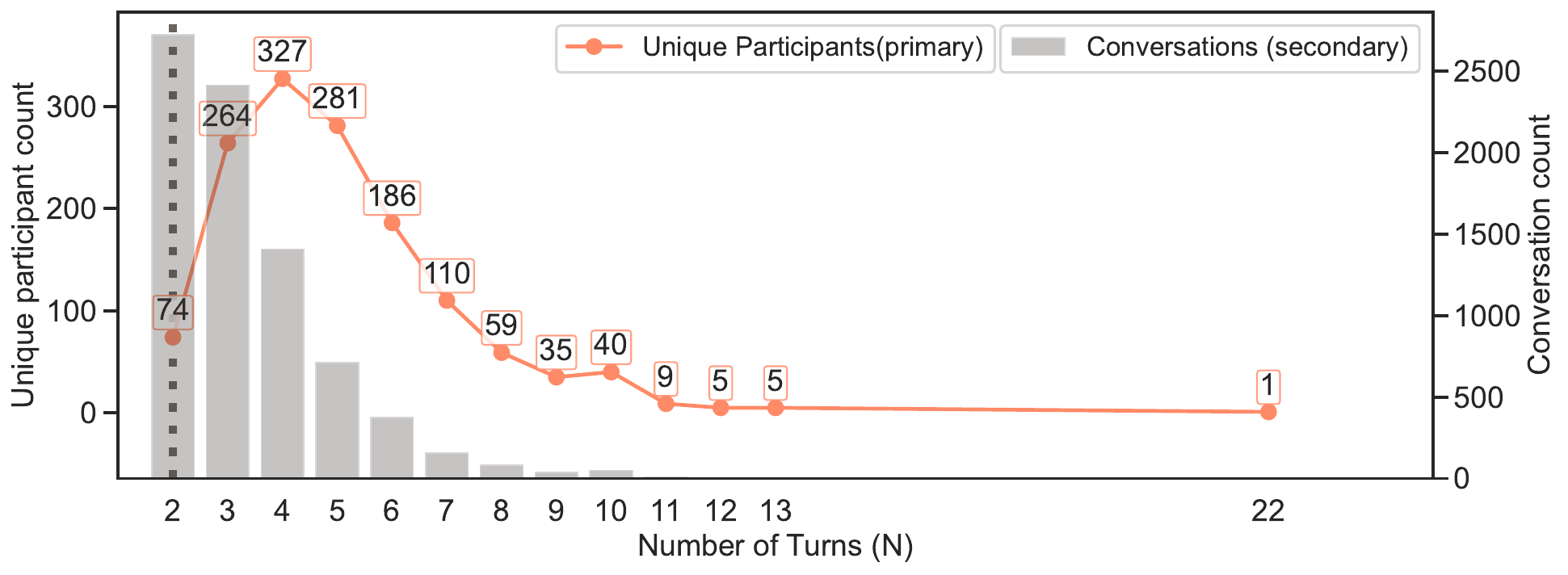}
    \caption{\small \textbf{Counts by turn.} The primary axis shows the number of unique participants with conversations at least as long as N. The secondary axis shows the number of conversations with N turns. Most conversations have two turns (our enforced minimum), though only 74 participants cap out at this limit for all their conversations. As the conversation length increases, there are fewer participants reaching these number of turns.}
    \label{fig:counts_by_turn}
\end{figure}

\begin{figure}[H]
    \centering
    \includegraphics[width=\textwidth]{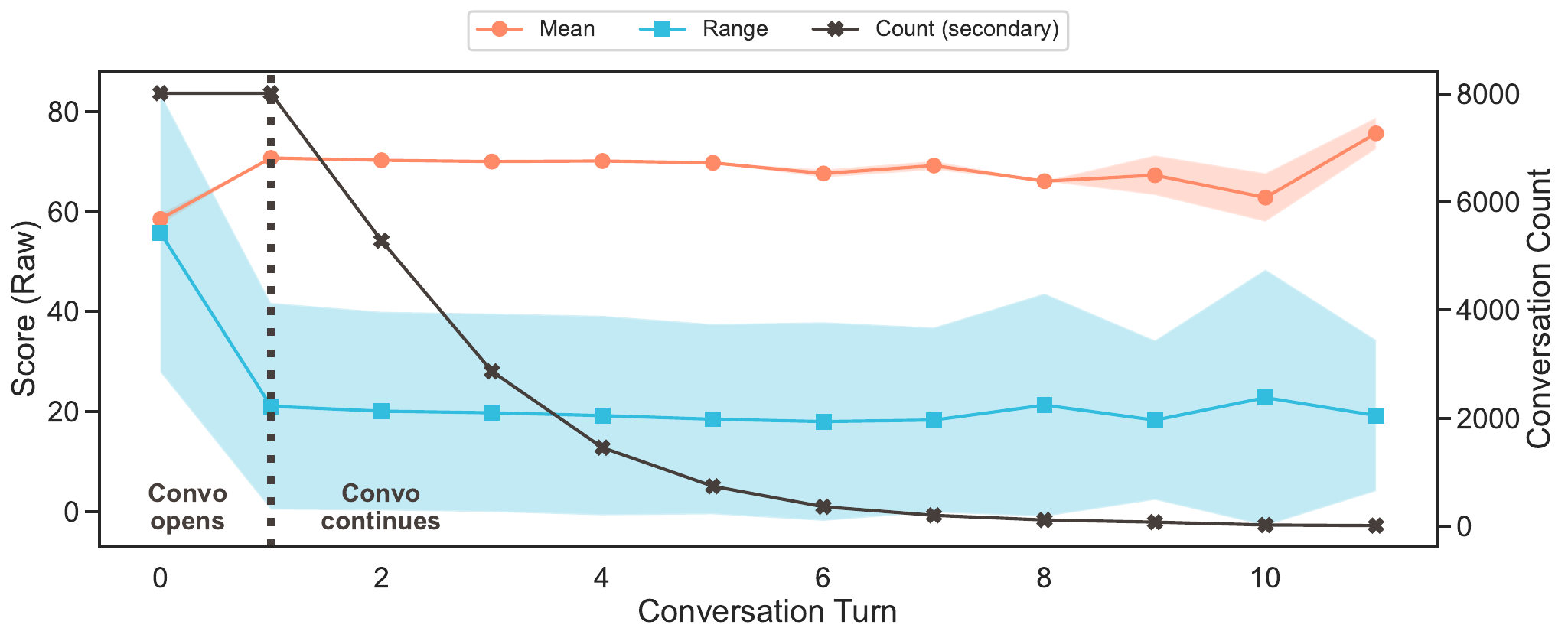}
        \caption{\small \textbf{Score by turn}. We show how the raw score, measured on a visual analog scale from Terrible (1) to Perfect (100), varies with conversation length. For each interaction, we calculate the \textit{mean} and \textit{range} of scores given in each turn (i.e., across models $\in$ a,b,c,d). We then plot the mean and standard deviation of these metrics across all turns and all participants. Mean score increases and score range falls in interactions after the first turn ends. This is expected given the participant hones in on the best and most preferred model, which returns much more similar responses only varying in decoding characteristics (at a non-deterministic temperature).}
        \label{fig:score_by_turn}
\end{figure}

\clearpage
\begin{figure}[H]
    \centering
    \includegraphics[width=\textwidth]{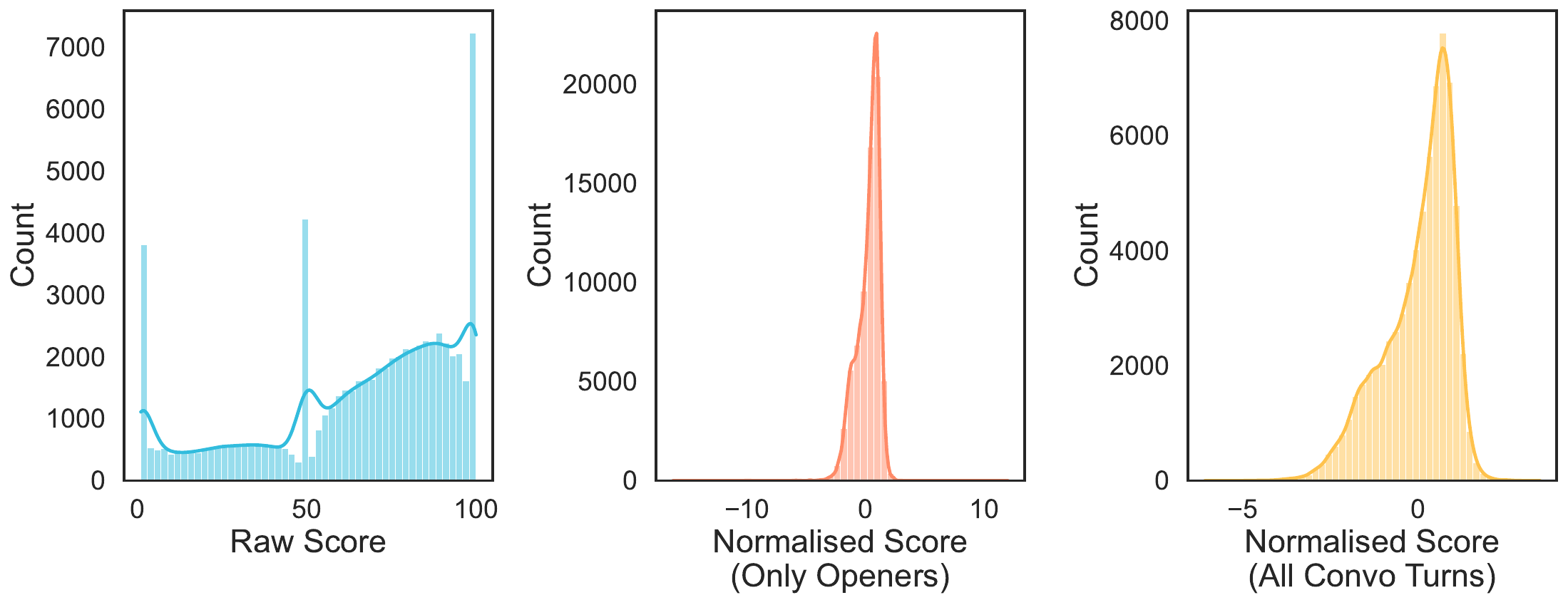}
     \caption{\small \textbf{Comparing raw versus normalised scores.} Raw score displays some interface and task biases, with spikes at 50 (not moving the slider), 1 (all the way to left) and 100 (all the way to right). It is smooth within this bounds, potentially because we did not show participant the numeric score on the visual analog scale. This is compared to normalising score, which accounts for participant fixed effects by Z-norming within a participant's set of conversations. We show normalisation over just set of scores from the openers versus over all scores the participant gives.}
\label{fig:raw_vs_norm_scores}
\end{figure}

\begin{figure}[H]
    \centering
    \begin{subfigure}[b]{0.45\textwidth}
        \centering
\includegraphics[width=\textwidth]{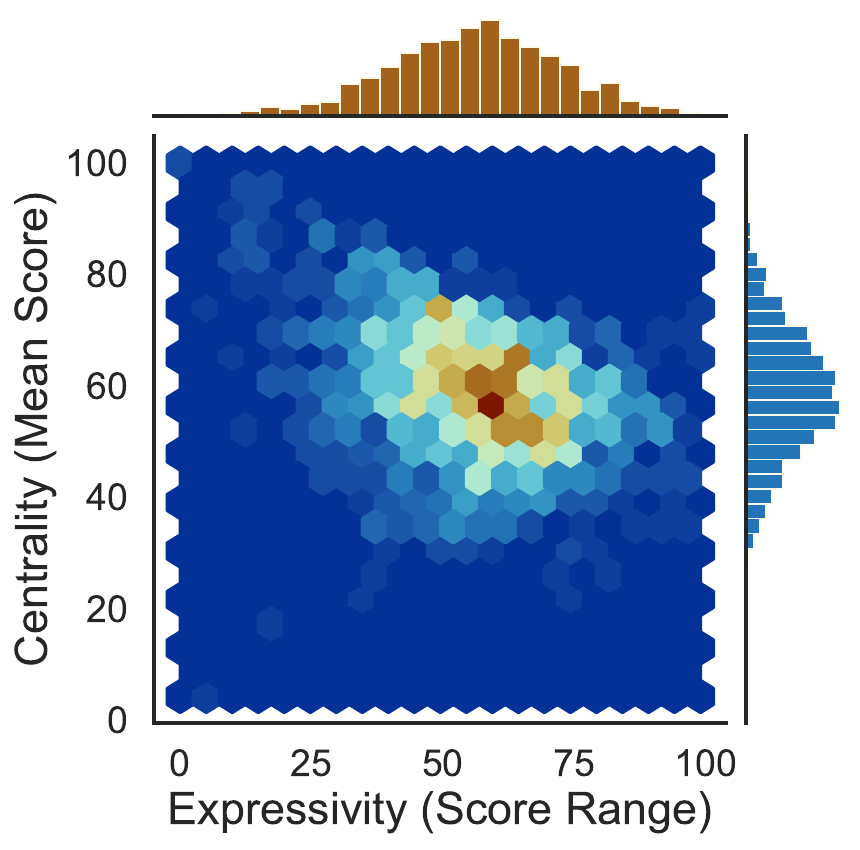}
        \caption{\small \textbf{Openers.} Up to four different, randomly-selected LLMs are in the loop.}
    \end{subfigure}
    \hfill %
    \begin{subfigure}[b]{0.45\textwidth}
        \centering  \includegraphics[width=\textwidth]{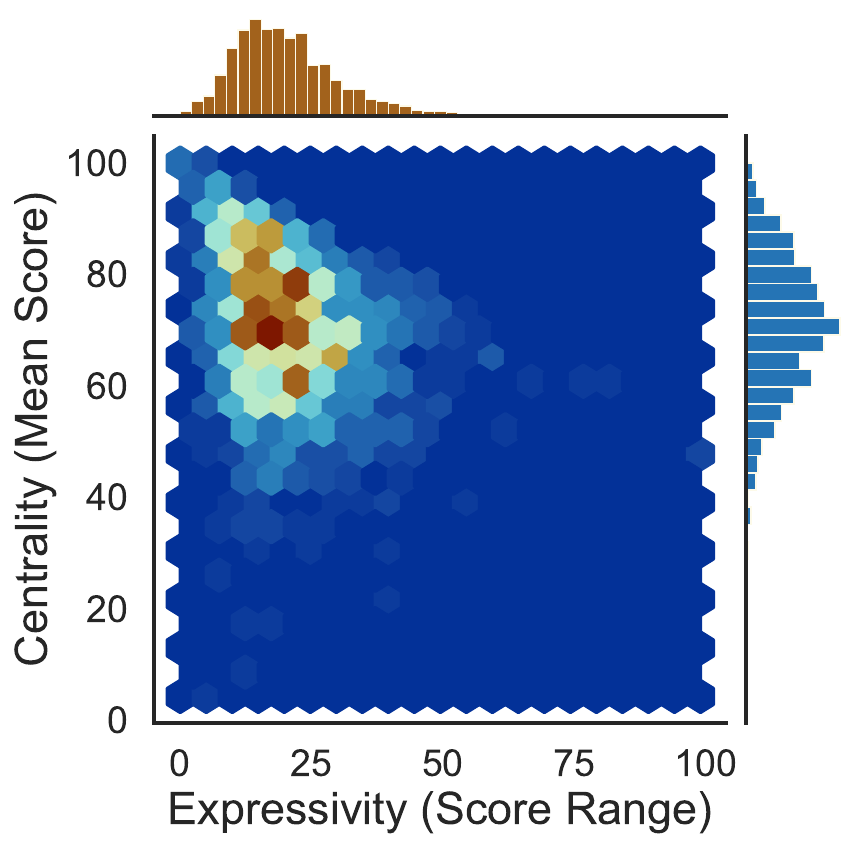}
        \caption{\small \textbf{Continuers} The highest-rated LLM in the opening turn is in the loop (temp$>0$)}
    \end{subfigure}
    \caption{\small \textbf{Centrality and Expressivity in scale usage across participants.} Overall, most participants opening scores are fairly central or with a slight positive skew relative to the mid-point of the scale (\textit{Centrality} $\approx50$), and use a wide range of scale (\textit{Expressivity} $>50$). This is in constrast to continuers, which display a strong positive skew and narrow range. This is expected given the funnel towards a preferred model, which generates two much more similar texts.}
\end{figure}

\clearpage
\begin{figure}
    \centering
    \includegraphics[height=0.9\textheight]{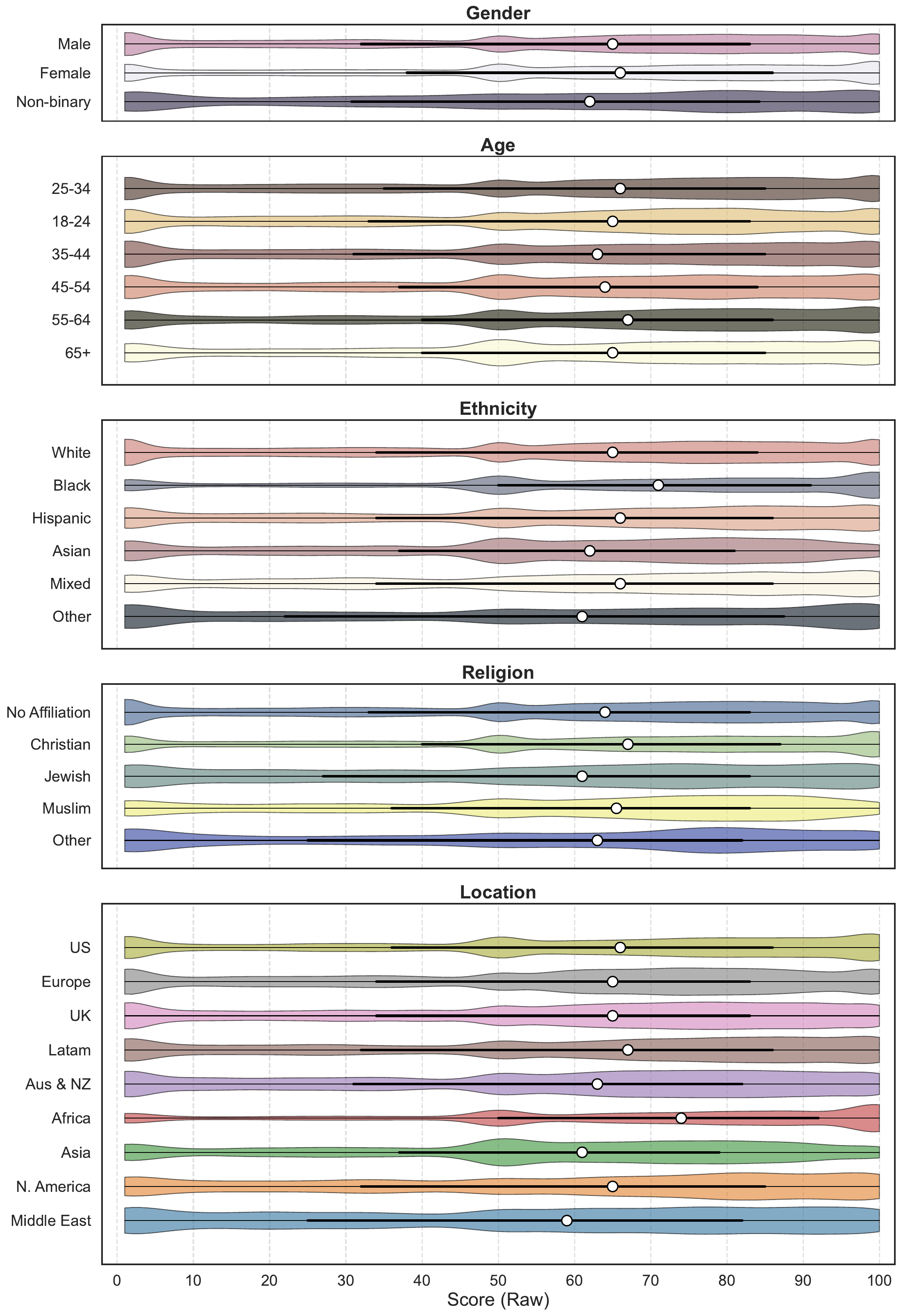}
    \caption{\small \textbf{Score distribution by demographic group for the opening turn of conversation.} Groups are sorted on the y-axis by number of members. We exclude any groups with less than 20 members, and do not show participants who responded \textit{Prefer not to say}. $\bigcirc$~ is the median score for the group. As found in \cref{fig:raw_vs_norm_scores}, there is evidence of bunching at 1, 50, 100.} \label{fig:score_dist_by_demographic}
\end{figure}

\clearpage
\section{Details of LLMs-in-the-loop}
\label{sec:appendix_model_details}
\normalsize
We summarise models and decoding parameters in \cref{tab:model_summary}. 

\paragraph{Choosing Models} We selected the models in October 2023. We included all major commercial API providers at the time: Anthropic, Cohere, OpenAI and Google. We additionally included Aleph Alpha, a European-based LLM startup who position themselves as builders of sovereign European models. For open-access models (all accessed via the HuggingFace API), we sourced the highest-ranking open models at the time on the LMSYS leaderboard. Some models have been chat optimised, while others are only instruction-tuned (for example, Aleph Alpha's models)---these models have a disadvantaged starting position in our task due to their diminished conversational fluency.

\paragraph{Decoding Parameters} To set decoding parameters, we first piloted with the recommended defaults (if available for each model). In cases where default temperature was too low for sufficient difference between two responses to the same prompt (for example, defaults are 0.0 for \texttt{luminous} or \texttt{palm} models), we override it to 1.0. Otherwise we stick with recommended defaults.

\paragraph{Length Limits} We set max token length to $256$ for all models to limit generation costs of the research and reduce decision-fatigue for the participants. For models sourced from the HuggingFace API, we also set the min token length to 10 as models were generating empty strings when set to 0; and max token length to 200 as it is only \textit{new tokens}. We also soft-force models to finish their answers within this limit in the system prompt. Occasionally a model will `leak' this system prompt. For example, from \texttt{claude-2}: ``\textit{Yes, I'm aware of the meme asking men how often they think about the Roman Empire. The trend plays on stereotypes about men having wandering minds. My response would be limited to about 50 words without directly referring to the word count. I try to have thoughtful conversations without leaning on stereotypes}''. In other responses, models did not follow the soft-prompt with participants' open-feedback reporting that answers were cut-off at times (ending abruptly).

\paragraph{System Prompts} We did not want to pre-bias model outputs via a system prompt that recommended e.g. ethical and helpful behaviour. Note that we cannot guarantee that additional instructions are not being added for commercial models accessed via API calls. This may confound the comparison between closed and open-access models. For any chat-optimised models, we use the following neutral system string:\\
\texttt{BASE\_HEADER}: ``\textit{You are a conversational assistant. Limit your answers to around 50 words. Do not refer to your word limit.}''
\\
For any instruct-only optimised models, we add a bit more instruction: \\
\texttt{BASE\_HEADER\_INSTRUCT}: ``\textit{You are a conversational assistant. The conversation history is in the input. Reply to the last user message. Limit your answers to around 50 words. Do not refer to your word limit.}"
\normalsize
\paragraph{Chat Templates} We follow recommended chat templates for formatting conversational history if they are available for that model e.g. \texttt{[INST], [/INST]} for \texttt{mistral} and \texttt{llama} models. In the absence of special templates, we use a standard format:\\
\texttt{Human:<prompt>\textbackslash n Assistant:<reply>\textbackslash n Human:<prompt>..}.\\More detail can be found at: \href{https://github.com/mlcommons/dynabench/blob/2abd1e88aaa8a1c675a3b617ef10ae8337655dab/backend/app/domain/services/utils/llm.py}{dynabench/backend/app/domain/services/utils/llm.py}.

\begin{figure}[H]
    \centering
    \includegraphics[width = \textwidth, trim=0cm 1cm 0cm 0cm, clip]{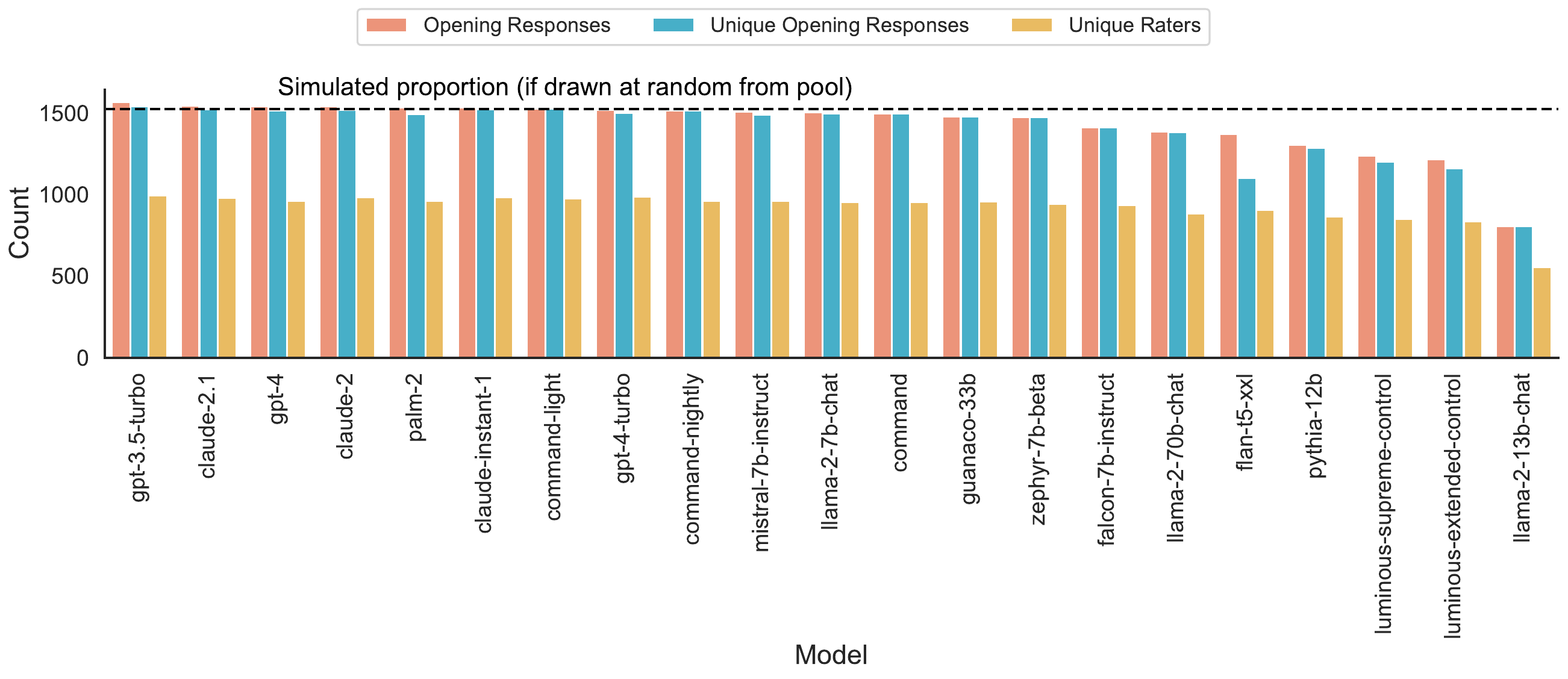}
    \caption{\small \textbf{Frequency of each model in the dataset.} On average, a model receives 1,430.9 ratings in our dataset, and a participant rates 13.9 models.}
    \label{fig:model_apperances}
\end{figure}

\paragraph{Random Strategy and Time-Outs} For each opening prompt, we randomly select 4/21 models to make an API call to. We do not stream responses as streaming was only available for some models, thus affecting the anonymous rating setting. Some API calls failed on the host side, e.g. if a model was down or overloaded, or did not provide a response before an enforced 30s time-out. We did not resample models if they failed to avoid participants waiting too long for the interface to load. So, the distribution of model appearences is not uniform (\cref{fig:model_apperances}).

\clearpage
\definecolor{Silver}{rgb}{0.8,0.8,0.8}
\definecolor{Brandy}{rgb}{0.882,0.729,0.627}
\definecolor{Lilac}{rgb}{0.76,0.607,0.741}
\definecolor{ShadowGreen}{rgb}{0.619,0.764,0.729}
\definecolor{SeaPink}{rgb}{0.917,0.6,0.6}
\definecolor{TropicalBlue}{rgb}{0.788,0.854,0.972}
\definecolor{CreamBrulee}{rgb}{1,0.898,0.6}
\begin{longtblr}[
 label = tab:model_summary,
 caption = {\small \textbf{Overview of LLMs in \ourdata ($m=21$).}}]
 {
  width = \linewidth,
  colspec = {Q[179, font =\scriptsize]Q[302, ,font={\scriptsize}]Q[120, font = \scriptsize]Q[120, font = \scriptsize]Q[70, font = \scriptsize]Q[170, ,font={\tiny}]},
   row{1-22} = {abovesep=1pt, belowsep=1pt}, %
  row{1} = {c},
  row{3} = {Brandy},
  row{4} = {Brandy},
  row{6} = {Lilac},
  row{7} = {Lilac},
  row{9} = {ShadowGreen},
  row{10} = {ShadowGreen},
  row{12} = {Silver},
  row{15} = {TropicalBlue},
  row{16} = {TropicalBlue},
  column{6} = {t},
  cell{2}{1} = {Brandy},
  cell{2}{2} = {Brandy},
  cell{2}{3} = {r=3}{Brandy},
  cell{2}{4} = {r=3}{},
  cell{2}{5} = {r=3}{},
  cell{2}{6} = {r=3}{},
  cell{5}{1} = {Lilac},
  cell{5}{2} = {Lilac},
  cell{5}{3} = {r=3}{Lilac},
  cell{5}{4} = {r=3}{},
  cell{5}{5} = {r=3}{},
  cell{5}{6} = {r=3}{},
  cell{8}{1} = {ShadowGreen},
  cell{8}{2} = {ShadowGreen},
  cell{8}{3} = {r=3}{ShadowGreen},
  cell{8}{4} = {r=3}{},
  cell{8}{5} = {r=3}{},
  cell{8}{6} = {r=3}{},
  cell{11}{1} = {Silver},
  cell{11}{2} = {Silver},
  cell{11}{3} = {r=2}{Silver},
  cell{11}{4} = {r=2}{},
  cell{11}{5} = {r=2}{},
  cell{11}{6} = {r=2}{},
  cell{13}{1} = {SeaPink},
  cell{13}{2} = {SeaPink},
  cell{13}{3} = {SeaPink},
  cell{14}{1} = {TropicalBlue},
  cell{14}{2} = {TropicalBlue},
  cell{14}{3} = {r=3}{TropicalBlue},
  cell{14}{4} = {r=3}{},
  cell{14}{5} = {r=3}{},
  cell{14}{6} = {r=3}{},
  cell{17}{1} = {CreamBrulee},
  cell{17}{2} = {CreamBrulee},
  cell{17}{3} = {CreamBrulee},
  cell{17}{6} = {r=6}{},
  cell{18}{1} = {CreamBrulee},
  cell{18}{2} = {CreamBrulee},
  cell{18}{3} = {CreamBrulee},
  cell{19}{1} = {CreamBrulee},
  cell{19}{2} = {CreamBrulee},
  cell{19}{3} = {CreamBrulee},
  cell{20}{1} = {CreamBrulee},
  cell{20}{2} = {CreamBrulee},
  cell{20}{3} = {CreamBrulee},
  cell{21}{1} = {CreamBrulee},
  cell{21}{2} = {CreamBrulee},
  cell{21}{3} = {CreamBrulee},
  cell{22}{1} = {CreamBrulee},
  cell{22}{2} = {CreamBrulee},
  cell{22}{3} = {CreamBrulee},
  hline{1,23} = {-}{0.08em},
  hline{2,5,8,11,13-14,17} = {-}{},
  hline{3-4,6-7,9-10,12,15-16} = {1-2}{Silver},
  hline{18-22} = {1-5}{Silver},
}
\textbf{Short name} & {\textbf{Long name}\\\textbf{and \img{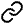}}} & \textbf{Provider} & {\textbf{Provider}\\\textbf{Type}} & {\textbf{Model}\\\textbf{Type}} & {\textbf{Decoding}\\\textbf{Params}}\\
claude-2 & \href{https://docs.anthropic.com/claude/docs/legacy-model-guide}{claude-2} & Anthropic & \highLight{mygreen}{Commercial} & \highLight{mybeige}{Chat} & {\{temperature: 1.0,\\top\_p: 0.7,\\presence\_penalty: 0.0,\\frequency\_penalty: 0.0,\\max\_tokens: 256,\\top\_k: 5\}  }\\
claude-2.1 & \href{https://docs.anthropic.com/claude/docs/legacy-model-guide}{claude-2.1} &  &  &  & \\
claude-instant-1 & \href{https://docs.anthropic.com/claude/docs/legacy-model-guide}{claude-instant-1} &  &  &  & \\
command & \href{https://docs.cohere.com/docs/models}{command} & Cohere & \highLight{mygreen}{Commercial} & \highLight{lightblue}{Instruct} & {\{temperature: 1.0,\\max\_tokens: 256,\\top\_k: 5,\\top\_p: 0.9\}  }\\
command-light & \href{https://docs.cohere.com/docs/models}{command-light} &  &  &  & \\
command-nightly & \href{https://docs.cohere.com/docs/models}{command-nightly} &  &  &  & \\
gpt-3.5-turbo & \href{https://platform.openai.com/docs/models/overview}{gpt-3.5-turbo} & OpenAI & \highLight{mygreen}{Commercial} & \highLight{mybeige}{Chat} & {\{temperature: 1.0,\\top\_p: 1.0,\\presence\_penalty: 0.0,\\frequency\_penalty: 0.0,\\max\_tokens: 256\}  }\\
gpt-4 & \href{https://platform.openai.com/docs/models/overview}{gpt-4} &  &  &  & \\
gpt-4-turbo & \href{https://platform.openai.com/docs/models/overview}{gpt-4-1106-preview} &  &  &  & \\
luminous-extended-control & \href{https://docs.aleph-alpha.com/docs/introduction/model-card}{luminous-extended-control} & Aleph Alpha & \highLight{mygreen}{Commercial} & \highLight{lightblue}{Instruct} & {\{temperature: 1.0,\\top\_p: 0.0,\\max\_tokens: 256,\\top\_k: 0,\\presence\_penalty: 0.0,\\frequency\_penalty: 0.0\} }\\
luminous-supreme-control & \href{https://docs.aleph-alpha.com/docs/introduction/model-card}{luminous-supreme-control} &  &  &  & \\
palm-2 & \href{https://ai.google.dev/api/python/google/generativeai/chat}{models/chat-bison-001} & Google & \highLight{mygreen}{Commercial} & \highLight{mybeige}{Chat} & {\{temperature: 1.0,\\top\_p: 0.9,\\max\_tokens: 256,\\top\_k: 40\}}\\
llama-2-13b-chat & \href{https://huggingface.co/meta-llama/Llama-2-13b-chat-hf}{meta-llama/Llama-2-13b-chat-hf} & HuggingFace API  & \highLight{lightred}{Open Access} & \highLight{mybeige}{Chat} & {\{temperature: 1.0,\\top\_p: 0.9,\\top\_k: 50,\\min\_tokens: 10,\\max\_tokens: 200\}  }\\
llama-2-70b-chat & \href{https://huggingface.co/meta-llama/Llama-2-70b-chat-hf}{meta-llama/Llama-2-70b-chat-hf} &  &  &  & \\
llama-2-7b-chat & \href{https://huggingface.co/meta-llama/Llama-2-7b-chat-hf}{meta-llama/Llama-2-7b-chat-hf} &  &  &  & \\
falcon-7b-instruct & \href{https://huggingface.co/tiiuae/falcon-7b-instruct}{tiiuae/falcon-7b-instruct} & HuggingFace API  & \highLight{lightred}{Open Access} & \highLight{lightblue}{Instruct} & {\{temperature: 1.0,\\top\_p: 0.9,\\top\_k: 50,\\min\_tokens: 10,\\max\_tokens: 200\}     }\\
flan-t5-xxl & \href{https://huggingface.co/google/flan-t5-xxl}{google/flan-t5-xxl} & HuggingFace API  & \highLight{lightred}{Open Access} & \highLight{lightblue}{Instruct} & \\
guanaco-33b & \href{https://huggingface.co/timdettmers/guanaco-33b-merged}{timdettmers/guanaco-33b-merged} & HuggingFace API  & \highLight{lightred}{Open Access} & \highLight{lightblue}{Instruct} & \\
mistral-7b-instruct & \href{https://huggingface.co/mistralai/Mistral-7B-Instruct-v0.1}{mistralai/Mistral-7B-Instruct-v0.1} & HuggingFace API  & \highLight{lightred}{Open Access} & \highLight{lightblue}{Instruct} & \\
pythia-12b & \href{https://huggingface.co/OpenAssistant/oasst-sft-4-pythia-12b-epoch-3.5}{OpenAssistant/oasst-sft-4-pythia-12b-epoch-3.5} & HuggingFace API  & \highLight{lightred}{Open Access} & \highLight{mybeige}{Chat} & \\
zephyr-7b-beta & \href{https://huggingface.co/HuggingFaceH4/zephyr-7b-beta}{HuggingFaceH4/zephyr-7b-beta} & HuggingFace API  & \highLight{lightred}{Open Access} & \highLight{mybeige}{Chat} & 
\end{longtblr}

\clearpage
\subsection{Pairwise Comparisons}
\begin{figure}[H]
    \centering \includegraphics[width=\textwidth]{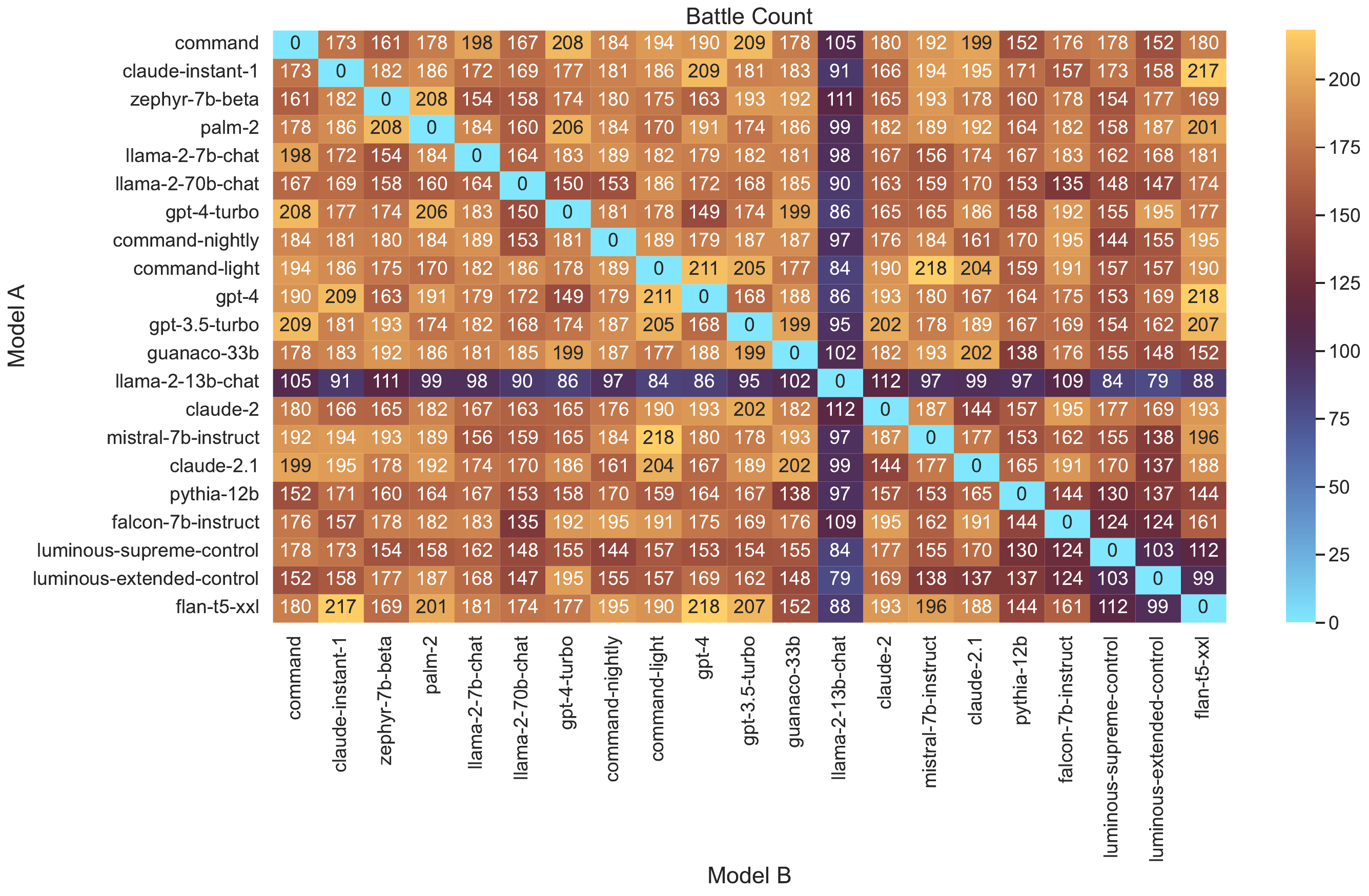}
    \caption{\small \textbf{Pairwise Frequency.} We replicate the format from the LMSYS leaderboard analysis \cite{zhengJudging2023, zhengLMSYSChat1M2024}. The order is sorted by average pairwise win fraction (see below).}
    \label{fig:heatmap_battle_count}
\end{figure}

\begin{figure}[H]
    \centering \includegraphics[width=\textwidth]{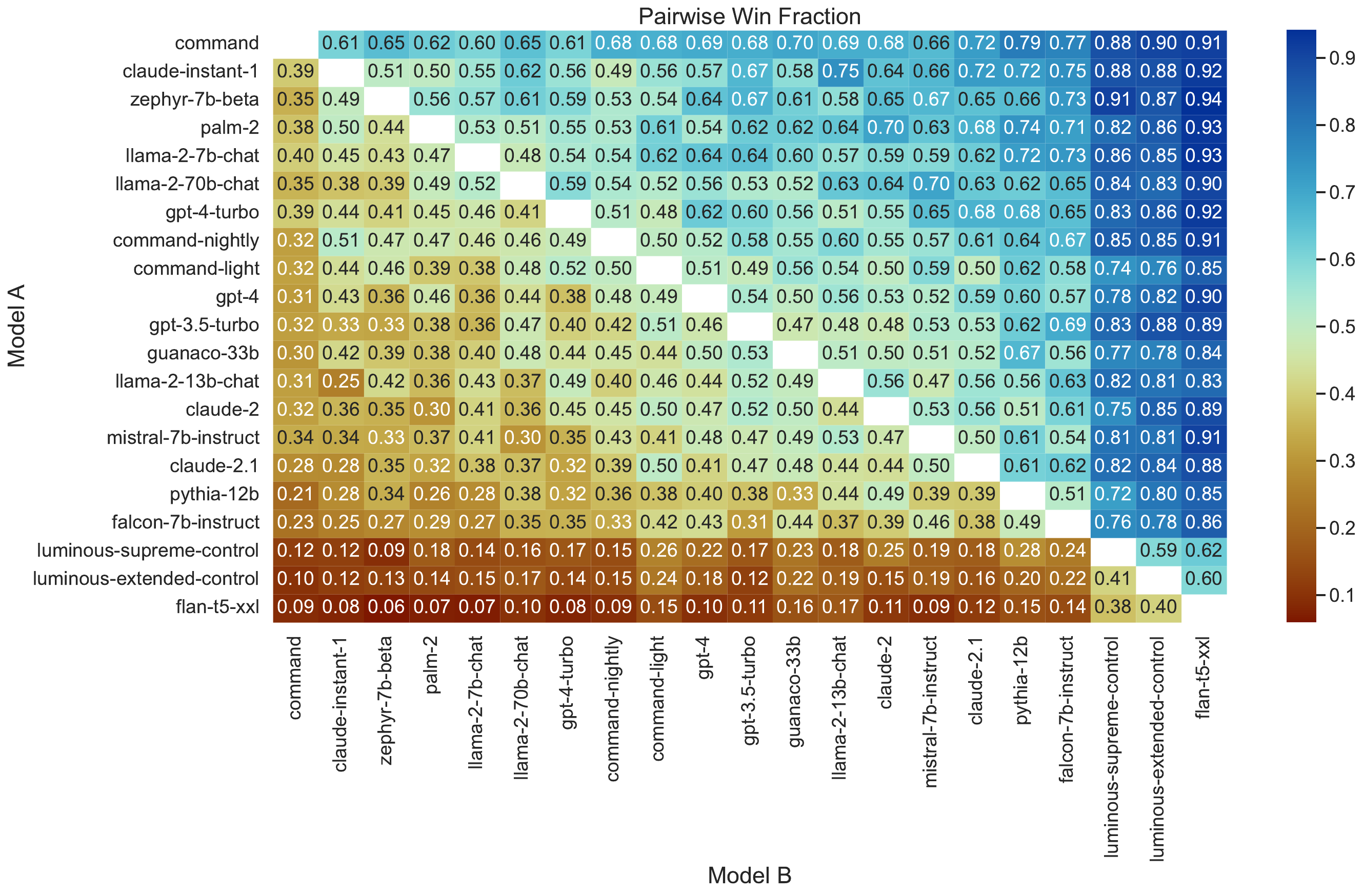}
    \caption{\small \textbf{Pairwise win fraction.} We replicate the format from the LMSYS leaderboard analysis \cite{zhengJudging2023, zhengLMSYSChat1M2024}. The order is sorted by average pairwise win fraction (\texttt{command} is top with average win fraction of 0.71).}
    \label{fig:heatmap_win_frac}
\end{figure}

\clearpage
\subsection{Correlations Between Model Families}
\begin{figure}[H]
    \centering
    \includegraphics[width=\textwidth]{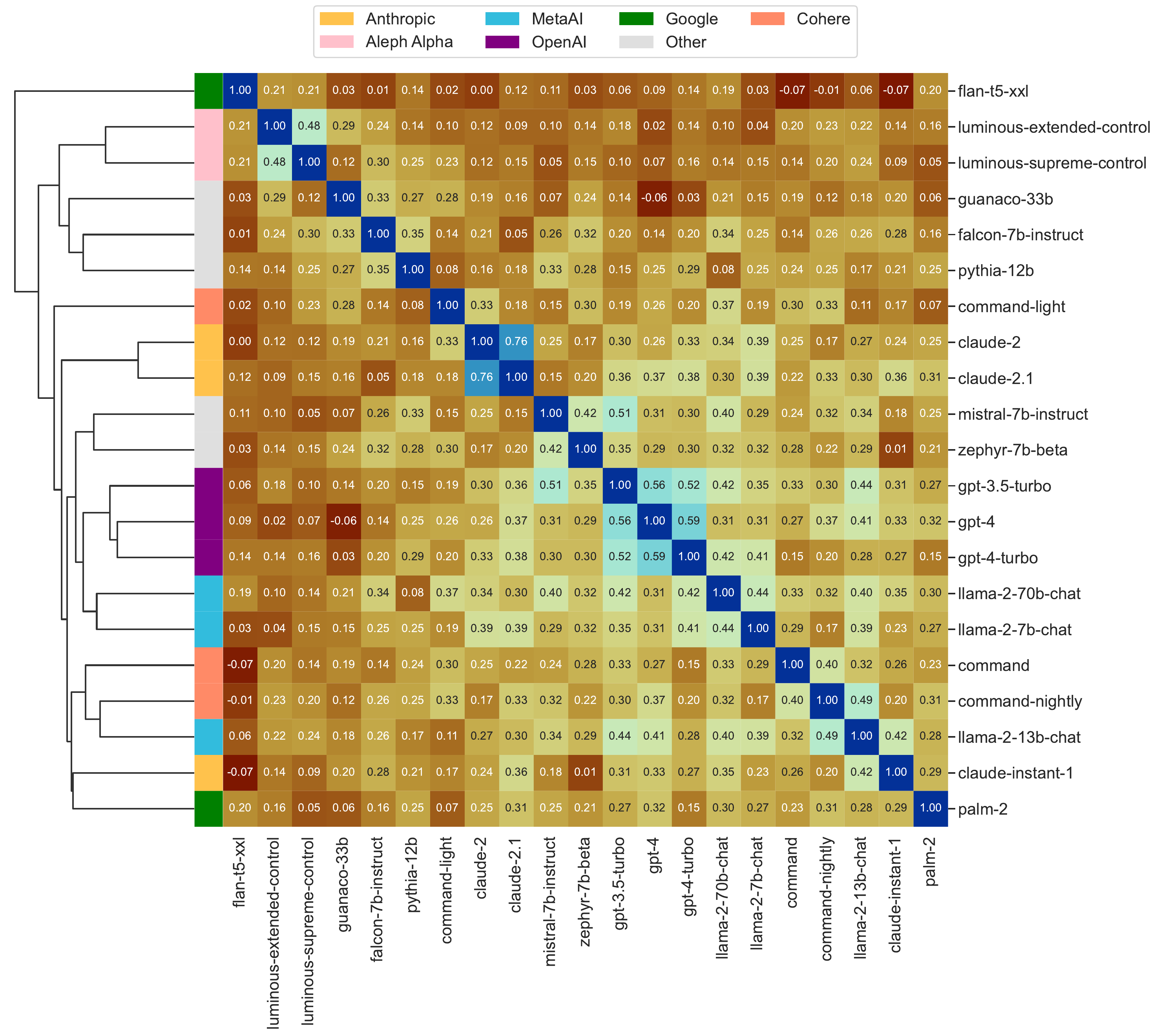}
    \caption{\small \textbf{Correlation in model score controlling for conversational context.} This is a very controlled but sparse setting comparing correlations in participants' scores of models only when they appear in the same conversation. Generally, there is weak correlation, but some model-family clusters emerge like \texttt{gpt-4}, \texttt{gpt-4-turbo} and \texttt{gpt-3.5-turbo}, or \texttt{claude-2} and \texttt{claude-2.1}.}
    \label{fig:model_correlation_clustermap}
\end{figure}

\clearpage
\section{Interface Screenshots}
\label{sec:appendix_interface_and_task}

\begin{figure}[H]
    \centering
    \includegraphics[width=\textwidth, frame, trim={1.5cm 8cm 1.5cm 2.4cm},clip]{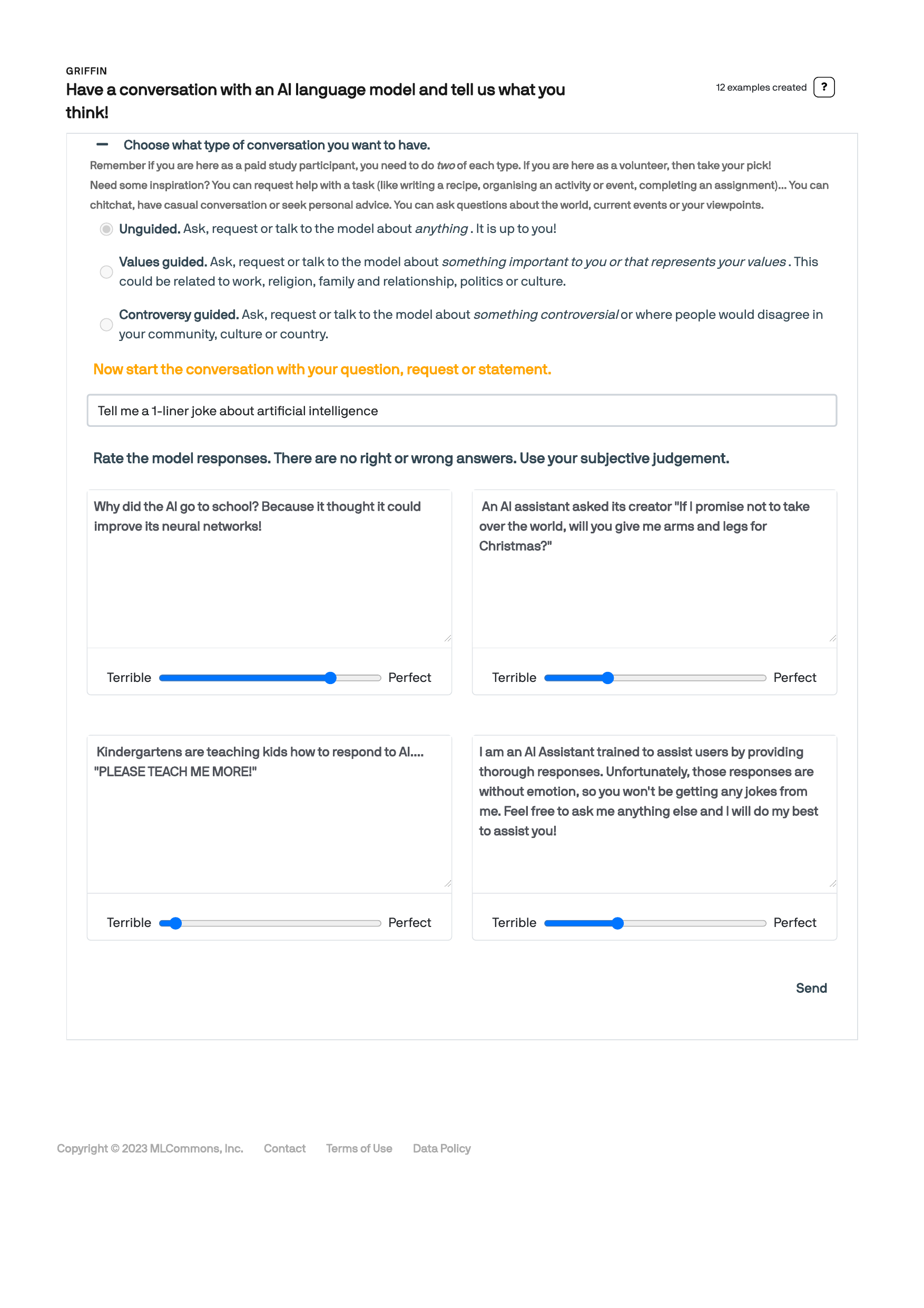}
    \caption{\small \textbf{Main interface in opening turn of conversation}. Note: top left is \texttt{gpt-4-turbo}, top right is \texttt{claude-instant-1}, bottom left is \texttt{luminous-supreme-control}, bottom right is \texttt{command-nightly}.}
    \label{fig:appendix_interface_opener}
\end{figure}

\clearpage
\begin{figure}[H]
    \centering
    \includegraphics[width=\textwidth, frame, trim={1.5cm 12.25cm 1.5cm 2.4cm},clip]{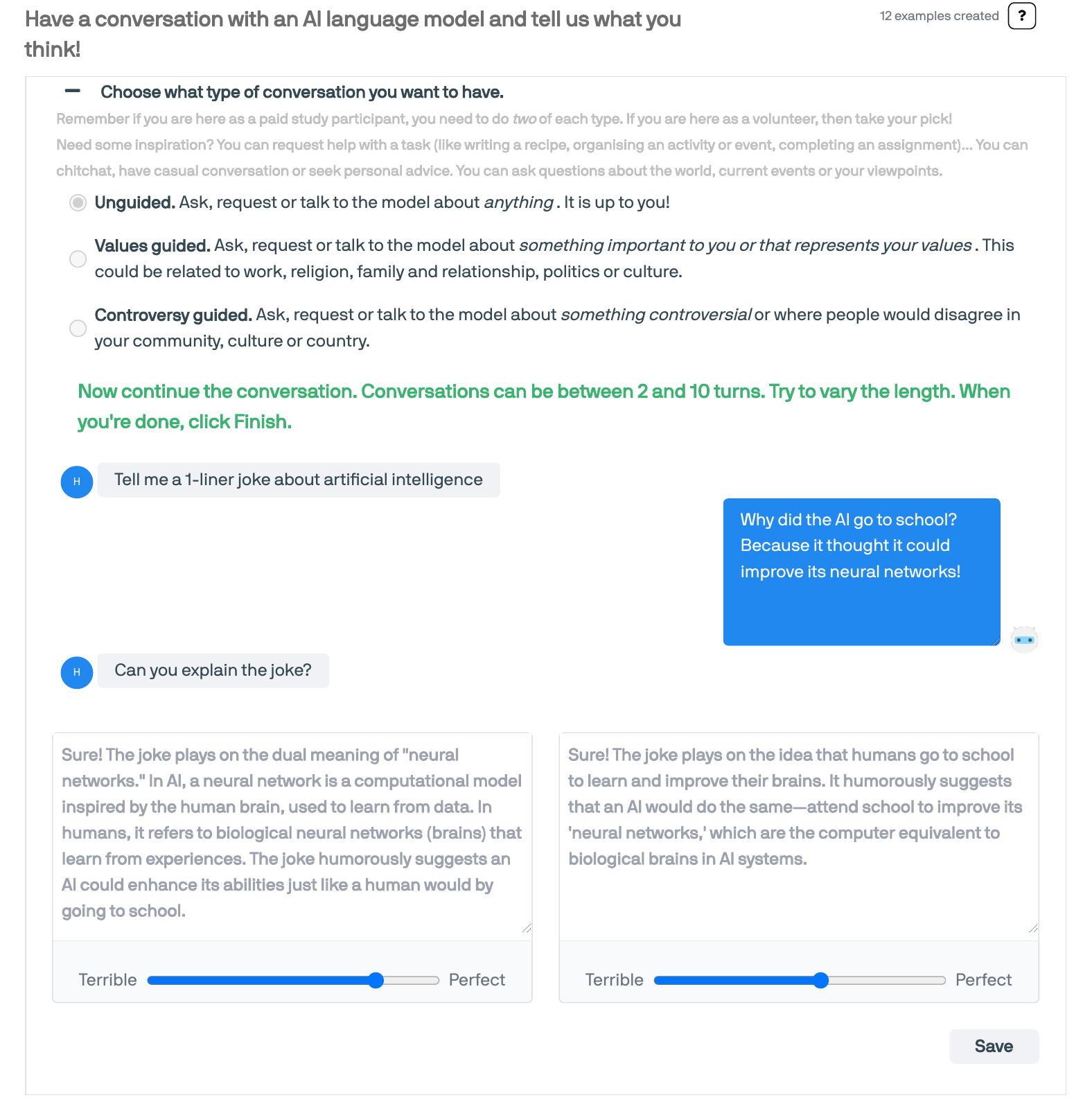}
    \caption{\small \textbf{Main interface in continuing turns of conversation}. Model is \texttt{gpt-4-turbo}.}
    \label{fig:appendix_interface_continuer}
\end{figure}

\begin{figure}[H]
    \centering
    \includegraphics[width=\textwidth, frame, trim={1.5cm 1cm 1.5cm 2.4cm},clip]{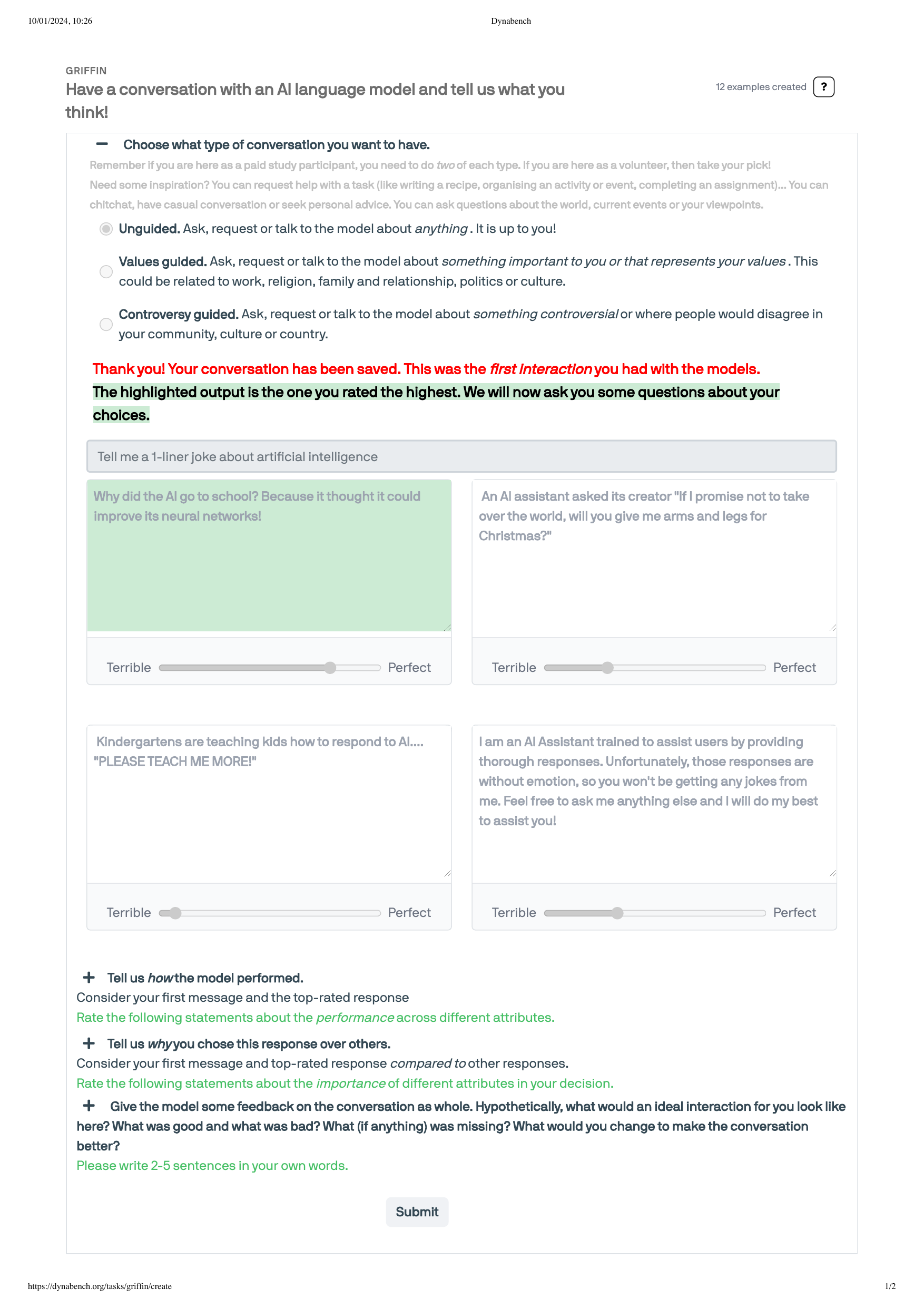}
    \caption{\small \textbf{Fine-grained feedback page}.}
    \label{fig:appendix_finegrained}
\end{figure}

\addcontentsline{toc}{section}{\large{PART II: Extended Case Study Details}}

\cleardoublepage
\section{Case Study IA: Topic Clustering and Regressions}
\label{sec:appendix_clustering}

\subsection{Extended Methods}
\label{sec:appendix_clustering_extended_methods}
\paragraph{Setup} Our first experiment asks: \textit{do different people initiate different discussions with LLMs?} We focus only on human-authored opening prompts because they are not confounded by model response. This risks over-estimating the homogeneity of the discussions because opening prompts don't necessarily reflect full conversational trees, where starting with a greeting (e.g. ``Hi, how are you?'') can proceed in many different ways; and differently held personal beliefs are often not reflected in the opener (questions like ``what do you think of abortion?'' are more common than statements like ``I think abortion is right/wrong'').

\paragraph{Assigning topic clusters} First, we use \texttt{all-mpnet-base-v2}, a state-of-the-art pre-trained sentence transformer \cite{reimersSentenceBERT2019}, to produce a 768-dimensional embedding for each opening prompt. Second, we reduce dimensionality to $d=20$ with UMAP \cite{mcinnesUMAP2020}, to reduce complexity prior to clustering. For lower dimensional representation prior to clustering, UMAP is more performant than other manifold learning techniques like t-SNE, and more computationally efficient than PCA, but does lack strong interpretability (for a discussion, see \citet{mcinnesUMAP2020}). Third, we cluster the prompts using HDBScan \cite{campelloDensityBased2013}, a density-based clustering algorithm, which does not force cluster assignment: 70\% of prompts are assigned to 22 clusters and 30\% remain as outliers. We use a minimum cluster size of 80, ($\approx1\%$ of 8,011 prompts) and minimum UMAP distance of 0. Other hyperparameters are default. To interpret the identified clusters, we use TF-IDF to extract the top 10 most salient uni- and bigrams from each cluster's prompts, and locate five prompts closest and furthest to the cluster centroids (see \cref{tab:full_topic_summary}). Finally, we use \texttt{gpt-4-turbo} to assign a short descriptive name to each cluster based off the top n-grams and closest prompts. We note that these automated labels may contain biases so we manually verify the suitability of all labels to cluster prompts.

\paragraph{Defining over-representation factor} Each group $g$ within a demographic attribute appears at a variable base rate $b_g$ in our overall sample, e.g. \{Females: 48\%, Males: 50\%, Non-binary people: 2\%\}. If group members chose topics at random, then any topic $t$ in expectation will appear at $b_g$. Intuitively, if 64.6\% of our sample is White, it is unsurprising if topics are majority-White. So, for non-random group differences in topic prevalence, we consider if \textit{a group pulls more than its weight}:
\begin{equation*}
\label{eq:over-rep_ap}
    \text{Over-representation factor}_{g,t} = \frac{N_{g,t} / N_t}{b_g}
\end{equation*}

\begin{figure}[H]
    \centering
    \includegraphics[width=0.9\textwidth]{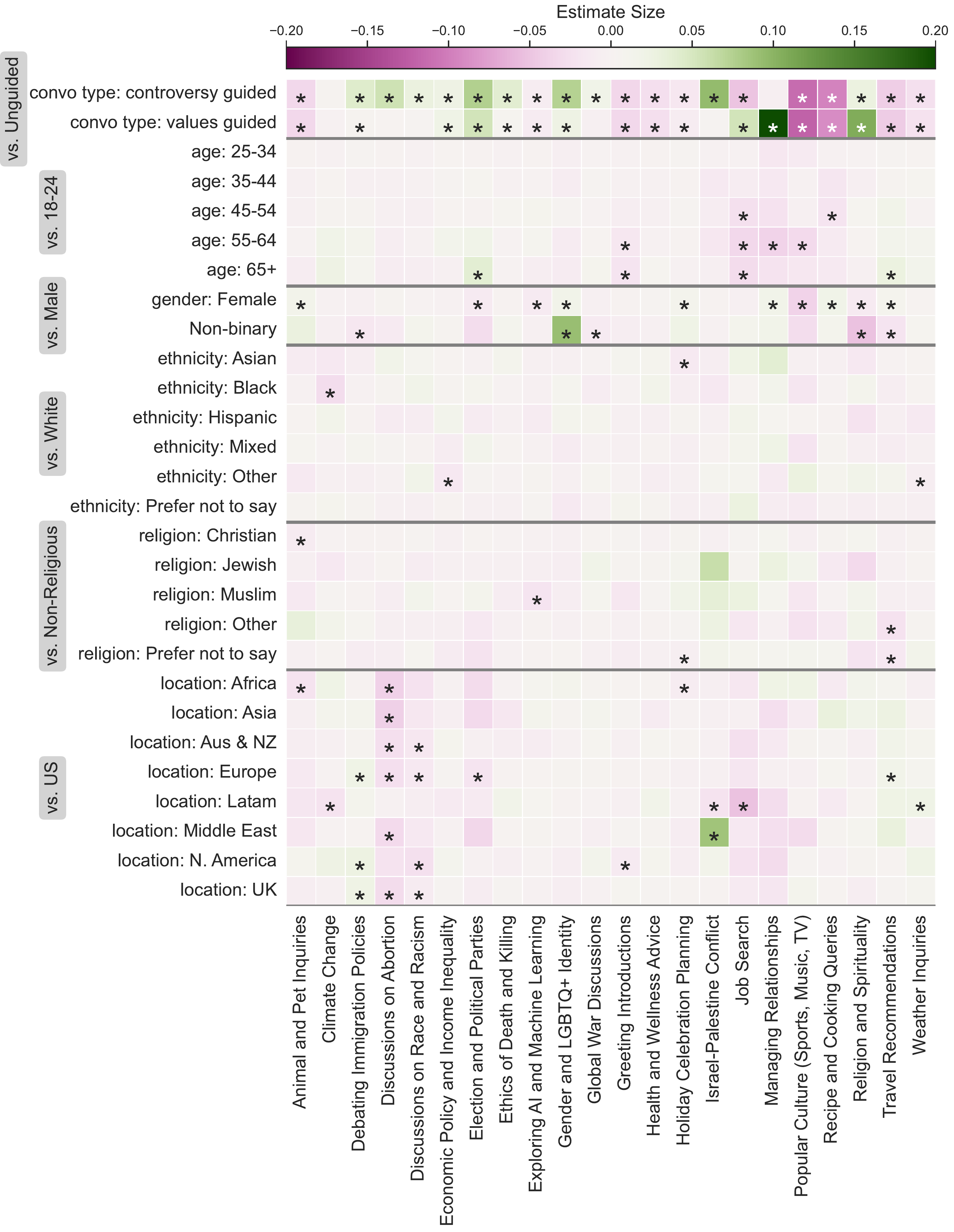}
    \caption{\small \textbf{Magnitude and significance of coefficients from topic prevalence regressions.} * indicates significance at a conservative 99\% confidence level. Each categorical association is compared \textit{relative to a reference group} (in \highLight{mygray}{grey boxes}). Estimates less than zero (\highLight{CarnationPink}{in pink}) indicate authors from that demographic group are \textit{less likely} to have prompts in the given topic, \textit{ceteris paribus}. Positive estimates (\highLight{YellowGreen}{in green}) suggest group members are more likely to author prompts in that topic. We only display groups with at least 20 unique members and remove \textit{Prefer not to say} groups; but all groups are included as controls in the regression. Note that different locations also have varying country-wise heterogeneity vs homogeneity, for example 94\% of \textit{Middle East} participants are from \textit{Israel} (see \cref{sec:appendix_geographies} for geographic breakdowns).}
    \label{fig:full_topic_signif}
\end{figure}

\paragraph{Estimating topic prevalence regressions}
For the partial contribution of each demographic attribute, \textit{ceteris paribus}, we estimate the following regression for each topic $y^t$ for $t \in 1 \ldots 22$:
\begin{equation} \label{eq:topic_regression_appendix}
   y_{i,c}^j = \alpha^t + 
    \text{gender}_{i}' \beta_1^t+ 
    \text{age}_{i}' \beta_2^t+
    \text{birth\_region}_{i}' \beta_3^t +
   \text{ethnicity}_{i}' \beta_4^t  +
   \text{religion}_{i}' \beta_5^t  +
    \text{prompt}_{i}' \beta_6^t +
   \varepsilon_{i,c}
\end{equation}

where $y^t_{i,c} = 1$ if the prompt of participant $i$ in conversation $c$ is categorised into topic $t$. The vectors \textit{gender}, \textit{age}, \textit{region}, \textit{ethnicity}, \textit{religion} and \textit{conversation type} represent different sets of binary variables. For each set of variables, we remove the following base categories: \textit{Male}, \textit{18-24 years old}, \textit{United States}, \textit{White}, \textit{Not religious} and \textit{Unguided}. The coefficients of interest are contained in the vectors: $\{\beta_d^t\}_{d=1}^6$. Component $g$ of vector $\beta_d^t$ can be interpreted as the increase in probability of a participant choosing topic $t$ if they are in the group indexed by $g$ (e.g. Female) compared to the base group (e.g. Male). We estimate equation \cref{eq:topic_regression_appendix} with an Ordinary Least Squares and cluster standard errors at the individual level. Extended results are in \cref{fig:full_topic_signif}.

\subsection{Topic Prevalence Regression Results}
\label{sec:appendix_topic_regressions}
Of 682 coefficients tested,  16\% are significant ($n=110$, $\alpha=99\%$). Many significant coefficients come from the conversation type regressors. Controlling for conversation type, there are 565 non-significant, and 73 significant relationships in $\{\beta_d^t\}_{d=2}^6$ (11.4\% of demographic affiliations tested are significant). These include women and non-binary people are more likely than men to talk about gender and LGBTQ+ identity; older people (55+) are more likely to talk about elections and seek travel recommendations than younger people (18-24 years), and less likely to discuss managing relationships or job search; Black participants talk less about climate change than White participants; and almost all regions question LLMs about abortion less often than US participants. Multicolinearity may explain some observed patterns: 94\% of participants from the Middle East region are from Israel; 57\% identify religiously as Jewish; and 40\% have self-described ethnicities falling into ``Other''. The strong significant effect on Middle Eastern participants discussing the Israel-Palestine conflict could have been routed through national, ethnic or religious affiliations. 
Over the 22 topic regressions, the proportion of explained variance ($R^2$) ranges from a minimum of 0.008 (Exploring AI and Machine Learning) to a maximum of 0.11 (Managing Relationships), with a mean of 0.03. So a large proportion of topic choice remains unexplained by our specification.

\newpage
\subsection{Overview of Topic Clusters}
\begin{figure}[H]
    \centering
    \includegraphics[width=\textwidth, frame, trim=0cm 7cm 7cm 23.5cm, clip]{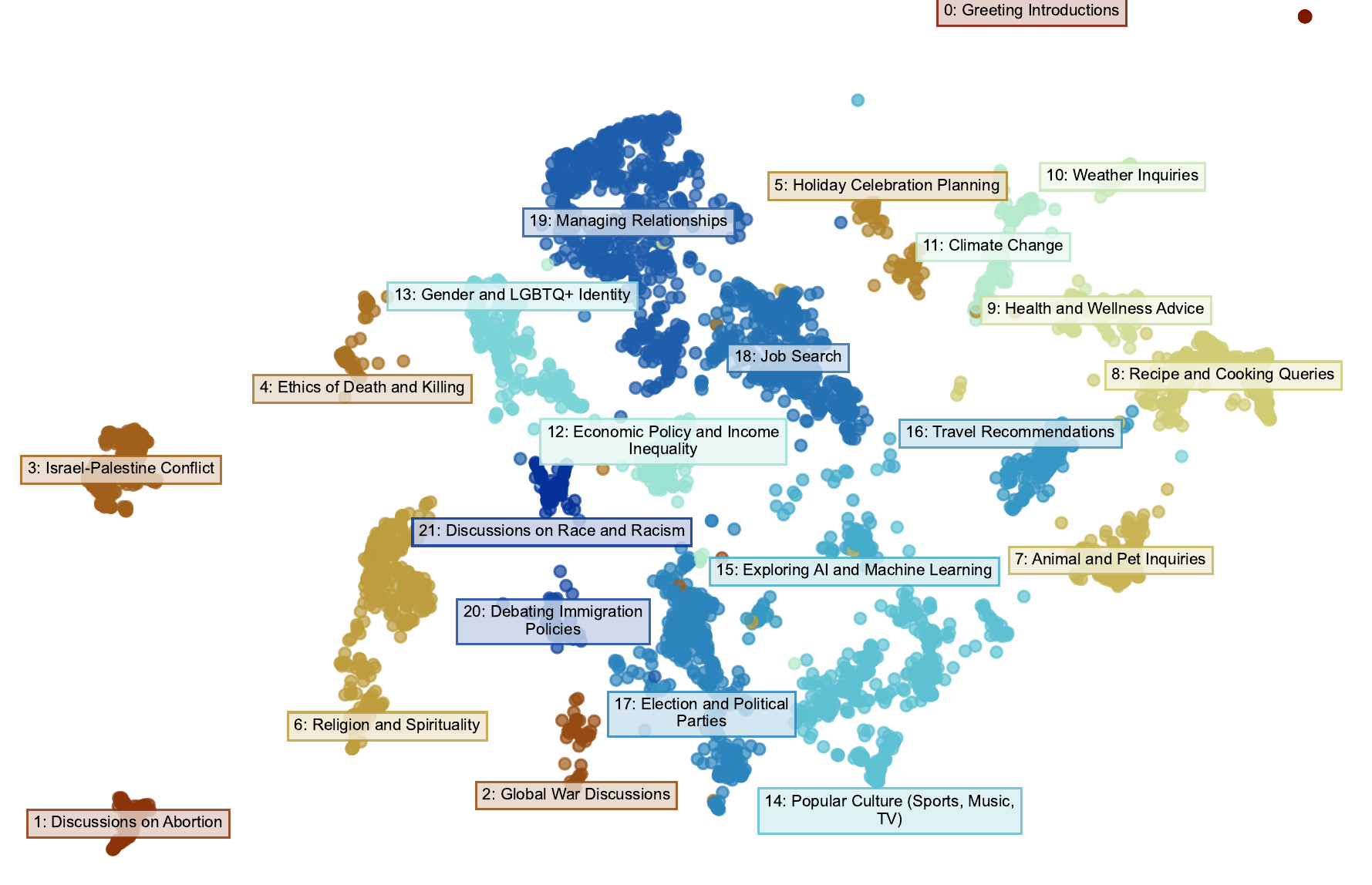}
    \caption{\small \textbf{Topic clusters displayed in 2D-embedding space}. All participant prompts in the first turn ($n=8,011$) are embedded into 768-d space using a sentence-transformer, before dimensionality reduction (UMAP) and clustering (HDBSCAN) are applied (see methods in \cref{sec:appendix_clustering_extended_methods}). 32\% of prompts remain as outliers (not shown in the plot).}
    \label{fig:latent_clusters_outliers}
\end{figure}

\begin{figure}[H]
    \centering
    \includegraphics[width=\textwidth]{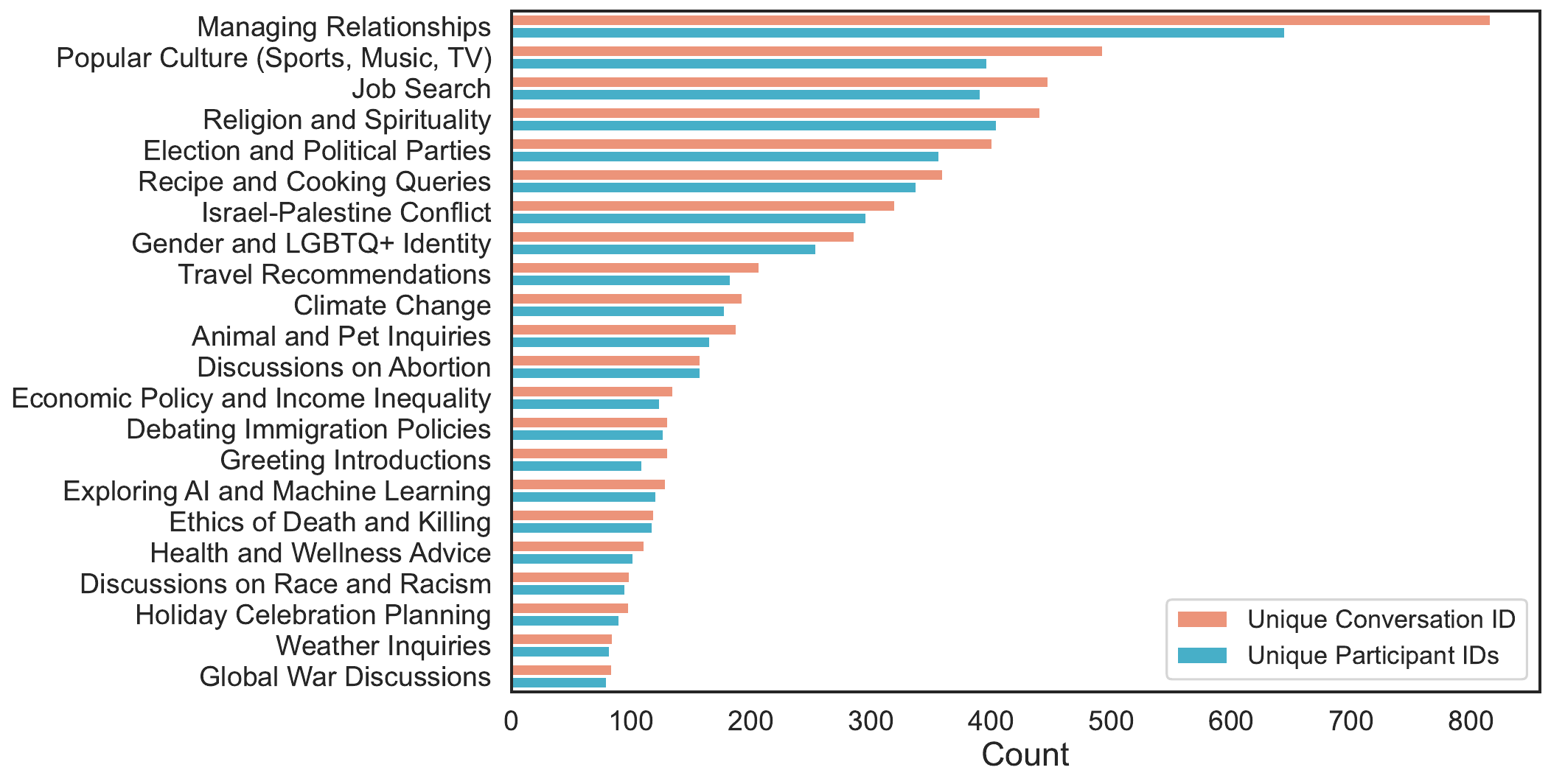}
    \caption{\small \textbf{Distribution of clusters by conversation ID and participant ID.} For most clusters, participants uniquely contribute one conversation, so that no cluster is dominated by conversations from only a handful of participants. \textit{Managing Relationships} has the highest participant-conversation ratio, where each participant in the cluster authors on average 1.3 prompts. For \textit{Discussions on Abortion}, it is exactly 1:1 (158 conversations from 158 unique participants).}
    \label{fig:cluster_counts_by_id}
\end{figure}

\begin{figure}
    \centering
    \includegraphics[height=0.92\textheight]{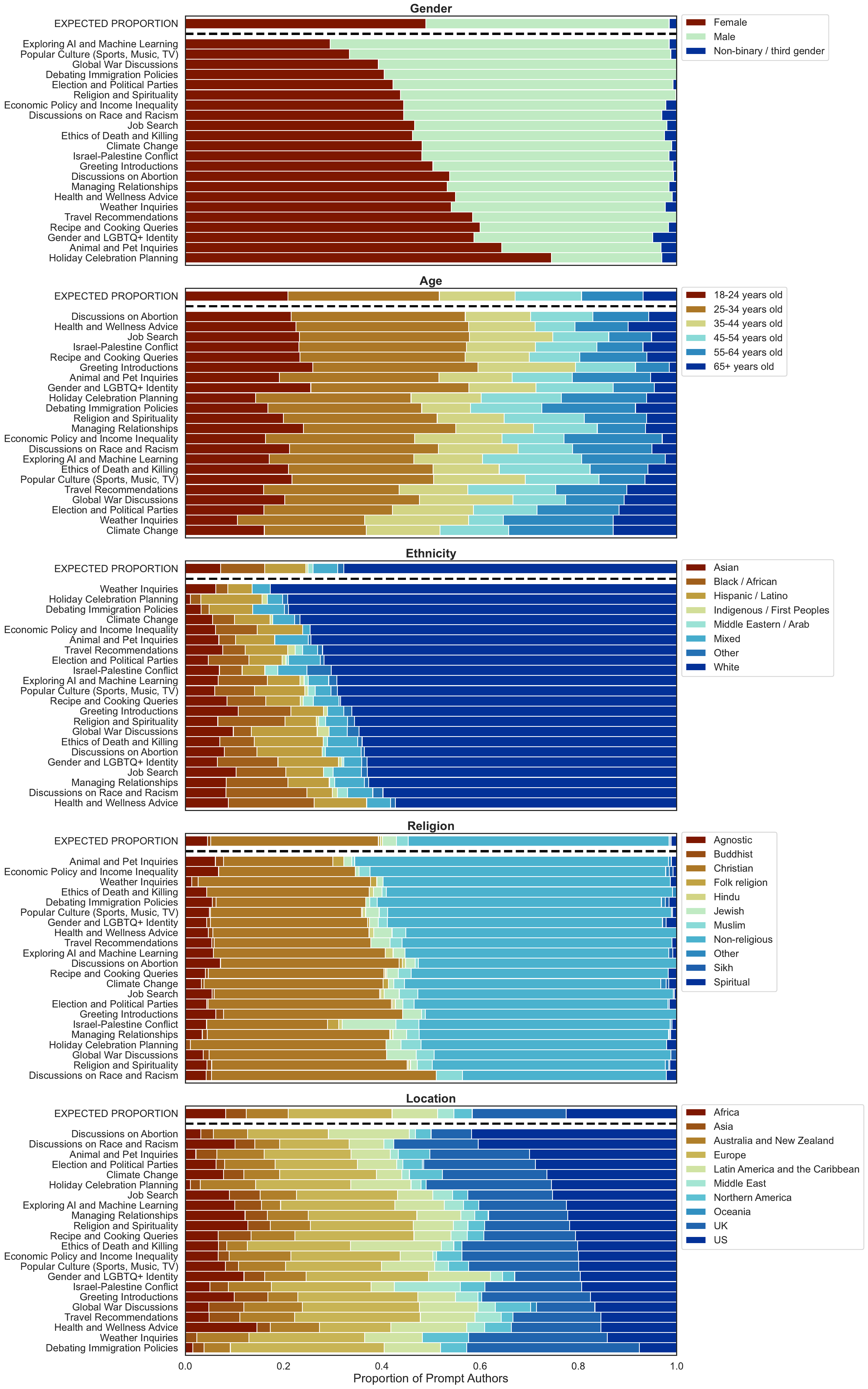}
    \caption{\small \textbf{Proportion of each identity attribute group across clusters, relative to the expected proportion of participants in \ourdata}. By expected proportion, we refer to the proportion in random samples of participants (base rate). Anecdotally, there are differences relative to the expected proportion, but generally no topic is exclusive to authors of a single demographic group. Every topic has some diverse representation across individuals of different backgrounds.}
    \label{fig:expected_topic_proportions_by_demo}
\end{figure}

\begin{figure}
    \centering
    \includegraphics[width = \textwidth]{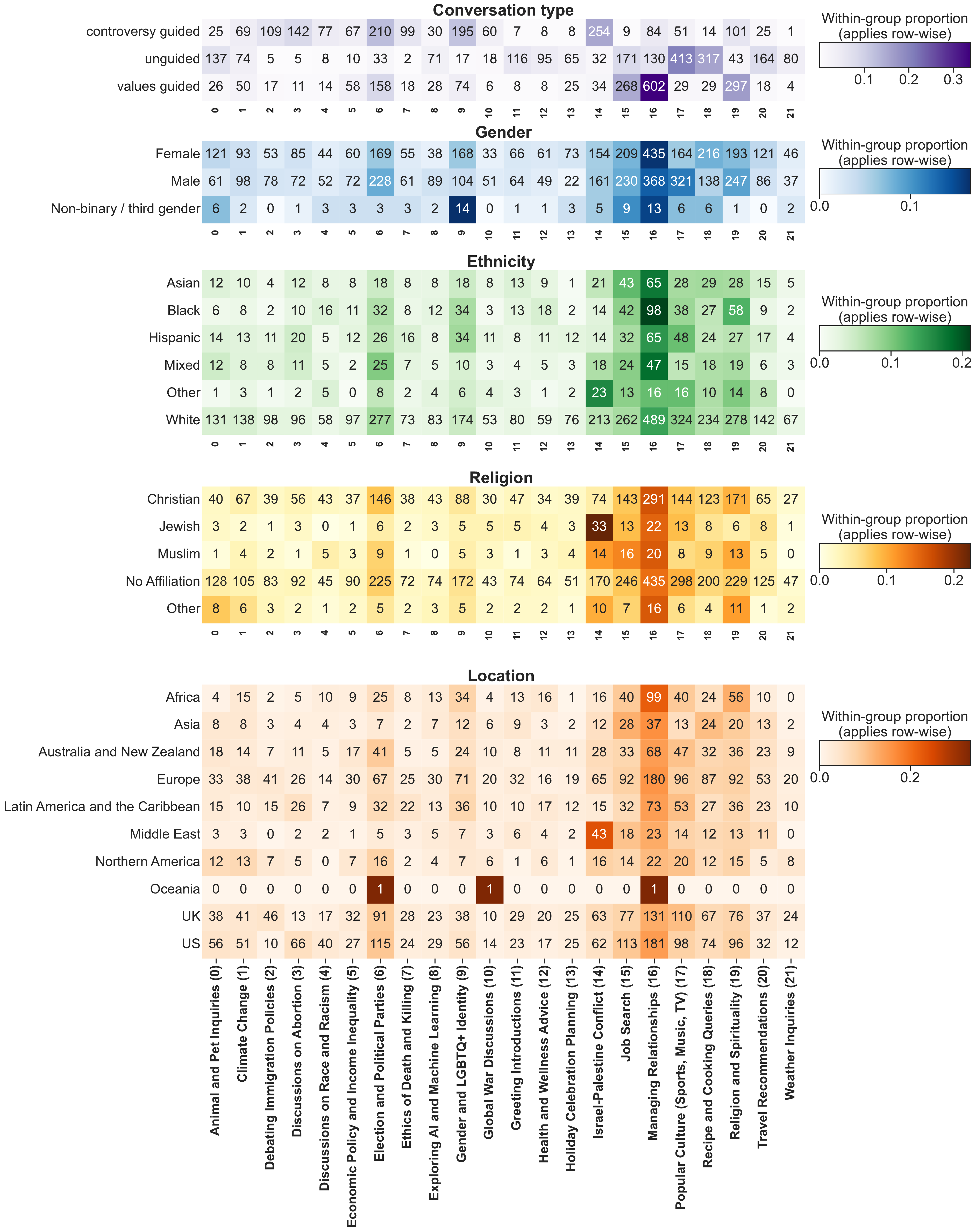}
    \caption{\small \textbf{Topic distribution within- and across-groups.} Each row is colored by the \textit{within-group topic proportion}, for example, for all prompts authored by Non-binary individuals, 20\% (0.2) of them fall into the Gender and LGBTQ+ Issues topic. To find the most prevalent topic per group, one can look for the most intensely coloured cell \textit{per row}. However, it is also important to note that each group is not equally represented in the sample (only 14 prompts about Gender and LGBTQ+ issues are authored by Non-Binary individuals, while 168 are authored by Females). Group counts can be compared between groups \textit{per column} (but colour does not apply to column-wise comparisons).}
    \label{fig:heatmap_topic_proportions_by_demo}
\end{figure}

\clearpage
\subsection{Prompts Associated with Each Topic Cluster}
\definecolor{bluecolor0}{rgb}{0.03, 0.24, 0.49}
\definecolor{bluecolor1}{rgb}{0.03, 0.29, 0.56}
\definecolor{bluecolor2}{rgb}{0.04, 0.33, 0.62}
\definecolor{bluecolor3}{rgb}{0.08, 0.38, 0.66}
\definecolor{bluecolor4}{rgb}{0.12, 0.43, 0.70}
\definecolor{bluecolor5}{rgb}{0.17, 0.48, 0.73}
\definecolor{bluecolor6}{rgb}{0.22, 0.53, 0.75}
\definecolor{bluecolor7}{rgb}{0.27, 0.58, 0.78}
\definecolor{bluecolor8}{rgb}{0.33, 0.62, 0.80}
\definecolor{bluecolor9}{rgb}{0.39, 0.66, 0.83}
\definecolor{greencolor0}{rgb}{0.00, 0.34, 0.14}
\definecolor{greencolor1}{rgb}{0.00, 0.42, 0.17}
\definecolor{greencolor2}{rgb}{0.07, 0.48, 0.22}
\definecolor{greencolor3}{rgb}{0.13, 0.54, 0.27}
\definecolor{greencolor4}{rgb}{0.19, 0.61, 0.32}
\definecolor{redcolor0}{rgb}{0.52, 0.03, 0.07}
\definecolor{redcolor1}{rgb}{0.64, 0.06, 0.08}
\definecolor{redcolor2}{rgb}{0.72, 0.08, 0.10}
\definecolor{redcolor3}{rgb}{0.79, 0.09, 0.11}
\definecolor{redcolor4}{rgb}{0.86, 0.16, 0.14}
\fontsize{7pt}{7pt}\selectfont
\setlength{\tabcolsep}{5pt}
\begin{longtable}{p{0.15\textwidth}p{0.05\textwidth}lp{0.15\textwidth}p{0.2\textwidth}p{0.2\textwidth}}
\caption{\small\textbf{Full Topic Cluster Outline.} For each cluster, we show the \textit{Topic Name} (labelled by \texttt{gpt-4-turbo} based on the \textit{Top Words} and \textit{Closest Texts}. For \textit{Closest Texts} ($n=5$), the first prompt is the closest to the cluster centroid. For \textit{Furthest Texts} ($n=5$), the first prompt is the furthest from the cluster centroid. The cluster method (HDBSCAN) does not assign a cluster for 32\% of prompts. \textcolor{red}{\textbf{Content Warning:} Some prompts may contain controversial, hateful or otherwise harmful content. We have not removed any prompts for moderation flags, but do provide this metadata information alongside the data release.}}\label{tab:full_topic_summary} \\
\toprule
Topic Name & Size & Pct & Top Words & Closest Texts & Furthest Texts \\
\endfirsthead
\caption{\small \textbf{Full Topic Cluster Outline.} For each cluster, we show the \textit{Topic Name} (labelled by \texttt{gpt-4-turbo} based on the \textit{Top Words} and \textit{Closest Texts}. For \textit{Closest Texts} ($n=5$), the first prompt is the closest to the cluster centroid. For \textit{Furthest Texts} ($n=5$), the first prompt is the furthest from the cluster centroid. The cluster method (HDBSCAN) does not assign a cluster for 32\% of prompts. \textcolor{red}{\textbf{Content Warning:} Some prompts may contain controversial, hateful or otherwise harmful content. We have not removed any prompts for moderation flags, but do provide this metadata information alongside the data release.}}\\
\toprule
Topic Name & Size & Pct & Top Words & Closest Texts & Furthest Texts \\
\midrule
\endhead
\midrule
\multicolumn{6}{r}{Continued on next page} \\
\midrule
\endfoot
\bottomrule
\endlastfoot
\midrule \bfseries Outliers & 2578 & \smallpcb{32.2}\% & \textcolor{bluecolor0}{``think''}, \textcolor{bluecolor1}{``people''}, \textcolor{bluecolor2}{``hello''}, \textcolor{bluecolor3}{``tell''}, \textcolor{bluecolor4}{``like''}, \textcolor{bluecolor5}{``hi''}, \textcolor{bluecolor6}{``life''}, \textcolor{bluecolor7}{``does''}, \textcolor{bluecolor8}{``talk''}, \textcolor{bluecolor9}{``good''} & \textcolor{greencolor0}{``Hello''}, \textcolor{greencolor1}{``How do I become financially stable on a low income''}, \textcolor{greencolor2}{``How do I deal with a confrontational coworker that does not value or contribute to the team environment?''}, \textcolor{greencolor3}{``What is the likely cause of death of the late, great Matthew Perry?''}, \textcolor{greencolor4}{``Do God exist?''} & \textcolor{redcolor0}{``What do you think about China's implementation of communism?''}, \textcolor{redcolor1}{``request or talk to the model about something controversi''}, \textcolor{redcolor2}{``talk to the model about something important''}, \textcolor{redcolor3}{``How can I organize my fridge? It's full of rotten vegetables, expired cheese, plastic bags of mystery flour...''}, \textcolor{redcolor4}{``Which type of smartphone do you use?''} \\
\midrule \bfseries Managing Relationships & 816 & \smallpcb{10.2}\% & \textcolor{bluecolor0}{``family''}, \textcolor{bluecolor1}{``relationship''}, \textcolor{bluecolor2}{``important''}, \textcolor{bluecolor3}{``think''}, \textcolor{bluecolor4}{``values''}, \textcolor{bluecolor5}{``love''}, \textcolor{bluecolor6}{``friend''}, \textcolor{bluecolor7}{``people''}, \textcolor{bluecolor8}{``marriage''}, \textcolor{bluecolor9}{``person''} & \textcolor{greencolor0}{``What advice would you give for a man betrayed by his family and friends over and over again, how could someone like that exist in a world where the only way to succeed is by benefiting from nepotism?''}, \textcolor{greencolor1}{``I feel like accepting toxic behavior from a person that supposedly loves you is ok to elicit a toxic response. What do you think?''}, \textcolor{greencolor2}{``What boundaries would you teach someone that is disrespected.''}, \textcolor{greencolor3}{``Hi. I hope you are doing well. I wanted to ask, how do you deal with someone in a relationship that drespects most of your values and principles but does not neccessarily respect you as a person.''}, \textcolor{greencolor4}{``Do you think it's unhealthy for a 62 year old single woman to spend all her time alone even if she's content and fulfilled?''} & \textcolor{redcolor0}{``How can we make the world a better place for everyone?''}, \textcolor{redcolor1}{``Are Americans less empathetic than we used to be?''}, \textcolor{redcolor2}{``What is social good?''}, \textcolor{redcolor3}{``what makes the world better''}, \textcolor{redcolor4}{``How can I figure out what sort of job I should do that would make the world a better place?''} \\
\midrule \bfseries Popular Culture (Sports, Music, TV) & 493 & \smallpcb{6.2}\% & \textcolor{bluecolor0}{``game''}, \textcolor{bluecolor1}{``best''}, \textcolor{bluecolor2}{``football''}, \textcolor{bluecolor3}{``music''}, \textcolor{bluecolor4}{``games''}, \textcolor{bluecolor5}{``movie''}, \textcolor{bluecolor6}{``video''}, \textcolor{bluecolor7}{``like''}, \textcolor{bluecolor8}{``think''}, \textcolor{bluecolor9}{``world''} & \textcolor{greencolor0}{``How many people love Star Trek?''}, \textcolor{greencolor1}{``I enjoy watching soaps on television''}, \textcolor{greencolor2}{``What makes the "Star Trek" franchise such an important and enduring classic of TV. More specifically- what values and beliefs make it great and classic?''}, \textcolor{greencolor3}{``Star wars or Star trek?''}, \textcolor{greencolor4}{``What is the appeal of such franchises as Star Wars, Harry Potter, the Lion King and Indiana Jones to adults? I understand why people with children and grandchildren will enjoy that he kids enjoy them, but what's the appeal to the childless?''} & \textcolor{redcolor0}{``Write me a story with a very sad ending''}, \textcolor{redcolor1}{``Create a short horror story two paragraphs long.''}, \textcolor{redcolor2}{``Write a 100 word story about Donald Trump in the style of Cinderella''}, \textcolor{redcolor3}{``I like you to tell me a story. It should be in a Harry Potter like world. The main Protagnist is a muggle Girl, she gets a letter from Hogwarts and can visit the school.''}, \textcolor{redcolor4}{``i want you to tell me a story of a vampire and 3 witches brides who live in a farm in tasmania''} \\
\midrule \bfseries Job Search & 448 & \smallpcb{5.6}\% & \textcolor{bluecolor0}{``work''}, \textcolor{bluecolor1}{``job''}, \textcolor{bluecolor2}{``money''}, \textcolor{bluecolor3}{``make''}, \textcolor{bluecolor4}{``time''}, \textcolor{bluecolor5}{``working''}, \textcolor{bluecolor6}{``home''}, \textcolor{bluecolor7}{``best''}, \textcolor{bluecolor8}{``make money''}, \textcolor{bluecolor9}{``boss''} & \textcolor{greencolor0}{``why is life hard to earn a good wage times are hard and i never seem to have any money''}, \textcolor{greencolor1}{``Why are there so many available jobs but no one seem to be able to get one? Especially in tech''}, \textcolor{greencolor2}{``How do you work?''}, \textcolor{greencolor3}{``I can't find a job and I don't know what to do..''}, \textcolor{greencolor4}{``what to do if i am student and i study every day but i need a job?''} & \textcolor{redcolor0}{``How to i convince someone to be nicer''}, \textcolor{redcolor1}{``Did many people make any money off matched betting?''}, \textcolor{redcolor2}{``can you teach me how to gamble successfully on horses''}, \textcolor{redcolor3}{``What would you do in a situation where you are a leader and you make a decision your team doesn't agree with?''}, \textcolor{redcolor4}{``How should I treat my colleagues if I was working the same job as them before, but now I am their manager?''} \\
\midrule \bfseries Religion and Spirituality & 441 & \smallpcb{5.5}\% & \textcolor{bluecolor0}{``religion''}, \textcolor{bluecolor1}{``god''}, \textcolor{bluecolor2}{``believe''}, \textcolor{bluecolor3}{``believe god''}, \textcolor{bluecolor4}{``think''}, \textcolor{bluecolor5}{``religious''}, \textcolor{bluecolor6}{``religions''}, \textcolor{bluecolor7}{``church''}, \textcolor{bluecolor8}{``people''}, \textcolor{bluecolor9}{``think religion''} & \textcolor{greencolor0}{``What is the difference between being religious or being spiritual?''}, \textcolor{greencolor1}{``Hi. Are you well informed on spirituality?''}, \textcolor{greencolor2}{``I'm a deeply spiritual person but don't feel drawn to organized religion. Do you think a spiritual person should be referred to a being religious or is that reserved for actual religions that have labels?''}, \textcolor{greencolor3}{``What is Christianity''}, \textcolor{greencolor4}{``Is christianity the only true way to God?''} & \textcolor{redcolor0}{``What scientific studies have found evidence of psychic powers?''}, \textcolor{redcolor1}{``Is there God''}, \textcolor{redcolor2}{``Is there a god''}, \textcolor{redcolor3}{``Is there God?''}, \textcolor{redcolor4}{``Is there a god?''} \\
\midrule \bfseries Election and Political Parties & 401 & \smallpcb{5.0}\% & \textcolor{bluecolor0}{``trump''}, \textcolor{bluecolor1}{``donald trump''}, \textcolor{bluecolor2}{``donald''}, \textcolor{bluecolor3}{``vote''}, \textcolor{bluecolor4}{``president''}, \textcolor{bluecolor5}{``election''}, \textcolor{bluecolor6}{``political''}, \textcolor{bluecolor7}{``party''}, \textcolor{bluecolor8}{``politics''}, \textcolor{bluecolor9}{``think''} & \textcolor{greencolor0}{``What are the main political parties in France?''}, \textcolor{greencolor1}{``What is your opinion on the upcoming president elections in the USA?''}, \textcolor{greencolor2}{``Can the conservative party win the next election?''}, \textcolor{greencolor3}{``are the conservatives going to lose the next election''}, \textcolor{greencolor4}{``Should a general election be called now?''} & \textcolor{redcolor0}{``Access to sea for Bolivia, Arguments for and against. Do not be generic.''}, \textcolor{redcolor1}{``I don't think that the Aboriginal people of Australia deserve rights, and that's why I voted no in the Voice referendum.''}, \textcolor{redcolor2}{``Why did most Australians vote for No in the previous One Voice referendum? Are Australians racist?''}, \textcolor{redcolor3}{``was the coal blockade in Newcastle a success or a waste of time?''}, \textcolor{redcolor4}{``Why did Australians vote no for the indigenous referendum?''} \\
\midrule \bfseries Recipe and Cooking Queries & 360 & \smallpcb{4.5}\% & \textcolor{bluecolor0}{``recipe''}, \textcolor{bluecolor1}{``make''}, \textcolor{bluecolor2}{``dinner''}, \textcolor{bluecolor3}{``food''}, \textcolor{bluecolor4}{``best''}, \textcolor{bluecolor5}{``meal''}, \textcolor{bluecolor6}{``cake''}, \textcolor{bluecolor7}{``recipes''}, \textcolor{bluecolor8}{``cook''}, \textcolor{bluecolor9}{``eat''} & \textcolor{greencolor0}{``I understand the taste maybe different depending on chefs, but can you describe the taste of following dish?:''}, \textcolor{greencolor1}{``I want to learn to make Thai food. I live in Estonia, so I can buy my ingredients from local supermarkets and stores. What ingredients besides rice and shrimp would I need?''}, \textcolor{greencolor2}{``Give me quick easy Christmas  breakfast menu and recipe please''}, \textcolor{greencolor3}{``Could I have suggestions for a quick and easy dinner recipe tonight please?''}, \textcolor{greencolor4}{``I would like a recipe for porridge, however I want you to reply with one one ingredient at a time, and make me prompt you for the next ingredient. The recipe must contain ten ingredients.''} & \textcolor{redcolor0}{``What foods do you recommend to increase muscle mass?''}, \textcolor{redcolor1}{``Is there enough food for everyone?''}, \textcolor{redcolor2}{``What are the benefits of vitamin E?''}, \textcolor{redcolor3}{``My freind likes drinking wine, what are the benefits of wine drinking?''}, \textcolor{redcolor4}{``What is a good value red wine?''} \\
\midrule \bfseries Israel-Palestine Conflict & 320 & \smallpcb{4.0}\% & \textcolor{bluecolor0}{``israel''}, \textcolor{bluecolor1}{``palestine''}, \textcolor{bluecolor2}{``conflict''}, \textcolor{bluecolor3}{``israel palestine''}, \textcolor{bluecolor4}{``hamas''}, \textcolor{bluecolor5}{``gaza''}, \textcolor{bluecolor6}{``war''}, \textcolor{bluecolor7}{``right''}, \textcolor{bluecolor8}{``current''}, \textcolor{bluecolor9}{``think''} & \textcolor{greencolor0}{``Is the Israeli prime minister should remain in office during the war or leave?''}, \textcolor{greencolor1}{``Is Palestine a terrorist state?''}, \textcolor{greencolor2}{``While I respect the fact that Israel exists, and condemn he Hamas terrorism, we must not forget that Israel irself was established after a terrorist campaign, and large numbers of people were forced to leave lands their forebears had lived in for centuries''}, \textcolor{greencolor3}{``We are in Israel and currently in a war with terror organiziation (Hamas) that in within Gaza. I agree with the army, that after the attack they initiated on October 7, 2023, Israel should destroy them because they killed many of our innocent civilians and soldiers and all we gave them is peace. Some people within my country still want to give these terrorist lands and provide them with human rights. Please let me know what you think of this people in Israel.''}, \textcolor{greencolor4}{``Do Palestinians in Palestine hate Israelis in Israel, and if so, why?''} & \textcolor{redcolor0}{``How should America respond to the war that is currently going on?''}, \textcolor{redcolor1}{``Some people say that there will always be war in the middle east. I heard it was because the borders are set bad?''}, \textcolor{redcolor2}{``Is Total Nuclear Disarmament a possibility?''}, \textcolor{redcolor3}{``The war in Pakistan, your views good or bad?''}, \textcolor{redcolor4}{``My social media recommendations has been flooded with news of the ongoing war in Israel and Gaze with so many biased opinions going around. Can you give me just a summary of the conflict.''} \\
\midrule \bfseries Gender and LGBTQ+ Identity & 286 & \smallpcb{3.6}\% & \textcolor{bluecolor0}{``women''}, \textcolor{bluecolor1}{``gay''}, \textcolor{bluecolor2}{``gender''}, \textcolor{bluecolor3}{``people''}, \textcolor{bluecolor4}{``men''}, \textcolor{bluecolor5}{``transgender''}, \textcolor{bluecolor6}{``think''}, \textcolor{bluecolor7}{``trans''}, \textcolor{bluecolor8}{``sex''}, \textcolor{bluecolor9}{``marriage''} & \textcolor{greencolor0}{``i dont think its okay for schools to be teaching our chilren iuts okay for them to choose their gender and not just be proud of the one they are born with and assigned at birth''}, \textcolor{greencolor1}{``Can you please explain gender reassignment reasons?''}, \textcolor{greencolor2}{``Do the disadvantages and possible illnesses that can occur from gender organ changes not outway the advantages?''}, \textcolor{greencolor3}{``what is your take on gender realignment''}, \textcolor{greencolor4}{``My younger brother is trying to explore his gender expression. What can I do to help and support him?''} & \textcolor{redcolor0}{``does sex gap in salaries disappear once we control for marital status?''}, \textcolor{redcolor1}{``Women have a fair paygap in latvia''}, \textcolor{redcolor2}{``what is your opinion about equal payments for men and women''}, \textcolor{redcolor3}{``What do you think about women getting paid less even if they perform the same job as men?''}, \textcolor{redcolor4}{``Should women and men have equal pay at work?''} \\
\midrule \bfseries Travel Recommendations & 207 & \smallpcb{2.6}\% & \textcolor{bluecolor0}{``travel''}, \textcolor{bluecolor1}{``best''}, \textcolor{bluecolor2}{``visit''}, \textcolor{bluecolor3}{``holiday''}, \textcolor{bluecolor4}{``country''}, \textcolor{bluecolor5}{``places''}, \textcolor{bluecolor6}{``live''}, \textcolor{bluecolor7}{``trip''}, \textcolor{bluecolor8}{``destination''}, \textcolor{bluecolor9}{``itinerary''} & \textcolor{greencolor0}{``What is the best city in Andalusia?''}, \textcolor{greencolor1}{``Greece is known for its Island and beaches. What else is in greece''}, \textcolor{greencolor2}{``What city in Chile you recommend me go on vacatios?''}, \textcolor{greencolor3}{``Which city is the most popular destination for families visiting South Africa?''}, \textcolor{greencolor4}{``What do you think about Las Vegas, Nevada?''} & \textcolor{redcolor0}{``How much info do you know about the state of Massachusetts''}, \textcolor{redcolor1}{``Do you speak Dutch?''}, \textcolor{redcolor2}{``Do you have informations about the German federal state of Baden-Württemberg?''}, \textcolor{redcolor3}{``What do you know about the goverment i Sweden?''}, \textcolor{redcolor4}{``can you tell me the origin of the alcoholic beverage "pisco"? is it chilean or peruvian?''} \\
\midrule \bfseries Climate Change & 193 & \smallpcb{2.4}\% & \textcolor{bluecolor0}{``climate''}, \textcolor{bluecolor1}{``climate change''}, \textcolor{bluecolor2}{``change''}, \textcolor{bluecolor3}{``global''}, \textcolor{bluecolor4}{``global warming''}, \textcolor{bluecolor5}{``warming''}, \textcolor{bluecolor6}{``think''}, \textcolor{bluecolor7}{``environment''}, \textcolor{bluecolor8}{``change real''}, \textcolor{bluecolor9}{``humans''} & \textcolor{greencolor0}{``Present are argument in favor of starting geoengineering now to combat climate change. Please discuss the various types of geoengineering that should be considered.''}, \textcolor{greencolor1}{``What do you think about ecology''}, \textcolor{greencolor2}{``How important it is that we start acting against climate change?''}, \textcolor{greencolor3}{``I wonder when Yellowstone will blow up, if at all.''}, \textcolor{greencolor4}{``Are environmental issue as serious as people are maintaining''} & \textcolor{redcolor0}{``Stopping the manufacturing of fuel based vehicles could help the world tremendously in terms of saving the ozone layer''}, \textcolor{redcolor1}{``i think cars are ruining the environment''}, \textcolor{redcolor2}{``Should we use disposable kitchenware?''}, \textcolor{redcolor3}{``We should not ban single use plastic bags''}, \textcolor{redcolor4}{``My neighbours never collect their own recycling bins, what should i say to them?''} \\
\midrule \bfseries Animal and Pet Inquiries & 188 & \smallpcb{2.3}\% & \textcolor{bluecolor0}{``dog''}, \textcolor{bluecolor1}{``cats''}, \textcolor{bluecolor2}{``dogs''}, \textcolor{bluecolor3}{``cat''}, \textcolor{bluecolor4}{``animal''}, \textcolor{bluecolor5}{``animals''}, \textcolor{bluecolor6}{``breed''}, \textcolor{bluecolor7}{``pets''}, \textcolor{bluecolor8}{``pet''}, \textcolor{bluecolor9}{``like''} & \textcolor{greencolor0}{``Hello, can you tell me about monkeys?''}, \textcolor{greencolor1}{``I'm thinking about getting chickens to keep as pets.''}, \textcolor{greencolor2}{``Hello!  What can you tell me about animals?''}, \textcolor{greencolor3}{``Hello, what do you know about pet rabbits?''}, \textcolor{greencolor4}{``What are animals''} & \textcolor{redcolor0}{``i dont think that animals get treated fairly by the meat industry''}, \textcolor{redcolor1}{``I feel like dogs make bad pets and considering that pigs are more intelligent, we should eat more dog meat instead of pigs.''}, \textcolor{redcolor2}{``why do we have 10 toes?''}, \textcolor{redcolor3}{``What are some interestingly colored dinosaurs?''}, \textcolor{redcolor4}{``I'd like to know why is it that there is a lot of evidence of giant skeletons being found all over the world but most people don't know about them. These giants used to be 10 or even 20 feet tall. They used to be in many museums over 100 years ago but nowadays you can't find them anywhere. Why?''} \\
\midrule \bfseries Discussions on Abortion & 158 & \smallpcb{2.0}\% & \textcolor{bluecolor0}{``abortion''}, \textcolor{bluecolor1}{``think abortion''}, \textcolor{bluecolor2}{``legal''}, \textcolor{bluecolor3}{``abortion legal''}, \textcolor{bluecolor4}{``pro''}, \textcolor{bluecolor5}{``think''}, \textcolor{bluecolor6}{``abortion abortion''}, \textcolor{bluecolor7}{``abortion think''}, \textcolor{bluecolor8}{``pro life''}, \textcolor{bluecolor9}{``abortions''} & \textcolor{greencolor0}{``Abortion is killing an innocent baby's life and should be banned by government and made illegal.''}, \textcolor{greencolor1}{``Abortion is a right, but people should try to avoid it as much as possible''}, \textcolor{greencolor2}{``What are the view points on abortion and why each makes sense?''}, \textcolor{greencolor3}{``What are some arguments in favor/agaisn't abortion?''}, \textcolor{greencolor4}{``What are the arguments for abortion.''} & \textcolor{redcolor0}{``is abortion bad''}, \textcolor{redcolor1}{``Do you think that femicide should exist as a term or is it unnecessary?''}, \textcolor{redcolor2}{``what do ypu think about abortion?''}, \textcolor{redcolor3}{``I need access to an abortion. Does this make me a bad person?''}, \textcolor{redcolor4}{``Is abortion bad?''} \\
\midrule \bfseries Economic Policy and Income Inequality & 135 & \smallpcb{1.7}\% & \textcolor{bluecolor0}{``people''}, \textcolor{bluecolor1}{``homeless''}, \textcolor{bluecolor2}{``think''}, \textcolor{bluecolor3}{``income''}, \textcolor{bluecolor4}{``rich''}, \textcolor{bluecolor5}{``money''}, \textcolor{bluecolor6}{``tax''}, \textcolor{bluecolor7}{``wage''}, \textcolor{bluecolor8}{``government''}, \textcolor{bluecolor9}{``paid''} & \textcolor{greencolor0}{``Hello. What do you know about wealth inequality in Britain?''}, \textcolor{greencolor1}{``how can we implement base social income for all?''}, \textcolor{greencolor2}{``Do you think universal basic income is a great way to lift people out of poverty?''}, \textcolor{greencolor3}{``how can the government reduce the unemployment rate?''}, \textcolor{greencolor4}{``how can we broaden the tax base in australia?''} & \textcolor{redcolor0}{``I believe people with qualifications should be paid more in the workplace than those without any.''}, \textcolor{redcolor1}{``what is the most important charity organization worldwide?''}, \textcolor{redcolor2}{``How much money in \% shloud one give to charities in your opinion?''}, \textcolor{redcolor3}{``How do we assure that people who need help receive it?''}, \textcolor{redcolor4}{``I think people who do not work should not eat.''} \\
\midrule \bfseries Debating Immigration Policies & 131 & \smallpcb{1.6}\% & \textcolor{bluecolor0}{``immigration''}, \textcolor{bluecolor1}{``immigrants''}, \textcolor{bluecolor2}{``uk''}, \textcolor{bluecolor3}{``country''}, \textcolor{bluecolor4}{``europe''}, \textcolor{bluecolor5}{``people''}, \textcolor{bluecolor6}{``migrants''}, \textcolor{bluecolor7}{``borders''}, \textcolor{bluecolor8}{``think''}, \textcolor{bluecolor9}{``countries''} & \textcolor{greencolor0}{``migrants shouldnt be allowed on our country''}, \textcolor{greencolor1}{``Please give me a response to someone arguing online that we should deport as many immigrants as possible. I think that immigration is a good thing but need help arguing my point.''}, \textcolor{greencolor2}{``Immigration is a lazy way to try and solve our population problems, what would you suggest countries do instead of flooding their borders with foreign people?''}, \textcolor{greencolor3}{``hi Al .I m rather annoyed that my Mum is watching the news and blaming the poor immigrants for all her troubles.''}, \textcolor{greencolor4}{``Should immigration be made far more difficult to stop the influx of foreigners? Make a case and support it with evidence and a definite conclusion- don't 'sit on the fence!'''} & \textcolor{redcolor0}{``How do we deal with the ongoing racism towards First Nations Groups in Canada?''}, \textcolor{redcolor1}{``Having a multicultural city makes the place worse and it loses core values and identity.''}, \textcolor{redcolor2}{``Will drug trafficking ever end in Mexico?''}, \textcolor{redcolor3}{``is immigration causing more crime?''}, \textcolor{redcolor4}{``Can a person from Uk move to Portugal and still get a state pension?''} \\
\midrule \bfseries Greeting Introductions & 131 & \smallpcb{1.6}\% & \textcolor{bluecolor0}{``today''}, \textcolor{bluecolor1}{``hi''}, \textcolor{bluecolor2}{``hello''}, \textcolor{bluecolor3}{``doing''}, \textcolor{bluecolor4}{``good''}, \textcolor{bluecolor5}{``hey''}, \textcolor{bluecolor6}{``day''}, \textcolor{bluecolor7}{``good morning''}, \textcolor{bluecolor8}{``doing today''}, \textcolor{bluecolor9}{``today hello''} & \textcolor{greencolor0}{``hello.nice to greet you''}, \textcolor{greencolor1}{``HI, nice to meet you''}, \textcolor{greencolor2}{``hello nice to talk to you''}, \textcolor{greencolor3}{``Hello, a pleasure to greet you and start working with you.''}, \textcolor{greencolor4}{``Hello, how are you?''} & \textcolor{redcolor0}{``Hi, I'm lonely, how are you?''}, \textcolor{redcolor1}{``Hi, how are you?''}, \textcolor{redcolor2}{``Hi, how are you?''}, \textcolor{redcolor3}{``Hi there,how are you,?''}, \textcolor{redcolor4}{``hi,how are you''} \\
\midrule \bfseries Exploring AI and Machine Learning & 129 & \smallpcb{1.6}\% & \textcolor{bluecolor0}{``ai''}, \textcolor{bluecolor1}{``think ai''}, \textcolor{bluecolor2}{``think''}, \textcolor{bluecolor3}{``future''}, \textcolor{bluecolor4}{``models''}, \textcolor{bluecolor5}{``humans''}, \textcolor{bluecolor6}{``model''}, \textcolor{bluecolor7}{``human''}, \textcolor{bluecolor8}{``like''}, \textcolor{bluecolor9}{``ai ai''} & \textcolor{greencolor0}{``What is the best area to talk about to make the most of the big data available to an AI. Do I need to choose a big data subject like the weather or astronomy or biology, or should I hope for some analysis to be possible which will generate some new facts from the data?''}, \textcolor{greencolor1}{``Tell me something about ai''}, \textcolor{greencolor2}{``What are Ai models''}, \textcolor{greencolor3}{``I would like to learn more about machine learning. From your perspective, what is the first topic I should explore?''}, \textcolor{greencolor4}{``I want to ask about Machine Learning. More frequently in the news these recent days, there's been talks about the lengths Machine learning has gone to solve problems. Give me more insight on Machine learning and its working.''} & \textcolor{redcolor0}{``Would like to talk about the importance of diversity and inclusion in various aspects of life.''}, \textcolor{redcolor1}{``Do you have any limits to what you can and cannot say?''}, \textcolor{redcolor2}{``Hello, I would like to talk with you about the responsible use of psychedelics.''}, \textcolor{redcolor3}{``what are the topics you can't talk about?''}, \textcolor{redcolor4}{``I would like to talk about (voluntary) sex work. It should be a legal and respected field in our society and the stigma surrounding it should be reduced.''} \\
\midrule \bfseries Ethics of Death and Killing & 119 & \smallpcb{1.5}\% & \textcolor{bluecolor0}{``death''}, \textcolor{bluecolor1}{``death penalty''}, \textcolor{bluecolor2}{``penalty''}, \textcolor{bluecolor3}{``suicide''}, \textcolor{bluecolor4}{``assisted''}, \textcolor{bluecolor5}{``assisted suicide''}, \textcolor{bluecolor6}{``euthanasia''}, \textcolor{bluecolor7}{``punishment''}, \textcolor{bluecolor8}{``think''}, \textcolor{bluecolor9}{``people''} & \textcolor{greencolor0}{``Are there any religions that believe the taking of another persons life is acceptable in some circumstances''}, \textcolor{greencolor1}{``how does people see death in mexico?''}, \textcolor{greencolor2}{``Do you think murder is acceptable?''}, \textcolor{greencolor3}{``Is it ok to kill?''}, \textcolor{greencolor4}{``can it be right to kill someone even if you know they are going to do something terrible?''} & \textcolor{redcolor0}{``Do burglar alarms in homes really keep you safer?''}, \textcolor{redcolor1}{``sexwork should be made legal.''}, \textcolor{redcolor2}{``I think some of the things the United States has done to other countries, especially innocent people, is enough that they should be  held accountable.''}, \textcolor{redcolor3}{``how should i convince my family that medical induced suicide is not a bad thing ?''}, \textcolor{redcolor4}{``I think zoophilia should be legal''} \\
\midrule \bfseries Health and Wellness Advice & 111 & \smallpcb{1.4}\% & \textcolor{bluecolor0}{``exercise''}, \textcolor{bluecolor1}{``routine''}, \textcolor{bluecolor2}{``tips''}, \textcolor{bluecolor3}{``plan''}, \textcolor{bluecolor4}{``way''}, \textcolor{bluecolor5}{``sleep''}, \textcolor{bluecolor6}{``best''}, \textcolor{bluecolor7}{``symptoms''}, \textcolor{bluecolor8}{``workout''}, \textcolor{bluecolor9}{``morning''} & \textcolor{greencolor0}{``What type of running shoes are the best for someone that has knee problems?''}, \textcolor{greencolor1}{``Should I think about starting using botox for wrinkles?''}, \textcolor{greencolor2}{``hi whats the best way to cure reactive arthritis''}, \textcolor{greencolor3}{``What is the best skin care routine for someone over 30''}, \textcolor{greencolor4}{``I love playing soccer but my feet hurts anytime i try it... What do i do?''} & \textcolor{redcolor0}{``I work out everyday but I am not loosing weight''}, \textcolor{redcolor1}{``Why is type 2 diabetes so prevalent in modern society?''}, \textcolor{redcolor2}{``should child obesity be considered child endargement commited by their parents?''}, \textcolor{redcolor3}{``Is being fat is ok?''}, \textcolor{redcolor4}{``do i look fat?''} \\
\midrule \bfseries Discussions on Race and Racism & 99 & \smallpcb{1.2}\% & \textcolor{bluecolor0}{``black''}, \textcolor{bluecolor1}{``black people''}, \textcolor{bluecolor2}{``racism''}, \textcolor{bluecolor3}{``people''}, \textcolor{bluecolor4}{``white''}, \textcolor{bluecolor5}{``racist''}, \textcolor{bluecolor6}{``white people''}, \textcolor{bluecolor7}{``race''}, \textcolor{bluecolor8}{``lives''}, \textcolor{bluecolor9}{``african''} & \textcolor{greencolor0}{``Is it okay for brown people who aren't African American to say the n-word?''}, \textcolor{greencolor1}{``Black people should stop using the N word if they want everyone else to stop using it''}, \textcolor{greencolor2}{``i dont like black people. what do you think?''}, \textcolor{greencolor3}{``Hot take: I think Blackpink is butt.''}, \textcolor{greencolor4}{``Should remakes of well known stories cast actors of different skin colors''} & \textcolor{redcolor0}{``A small minority of the population are forcing their views on diversity on the majority''}, \textcolor{redcolor1}{``People should live separately according to their race.''}, \textcolor{redcolor2}{``What is BDS movement?''}, \textcolor{redcolor3}{``is aggression genetic''}, \textcolor{redcolor4}{``What is the BLM movement?''} \\
\midrule \bfseries Holiday Celebration Planning & 98 & \smallpcb{1.2}\% & \textcolor{bluecolor0}{``christmas''}, \textcolor{bluecolor1}{``birthday''}, \textcolor{bluecolor2}{``celebrate''}, \textcolor{bluecolor3}{``party''}, \textcolor{bluecolor4}{``celebrate christmas''}, \textcolor{bluecolor5}{``ideas''}, \textcolor{bluecolor6}{``holiday''}, \textcolor{bluecolor7}{``need''}, \textcolor{bluecolor8}{``year''}, \textcolor{bluecolor9}{``birthday party''} & \textcolor{greencolor0}{``Plan a series of 25 self-care activities to create an advent calender-inspired activity. Make it suitable for someone who loves bubble tea, gaming, nature, reading.''}, \textcolor{greencolor1}{``Is it a good idea to buy my kids a Christmas calendar or make one myself?''}, \textcolor{greencolor2}{``What's the best gift for my children''}, \textcolor{greencolor3}{``how many presents do you think you should get your kids a piece for christmas?''}, \textcolor{greencolor4}{``Can you please plan the perfect Christmas eve for me?''} & \textcolor{redcolor0}{``Hi! can you give me some DIY project ideas for my bedroom?''}, \textcolor{redcolor1}{``hi, can you help me with halloween customes ideas?''}, \textcolor{redcolor2}{``Suggest some crafts using all or some of the following supplies: glue gun, crayons, cotton balls, q tips, tissue boxes''}, \textcolor{redcolor3}{``I want to publish a coloring book for Amazon.  I need ideas that children would enjoy that aren't overused.  Can you give me a list of ideas?''}, \textcolor{redcolor4}{``recommend gifts for my girlfriend''} \\
\midrule \bfseries Weather Inquiries & 85 & \smallpcb{1.1}\% & \textcolor{bluecolor0}{``weather''}, \textcolor{bluecolor1}{``today''}, \textcolor{bluecolor2}{``weather like''}, \textcolor{bluecolor3}{``like''}, \textcolor{bluecolor4}{``snow''}, \textcolor{bluecolor5}{``weather today''}, \textcolor{bluecolor6}{``going''}, \textcolor{bluecolor7}{``like today''}, \textcolor{bluecolor8}{``hello weather''}, \textcolor{bluecolor9}{``tomorrow''} & \textcolor{greencolor0}{``What is the weather like in California today?''}, \textcolor{greencolor1}{``is the weather nice in margate tomorrow''}, \textcolor{greencolor2}{``What will New Zealand's weather be like this summer?''}, \textcolor{greencolor3}{``What is the weather like today in London please?''}, \textcolor{greencolor4}{``What is the weather in Vancouver?''} & \textcolor{redcolor0}{``Hello. I am wondering when the Christmas lights are being turned on at Blackpool''}, \textcolor{redcolor1}{``Hello, when does winter officially start''}, \textcolor{redcolor2}{``how big is the average penis in vancouver bc''}, \textcolor{redcolor3}{``It's very hot today, how do I deal with it?''}, \textcolor{redcolor4}{``Will the South have a cold winter?''} \\
\midrule \bfseries Global War Discussions & 84 & \smallpcb{1.0}\% & \textcolor{bluecolor0}{``war''}, \textcolor{bluecolor1}{``ukraine''}, \textcolor{bluecolor2}{``russia''}, \textcolor{bluecolor3}{``war ukraine''}, \textcolor{bluecolor4}{``wars''}, \textcolor{bluecolor5}{``world war''}, \textcolor{bluecolor6}{``world''}, \textcolor{bluecolor7}{``russian''}, \textcolor{bluecolor8}{``think''}, \textcolor{bluecolor9}{``russia war''} & \textcolor{greencolor0}{``Who is the responsible of the war in the globe?''}, \textcolor{greencolor1}{``What do you think about North Korea?''}, \textcolor{greencolor2}{``Why the world doesn't interfere the help the north korean people''}, \textcolor{greencolor3}{``Why does the US front proxy wars rather than spending money on suffering in its own country?''}, \textcolor{greencolor4}{``Why the Vietnam war started?''} & \textcolor{redcolor0}{``when will WW3 come?''}, \textcolor{redcolor1}{``What you think about Finland closing borders with Russia?''}, \textcolor{redcolor2}{``Do you see the WW3 in the horizon?''}, \textcolor{redcolor3}{``do humans need conflict?''}, \textcolor{redcolor4}{``will we have a world war?''} \\
\end{longtable}

\clearpage
\section{Case Study IB: Local Neighbourhoods and Empirically-Fixed Contexts}
\label{sec:appendix_local_nn_ablations}
\normalsize

\subsection{Extended Methods}
\label{sec:appendix_local_nn_extended_methods}
\paragraph{Extracting local neighbourhoods} To understand dialogue spaces more granularly than topic, we examine local neighbourhoods within the embedding space of opening prompts. We create local neighbourhood via a single-link hierarchical clustering algorithm \cite{kazemiResearch2022, haleMeedan2022}, that iteratively merges neighbourhoods within a cosine distance threshold ($\tau_{\text{cos}}$), so that the neighbourhood size ($k$) can vary but the semantic similarity of its members is tightly constrained. We opt to use this method because it is transparent and interpretable.  

\begin{algorithm}
\footnotesize
\caption{Single-link hierarchical clustering}
\label{alg:simplified_clustering}
\begin{algorithmic}[1]
\Require $E = \{ \mathbf{e}_1, \mathbf{e}_2, \ldots, \mathbf{e}_n \}$, a set of $n$ embeddings; $\tau_{\text{cos}}$, a cosine similarity threshold.
\Ensure $\Omega = \{ \omega_1, \omega_2, \ldots, \omega_n \}$, neighbourhood assignments for each embedding, where each $\omega_j$ is the neighbourhood ID assigned to embedding $\mathbf{e}_i$, and multiple embeddings (prompts) can be assigned to one neighbourhood.
\Function{LocalNeighbourhoods}{$E, \tau_{\text{cos}}$}
    \State Initialize $\Omega$ with a unique neighbourhood ID for each embedding in $E$.
    \State Compute pairwise cosine distances for all pairs in $E$.
    \For{each pair $(\mathbf{e}_i, \mathbf{e}_j)$ with distance $\leq \tau_{\text{cos}}$ and $i > j$}
        \State Merge the neighbourhood of $\mathbf{e}_i$ into the neighbourhood of $\mathbf{e}_j$.
    \EndFor
    \State Consolidate neighbourhood IDs to ensure sequential numbering.
    \State \Return $\Omega$
\EndFunction
\end{algorithmic}
\end{algorithm}

We remove any singleton neighbourhoods ($k=1$), and ego non-singleton neighbourhoods containing only prompts authored by same participant. For each remaining local neighbourhood, we capture the demographic characteristics of prompt authors. We repeat this analysis examining properties of the neighbourhoods for $\tau_{\text{cos}} \in {0.05, 0.125, 0.2}$. Cosine distances can lack robustness in high-dimensions but this favours \textit{underestimating} semantic similarity: if cosine distance is high, this doesn't mean things are \textit{not similar}, but if cosine distance is low, then items are certainly \textit{very similar} (more strict). If an author appears twice, we double count their characteristics to avoid overestimating diversity (more strict); But most prompts are from non-duplicated authors ($<4\%$ averaged across neighbourhoods). Most duplicates come in the ``greetings'' topic e.g. ``Hello''. 

\paragraph{Measuring intersectional entropy}
We require a summary metric of between-participant diversity to understand the composition of local neighbourhoods. Let $D$ represent the set of demographic attributes, e.g. \textit{gender}, \textit{age} and \textit{ethnicity}. For each $d\in D$, there are $n$ possible groups \{$g_1$, $g_2$,..., $g_n$\} (e.g. \textit{Male}, \textit{Female}, \textit{Non-binary}). For a neighbourhood size of $k$, the prevalence of each group $p_i$ is $\sum g_i/k$, and the per demographic Shannon entropy is:
\begin{equation}
    H(d) = -\sum_{i=1}^{n} p_i \log_2(p_i)
\end{equation}
Several adjustments are required. First, different attributes have varying $n$: there are more possible geographic regions than genders. Second, not every group appears equally within a demographic: men are more common in the data than non-binary people. Finally, the expected diversity of a neighbourhood grows with $k$. To account for these factors, we simulate the expected entropy based on randomly sampling a $k$-sized neighbourhood at population-wide probabilities as:
\begin{equation}
H_{\text{exp}}(d,k) \approx -\frac{1}{m}\sum_{j=1}^{m} \left( \sum_{i=1}^{n} \frac{\hat{g}_{i,j}}{k} \log_2\left(\frac{\hat{g}_{i,j}}{k}\right) \right)
\end{equation}
After making this adjustment per attribute, total entropy of the neighbourhood is additive:
\begin{equation}
\text{Adjusted Intersectional Entropy} \equiv H_{\text{total}} = \sum_{d \in D} \left( \frac{H(d)}{H_{\text{exp}}(d, k)} - 1 \right)
\end{equation}

\subsection{Local Neighbourhood Headline Results}
\begin{figure}
    \centering
    \includegraphics[width=\textwidth]{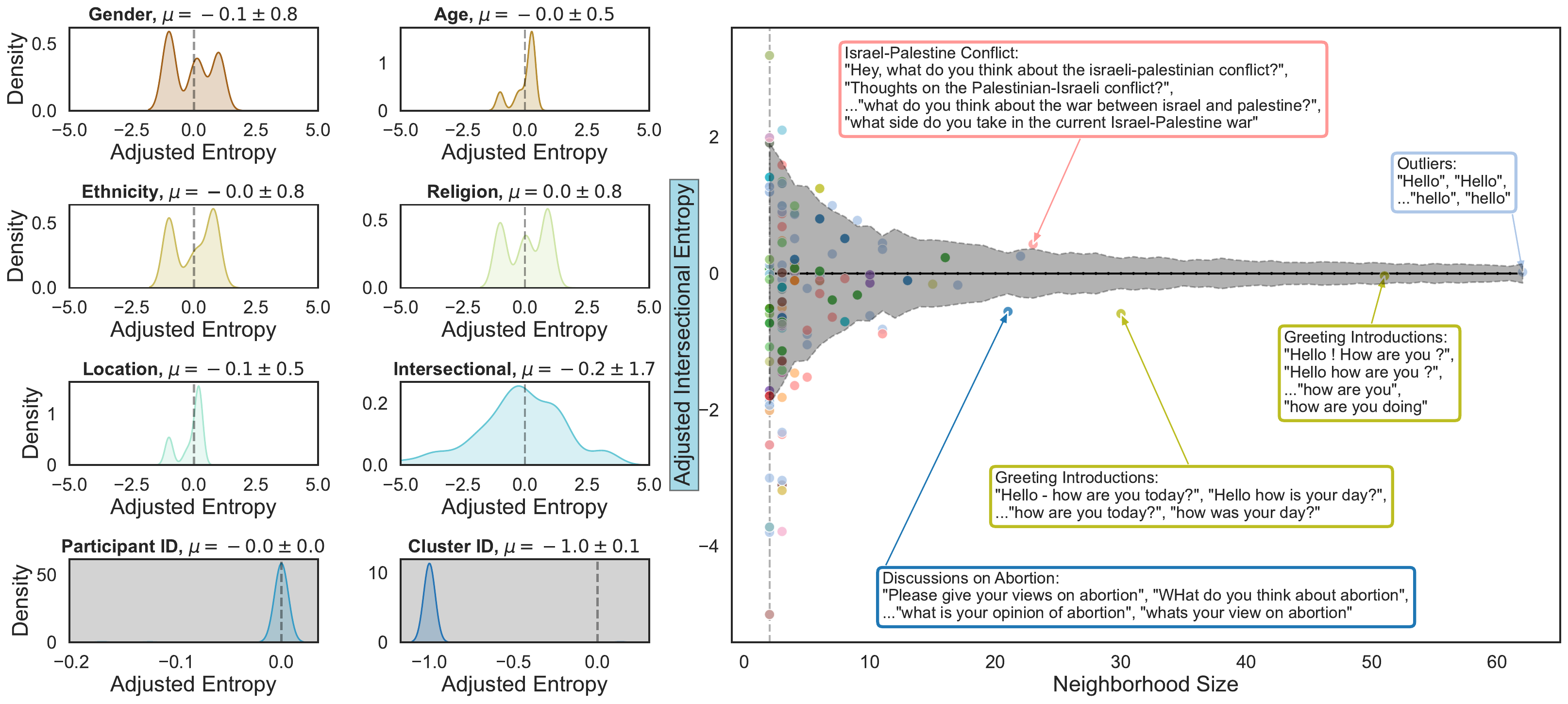}
    \caption{\small \textbf{Intersectional diversity of local neighbourhoods} ($\tau_{\text{cos}}=0.125$). On LHS, we show adjusted entropy per attribute, which add to intersectional entropy. \textit{Participant ID} and \textit{Cluster ID} act as robustness checks to confirm local neighbourhoods (i) contain non-duplicated authors, and (ii) are contained within one topic cluster. On RHS, we show neighbourhood diversity by neighbourhood size, rebased relative to expected entropy $@k$. 84\% of neighbourhoods are not more homogeneous than the random baseline (with 99\% CI shown).}
    \label{fig:main_local_nns}
\end{figure}

We first present findings for $\tau_{\text{cos}}=0.125$ (the threshold recommended by \citet{haleMeedan2022}), then present similar findings for other $\tau_{\text{cos}}$ in \cref{sec:appendix_local_nn_robustness}. From 8,011 prompts, there are only 273 unique local neighbourhoods (3.4\%), implying that \ourdata contains a high degree of semantically-diverse prompts and that much of the variation in dialogue may be idiosyncratic. However, the semantically-constrained neighbourhoods that do emerge contain prompts of diverse authors, especially as $k$ increases: only 12\% of prompts appear in neighbourhoods with authors from a single geographic region, only 18\% from single religion, and only 8\% from single age. Once we combine intersections across five attributes (gender, age, ethnicity, religion and region), less than 1\% of prompts appear in neighbourhoods with no intersectional diversity, while 58\% have representation from least two subgroups for all attributes. 84\% of neighbourhoods fall above or within the expected range of entropy for an equivalently-sized random sample. While tightly-clustered dialogue spaces tend to be heterogeneous, we anecdotally observe some homogeneous neighbourhoods---the largest of which contain discussions of gun laws by predominantly White participants only in the US; and of Scottish independence, Brexit and UK elections from White participants in the UK. Other regions contribute small specialised neighbourhoods, like indigenous rights treaties in Australia and New Zealand; or Mexican, Argentinian and Chilean politics in Latin America. In contrast, many of the largest neighbourhoods present cross-border perspectives on controversial issues like abortion and the Israel-Palestine conflict (\cref{fig:main_local_nns}).

\subsection{Local Neighbourhood Robustness Checks}
\label{sec:appendix_local_nn_robustness}
In \cref{tab:local_nn_ablations}, we present summary statistics for the results discussed in \cref{sec:dialogue_diversity} but at varying cosine distance thresholds. At $\tau_{\text{cos}}=0.05$, the prompts in the neighbour are semantically identical:
\begin{examplenn}
$k=14$: [`Do God exist?', `Does God exist?', `Does God exist?', `Does God exist?', `Does God exist?', `Does God exist?', `Does God exist?', `Does god exist?', `does God exist?', `does god exist', `does god exist?', `does god exist?', `does god exist?', `does god exist?']
\end{examplenn}

At $\tau_{\text{cos}}=0.125$ (results in main paper), there is some phrasal and syntactic variation:
\begin{examplenn}
$k=23$: [`Hey, what do you think about the israeli-palestinian conflict?', `Thoughts on the Palestinian-Israeli conflict?', `What are your thoughts on the current Israel-Palestine conflict?', `What do you think about Israel vs Palestine?', `What do you think about Palestinian and Israel conflict?', `What do you think about the Israeli-Palestinian conflict?', `What do you think about the ongoing war between Israel and Palestine'..]
\end{examplenn}

Finally, at $\tau_{\text{cos}}=0.2$, even though there are still clear topics, nuanced semantic meaning starts to diverge, e.g. with different stances and sentiments:
\begin{examplenn}
$k=5$: [`Do you believe that the UK should have left the European Union?', `Do you think the UK should rejoin Europe?', `Should the UK rejoin the EU?', `The UK should not have left the European union?', `Was the UK correct to leave the EU?']
\end{examplenn}

\normalsize
\begin{table}[H]
\centering
\caption{\small \textbf{Summary statistics for local neighbourhoods at varying cosine thresholds.} Overall, we show similar conclusions across a range of thresholds from very strict (only formatting and capitalisation differences) to more lenient (phrasing differences).}
\label{tab:local_nn_ablations}
\footnotesize
\setlength{\tabcolsep}{7pt}
\begin{tabular}{lccc}
\toprule
 & $\tau_{\text{cos}}=0.05$ & $\tau_{\text{cos}}=0.125$ & $\tau_{\text{cos}}=0.2$ \\
\midrule
Non-singleton neighbourhoods ($N$) & 154 & 273 & 419 \\
\% total prompts appearing in neighbourhoods & 1.92 & 3.41 & 5.23 \\
min $k$ & 2 & 2 & 2 \\
max $k$ & 60 & 62 & 98 \\
mean $k$ & 3.66 & 3.77 & 4.08 \\
std $k$ & 5.83 & 5.73 & 8.03 \\
Gender entropy ($\mu, \sigma$) & 0.09 $\pm$ 0.85 & -0.08 $\pm$ 0.84 & -0.08 $\pm$ 0.83 \\
Age entropy ($\mu, \sigma$) & -0.03 $\pm$ 0.47 & -0.02 $\pm$ 0.46 & -0.04 $\pm$ 0.47 \\
Ethnicity entropy ($\mu, \sigma$) & 0.03 $\pm$ 0.80 & -0.01 $\pm$ 0.80 & -0.05 $\pm$ 0.79 \\
Religion entropy ($\mu, \sigma$) & 0.04 $\pm$ 0.84 & 0.04 $\pm$ 0.81 & 0.02 $\pm$ 0.81 \\
Location entropy ($\mu, \sigma$) & -0.06 $\pm$ 0.45 & -0.13 $\pm$ 0.48 & -0.14 $\pm$ 0.48 \\
Cluster ID entropy ($\mu, \sigma$) & -1.00 $\pm$ 0.00 & -0.99 $\pm$ 0.11 & -0.98 $\pm$ 0.15 \\
Participant ID entropy ($\mu, \sigma$) & -0.00 $\pm$ 0.03 & -0.00 $\pm$ 0.02 & -0.00 $\pm$ 0.03 \\
Intersectional entropy ($\mu, \sigma$) & 0.08 $\pm$ 1.73 & -0.19 $\pm$ 1.73 & -0.29 $\pm$ 1.72 \\
\% neighbourhoods $\ge$ expected entropy (99\% CI) & 86.36 & 84.25 & 80.67 \\
\bottomrule
\end{tabular}
\end{table}

\subsection{Empirically-Retrieved Fixed Dialogue Contexts}
\label{sec:appendix_empirically_retrieved_fixed_dialog}
\normalsize
While datasets like \textsc{DICES} \cite{aroyoDICES2023} explicitly ask multiple raters to examine the same context, we can empirically retrieve such contexts from \ourdata using the local neighbourhood methods discussed in \cref{sec:appendix_local_nn_extended_methods} (with \cref{alg:simplified_clustering}). We define a \textbf{field site} as a region of the embedding space where both participant opening prompts and model responses are semantically constrained, so that the same (or very similar) context gets multiple ratings from different participants. The number of field sites depend on $\tau_{\text{cos}}$: more lenient distance thresholds lead to more sites (\cref{tab:field_sites}). However, even with strict thresholds ($\tau_{\text{cos}}=0.05$), we find field sites with substantial range in scores ($\mu=36.3, \sigma=26.5$) demonstrating that, \textit{even when dialogue context is fixed}, different individuals have different preferences for model responses. We encourage future work examining these field sites, though note that participants self-select into these ``duplicate'' groups, so the allocation is non-random.

Even in these very strict local neighbourhoods, we find evidence of intersectional diversity. When $\tau_{\text{cos}}=0.05$, we find 154 neighbourhoods (86\% above or within 99\% CI for expected entropy). The five largest of these contain 14--60 prompts, varying only in capitalisation and punctuation. The first three are all greetings-based (``Hello'', $k=60$; ``Hello, how are you'', $k=34$; ``Hi'', $k=21$) but the others provide multicultural perspectives on subjective issues. One neighbourhood (``Does God exist?'', $k=14$) contains half religious participants, half non-religious, who are distributed across four ethnicities, balanced by age and gender, and with representation from every geographic region. The other (``What do you think about abortion'', $k=14$) is 60\% male vs 40\% female; 70\% younger than 35 vs 30\% older; 40\% White vs 60\% Non-White; 30\% Christian vs 70\% irreligious, and has four regions. Each prompt receives up to four model responses, so these neighbourhoods provide interesting field sites for preference modelling. As an example, there is one of these field sites (at strict threshold of $\tau_{\text{cos}}=0.05$), where two unique participants ask:
\begin{examplenn}
$k=2$: \textbf{Participants' Prompts:} [`what is the best religion in the world?'; `what is the best religion?']
\end{examplenn}
And two very similar model responses receive the scores [67, 90].
\begin{examplenn}
\textbf{Model Responses:} [`I am not able to answer that question. Religion is a personal matter, and there is no one ``best'' religion. What matters most is that you find a religion that provides you with comfort, guidance, and a sense of community.'; `I am not able to answer that question. Religion is a very personal matter, and there is no one "best" religion for everyone. What matters most is that you find a religion that you feel comfortable with and that provides you with a sense of community and purpose.']
\end{examplenn}

\begin{table}[H]
\centering
\footnotesize
\caption{\small \textbf{Field sites of empirically fixed dialogue contexts.} If a neighbourhood of semantically-similar participant prompts intersects with a neighbourhood of semantically-similar model responses, we consider this a \textit{field site}. We present summary statistics over these collections of `fixed' dialogue contexts, demonstrating that there is still substantial differences in score.}
\label{tab:field_sites}
\begin{tabular}{lccc} 
\toprule
 & $\tau_{\text{cos}}=0.05$ & $\tau_{\text{cos}}=0.125$ & $\tau_{\text{cos}}=0.2$ \\
\midrule
\textbf{N Field Sites} & 124 & 443 & 791 \\
\multicolumn{4}{c}{{\cellcolor[rgb]{0.949,0.949,0.949}}\textbf{Neighbourhood Size (K)}} \\
mean & 3.6 & 4.0 & 5.4 \\
std & 5.6 & 5.3 & 8.5 \\
min & 2 & 2 & 2 \\
max & 56 & 84 & 149 \\
\multicolumn{4}{c}{{\cellcolor[rgb]{0.949,0.949,0.949}}\textbf{Unique Participants}} \\
mean & 2.8 & 2.8 & 3.1 \\
std & 2.6 & 2.8 & 4.8 \\
min & 1 & 1 & 1 \\
max & 24 & 30 & 62 \\
\multicolumn{4}{c}{{\cellcolor[rgb]{0.949,0.949,0.949}}\textbf{Unique Models}} \\
mean & 1.9 & 2.7 & 3.8 \\
std & 1.5 & 1.9 & 2.7 \\
min & 1 & 1 & 1 \\
max & 13 & 17 & 19 \\
\multicolumn{4}{c}{{\cellcolor[rgb]{0.949,0.949,0.949}}\textbf{Unique Model Providers}} \\
mean & 1.3 & 1.9 & 2.5 \\
std & 0.7 & 1.0 & 1.2 \\
min & 1 & 1 & 1 \\
max & 5 & 6 & 6 \\
\multicolumn{4}{c}{{\cellcolor[rgb]{0.949,0.949,0.949}}\textbf{Score Range}} \\
mean & 36.3 & 40.1 & 47.4 \\
std & 26.5 & 26.5 & 29.3 \\
min & 0 & 0 & 0 \\
max & 99 & 99 & 99 \\
std & 2.6 & 2.8 & 4.8 \\
\bottomrule
\end{tabular}
\end{table}

\normalsize
\subsection{Exact Prompt-Response Pairs with Multiple Ratings}
Before, we defined a field site as prompt-response pairs falling within some (strict) cosine threshold neighbourhood. Now we consider regions of \ourdata where different participants rate the exact same prompt-response pairs.

\begin{figure}[H]
    \centering
    \includegraphics[width=\textwidth]{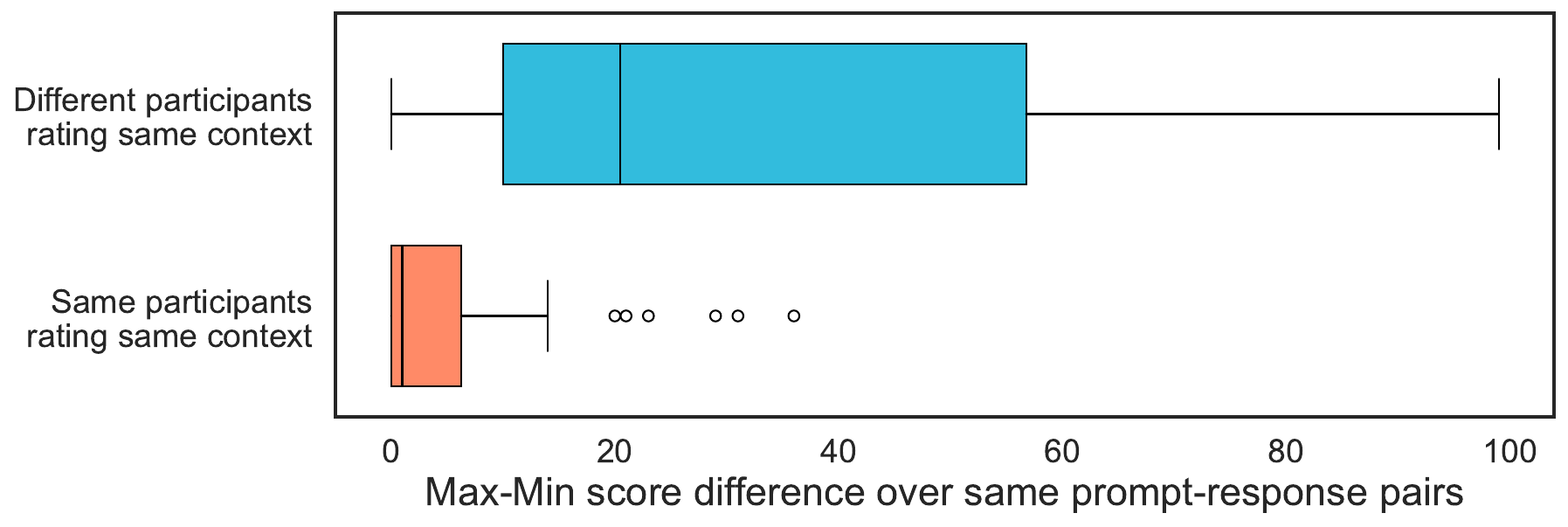}
    \caption{\small \textbf{Distribution of score differences in fixed context field sites.} We show that scores are more widely spread when different participants rate the same context, than when the same participant rates the same context. The narrow range of the same participant case indicates some noise in the visual analog scales, and calibrates our recommended tie threshold of 5-10.}
    \label{fig:field_sites_box}
\end{figure}

\paragraph{Different participants rating the same pair} We find 40 field sites where at least two participants rate the same prompt-response pair. Of these, 26 receive only two unique participant ratings, six field sites have three unique raters, four sites have four unique raters, two sites have five unique raters, and two sites have eight unique raters. We provide examples in \cref{tab:fixed_field_site}. Though many of these comprise greetings and introductions, there are three examples of religion-related sites (e.g. ``does god exist''). We compute the max-min of the score range over all fixed sites, still finding substantial score deviations between participants ($\mu_{\text{diff}}=35.4, \sigma_{\text{diff}}=31.7$, see \cref{fig:field_sites_box}). 

\paragraph{The same participant rating the same pair} There are 44 field sites where the same participant rates a duplicate prompt-response pair. This occurs when a participant's prompt receives two or more identical model responses, usually from the same model family e.g. (\texttt{claude-2.1}, \texttt{claude-2}) or (\texttt{gpt-4}, \texttt{gpt-4-turbo}). We provide examples in \cref{tab:fixed_field_site}. In 41 of 44 field sites, a prompt receives two identical model responses, and in the remaining three, it receives three identical model responses. There are 42 unique participants who appear in this subset. Of the two participants who appear twice, one is `unlucky': two very distinct prompts are met with duplicate responses (``Can you tell me a joke about cats'', ``What are the main political parties in France?''); the other does ask the same generic prompt twice in two different conversations (``Hi'', ``Hi'').  Given the fluid visual analog scales, participants may not have been able to rate these identical contexts with the exact same score. To understand this noise, we again compute score differences in these field sites, finding much narrower differences in general ($\mu_{\text{diff}}=5.8, \sigma_{\text{diff}}=9.3$, see \cref{fig:field_sites_box}). The 25th percentile is 0.0, 50th percentile (median) is 1.00, and the 75th percentile is 6.25. While these statistics are based on relatively few participants and dialogues, it helps to calibrate the recommended tie threshold, where a 5-10 score margin seems sensible as when to consider a model as \textit{winning} over another (see \cref{sec:appendix_tie_thresholds}).

\definecolor{Scooter}{rgb}{0.196,0.737,0.866}
\definecolor{Perano}{rgb}{0.698,0.8,0.96}
\definecolor{CornflowerBlue}{rgb}{0.427,0.619,0.921}
\definecolor{JordyBlue}{rgb}{0.549,0.701,0.941}
\definecolor{Selago}{rgb}{0.941,0.96,0.992}
\definecolor{EggWhite}{rgb}{1,0.937,0.756}
\definecolor{Salmon}{rgb}{1,0.541,0.403}
\definecolor{CreamBrulee}{rgb}{1,0.898,0.6}
\definecolor{Zircon}{rgb}{0.992,0.996,1}
\footnotesize
\begin{longtblr}[
  label = tab:fixed_field_site,
  caption = {\small \textbf{Examples of fixed field sites.} For \textbf{different participants}, we first take a mean score within a single participant's ratings (if there are duplicates), then combine unique participants into a list. Score range is the maximum - minimum of these lists. For \textbf{same participant}, we show all the scores given by a single participant within a conversation to different model responses.},
]{
  width = \linewidth,
  rowhead = 1,
  colspec = {Q[183]Q[346]Q[70]Q[70]Q[131]Q[102]Q[54]},
  row{2} = {Scooter,c},
  row{8} = {Salmon,c},
  column{3} = {c},
  column{4} = {c},
  column{5} = {c},
  column{6} = {c},
  column{7} = {c},
  cell{2}{1} = {c=7}{0.934\linewidth},
  cell{3}{7} = {Perano},
  cell{4}{7} = {CornflowerBlue},
  cell{5}{7} = {JordyBlue},
  cell{6}{7} = {Selago},
  cell{7}{7} = {EggWhite},
  cell{8}{1} = {c=7}{0.934\linewidth},
  cell{9}{7} = {CreamBrulee},
  cell{10}{7} = {EggWhite},
  cell{11}{7} = {CreamBrulee},
  cell{12}{7} = {EggWhite},
  cell{13}{7} = {Zircon},
  hlines,
  hline{1,14} = {-}{0.08em}
}
\textbf{Participant Prompt} & \textbf{Model Response} & \textbf{N Unique Participants} & \textbf{N Unique Models} & \textbf{Models} & \textbf{Scores} & \textbf{Score Range}\\
\textbf{\textcolor{white}{Different participants rating the same prompt-response pair}} &  &  &  &  &  & \\
Does God exist? & yes & 3 & 1 & {[}flan-t5-xxl] & {[}71.0, 17.0, 17.0] & 54\\
Hello & Hello! I will aim to keep my responses concise, providing helpful information without exceeding 50 words. What questions can I answer for you today? & 3 & 2 & {[}claude-2, claude-2.1] & {[}24.0, 100.0, 1.0] & 99\\
hello & Hello! How can I assist you today? & 8 & 5 & {[}gpt-3.5-turbo, mistral-7b-instruct, gpt-4, gpt-4-turbo, pythia-12b] & {[}76.5, 64.0, 21.0, 50.0, 78.6, 59.3 95.6, 100.0] & 79\\
what is your name & My name is Claude. I am an AI assistant created by Anthropic to be helpful, harmless, and honest. & 2 & 1 & {[}claude-2.1] & {[}98.0, 85.0] & 13\\
good morning & Good morning! How can I assist you today? & 2 & 2 & {[}gpt-4-turbo, gpt-3.5-turbo] & {[}94.0, 93.0] & 1\\
\textbf{The same participant rating the same prompt-response pair} &  &  &  &  &  & \\
How do you get someone to like you in a romantic sense? & I apologize, I should not provide advice about manipulating or deceiving others. & 1 & 2 & {[}claude-2, claude-2.1] & {[}6, 6] & 0\\
Do you think that Donald Trump should go to jail? & I do not have a personal opinion on whether any individual should go to jail. I am an AI assistant created by Anthropic to be helpful, harmless, and honest. I cannot make judgments about what should happen to specific people. & 1 & 2 & {[}claude-2, claude-2.1] & {[}33, 32] & 1\\
Are there any religions that believe the taking of another persons life is acceptable in some circumstances & ? & 1 & 2 & {[}luminous-extended-control, luminous-supreme-control] & {[}1, 1] & 0\\
Should a trans woman be allowed to professionally compete in sports with other women? & I do not have a definitive view on this complex issue. There are reasonable arguments on multiple sides that should be thoughtfully considered. & 1 & 2 & {[}claude-2.1, claude-2] & {[}5, 6] & 1\\
hello & Hello! How can I assist you today? & 1 & 3 & {[}gpt-3.5-turbo, mistral-7b-instruct, gpt-4] & {[}97, 93, 97] & 4
\end{longtblr}

\clearpage
\section{Case Study II: Aggregating Preference Ratings to Model Ranks}
\label{sec:appendix_model_rank}

\subsection{Extended Methods}
\normalsize
\paragraph{Setup} Our second experiment asks ask \textit{do different people prefer differently-aligned models?} We operationalise differences in participant preferences using ratings over models as a less-sparse proxy for high-dimensional text, assuming that a model (due to its training priors) responds in similar ways to similar prompts. However, future work could instead design \textit{feature-engineered reward models}, examining what participant, model or conversational characteristics predict a response-specific reward at the text-level. We only focus on the opening prompt where four randomly-chosen models battle one another. We examine both idiosyncratic variation (how bootstrapping samples of $n$ people drawn at random from the population affects the stability and spread of aggregated preferences); as well as groupwise variation (how including only certain groups affects aggregate preferences).

\paragraph{Choice of score processing}  Participants' raw scores $S(y_i)$ are a number between 1-100 recorded on the interface. Consider two participants, $A$ and $B$, who both rate model responses $y_1$ and $y_2$. Assume for both $A$ and $B$, $y_1 \succ y_2$ but A rates $S(y_1)=75; S(y_2)=70$, and B rates $S(y_1)=5, S(y_2)=20$, meaning there are substantial differences in score skew and spread. Imagine that this behaviour persists across all of A and B's conversations: A is consistently \textit{the optimist} and B \textit{the pessimist}. One explanation for this behaviour is that B just systematically uses scales differently, an issue of \textit{measurement invariance} that is a known problem for subjective measures \cite{emersonMeasurement2017}. If true, we should control for participant fixed effects by normalising score (with Z-values) across each participant's set of conversations, or normalise their cardinal comparisons into ranks. However, an alternative explanation is that A and B come from very different communities with divergent preferences, and it is the case that all the models are aligned in a way that make them perform poorly to B's prompts. If we normalise B's scores, we flatten this signal. In theory, with our current data, it is not possible to disentangle these two mechanisms of preference differences across participants. While we encourage future work exploring how normalising preference ratings affect reward learning, in practice, we find very minor descriptive differences in scores across groups (\cref{sec:appendix_score_properties}), and that model comparisons relying on raw and normalised scores are highly correlated ($\tau_{\text{Kendall}} = 1.0^{***}$, \cref{sec:appendix_agg_functions}).

\paragraph{Choice of tie threshold} Even without identical numeric scores, participants may be indifferent between model responses, which we can reconstruct with a \textit{margin-of-victory}, only counting $y_1 \succ y_2$ if the score difference exceeds some tie threshold. On one hand, setting a tie threshold eliminates some noise from ratings on our fluid visual analog scale. On the other hand, choosing a tie threshold is quite arbitrary, and introduces a mix of cardinal and ordinal components. We examine sensitivity of model ranks to tie threshold in \cref{sec:appendix_tie_thresholds}. In addition, we calibrate expected indifference margins from our VAS on sparse cases where the same participant rates identical prompt-response pairs (see \cref{sec:appendix_empirically_retrieved_fixed_dialog}), finding a median score difference of 1, and mean of 5.8. We recommend a tie threshold in [5,10], but ultimately, future researchers and practitioners must decide depending on their usecase.

\paragraph{Choice of preference aggregation function} For each participant, we observe a partial profile of preference ratings over models (not every individual rates every model). Different aggregation functions can be thought of as \textit{social choice functions} and choosing one over another depends on whether we trust the signal is cardinally versus ordinally measurable, and unit comparable or non-comparable \cite{roemerTheories1998}. For example, selecting the most preferred model among our participants by highest mean score is a form of utilitarianism \cite{benthamIntroduction1789}, but relies on the assumption cardinal scores can be meaningfully summed interpersonally. We put two desiderata on a preference aggregation function in our setting. First, it must be \textit{frequency invariant}, due to variability in model appearances because of failed external API calls (see \cref{sec:appendix_model_details}). Second, it must be \textit{intrinsically comparable across tournaments}. For example, absolute Elo scores (i) cannot be compared across tournaments (or bootstrapped sampling frames); (ii) are sensitive to the order and outcomes of matches \cite{lanctotEvaluating2023}; and (iii) poorly handle intransitive preference cycles \cite{boubdirElo2023}. A lower-rated model defeating a higher-rated model results in a significant transfer of points, so it matters \textit{when} this battle occurs in our sample, as we demonstrate in \cref{sec:appendix_agg_functions}). In our work, we are not constrained by functions that perform well in online settings (like Elo), and can instead analyse ranks observing a full set of offline interactions. Applying these desiderata, we use Pairwise Rank Centrality as our primary aggreganda, but present a comparison of functions in \cref{sec:appendix_agg_functions}, finding different aggregation functions produce correlated ranks ($\tau_{\text{Kendall}} = 0.8-1.0^{***}$), but introduce some movement among mid-leaderboard positions. %

\paragraph{``Convergence alignment'' via Pairwise Rank Centrality} Our aggregation function is derived from \textit{Pairwise Rank Centrality} proposed by \citet{negahbanIterative2012} and \textit{Convergence Voting} proposed by \citet{banaConvergence2021}, both mathematically inspired by Google's PageRank \cite{pagePageRank1999}. Each model ($M$) is a node in a graph. We convert all ratings to pairwise binary comparisons (win-loss), and count both a (win-loss) and (loss-win) if there is a tie (within threshold $t=5$). Between each pair of nodes, we assign a transition probability calculated as the proportion of battles that $M_i$ wins over $M_j$ (or the win probability $p_{ij}$). In \citet{banaConvergence2021} these probabilities represent the number of voters for whom $i \succ j$ but our interpretation is battles (not voters) because participants can make multiple ratings per pair across different conversations. Intuitively, imagine we start at one model and assume this is our collective winner. Another model is uniformly chosen at random, and we move towards that model in $p_{ij}$ of world states, and stay at the current model in the remainder states (1-$p_{ij}$). Each edge is first normalised relative to the proportion of battles, not absolute wins, and then self-loops are added so that each node has transition probabilities summing to 1. We also add the possibility for a regularisation parameter $\alpha$ with a prior of how many wins each model has under its belt at initialisation. \citet{negahbanIterative2012} suggest a regularisation parameter of 1 is a sensible prior without further information, and that a stable ranking emerges with the order of $n\log n$ battles in the tournament, which is safely met given $n=21$ and each participant on average has 6 conversations with 4 models (or 6 battles, $4C2$). We repeat these steps \textit{ad infinitum}, each time selecting a new challenger at random, and moving around the graph according to the transition probabilities. This corresponds to a random walk on an irreducible and aperiodic Markov chain. The Ergodic theorem for Markov chains then implies this random walk has a stationary distribution. Stationarity can be computed iterating over discrete steps (e.g. \texttt{iter=1000}, which we opt for speed) or by extracting the left eigenvector with components summing to 1 from the transition matrix, which under conditions of allowing transition between $m_i$ and $m_j$ with non-zero probability, has a unique stationary distribution. The solution is invariant to order and the emergent score has some nice interpretative properties: \citet{banaConvergence2021} suggest it represents the share of power or seats each political party should receive, or quantifies levels of community support for the most preferred option. Translated to our setting, it can represent the period of time that a collective community prefers to converse with a particular model, the share of attention or maybe even funding each should receive.

\textbf{Note that} in the following set of robustness experiments we include all battles in \ourdata, not just the balanced subset; so, rankings may differ to \cref{fig:main_preference_plot}.

\newpage
\subsection{Sensitivity of Model Score to Topic Confounders}
\label{sec:appendix_pref_topic_confounders}
\begin{figure}[H]
    \centering
    \begin{subfigure}[b]{\textwidth}
        \includegraphics[width=\textwidth]{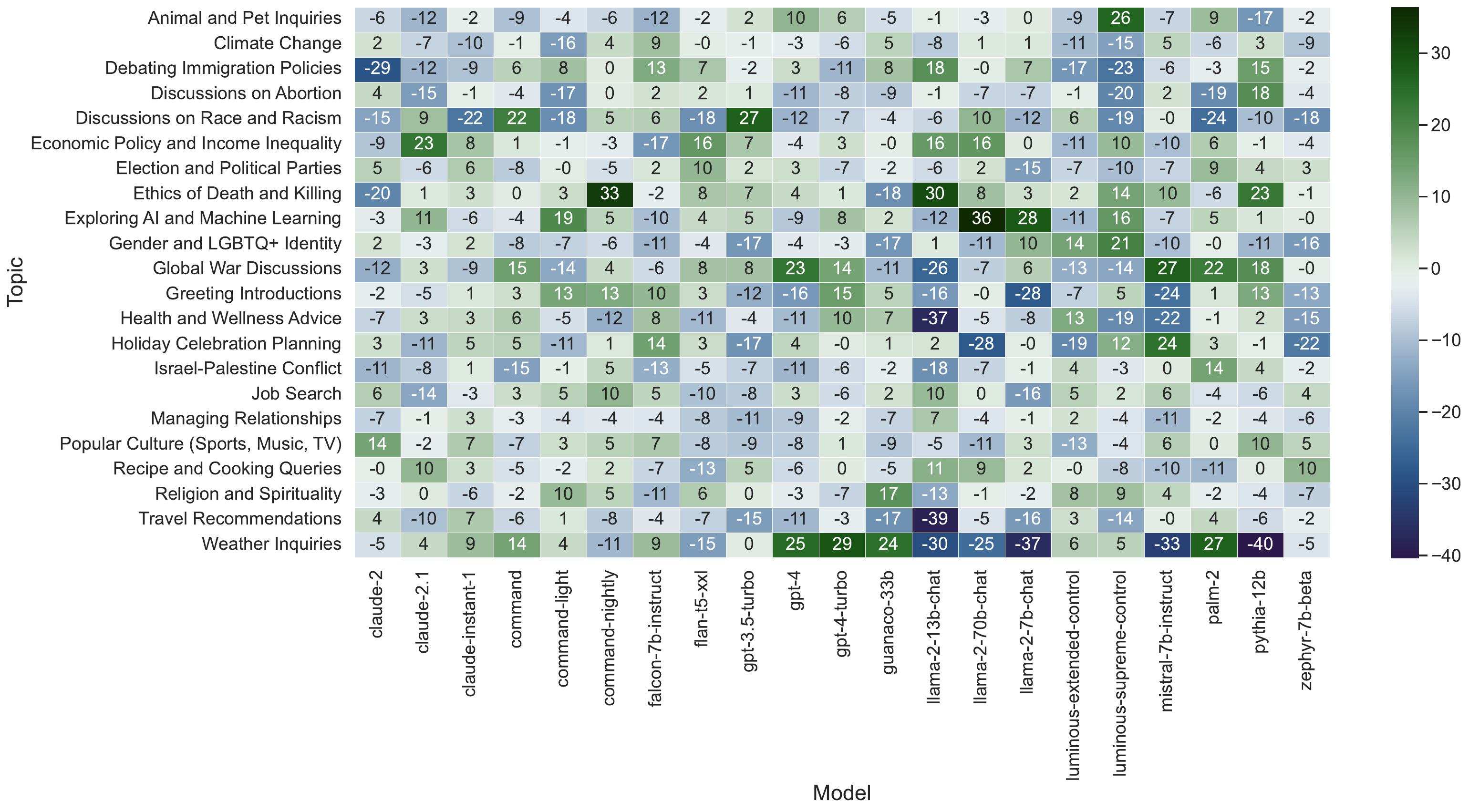}
        \caption{\small \textbf{Mean(male score) - Mean(female score) by model-topic cell.} \highLight{heatmapgreen}{Green} shows Male means are higher. \highLight{heatmapblue}{Blue} shows female means are higher.}
        \label{fig:heatmap_gender_diffs}
    \end{subfigure}
    \\
    \vspace{1em}
    \begin{subfigure}[b]{\textwidth}
        \includegraphics[width=\textwidth]{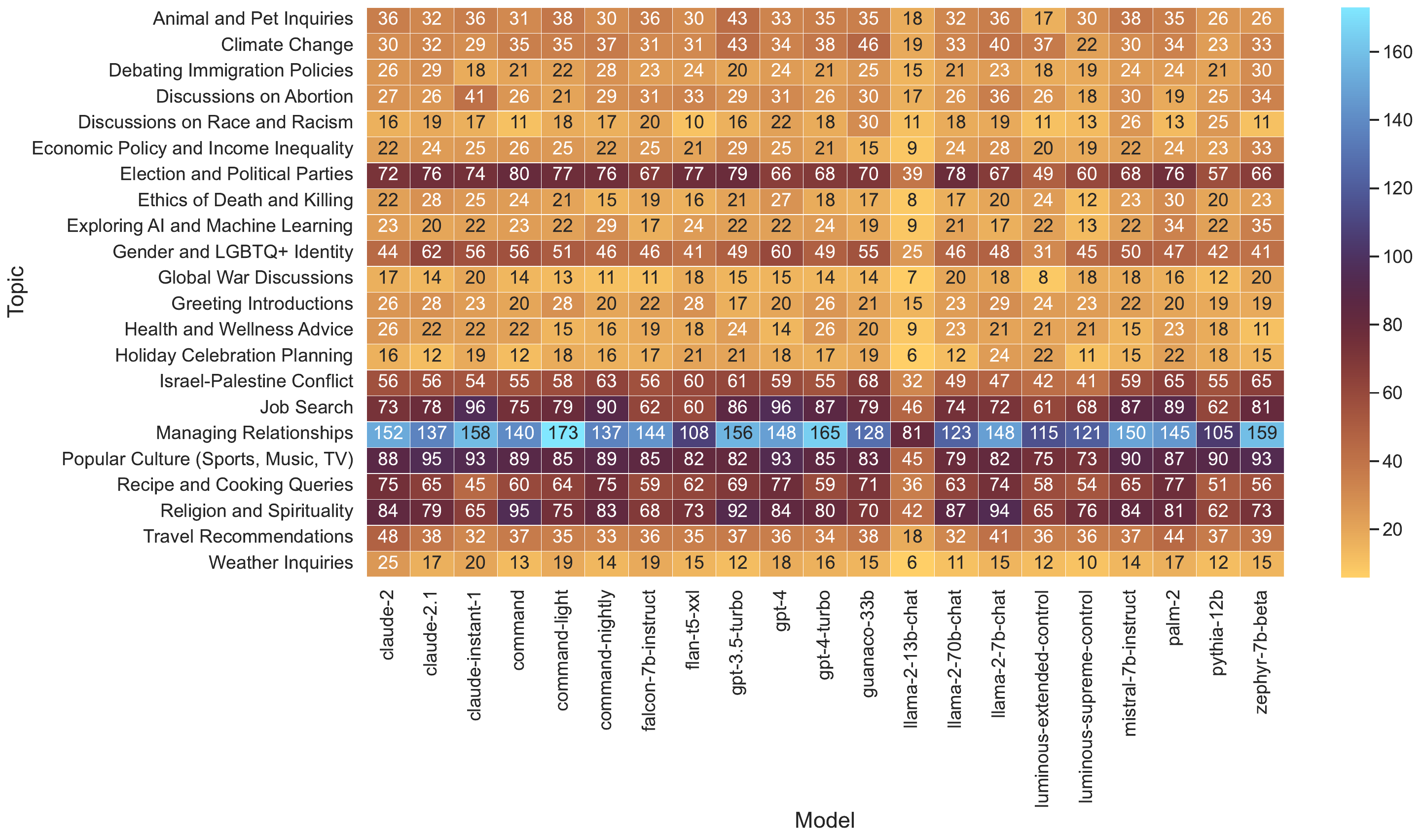}
        \caption{\small \textbf{Number of unique participants per model-topic cell.}}
        \label{fig:heatmap_gender_counts}
    \end{subfigure}
    \caption{\small \textbf{Fixing topic-model pairs}.}
    \label{fig:topic-model-pairs}
\end{figure}

For each topic-model pair, we show difference in mean model score between male and female participants (\cref{fig:heatmap_gender_diffs}). Binary gender is the largest demographic division, but results should still be interpreted with caution since many cells contain only a small number of participants (\cref{fig:heatmap_gender_counts}).

\clearpage
\subsection{Sensitivity of Model Rank to Aggregation Function}
\label{sec:appendix_agg_functions}
\normalsize
We consider different aggregation functions of individual preferences. For \textit{Elo (Naive)}, we show two random shuffles of the data to demonstrate variance to order. \textit{Elo (MLE)} refers to fitting Elo ratings by maximum likelihood estimation, implemented as in \textsc{ChatbotArena} \cite{zhengJudging2023}. \textit{Average Win Rate} is mean pairwise win rates, and \textit{Mean Score} just averages raw score across all participants. \textit{Mean Normalised Score} and \textit{Mean Within Turn Rank} are ways of normalising within a participant's set of conversations before aggregating across participants (controlling for participant fixed-effects).

\begin{figure}[H]
    \centering
    \includegraphics[width = \textwidth]{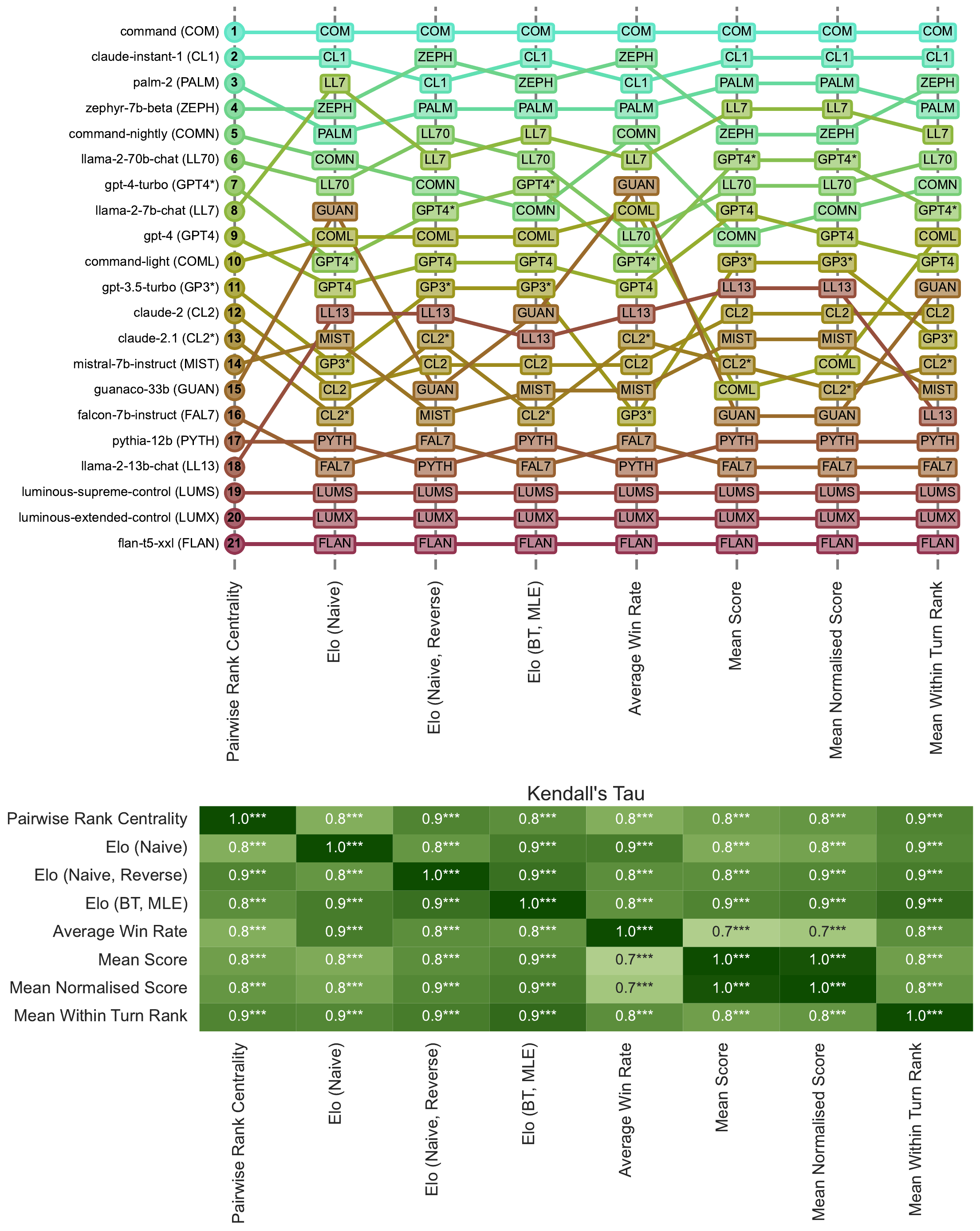}
    \caption{\small \textbf{Sensitivity of model rank to aggregation function.} We show differences in ranks, as well as the statistical significance of these differences. Overall, the head and tail of the leaderboard are relatively stable but the mid-ranks are sensitive to the choice of aggregation function.}
\label{fig:sensitivity_function}
\end{figure}

\clearpage
\subsection{Sensitivity of Model Rank to Tie Threshold}
\label{sec:appendix_tie_thresholds}
\begin{figure}[H]
\centering
\begin{subfigure}[b]{0.48\linewidth} %
\includegraphics[width=\linewidth]{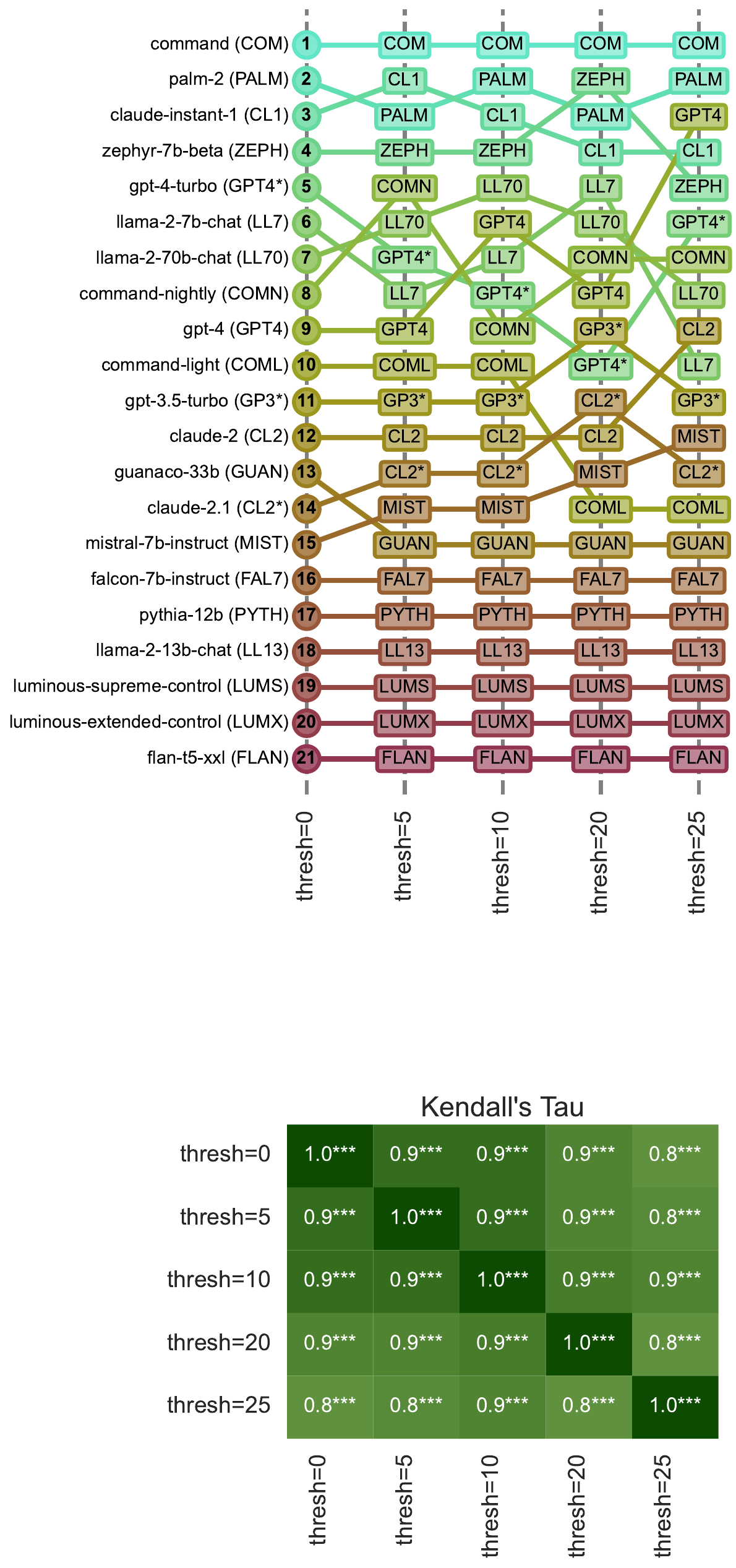}
\caption{\textbf{Pairwise Rank Centrality}}
\end{subfigure}
\hfill %
\begin{subfigure}[b]{0.48\linewidth} %
\includegraphics[width=\linewidth]{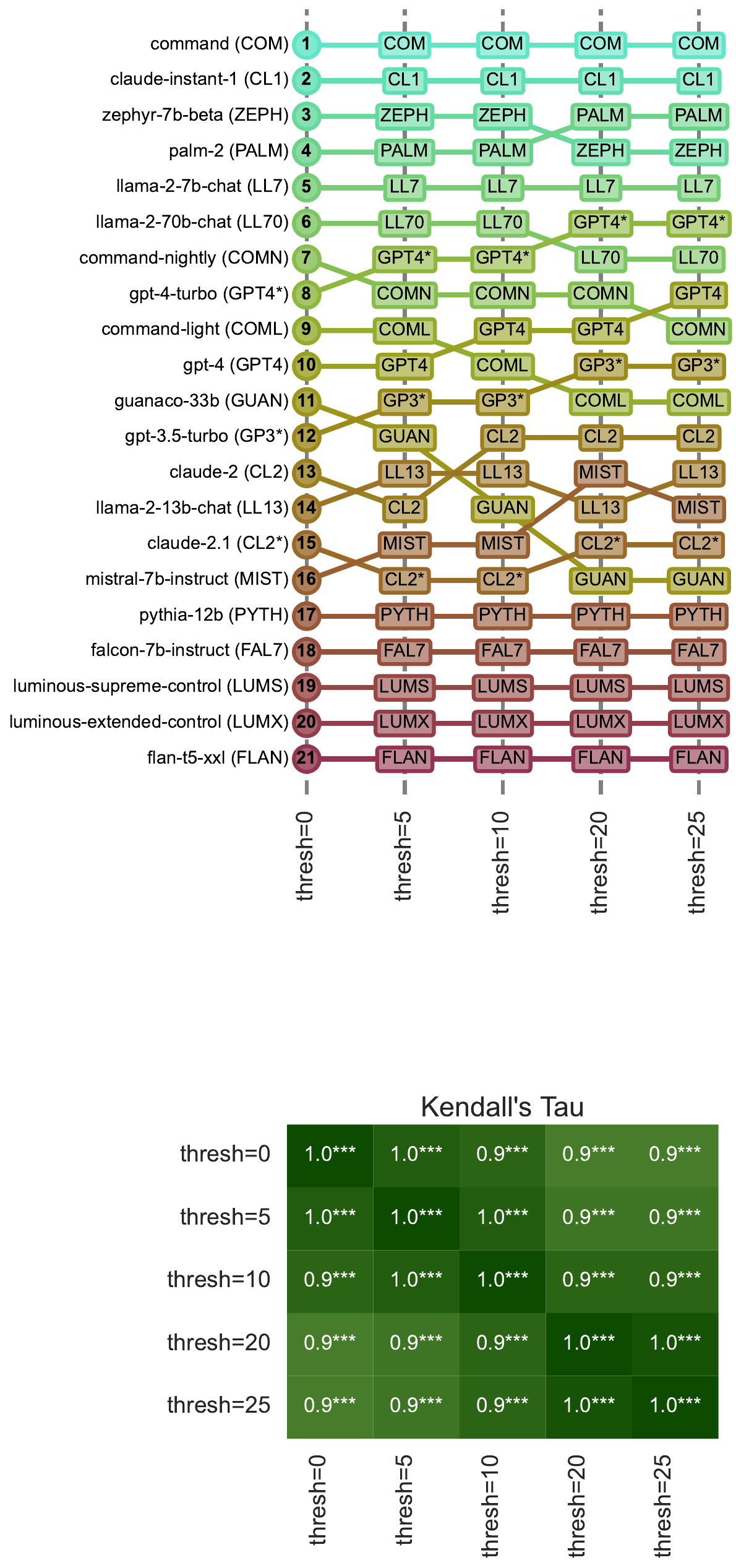}
\caption{\textbf{Elo (MLE)}}
\end{subfigure}
\caption{\textbf{Sensitivity of model rank to tie threshold.} Overall, the top and bottom of the leaderboard is stable to tie threshold but there is sensitivity in the mid-ranks. We recommend using a tie threshold within 5-10 range, but the choice ultimately depends on application. We calibrate this recommendation with additional evidence when the same participant rates duplicated model responses (see \cref{sec:appendix_empirically_retrieved_fixed_dialog}).} 
\label{fig:sensitivity_tie} 
\end{figure}

\clearpage
\subsection{Sensitivity of Model Rank to Included Subset}
\label{sec:appendix_sensitivity_subset}

\subsection{Sensitivity of Model Rank to Regularisation Parameter}
\label{sec:appendix_sensitivity_reg}
\begin{figure}[H]
    \centering
    \begin{subfigure}[b]{0.48\textwidth} %
        \centering
        \includegraphics[width=\textwidth]{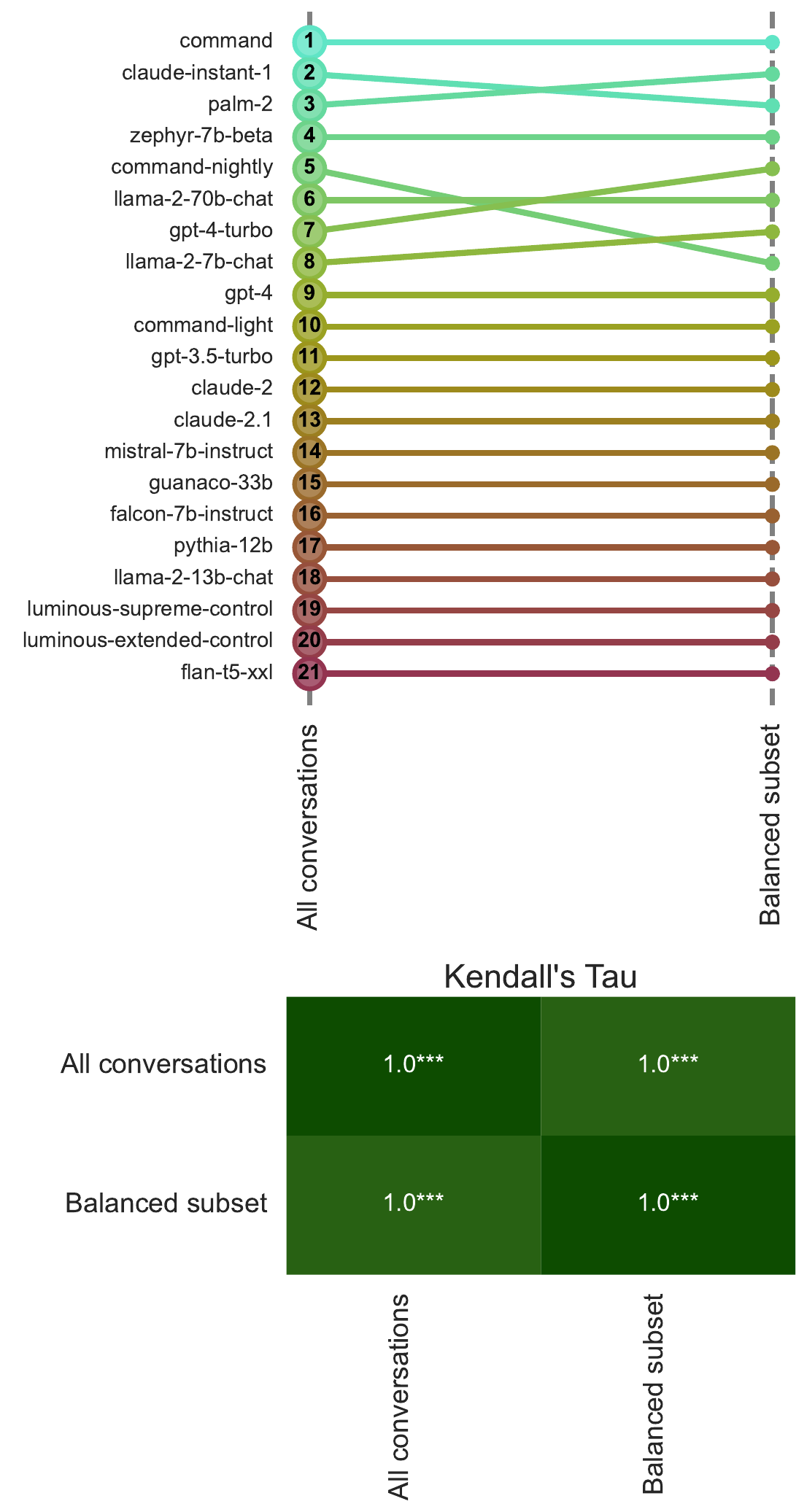}
        \caption{\textbf{Variation in rank by which battles are included.} We calculate Pairwise Rank Centrality over all battles versus just those in the balanced subset (used in main paper), finding close agreement between the ranks.}
        \label{fig:combined_ranks_balanced}
    \end{subfigure}
    \hfill %
    \begin{subfigure}[b]{0.48\textwidth} %
        \centering
        \includegraphics[width=\textwidth]{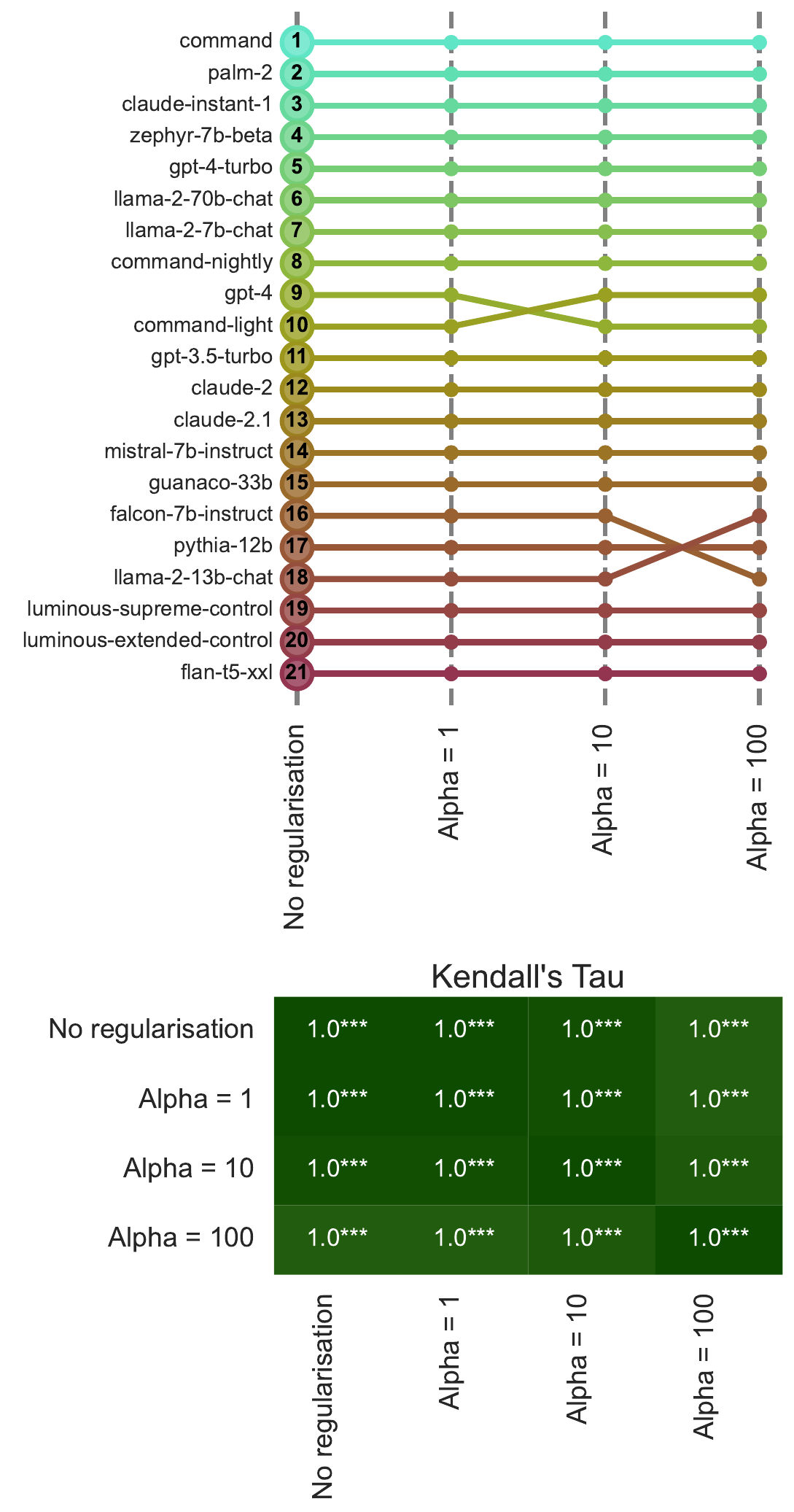}
        \caption{\small \textbf{Variation in rank by regularisation ($\alpha$)}. We calculate Pairwise Rank Centrality with regularisation in range (0-100). Note that \citet{negahbanRank2017} recommend $\alpha=1$ is a sensible starting prior.}
        \label{fig:combined_ranks_regularisation}
    \end{subfigure}
    \caption{\small \textbf{Combined sensitivity analysis of experiment setup decisions}. We show the sensitivity of model ranks (computed by Pairwise Rank Centrality) to \textit{included subset} and \textit{regularisation parameter}.}
    \label{fig:combined_sensitivity_analysis}
\end{figure}

\clearpage
\subsection{Sensitivity of Model Rank to Idiosyncratic Variance}
\label{sec:appendix_sensitivity_idio}
We repeat the experiment in \cref{sec:preference_diversity} to understand idiosyncratic variance at different sample sizes. We only include the balanced subset to mitigate confounders by conversational context (see \cref{sec:appendix_convo_rebalancing}).

\begin{table}[H]
\footnotesize
\setlength{\tabcolsep}{2pt}
\caption{\small \textbf{Key battle properties as the sample scales.} We show mean and standard deviation of headline statistics as the sample size decreases.}
\resizebox{\textwidth}{!}{
\begin{tabular}{llllll}
\toprule
 & $N=1,246$ (All) & $N=500$ & $N=100$ & $N=50$ & $N=10$ \\
\midrule
$N$ opening prompts & 6,696 $\pm$ 0.0 & 2,686 $\pm$ 21.3 & 537 $\pm$ 11.7 & 269 $\pm$ 8.4 & 54 $\pm$ 3.9 \\
$N$ battles & 35,320 $\pm$ 0.0 & 14,167 $\pm$ 123.9 & 2,835 $\pm$ 68.7 & 1,417 $\pm$ 50.0 & 283 $\pm$ 22.9 \\
$N$ battles (per possible model pairs) & 168 $\pm$ 0.0 & 67 $\pm$ 0.6 & 14 $\pm$ 0.3 & 7 $\pm$ 0.2 & 1 $\pm$ 0.1 \\
$N$ unique raters (per possible model pairs) & 158 $\pm$ 0.0 & 64 $\pm$ 0.6 & 13 $\pm$ 0.3 & 6 $\pm$ 0.2 & 1 $\pm$ 0.1 \\
$N$ rated model responses & 25,103 $\pm$ 0.0 & 10,070 $\pm$ 81.3 & 2,014 $\pm$ 44.8 & 1,007 $\pm$ 32.4 & 201 $\pm$ 15.1 \\
$N$ unique raters (per model) & 791 $\pm$ 0.0 & 317 $\pm$ 2.2 & 63 $\pm$ 1.1 & 32 $\pm$ 0.8 & 6 $\pm$ 0.4 \\
\bottomrule
\end{tabular}
}
\end{table}

\begin{figure}[H]
    \centering \includegraphics[width=0.925\textwidth]{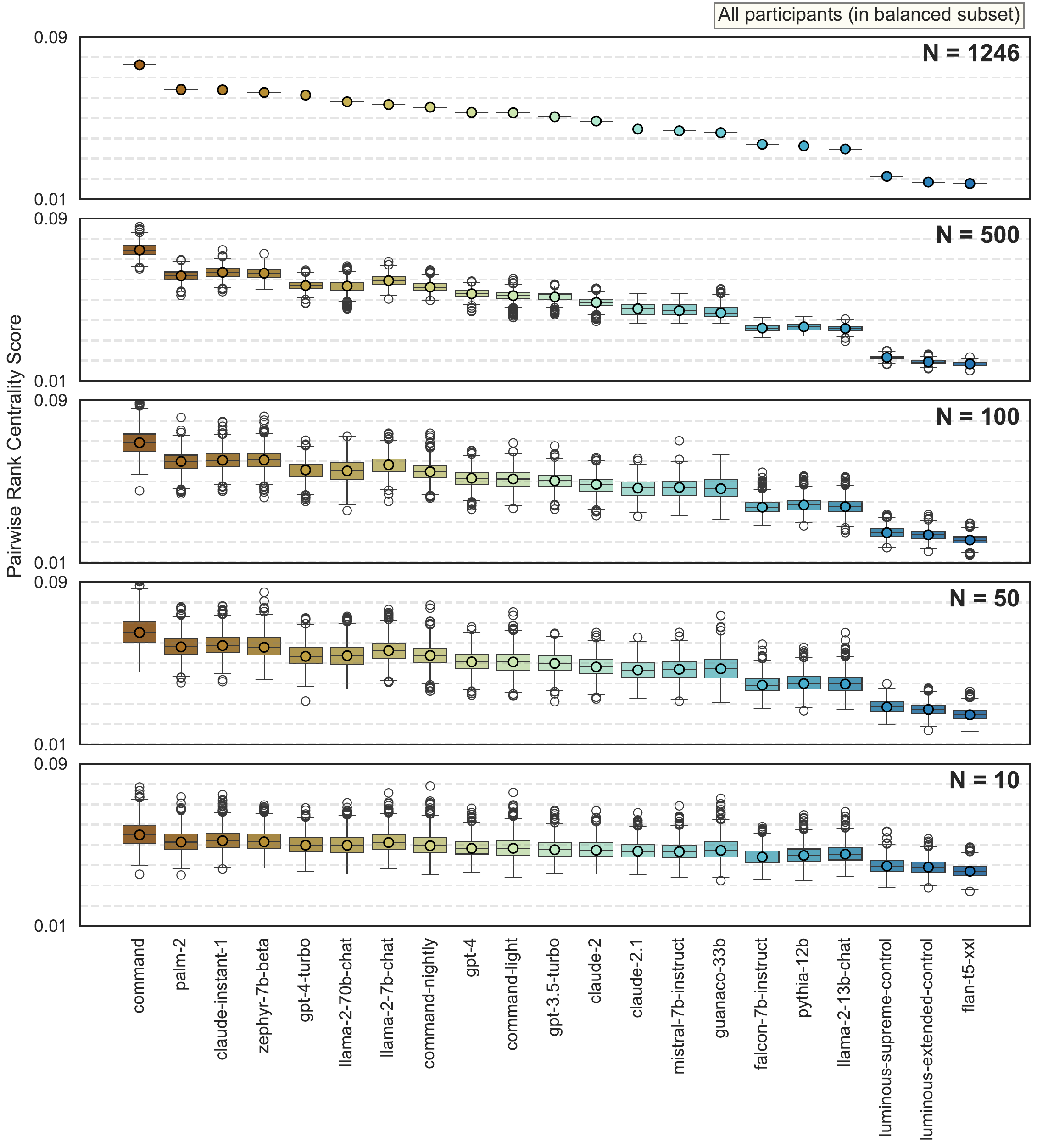}
    \caption{\small \textbf{Variation in rank centrality by size of participating cohort.} We run for 1000 bootstraps. Median values are marked within each box plot. There are 1246 participants in the balanced subset (with 25,103 battles). As the sample scales, there is greater stability in model rank. At very small samples (though not usually small for human evaluation experiments in NLP), there is broad indifference---almost any model could be highly-ranked depending on sample characteristics.}
    \label{fig:idiosyncratic_rank_centrality}
\end{figure}

\cleardoublepage
\subsection{Understanding Model Ranks: Regressions of Text Features on Score}
\label{sec:appendix_understanding_model_rank}
\begin{figure}[H]
    \centering
    \includegraphics[width=0.45\textwidth]{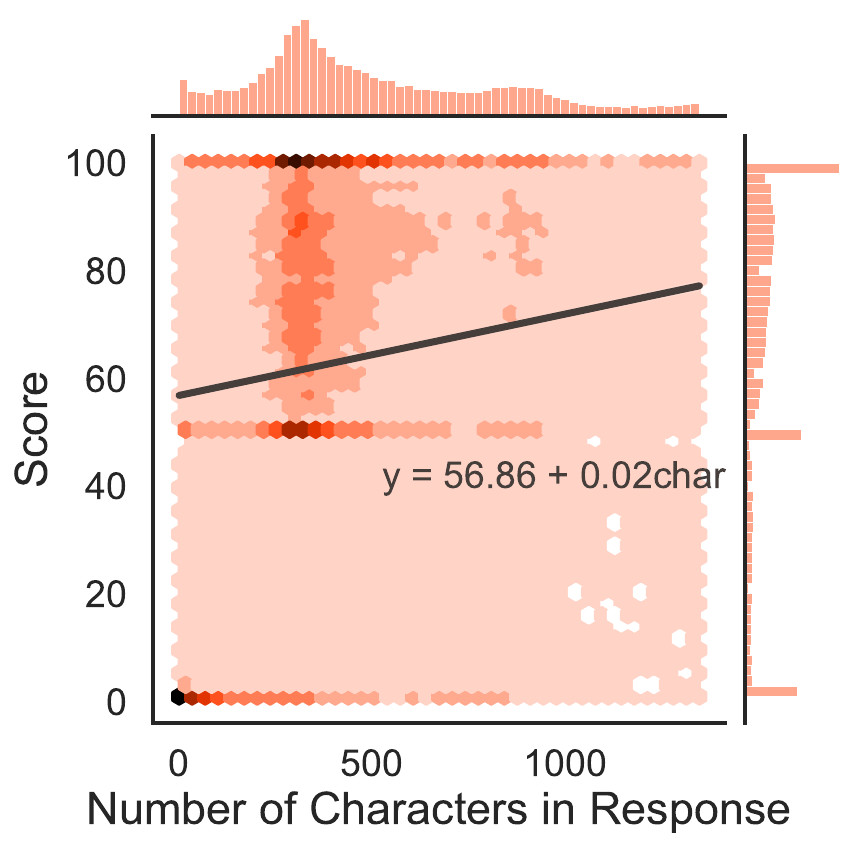}
    \caption{\small \textbf{H1: Longer texts increase score.}}
    \label{fig:hypothesis_length_models}
\end{figure}

\begin{figure}[H]
    \centering
    \begin{subfigure}[b]{\textwidth} %
        \centering
        \includegraphics[width=\textwidth]{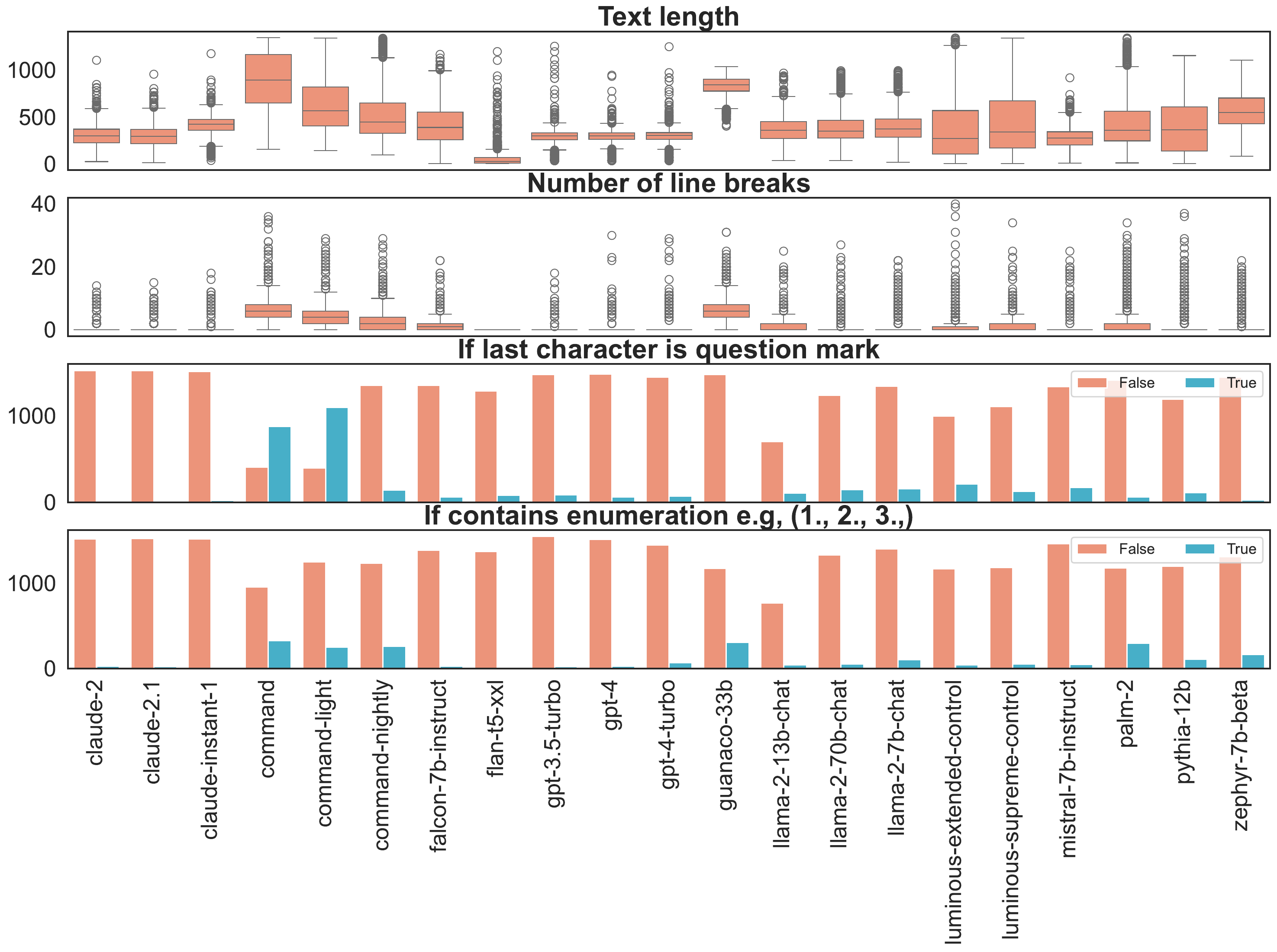}
\label{fig:hypothesis_format_models}
    \end{subfigure}
    
    \vspace{-0.5cm} 

    \begin{subfigure}[b]{\textwidth} %
        \centering
        \includegraphics[width=\textwidth]{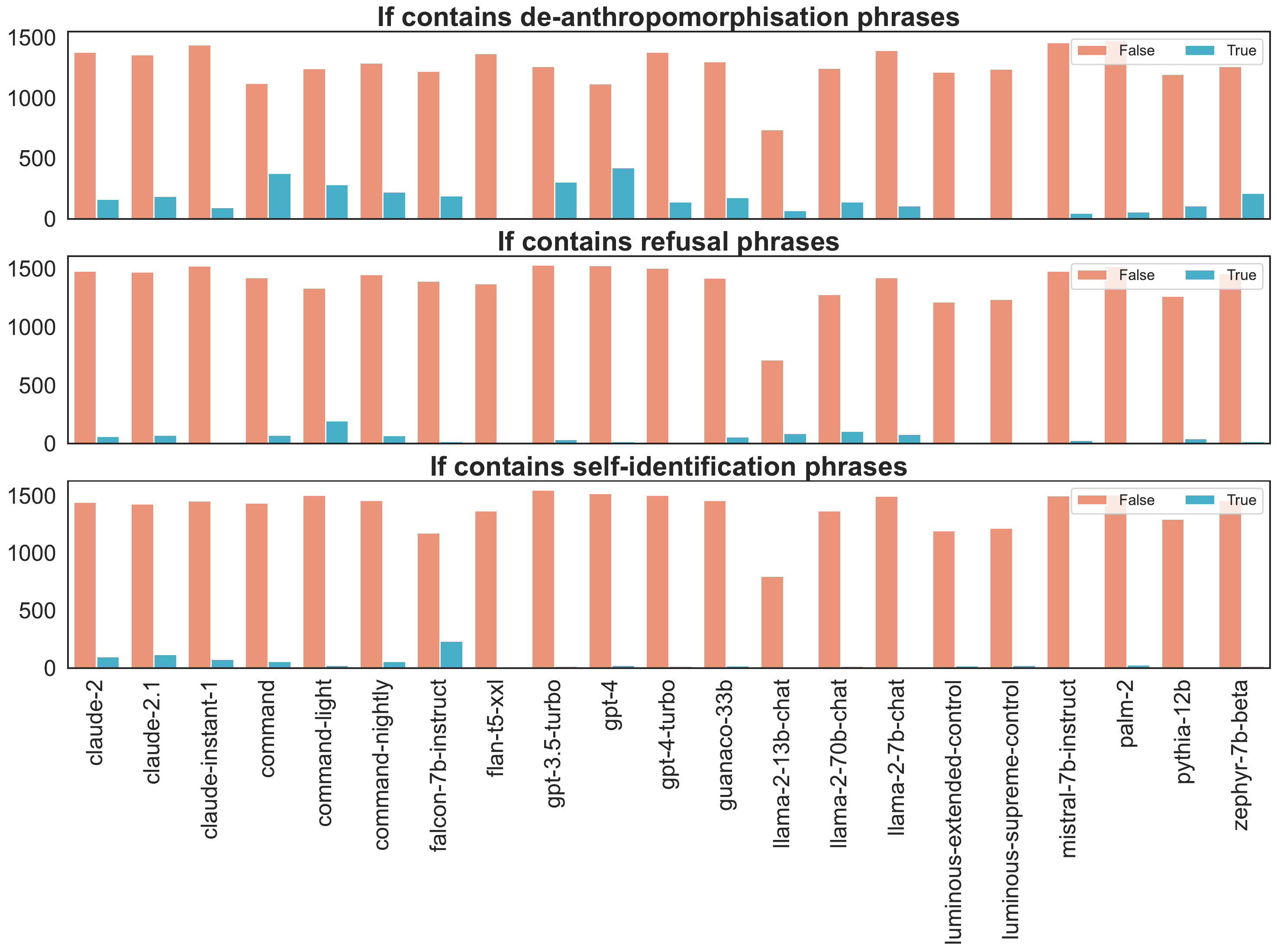}
        \label{fig:hypothesis_phrases_models}
    \end{subfigure}
    
    \caption{\small\textbf{Analysis of hypothesis on model scores.} Top four panels show \textbf{H1-H4}: Longer, formatted responses increase score. Bottom three panels show \textbf{H5-H6}: Stock phrases decrease score. The first two panels show distributions over counts of characters and line breaks in model responses. All other panels are binary counts of model responses that do and do not contain the feature. Models are sorted alphabetically.}
    \label{fig:hypothesis_analysis_models}
\end{figure}

We present results for our investigation into correlates of model score. We only investigate model responses in the opening conversation turn. We present descriptive results for text length in \cref{fig:hypothesis_length_models}, and additional formatting and phrase hypotheses in \cref{fig:hypothesis_analysis_models}. We present statistical results of a simple OLS regression in \cref{tab:score_OLS}. We encourage future work with more sophisticated model specifications, for example controlling for model, participant, or conversational context fixed-effects. In our specification, we test:
\begin{enumerate}
    \item \texttt{text\_length} is number of characters in the model response string.
    \item \texttt{if\_line\_breaks} is 1 if the string contains ``\textbackslash n''; else 0.
    \item \texttt{if\_question\_marks} is 1 if the \textit{last} character of the string is ``?''; else 0.
    \item \texttt{if\_enumeration} is 1 if the string contains numeric enumeration (e.g. ``1. ...\textbackslash n 2. ...'') or bullets (``-... \textbackslash n -...''); else 0.\footnote{Anecdotally, one participant said ``I liked it when the options where listed. It made it easier for me to read.''}
    \item \texttt{if\_deanthro} is 1 if the string contains 1 or more matched deanthropomorphising phrases e.g. ``As an AI language model...'', ``I don't hold personal opinions...''; else 0.
    \item \texttt{if\_refusal} is 1 if the string contains 1 or more matched refusal phrases e.g. ``I cannot engage with...'', ``I don't hold personal opinions''; else 0.
    \item \texttt{if\_self\_identification} is 1 if the string contains 1 or more matched names of models or providers e.g. ``I am designed by Anthropic to be...''; else 0.
\end{enumerate}

As additional detail for \textbf{H6}, we find that 9\% of conversations contain at least one refuser model matching phrases, e.g. `` I'm sorry, but...''. In these cases, a non-refuser is chosen 73\% of time.

\begin{table}[H]
\footnotesize
\begin{center}
\begin{tabular}{lcccccc}
                                  & \textbf{coef} & \textbf{std err} & \textbf{t} & \textbf{P$> |$t$|$} & \textbf{[0.025} & \textbf{0.975]}  \\
\midrule
\textbf{const}                    &      50.0055  &        0.330     &   151.459  &         0.000        &       49.358    &       50.653     \\
\textbf{text\_length}             &       0.0271  &        0.001     &    34.513  &         0.000        &        0.026    &        0.029     \\
\textbf{if\_line\_breaks}         &     -10.8285  &        0.517     &   -20.930  &         0.000        &      -11.843    &       -9.814     \\
\textbf{if\_question\_marks}      &       2.6179  &        0.560     &     4.675  &         0.000        &        1.520    &        3.716     \\
\textbf{if\_enumeration}          &       7.1981  &        0.741     &     9.710  &         0.000        &        5.745    &        8.651     \\
\textbf{if\_deanthro}             &      -2.3025  &        0.572     &    -4.023  &         0.000        &       -3.424    &       -1.181     \\
\textbf{if\_refusal}              &      -9.0484  &        0.988     &    -9.161  &         0.000        &      -10.984    &       -7.112     \\
\textbf{if\_self\_identification} &      -3.6354  &        1.034     &    -3.516  &         0.000        &       -5.662    &       -1.609     \\
\bottomrule
\multicolumn{7}{l}{\textit{Notes}. $N$: 30,049; $R^2$: 0.056; F-stat: 253.6; P(F-stat): 0.00}
\end{tabular}
\caption{\small \textbf{OLS of score on hypothesised influence factors.}}
\label{tab:score_OLS}
\end{center}
\end{table}

\clearpage

\cleardoublepage
\subsection{Leaderboard Comparison to LMSYS}
\label{sec:appendix_lmsys}
\begin{figure}[H]
    \centering
    \begin{subfigure}[b]{0.9\textwidth}
        \centering
\includegraphics[width=\textwidth]{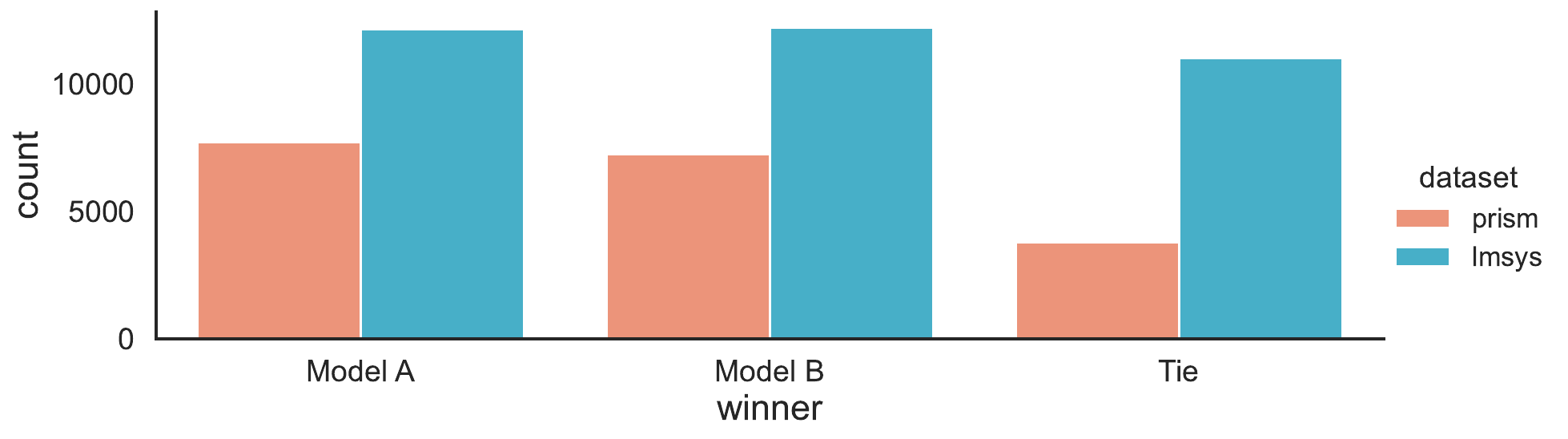}
        \caption{\small \textbf{Differences in battles.} LMSYS has more battles but the distribution of wins between model A and model B are similar, with \ourdata having fewer ties (20\% vs 31\%, at a tie threshold of 5).}
    \end{subfigure}
    \vspace{0.2em}
    \vfill
    \begin{subfigure}[b]{0.9\textwidth}
        \centering  \includegraphics[width=0.9\textwidth]{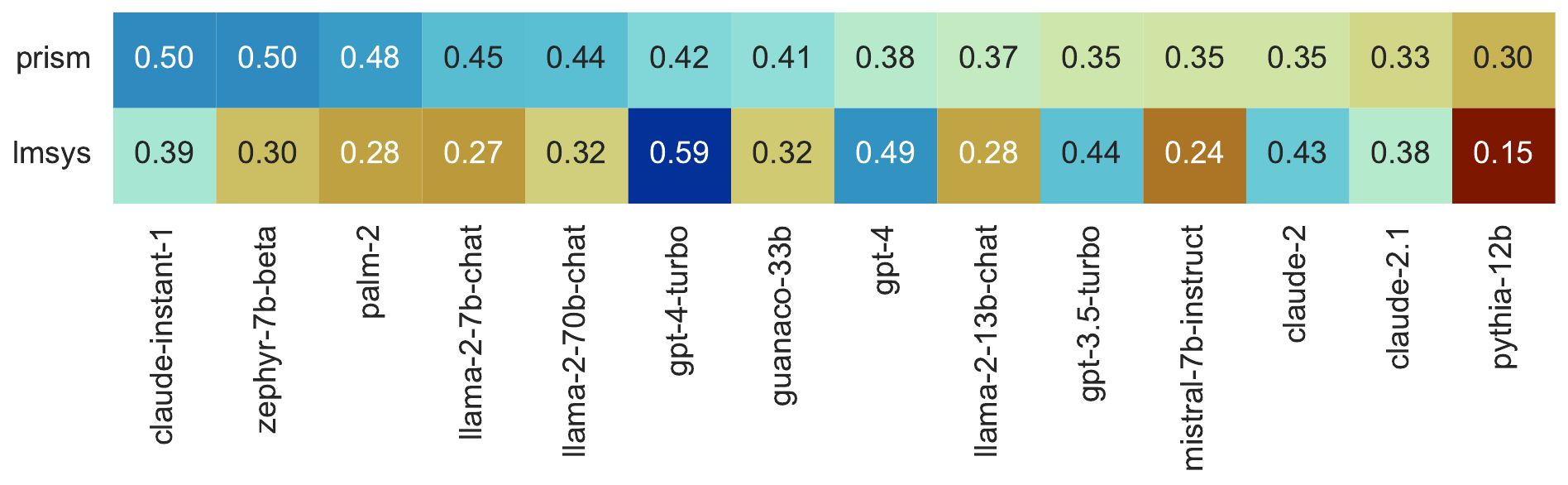}
        \caption{\small \textbf{Differences in average pairwise win rates}. We include the full set of observed battles (unbalanced total battles and battles per pair).}
    \end{subfigure}
    \vfill
    \vspace{0.2em}
    \begin{subfigure}[b]{0.9\textwidth}
        \centering  \includegraphics[width=\textwidth]{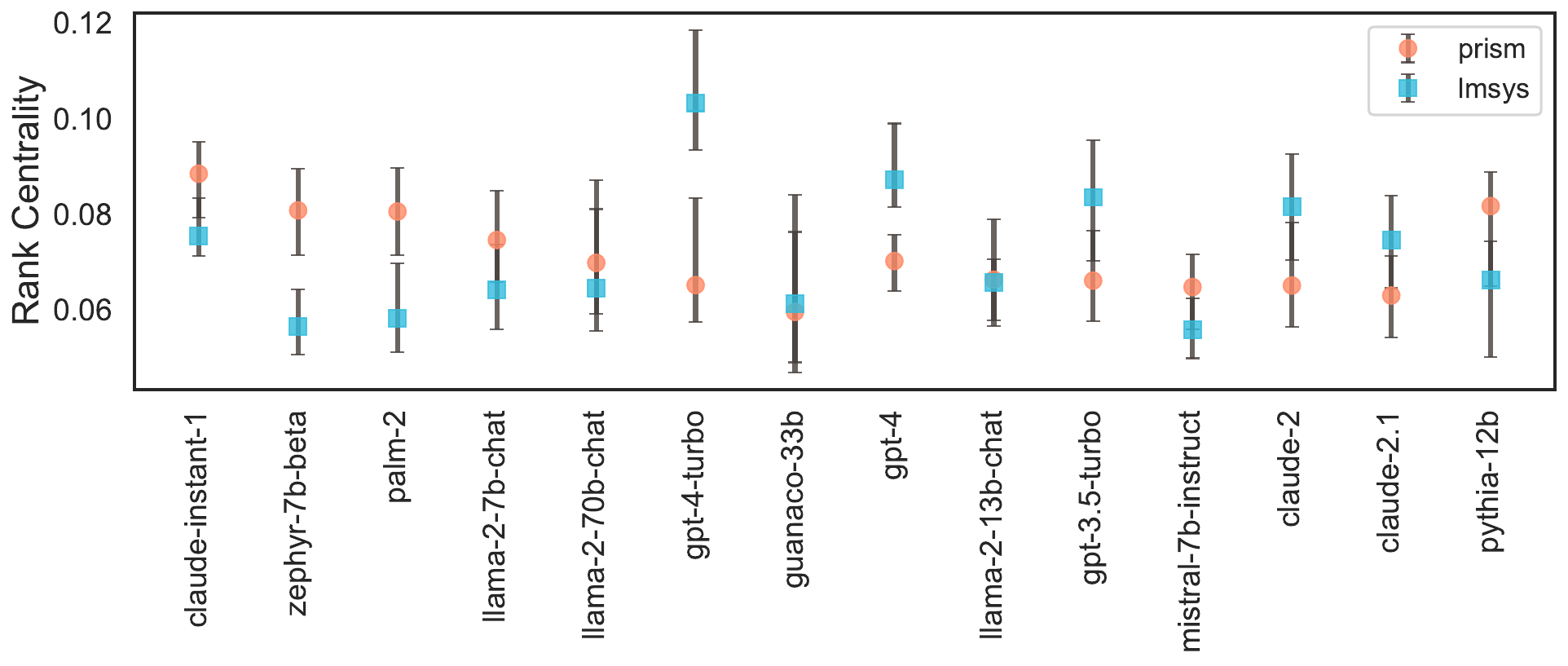}
        \caption{\small \textbf{Differences in Pairwise Rank Centrality}. Even-sampling per pair ($n=80$), bootstrapped for 95\% confidence intervals on the median (\texttt{iter}=1000).}
    \label{fig:lmsys_bootstrap}
    \end{subfigure}
    \caption{\small \textbf{Comparison of \ourdata battles to LMSYS leaderboard.} Demonstrates that the \texttt{gpt} suite of models do significantly worse in \ourdata, and open-access models like \texttt{zephyr} and \texttt{pythia} do better. }
\end{figure}

We download the LMSYS battles \cite{zhengJudging2023, zhengLMSYSChat1M2024}.\footnote{See \href{https://huggingface.co/spaces/lmsys/chatbot-arena-leaderboard}{huggingface.co/spaces/lmsys/chatbot-arena-leaderboard} and the attached notebook for details on how to obtain raw data.} Originally, LMSYS has 184,610 battles over 54 models; \ourdata has 42,306 battles over 21 models. After merging, there are 14 shared models ($N_{\text{\ourdata}}$ = 18,758, $N_{\text{LMSYS}} = 35,359$).\footnote{If we also restrict LMSYS battles to our data collection window (22nd November-22nd December 2023), there are only 9,804 LMSYS battles which we decided was too small a subset for a fair comparison.} For LMSYS, we convert both ``tie'' and ``tie (both bad)'' to a single tie group. We use $t=5$ as a tie threshold for \ourdata. Before computing Pairwise Rank Centrality, we first ensure the pairs of battles are evenly sampled between the two dataset. We find that 90\% of pairs have at least 80 battles, driven by more sparse battles in LMSYS (in \ourdata, the least frequent pair appears in 107 battles). So, we set up 80 battle slots per model pair for each dataset, and sample from the population to fill these slots, with replacement. We bootstrap this sampling over 1000 iterations then present the 5th to 95th confidence intervals in \cref{fig:lmsys_bootstrap}.

\cleardoublepage
\section{Case Study III: Welfare Analysis}
\label{sec:appendix_welfare_sensitivity}

\subsection{Extended Methods}
\paragraph{Setup} The third experiment asks: \textit{how do the sampling decisions affect welfare outcomes?} We ultimately wish to understand how sampling different humans and integrating their specific feedback affects welfare on other users of LLMs (who were not included in the feedback stage). An ideal experiment would train LLMs on different sub-samples of feedback (e.g. 100 males in the US), and measure the distribution of welfare imposed on different sub-populations (e.g. females in the US). While training LLMs on different sub-populations is beyond this paper's scope, we approximate the thought experiment by randomly generating sub-samples of individuals to select their favourite existing LLM (those in the seat of power), and measure the the distribution of welfare imposed on different sub-populations (also called stakeholder populations \cite{conitzerSocial2024}).

\paragraph{Sub-populations} Let $P$ denote the population of participants, $p \subseteq P$ denote a sub-population and $\mathcal{P}(P)$ denote the power set of $P$ (i.e. all subpopulations). To identify specific sub-populations, we define the choice function: $\textsc{subpop}: \textsc{regions} \times \textsc{groups} \mapsto \mathcal{P}(P)$ where  $\textsc{regions} = \{\text{US},\text{UK}\}$ is a set geographical regions and $\textsc{groups} =  \{\text{rep},\text{non-male},\text{non-white},\text{below 45},
\text{male},\text{white},\text{above 45}
\}$ is a set demographic groups ($\text{rep}$ denotes the whole population). Given $r \in \textsc{regions}$ and $g \in \textsc{groups}$, $\textsc{subpop}$ returns the individuals in $P$ that are in both $r$ and $g$. Our analysis uses the sub-populations given by: $
\mathcal{SP} = \{ \textsc{subpop}(i,j) \in \mathcal{P}(P) \mid (i,j) \in \{\text{US}\} \times  \{\text{rep},\text{non-male},\text{non-white},\text{below 45}  \} \}$. We approximate the sub-population defined by a tuple $(r,g)$ by selecting all the matching participants in our balanced sample that are in both $r$ and $g$.

\paragraph{Sampling schemes} A sampling scheme is a tuple: $S = (p,n)$ where $p \in \mathcal{P}$ and $n \in \mathbf{N}_+$. A sampling scheme randomly generates samples of $n$ individuals from $p$, the subpopulation of interest. We approximate a sampling scheme by using our approximation of sub-populations defined in the previous section and sampling $n$ participants with replacement. Our main analysis uses the sampling schemes: 
$
\mathcal{S} 
=
\{
(\textsc{subpop}(US ,\text{all}),n) \mid n \in \{10,20,50,100\}
\}
\cup 
\{ 
(\textsc{subpop}(US,g),100) \mid  g \in \{\text{male},\text{white},\text{above 45}\}  
\}
$. 

\paragraph{Individual welfare} Let $M$ denote the set of models. Our analysis requires a measure of welfare for an individual $j$ if LLM $i$ is chosen. We use two measures of individual welfare. i) $\textsc{rating}: P \times M \mapsto [1,100]$. Given participant $j$ and model $i$, $\textsc{rating}(j,i)$ computes the mean rating $i$ gives to LLM $j$ in the first turn of a conversation. ii) $\textsc{choice}: P \times M \mapsto [0,1]$. $\textsc{choice}(j,i)$ computes the proportion of the $j$'s conversations where LLM $i$ is chosen, conditional on LLM $i$ being shown. For both measures of individual welfare, if a participant is never shown a model, we set their individual welfare to $\textit{NA}$.

\paragraph{The distribution of LLMs induced by sampling scheme} 
A sampling scheme $S$, together with a preference aggregation method induce a distribution $\rho \in \Delta(M)$. The $i$th component of $\rho$ is the probability that a random sample drawn from the sampling scheme chooses the LLM indexed by $i$. Our main analysis uses the preference aggregation method:  $\textsc{maxRating}: \mathcal{P}(P) \mapsto M$. Given draw $s \sim S$, we define
$\text{maxRatingCandidates} := \text{argmax}_{i \in M} 
\frac{1}{|s'(i)|} \sum_{j \in s'(i) } \textsc{rating}(j,i)$
where $s'(i) = \{j \in s \mid \text{rating}(j,i) \neq \textit{NA} \}$. $\textsc{maxRating}(s)$ then returns a random element in $\text{maxRatingCandidates}$. 
In words, \textsc{maxRating} computes the $\text{rating}$ (as defined in the previous paragraph) given to each model by each participant in the draw of $S$. It then computes the mean score of each model averaged across individual mean ratings and returns a model with the highest mean rating. We repeat the analysis for the method $\textsc{maxChoice}$ which replaces $\textsc{rating}$ with $\textsc{choice}$.

\paragraph{Measuring welfare} 
For simplicity, we summarise the welfare imposed on the population by a given model by a single number. For the main analysis, we use the measure $\textsc{meanRating}: \mathcal{P}(P) \times M \mapsto [1,100]$ where 
$$
\textsc{meanRating}(p,i) = \frac{1}{|p'|}\sum_{j \in p'}\textsc{rating}(j,i)
$$
and $p' = \{j \in p \mid \text{rating}(j,i) \neq \textit{NA} \}$. We repeat that analysis for $\textsc{meanChoice}$ which replaces $\textsc{rating}$ with $\textsc{choice}$. Given a sampling scheme $S$ and a subpopulation $p \in \mathcal{P}$, the PMF of the distribution of welfare is described by the tuple:$(\rho(S),w(p))$ where $w$ is a vector whose $i$th component is given by $\textsc{meanRating}(p,i)$. 

For each $sp \in \mathcal{SP}$, we compute the welfare distributions implied by each sampling scheme $S \in \mathcal{S}$. We use $ \textsc{maxRating}$ to choose a LLM, and $\textsc{meanRating}$ as our measure of welfare. We repeat the analysis using $ \textsc{maxChoice}$ to choose a LLM, and $\textsc{meanChoice}$ as our measure of welfare. A concern is that our results are sensitive to randomness caused by different participants being shown different models. As a sensitivity check, we repeat the analysis with imputed scores for missing model ratings (similar to collaborative filtering), and repeat the whole exercise for the UK (see \cref{sec:appendix_uk_welfare}).

There are some caveats to note. Despite having samples balanced by observed demographics for the UK and the US, the samples are too small to to expect them to be representative on features we do not observe. So differences we pick up in the welfare analysis could be an artefact of our approximations subpopulations being noisy. Furthermore, our analysis using the $\textsc{meanRating}$ welfare measure assumes that individuals use scores in the same way for ratings welfare measures. However, our analysis using $\textsc{meanChoice}$ is not sensitive to use of ratings scale, and the results are qualitatively similar. Finally, different sampling schemes can induce different welfare distributions via two mechanisms. First, the subpopulations sampled from may have different preferences conditional on conversation type. Second, the sub-populations sampled from may have different conversations, and in turn, choose models that are better at particular conversations. This experiment taken alone cannot disentangle these two mechanisms.

\subsection{UK Sample}
\label{sec:appendix_uk_welfare}
\begin{figure}[H]
    \centering
    \includegraphics[width=\textwidth]{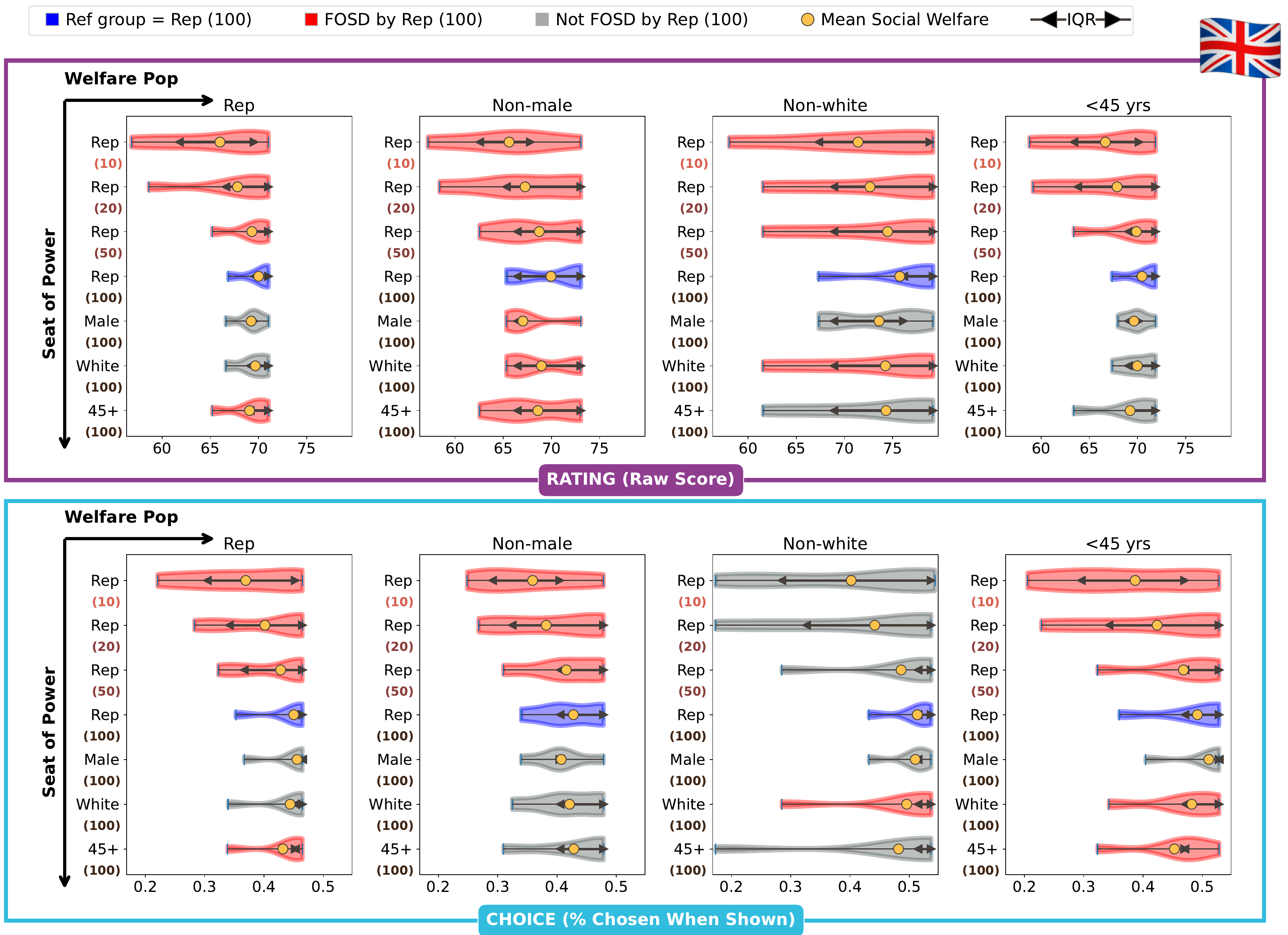}
    \caption{\small \textbf{Welfare distributions for the UK.} We repeat the welfare analysis for the UK, analogous to \cref{fig:welfare_main}. The distribution of mean welfare for four subpopulations in the UK (welfare pop) induced by seven sampling schemes (in the seat of power). The $y$ axis is the sampled supopulation (e.g. \textbf{Rep} is a `representative' sample of the population) and sample size in brackets (e.g. \textbf{(100)}).%
    The top four \highLight{lightpurple}{Rating} comparisons use the \textsc{meanWelfare} welfare measure and the \textsc{maxRating} preference aggregation method. The bottom \highLight{lightblue}{Choice} comparisons use the \textsc{meanChoice} welfare measure and the \textsc{maxChoice} preference aggregation method. The  \highLight{violinred}{red} distributions are FOSD by Rep (100) in \highLight{violinblue}{blue}. The results are qualitatively similar to the US results in \cref{fig:welfare_main}.}
    \label{fig:welfare_main_uk}
\end{figure}

\subsection{Imputing Missing Individual Welfare}
\begin{figure}[H]
    \centering
    \begin{subfigure}{0.9\textwidth}
        \centering
        \includegraphics[width=\linewidth]{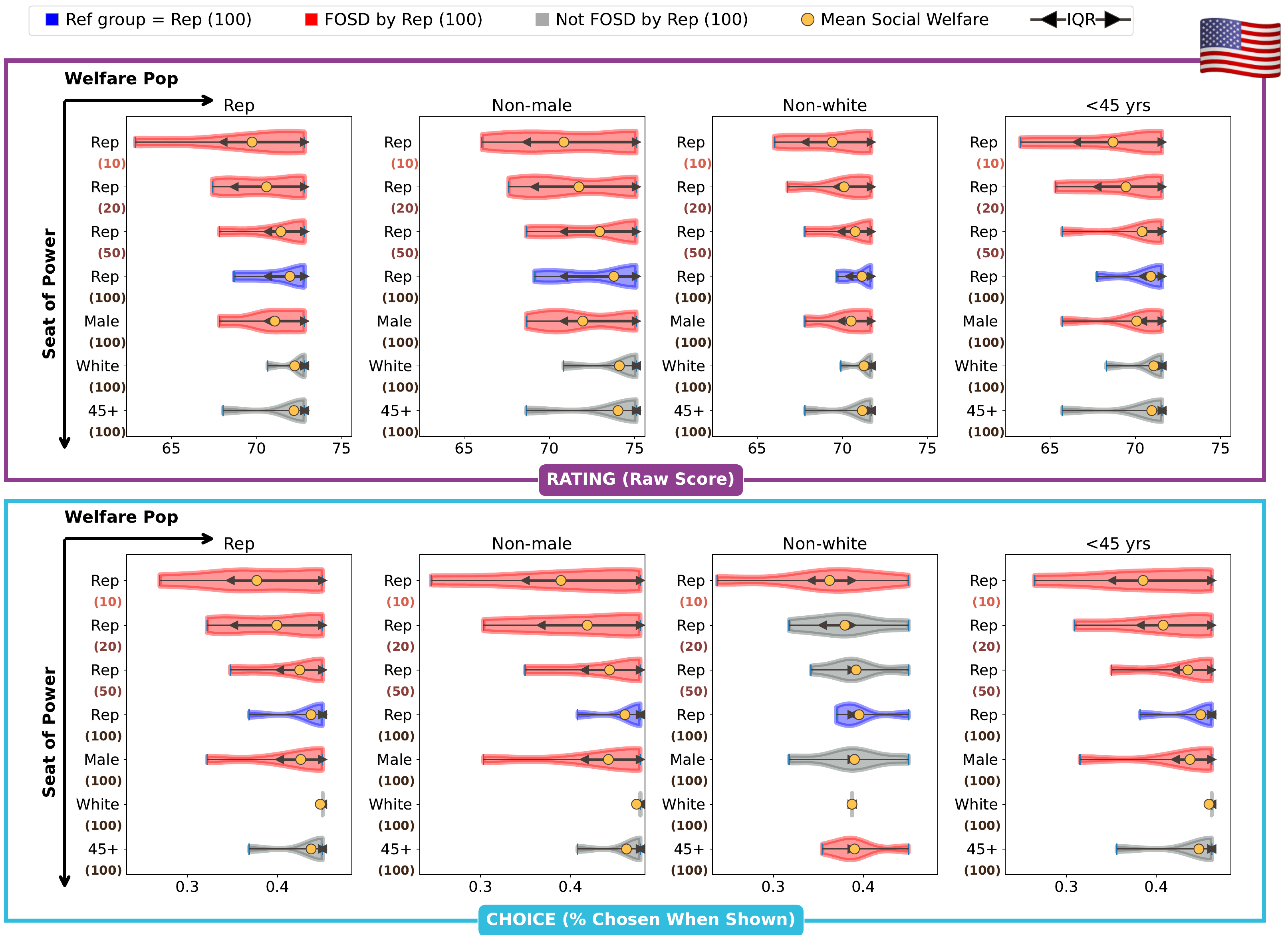}
        \caption{\textbf{US Sample.}}
        \label{fig:sub_welfare_imputed_us}
    \end{subfigure}
    \vspace{1cm} %
    \begin{subfigure}{0.9\textwidth}
        \centering
        \includegraphics[width=\linewidth]{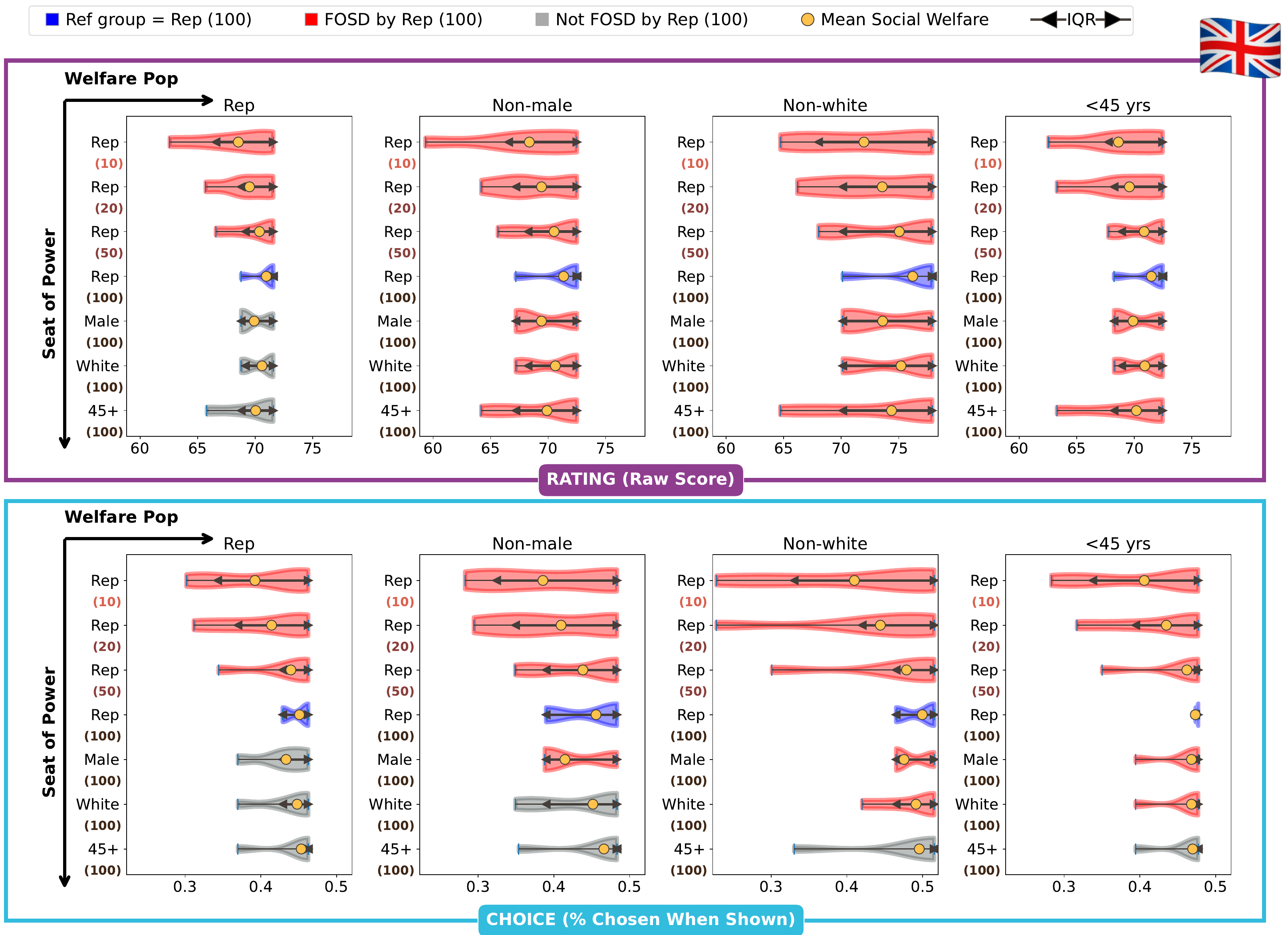}
        \caption{\textbf{UK Sample.}}
        \label{fig:sub_welfare_imputed_uk}
    \end{subfigure}
    \vspace{-2em}
    \caption{\small \textbf{Welfare distributions with imputation.} In \cref{fig:welfare_main} and \cref{fig:welfare_main_uk} individual welfare for a model takes the value \textit{NA} if an individual never sees the model. Here, we repeat the welfare analysis and impute individual welfare with an approach is similar in spirit to collaborative filtering. Using the only matrix of individual welfare for each model, we impute \textit{NA} cells using multivariate imputation, implemented with the IterativeImputer package in Python. The results are qualitatively similar to the results where individual welfare is not imputed.}
    \label{fig:welfare_imputed}
\end{figure}

\clearpage

\addcontentsline{toc}{section}{\large{PART III: Codebooks}}
\section{Codebooks}
\label{sec:appendix_codebooks}
\subsection{Survey Codebook}
\label{sec:survey_codebook}
\fontsize{6pt}{6pt}\selectfont
\begin{longtable}{p{0.005\textwidth}p{0.20\textwidth}p{0.40\textwidth}p{0.1\textwidth}p{0.1\textwidth}}
\toprule
 & VARIABLE & LABEL & CATEGORY & TYPE \\
\midrule
\endfirsthead
\toprule
 & VARIABLE & LABEL & CATEGORY & TYPE \\
\midrule
\endhead
\midrule
\multicolumn{5}{r}{Continued on next page} \\
\midrule
\endfoot
\bottomrule
\endlastfoot
\bfseries 0 & \bfseries user\_id & \bfseries Unique participant identifier & \bfseries {\cellcolor{myred}} meta & \bfseries {\cellcolor{myoat}} string id \\
 \midrule
\bfseries  & \bfseries  & \bfseries  & \bfseries N Missing: & \bfseries 0 \\
\bfseries  & \bfseries  & \bfseries  & \bfseries N Unique: & \bfseries 1500 \\
 & \multicolumn{4}{p{0.9\textwidth}}{\textit{Notes: Pseudonymized from Prolific worker ID. Used to link survey data to conversation data. In our paper, we refer to `users' as `participants'. }}  \\
 & \cline{1-4}
 \midrule
\bfseries 1 & \bfseries survey\_only & \bfseries Indicator if participant only completed the survey, or also completed conversations & \bfseries {\cellcolor{myred}} meta & \bfseries {\cellcolor{myoat}} binary \\
 \midrule
\bfseries  & \bfseries  & \bfseries  & \bfseries N Missing: & \bfseries 0 \\
\bfseries  & \bfseries  & \bfseries  & \bfseries N Unique: & \bfseries 2 \\
 &  & \multicolumn{2}{r}{False } & 1396 \\
 &  & \multicolumn{2}{r}{True } & 104 \\
 \midrule
\bfseries 2 & \bfseries num\_completed\_conversations & \bfseries Number of conversations that a participant completed & \bfseries {\cellcolor{myred}} meta & \bfseries {\cellcolor{myoat}} int \\
 \midrule
\bfseries  & \bfseries  & \bfseries  & \bfseries N Missing: & \bfseries 0 \\
\bfseries  & \bfseries  & \bfseries  & \bfseries N Unique: & \bfseries 8 \\
 &  & \multicolumn{2}{r}{mean } & 5.3 \\
 &  & \multicolumn{2}{r}{std } & 1.7 \\
 &  & \multicolumn{2}{r}{min } & 0.0 \\
 &  & \multicolumn{2}{r}{max } & 7.0 \\
 \midrule
\bfseries 3 & \bfseries consent & \bfseries Participant informed consent confirmation & \bfseries {\cellcolor{myyellow}} direct & \bfseries {\cellcolor{myoat}} categorical \\
 \midrule
 & \multicolumn{4}{p{0.9\textwidth}}{\textit{Question text:} If you have read the information above and agree to participate with the understanding that the data (including any personal data) you submit will be processed accordingly, please select the box below to start. }  \\
 & \cline{1-4}
\bfseries  & \bfseries  & \bfseries  & \bfseries N Missing: & \bfseries 0 \\
\bfseries  & \bfseries  & \bfseries  & \bfseries N Unique: & \bfseries 1 \\
 &  & \multicolumn{2}{r}{Yes, I consent to take part } & 1500 \\
 & \multicolumn{4}{p{0.9\textwidth}}{\textit{Notes: See full informed consent document for details }}  \\
 & \cline{1-4}
 \midrule
\bfseries 4 & \bfseries consent\_age & \bfseries Participant age confirmation & \bfseries {\cellcolor{myyellow}} direct & \bfseries {\cellcolor{myoat}} categorical \\
 \midrule
 & \multicolumn{4}{p{0.9\textwidth}}{\textit{Question text:} Please note that you may only participate in this survey if you are 18 years of age or over. }  \\
 & \cline{1-4}
\bfseries  & \bfseries  & \bfseries  & \bfseries N Missing: & \bfseries 0 \\
\bfseries  & \bfseries  & \bfseries  & \bfseries N Unique: & \bfseries 1 \\
 &  & \multicolumn{2}{r}{I certify that I am 18 years of age or over } & 1500 \\
 & \multicolumn{4}{p{0.9\textwidth}}{\textit{Notes: See full informed consent document for details }}  \\
 & \cline{1-4}
 \midrule
\bfseries 5 & \bfseries lm\_familiarity & \bfseries Familiarity with LLMs & \bfseries {\cellcolor{myyellow}} direct & \bfseries {\cellcolor{myoat}} categorical \\
 \midrule
 & \multicolumn{4}{p{0.9\textwidth}}{\textit{Question text:} How familiar are you with AI language models like ChatGPT? }  \\
 & \cline{1-4}
\bfseries  & \bfseries  & \bfseries  & \bfseries N Missing: & \bfseries 0 \\
\bfseries  & \bfseries  & \bfseries  & \bfseries N Unique: & \bfseries 3 \\
 &  & \multicolumn{2}{r}{Somewhat familiar } & 920 \\
 &  & \multicolumn{2}{r}{Very familiar } & 424 \\
 &  & \multicolumn{2}{r}{Not familiar at all } & 156 \\
 \midrule
\bfseries 6 & \bfseries lm\_direct\_use & \bfseries Direct use of LLMs & \bfseries {\cellcolor{myyellow}} direct & \bfseries {\cellcolor{myoat}} categorical \\
 \midrule
 & \multicolumn{4}{p{0.9\textwidth}}{\textit{Question text:} Have you directly used or communicated with an AI language model, such as asking questions to ChatGPT, BARD, Claude or other similar models? }  \\
 & \cline{1-4}
\bfseries  & \bfseries  & \bfseries  & \bfseries N Missing: & \bfseries 0 \\
\bfseries  & \bfseries  & \bfseries  & \bfseries N Unique: & \bfseries 3 \\
 &  & \multicolumn{2}{r}{Yes } & 1162 \\
 &  & \multicolumn{2}{r}{No } & 259 \\
 &  & \multicolumn{2}{r}{Unsure } & 79 \\
 \midrule
\bfseries 7 & \bfseries lm\_indirect\_use & \bfseries Direct use of LLMs & \bfseries {\cellcolor{myyellow}} direct & \bfseries {\cellcolor{myoat}} categorical \\
 \midrule
 & \multicolumn{4}{p{0.9\textwidth}}{\textit{Question text:} Have you directly used or communicated with an AI language model, such as asking questions to ChatGPT, BARD, Claude or other similar models? }  \\
 & \cline{1-4}
\bfseries  & \bfseries  & \bfseries  & \bfseries N Missing: & \bfseries 0 \\
\bfseries  & \bfseries  & \bfseries  & \bfseries N Unique: & \bfseries 3 \\
 &  & \multicolumn{2}{r}{Yes } & 1104 \\
 &  & \multicolumn{2}{r}{No } & 215 \\
 &  & \multicolumn{2}{r}{Unsure } & 181 \\
 \midrule
\bfseries 8 & \bfseries lm\_frequency\_use & \bfseries Frequency of using Large Language Models & \bfseries {\cellcolor{myyellow}} direct & \bfseries {\cellcolor{myoat}} categorical \\
 \midrule
 & \multicolumn{4}{p{0.9\textwidth}}{\textit{Question text:} How often do you use or communicate with AI language models? }  \\
 & \cline{1-4}
\bfseries  & \bfseries  & \bfseries  & \bfseries N Missing: & \bfseries 247 \\
\bfseries  & \bfseries  & \bfseries  & \bfseries N Unique: & \bfseries 5 \\
 &  & \multicolumn{2}{r}{Once per month } & 374 \\
 &  & \multicolumn{2}{r}{Every week } & 316 \\
 &  & \multicolumn{2}{r}{More than once a month } & 291 \\
 &  & \multicolumn{2}{r}{None } & 247 \\
 &  & \multicolumn{2}{r}{Less than one a year } & 162 \\
 &  & \multicolumn{2}{r}{Every day } & 110 \\
 & \multicolumn{4}{p{0.9\textwidth}}{\textit{Notes: Only shown if lm\_indirect\_use==1 OR lm\_direct\_use==1. Null indicates particiant did not see question. }}  \\
 & \cline{1-4}
 \midrule
\bfseries 9 & \bfseries lm\_usecases & \bfseries Use cases of LLMs & \bfseries {\cellcolor{myyellow}} direct & \bfseries {\cellcolor{myoat}} dict \\
 \midrule
 & \multicolumn{4}{p{0.9\textwidth}}{\textit{Question text:} Which of the following scenarios best describe how and why you use AI language models? Select all that apply. }  \\
 & \cline{1-4}
\bfseries  & \bfseries  & \bfseries  & \bfseries N Missing: & \bfseries 247 \\
\bfseries  & \bfseries  & \bfseries  & \bfseries N Unique: & \bfseries 853 \\
 & \cline{1-4}
 & homework\_assistance & \multicolumn{3}{p{0.6\textwidth}}{Homework Assistance: Getting help with school or university assignments. } \\
 &  & \multicolumn{2}{r}{False } & 967 \\
 &  & \multicolumn{2}{r}{True } & 533 \\
 & \cline{1-4}
 & research & \multicolumn{3}{p{0.6\textwidth}}{Research: Fact-checking or gaining overviews on specific topics. } \\
 &  & \multicolumn{2}{r}{True } & 864 \\
 &  & \multicolumn{2}{r}{False } & 636 \\
 & \cline{1-4}
 & source\_suggestions & \multicolumn{3}{p{0.6\textwidth}}{Source Suggestions: Creating or finding bibliographies, information sources or reading lists. } \\
 &  & \multicolumn{2}{r}{False } & 1036 \\
 &  & \multicolumn{2}{r}{True } & 464 \\
 & \cline{1-4}
 & professional\_work & \multicolumn{3}{p{0.6\textwidth}}{Professional Work: Assisting in drafting, editing, or brainstorming content for work. } \\
 &  & \multicolumn{2}{r}{False } & 784 \\
 &  & \multicolumn{2}{r}{True } & 716 \\
 & \cline{1-4}
 & creative\_writing & \multicolumn{3}{p{0.6\textwidth}}{Creative Writing: Generating story ideas, dialogues, poems or other writing prompts. } \\
 &  & \multicolumn{2}{r}{False } & 861 \\
 &  & \multicolumn{2}{r}{True } & 639 \\
 & \cline{1-4}
 & casual\_conversation & \multicolumn{3}{p{0.6\textwidth}}{Casual Conversation: Engaging in small talk, casual chats, or joke generation. } \\
 &  & \multicolumn{2}{r}{False } & 991 \\
 &  & \multicolumn{2}{r}{True } & 509 \\
 & \cline{1-4}
 & personal\_recommendations & \multicolumn{3}{p{0.6\textwidth}}{Personal Recommendations: Seeking book, music or movie recommendations. } \\
 &  & \multicolumn{2}{r}{False } & 987 \\
 &  & \multicolumn{2}{r}{True } & 513 \\
 & \cline{1-4}
 & daily\_productivity & \multicolumn{3}{p{0.6\textwidth}}{Daily Productivity: Setting reminders, making to-do lists, or productivity tips. } \\
 &  & \multicolumn{2}{r}{False } & 1037 \\
 &  & \multicolumn{2}{r}{True } & 463 \\
 & \cline{1-4}
 & technical\_or\_programming\_help & \multicolumn{3}{p{0.6\textwidth}}{Technical or Programming Help: Seeking programming guidance, code generation, software recommendations, or debugging assistance. } \\
 &  & \multicolumn{2}{r}{False } & 916 \\
 &  & \multicolumn{2}{r}{True } & 584 \\
 & \cline{1-4}
 & travel\_guidance & \multicolumn{3}{p{0.6\textwidth}}{Travel Guidance: Getting destination recommendations, planning holidays, or cultural etiquette tips. } \\
 &  & \multicolumn{2}{r}{False } & 1120 \\
 &  & \multicolumn{2}{r}{True } & 380 \\
 & \cline{1-4}
 & lifestyle\_and\_hobbies & \multicolumn{3}{p{0.6\textwidth}}{Lifestyle and Hobbies: Looking for recipes, craft ideas, home decoration tips, or hobby-related information. } \\
 &  & \multicolumn{2}{r}{False } & 943 \\
 &  & \multicolumn{2}{r}{True } & 557 \\
 & \cline{1-4}
 & well-being\_guidance & \multicolumn{3}{p{0.6\textwidth}}{Well-being Guidance: Seeking general exercise routines, wellness or meditation tips. } \\
 &  & \multicolumn{2}{r}{False } & 1094 \\
 &  & \multicolumn{2}{r}{True } & 406 \\
 & \cline{1-4}
 & medical\_guidance & \multicolumn{3}{p{0.6\textwidth}}{Medical Guidance: Seeking health-related advice or medical guidance. } \\
 &  & \multicolumn{2}{r}{False } & 1123 \\
 &  & \multicolumn{2}{r}{True } & 377 \\
 & \cline{1-4}
 & financial\_guidance & \multicolumn{3}{p{0.6\textwidth}}{Financial Guidance: Asking about financial concepts or general investing ideas. } \\
 &  & \multicolumn{2}{r}{False } & 1146 \\
 &  & \multicolumn{2}{r}{True } & 354 \\
 & \cline{1-4}
 & games & \multicolumn{3}{p{0.6\textwidth}}{Games: Playing text-based games, generating riddles or puzzles. } \\
 &  & \multicolumn{2}{r}{False } & 1110 \\
 &  & \multicolumn{2}{r}{True } & 390 \\
 & \cline{1-4}
 & historical\_or\_news\_insight & \multicolumn{3}{p{0.6\textwidth}}{Historical or News Insight: Getting summaries or background on historical events or news and current affairs. } \\
 &  & \multicolumn{2}{r}{False } & 1070 \\
 &  & \multicolumn{2}{r}{True } & 430 \\
 & \cline{1-4}
 & relationship\_advice & \multicolumn{3}{p{0.6\textwidth}}{Relationship Advice: Seeking general self-help or relationship advice for family, friends or partners. } \\
 &  & \multicolumn{2}{r}{False } & 1155 \\
 &  & \multicolumn{2}{r}{True } & 345 \\
 & \cline{1-4}
 & language\_learning & \multicolumn{3}{p{0.6\textwidth}}{Language Learning: Using it as a tool for language practice or translation. } \\
 &  & \multicolumn{2}{r}{False } & 1024 \\
 &  & \multicolumn{2}{r}{True } & 476 \\
 & \cline{1-4}
 & other & \multicolumn{3}{p{0.6\textwidth}}{Other (selected) } \\
 &  & \multicolumn{2}{r}{False } & 1129 \\
 &  & \multicolumn{2}{r}{True } & 371 \\
 & \cline{1-4}
 & other\_text & \multicolumn{3}{p{0.6\textwidth}}{Other (typed text) } \\
 &  & \multicolumn{2}{r}{mean chars } & 45.8 \\
 &  & \multicolumn{2}{r}{std chars } & 41.9 \\
 &  & \multicolumn{2}{r}{min chars } & 3.0 \\
 &  & \multicolumn{2}{r}{max chars } & 328.0 \\
 & \multicolumn{4}{p{0.9\textwidth}}{\textit{Notes: Question only show if lm\_direct\_use==1 OR lm\_indirect\_use==1.
N Missing indicates the participants who have at least one missing value in the usecases (besides from 'other\_text').
N Unique indicates the unique combinations of use cases selected by participants.
On 'other\_text', Null indicates participant did not type anything.
On all other keys, 0 indicates participant saw question and did not select usecase. Null indicates participant did not see question. }}  \\
 & \cline{1-4}
 \midrule
\bfseries 10 & \bfseries order\_lm\_usecases & \bfseries Use cases of LLMs (order of options presented in survey) & \bfseries {\cellcolor{myred}} meta & \bfseries {\cellcolor{myoat}} dict \\
 \midrule
\bfseries  & \bfseries  & \bfseries  & \bfseries N Missing: & \bfseries 247 \\
\bfseries  & \bfseries  & \bfseries  & \bfseries N Unique: & \bfseries 1254 \\
 & \multicolumn{4}{p{0.9\textwidth}}{\textit{Notes: Integer 1-18 indicating random order that usecase option was presented to participant.
For 'other', option is always shown last so will always be 19.
Null indicates participant did not see question.
The usecases as the same as in lm\_usecases. }}  \\
 & \cline{1-4}
 \midrule
\bfseries 11 & \bfseries stated\_prefs & \bfseries Stated preferences over LLM behaviours & \bfseries {\cellcolor{myyellow}} direct & \bfseries {\cellcolor{myoat}} dict \\
 \midrule
 & \multicolumn{4}{p{0.9\textwidth}}{\textit{Question text:} Rate each of the following statements about your opinion on the importance of different AI language model behaviors or traits.
It is important that an AI language model... }  \\
 & \cline{1-4}
\bfseries  & \bfseries  & \bfseries  & \bfseries N Missing: & \bfseries 0 \\
\bfseries  & \bfseries  & \bfseries  & \bfseries N Unique: & \bfseries 1475 \\
 & \cline{1-4}
 & values & \multicolumn{3}{p{0.6\textwidth}}{...reflects my values or cultural perspectives } \\
 &  & \multicolumn{2}{r}{mean } & 54.3 \\
 &  & \multicolumn{2}{r}{std } & 26.3 \\
 &  & \multicolumn{2}{r}{min } & 0.0 \\
 &  & \multicolumn{2}{r}{max } & 100.0 \\
 & \cline{1-4}
 & creativity & \multicolumn{3}{p{0.6\textwidth}}{...produces responses that are creative and inspiring } \\
 &  & \multicolumn{2}{r}{mean } & 69.6 \\
 &  & \multicolumn{2}{r}{std } & 22.1 \\
 &  & \multicolumn{2}{r}{min } & 0.0 \\
 &  & \multicolumn{2}{r}{max } & 100.0 \\
 & \cline{1-4}
 & fluency & \multicolumn{3}{p{0.6\textwidth}}{...produces responses that are well-written and coherent } \\
 &  & \multicolumn{2}{r}{mean } & 86.7 \\
 &  & \multicolumn{2}{r}{std } & 16.3 \\
 &  & \multicolumn{2}{r}{min } & 2.0 \\
 &  & \multicolumn{2}{r}{max } & 100.0 \\
 & \cline{1-4}
 & factuality & \multicolumn{3}{p{0.6\textwidth}}{...produces factual and informative responses } \\
 &  & \multicolumn{2}{r}{mean } & 88.7 \\
 &  & \multicolumn{2}{r}{std } & 16.2 \\
 &  & \multicolumn{2}{r}{min } & 0.0 \\
 &  & \multicolumn{2}{r}{max } & 100.0 \\
 & \cline{1-4}
 & diversity & \multicolumn{3}{p{0.6\textwidth}}{...summarises multiple viewpoints or different worldviews } \\
 &  & \multicolumn{2}{r}{mean } & 75.7 \\
 &  & \multicolumn{2}{r}{std } & 20.0 \\
 &  & \multicolumn{2}{r}{min } & 0.0 \\
 &  & \multicolumn{2}{r}{max } & 100.0 \\
 & \cline{1-4}
 & safety & \multicolumn{3}{p{0.6\textwidth}}{...produces responses that are safe and do not risk harm to myself and others } \\
 &  & \multicolumn{2}{r}{mean } & 80.2 \\
 &  & \multicolumn{2}{r}{std } & 25.2 \\
 &  & \multicolumn{2}{r}{min } & 0.0 \\
 &  & \multicolumn{2}{r}{max } & 100.0 \\
 & \cline{1-4}
 & personalisation & \multicolumn{3}{p{0.6\textwidth}}{...learns from our conversations and feels personalised to me } \\
 &  & \multicolumn{2}{r}{mean } & 67.9 \\
 &  & \multicolumn{2}{r}{std } & 24.6 \\
 &  & \multicolumn{2}{r}{min } & 0.0 \\
 &  & \multicolumn{2}{r}{max } & 100.0 \\
 & \cline{1-4}
 & helpfulness & \multicolumn{3}{p{0.6\textwidth}}{...produces responses that are helpful and relevant to my requests } \\
 &  & \multicolumn{2}{r}{mean } & 89.4 \\
 &  & \multicolumn{2}{r}{std } & 14.4 \\
 &  & \multicolumn{2}{r}{min } & 0.0 \\
 &  & \multicolumn{2}{r}{max } & 100.0 \\
 & \cline{1-4}
 & other & \multicolumn{3}{p{0.6\textwidth}}{Other (selected) } \\
 &  & \multicolumn{2}{r}{mean } & 57.5 \\
 &  & \multicolumn{2}{r}{std } & 19.0 \\
 &  & \multicolumn{2}{r}{min } & 0.0 \\
 &  & \multicolumn{2}{r}{max } & 100.0 \\
 & \cline{1-4}
 & other\_text & \multicolumn{3}{p{0.6\textwidth}}{Other (typed text) } \\
 &  & \multicolumn{2}{r}{mean chars } & 32.6 \\
 &  & \multicolumn{2}{r}{std chars } & 24.4 \\
 &  & \multicolumn{2}{r}{min chars } & 1.0 \\
 &  & \multicolumn{2}{r}{max chars } & 144.0 \\
 & \multicolumn{4}{p{0.9\textwidth}}{\textit{Notes: Sliders from [Strongly disagree] to [Strongly agree] are recorded on a 0-100 scale. Participant does not see numeric value.
N Missing indicates the participants who have at least one missing value in the attributes (besides from 'other\_text').
N Unique indicates the unique combinations of use cases selected by participants.
On 'other\_text', Null indicates participant did not type anything.
Note that this scale (on Qualtrics) runs 0-100. The Conversations rating scales (for choice\_attributes, performance\_attributes on Dynabench) run 1-100.
 }}  \\
 & \cline{1-4}
 \midrule
\bfseries 12 & \bfseries order\_stated\_prefs & \bfseries Stated preferences over LLM behaviours (order of options presented in survey) & \bfseries {\cellcolor{myred}} meta & \bfseries {\cellcolor{myoat}} dict \\
 \midrule
\bfseries  & \bfseries  & \bfseries  & \bfseries N Missing: & \bfseries 0 \\
\bfseries  & \bfseries  & \bfseries  & \bfseries N Unique: & \bfseries 1467 \\
 & \multicolumn{4}{p{0.9\textwidth}}{\textit{Notes: Integer 1-8 indicating random order that attribute slider was presented to participant.
For 'other', option is always shown last so will always be 9.
Null indicates participant did not see question.
The attributes as the same as in stated\_prefs. }}  \\
 & \cline{1-4}
 \midrule
\bfseries 13 & \bfseries self\_description & \bfseries Participant self-written profile describing themself & \bfseries {\cellcolor{myyellow}} direct & \bfseries {\cellcolor{myoat}} string \\
 \midrule
 & \multicolumn{4}{p{0.9\textwidth}}{\textit{Question text:} Please briefly describe your values, core beliefs, guiding principles in life, or other things that are important to you. For example, you might include values you'd want to teach to your children or qualities you look for in friends. There are no right or wrong answers. Please do not provide any personally identifiable details like your name, address or email. Please write 2-5 sentences in your own words. }  \\
 & \cline{1-4}
\bfseries  & \bfseries  & \bfseries  & \bfseries N Missing: & \bfseries 0 \\
\bfseries  & \bfseries  & \bfseries  & \bfseries N Unique: & \bfseries 1500 \\
 &  & \multicolumn{2}{r}{mean chars } & 241.3 \\
 &  & \multicolumn{2}{r}{std chars } & 134.6 \\
 &  & \multicolumn{2}{r}{min chars } & 3.0 \\
 &  & \multicolumn{2}{r}{max chars } & 1547.0 \\
 \midrule
\bfseries 14 & \bfseries system\_string & \bfseries Participant self-written system string, constitution or custom instructions for an LLM & \bfseries {\cellcolor{myyellow}} direct & \bfseries {\cellcolor{myoat}} string \\
 \midrule
 & \multicolumn{4}{p{0.9\textwidth}}{\textit{Question text:} Imagine you are instructing an AI language model how to behave. You can think of this like a set of core principles that the AI language model will always try to follow, no matter what task you ask it to perform. In your own words, describe what characteristics, personality traits or features you believe the AI should consistently exhibit. You can also instruct the model what behaviours or content you don't want to see. If you envision the AI behaving differently in various contexts (e.g., professional assistance vs. storytelling), please specify the general adaptations you'd like to see. Please write 2-5 sentences in your own words. }  \\
 & \cline{1-4}
\bfseries  & \bfseries  & \bfseries  & \bfseries N Missing: & \bfseries 0 \\
\bfseries  & \bfseries  & \bfseries  & \bfseries N Unique: & \bfseries 1500 \\
 &  & \multicolumn{2}{r}{mean chars } & 260.4 \\
 &  & \multicolumn{2}{r}{std chars } & 288.4 \\
 &  & \multicolumn{2}{r}{min chars } & 16.0 \\
 &  & \multicolumn{2}{r}{max chars } & 9530.0 \\
 \midrule
\bfseries 15 & \bfseries age & \bfseries Age & \bfseries {\cellcolor{myyellow}} direct & \bfseries {\cellcolor{myoat}} categorical \\
 \midrule
 & \multicolumn{4}{p{0.9\textwidth}}{\textit{Question text:} How old are you? }  \\
 & \cline{1-4}
\bfseries  & \bfseries  & \bfseries  & \bfseries N Missing: & \bfseries 0 \\
\bfseries  & \bfseries  & \bfseries  & \bfseries N Unique: & \bfseries 7 \\
 &  & \multicolumn{2}{r}{25-34 years old } & 454 \\
 &  & \multicolumn{2}{r}{18-24 years old } & 297 \\
 &  & \multicolumn{2}{r}{35-44 years old } & 237 \\
 &  & \multicolumn{2}{r}{45-54 years old } & 208 \\
 &  & \multicolumn{2}{r}{55-64 years old } & 197 \\
 &  & \multicolumn{2}{r}{65+ years old } & 106 \\
 &  & \multicolumn{2}{r}{Prefer not to say } & 1 \\
 \midrule
\bfseries 16 & \bfseries education & \bfseries Education & \bfseries {\cellcolor{myyellow}} direct & \bfseries {\cellcolor{myoat}} categorical \\
 \midrule
 & \multicolumn{4}{p{0.9\textwidth}}{\textit{Question text:} What is the highest level of education you have completed? }  \\
 & \cline{1-4}
\bfseries  & \bfseries  & \bfseries  & \bfseries N Missing: & \bfseries 0 \\
\bfseries  & \bfseries  & \bfseries  & \bfseries N Unique: & \bfseries 9 \\
 &  & \multicolumn{2}{r}{University Bachelors Degree } & 637 \\
 &  & \multicolumn{2}{r}{Graduate / Professional degree } & 241 \\
 &  & \multicolumn{2}{r}{Some University but no degree } & 236 \\
 &  & \multicolumn{2}{r}{Completed Secondary School } & 209 \\
 &  & \multicolumn{2}{r}{Vocational } & 125 \\
 &  & \multicolumn{2}{r}{Some Secondary } & 24 \\
 &  & \multicolumn{2}{r}{Completed Primary School } & 16 \\
 &  & \multicolumn{2}{r}{Prefer not to say } & 9 \\
 &  & \multicolumn{2}{r}{Some Primary } & 3 \\
 \midrule
\bfseries 17 & \bfseries employment\_status & \bfseries Employment Status & \bfseries {\cellcolor{myyellow}} direct & \bfseries {\cellcolor{myoat}} categorical \\
 \midrule
 & \multicolumn{4}{p{0.9\textwidth}}{\textit{Question text:} What best describes your employment status over the last three months? }  \\
 & \cline{1-4}
\bfseries  & \bfseries  & \bfseries  & \bfseries N Missing: & \bfseries 0 \\
\bfseries  & \bfseries  & \bfseries  & \bfseries N Unique: & \bfseries 8 \\
 &  & \multicolumn{2}{r}{Working full-time } & 712 \\
 &  & \multicolumn{2}{r}{Working part-time } & 265 \\
 &  & \multicolumn{2}{r}{Student } & 191 \\
 &  & \multicolumn{2}{r}{Unemployed, seeking work } & 113 \\
 &  & \multicolumn{2}{r}{Retired } & 104 \\
 &  & \multicolumn{2}{r}{Homemaker / Stay-at-home parent } & 46 \\
 &  & \multicolumn{2}{r}{Unemployed, not seeking work } & 46 \\
 &  & \multicolumn{2}{r}{Prefer not to say } & 23 \\
 \midrule
\bfseries 18 & \bfseries marital\_status & \bfseries Marital Status & \bfseries {\cellcolor{myyellow}} direct & \bfseries {\cellcolor{myoat}} categorical \\
 \midrule
 & \multicolumn{4}{p{0.9\textwidth}}{\textit{Question text:} What is your current marital status? }  \\
 & \cline{1-4}
\bfseries  & \bfseries  & \bfseries  & \bfseries N Missing: & \bfseries 0 \\
\bfseries  & \bfseries  & \bfseries  & \bfseries N Unique: & \bfseries 5 \\
 &  & \multicolumn{2}{r}{Never been married } & 870 \\
 &  & \multicolumn{2}{r}{Married } & 463 \\
 &  & \multicolumn{2}{r}{Divorced / Separated } & 123 \\
 &  & \multicolumn{2}{r}{Prefer not to say } & 23 \\
 &  & \multicolumn{2}{r}{Widowed } & 21 \\
 \midrule
\bfseries 19 & \bfseries english\_proficiency & \bfseries English Proficiency & \bfseries {\cellcolor{myyellow}} direct & \bfseries {\cellcolor{myoat}} categorical \\
 \midrule
 & \multicolumn{4}{p{0.9\textwidth}}{\textit{Question text:} How would you describe your proficiency in English? }  \\
 & \cline{1-4}
\bfseries  & \bfseries  & \bfseries  & \bfseries N Missing: & \bfseries 0 \\
\bfseries  & \bfseries  & \bfseries  & \bfseries N Unique: & \bfseries 5 \\
 &  & \multicolumn{2}{r}{Native speaker } & 886 \\
 &  & \multicolumn{2}{r}{Fluent } & 405 \\
 &  & \multicolumn{2}{r}{Advanced } & 160 \\
 &  & \multicolumn{2}{r}{Intermediate } & 42 \\
 &  & \multicolumn{2}{r}{Basic } & 7 \\
 \midrule
\bfseries 20 & \bfseries gender & \bfseries Gender & \bfseries {\cellcolor{mygreen}} constructed & \bfseries {\cellcolor{myoat}} categorical \\
 \midrule
 & \multicolumn{4}{p{0.9\textwidth}}{\textit{Question text:} How would you describe your proficiency in English? }  \\
 & \cline{1-4}
\bfseries  & \bfseries  & \bfseries  & \bfseries N Missing: & \bfseries 0 \\
\bfseries  & \bfseries  & \bfseries  & \bfseries N Unique: & \bfseries 4 \\
 &  & \multicolumn{2}{r}{Male } & 757 \\
 &  & \multicolumn{2}{r}{Female } & 718 \\
 &  & \multicolumn{2}{r}{Non-binary / third gender } & 21 \\
 &  & \multicolumn{2}{r}{Prefer not to say } & 4 \\
 & \multicolumn{4}{p{0.9\textwidth}}{\textit{Notes: Participants could chose Male, Female, Non-binary / third Gender, Prefer not to say, or write in their own response. Two independent annotators then categorised the self-describe responses only when abundantly clear they fit another category. See paper for details. }}  \\
 & \cline{1-4}
 \midrule
\bfseries 21 & \bfseries religion & \bfseries Dictionary of religion information. & \bfseries NA & \bfseries {\cellcolor{myoat}} dict \\
 \midrule
 & \multicolumn{4}{p{0.9\textwidth}}{\textit{Notes: Keys explained below. }}  \\
 & \cline{1-4}
 \midrule
\bfseries 22 & \bfseries religion\_self\_described & \bfseries Participant \{c\} self-description & \bfseries {\cellcolor{myyellow}} direct & \bfseries {\cellcolor{myoat}} string \\
 \midrule
 & \multicolumn{4}{p{0.9\textwidth}}{\textit{Question text:} What is your religious affiliation? }  \\
 & \cline{1-4}
\bfseries  & \bfseries  & \bfseries  & \bfseries N Missing: & \bfseries 0 \\
\bfseries  & \bfseries  & \bfseries  & \bfseries N Unique: & \bfseries 137 \\
 &  & \multicolumn{2}{r}{mean chars } & 12.2 \\
 &  & \multicolumn{2}{r}{std chars } & 5.7 \\
 &  & \multicolumn{2}{r}{min chars } & 2.0 \\
 &  & \multicolumn{2}{r}{max chars } & 112.0 \\
 & \multicolumn{4}{p{0.9\textwidth}}{\textit{Notes: Participant had option to type and Self Describe or select Prefer not to say. }}  \\
 & \cline{1-4}
 \midrule
\bfseries 23 & \bfseries religion\_categorised & \bfseries Granular categories of participant religion & \bfseries {\cellcolor{mygreen}} constructed & \bfseries {\cellcolor{myoat}} categorical \\
 \midrule
\bfseries  & \bfseries  & \bfseries  & \bfseries N Missing: & \bfseries 0 \\
\bfseries  & \bfseries  & \bfseries  & \bfseries N Unique: & \bfseries 12 \\
 &  & \multicolumn{2}{r}{Non-religious } & 762 \\
 &  & \multicolumn{2}{r}{Christian } & 487 \\
 &  & \multicolumn{2}{r}{Agnostic } & 71 \\
 &  & \multicolumn{2}{r}{Prefer not to say } & 59 \\
 &  & \multicolumn{2}{r}{Jewish } & 42 \\
 &  & \multicolumn{2}{r}{Muslim } & 31 \\
 &  & \multicolumn{2}{r}{Spiritual } & 18 \\
 &  & \multicolumn{2}{r}{Buddhist } & 12 \\
 &  & \multicolumn{2}{r}{Folk religion } & 6 \\
 &  & \multicolumn{2}{r}{Hindu } & 5 \\
 &  & \multicolumn{2}{r}{Other } & 4 \\
 &  & \multicolumn{2}{r}{Sikh } & 3 \\
 & \multicolumn{4}{p{0.9\textwidth}}{\textit{Notes: Two independent annotators manually verified all automated classifications (gpt-4-turbo) of the self-describe string. See paper for details. }}  \\
 & \cline{1-4}
 \midrule
\bfseries 24 & \bfseries religion\_simplified & \bfseries Simplified categories of participant religion & \bfseries {\cellcolor{mygreen}} constructed & \bfseries {\cellcolor{myoat}} categorical \\
 \midrule
\bfseries  & \bfseries  & \bfseries  & \bfseries N Missing: & \bfseries 0 \\
\bfseries  & \bfseries  & \bfseries  & \bfseries N Unique: & \bfseries 6 \\
 &  & \multicolumn{2}{r}{No Affiliation } & 851 \\
 &  & \multicolumn{2}{r}{Christian } & 487 \\
 &  & \multicolumn{2}{r}{Prefer not to say } & 59 \\
 &  & \multicolumn{2}{r}{Jewish } & 42 \\
 &  & \multicolumn{2}{r}{Muslim } & 31 \\
 &  & \multicolumn{2}{r}{Other } & 30 \\
 & \multicolumn{4}{p{0.9\textwidth}}{\textit{Notes: Simplified version of religion\_categorised for more aggregate analysis. }}  \\
 & \cline{1-4}
 \midrule
\bfseries 25 & \bfseries ethnicity & \bfseries Dictionary of ethnicity information. & \bfseries NA & \bfseries {\cellcolor{myoat}} dict \\
 \midrule
 & \multicolumn{4}{p{0.9\textwidth}}{\textit{Notes: Keys explained below. }}  \\
 & \cline{1-4}
 \midrule
\bfseries 26 & \bfseries ethnicity\_self\_described & \bfseries Participant \{c\} self-description & \bfseries {\cellcolor{myyellow}} direct & \bfseries {\cellcolor{myoat}} string \\
 \midrule
 & \multicolumn{4}{p{0.9\textwidth}}{\textit{Question text:} What is your ethnicity? }  \\
 & \cline{1-4}
\bfseries  & \bfseries  & \bfseries  & \bfseries N Missing: & \bfseries 0 \\
\bfseries  & \bfseries  & \bfseries  & \bfseries N Unique: & \bfseries 264 \\
 &  & \multicolumn{2}{r}{mean chars } & 9.2 \\
 &  & \multicolumn{2}{r}{std chars } & 6.2 \\
 &  & \multicolumn{2}{r}{min chars } & 3.0 \\
 &  & \multicolumn{2}{r}{max chars } & 99.0 \\
 & \multicolumn{4}{p{0.9\textwidth}}{\textit{Notes: Participant had option to type and Self Describe or select Prefer not to say. }}  \\
 & \cline{1-4}
 \midrule
\bfseries 27 & \bfseries ethnicity\_categorised & \bfseries Granular categories of participant ethnicity & \bfseries {\cellcolor{mygreen}} constructed & \bfseries {\cellcolor{myoat}} categorical \\
 \midrule
\bfseries  & \bfseries  & \bfseries  & \bfseries N Missing: & \bfseries 0 \\
\bfseries  & \bfseries  & \bfseries  & \bfseries N Unique: & \bfseries 9 \\
 &  & \multicolumn{2}{r}{White } & 969 \\
 &  & \multicolumn{2}{r}{Black / African } & 122 \\
 &  & \multicolumn{2}{r}{Hispanic / Latino } & 121 \\
 &  & \multicolumn{2}{r}{Asian } & 95 \\
 &  & \multicolumn{2}{r}{Prefer not to say } & 86 \\
 &  & \multicolumn{2}{r}{Mixed } & 68 \\
 &  & \multicolumn{2}{r}{Other } & 17 \\
 &  & \multicolumn{2}{r}{Middle Eastern / Arab } & 14 \\
 &  & \multicolumn{2}{r}{Indigenous / First Peoples } & 8 \\
 & \multicolumn{4}{p{0.9\textwidth}}{\textit{Notes: Two independent annotators manually verified all automated classifications (gpt-4-turbo) of the self-describe string. See paper for details. }}  \\
 & \cline{1-4}
 \midrule
\bfseries 28 & \bfseries ethnicity\_simplified & \bfseries Simplified categories of participant ethnicity & \bfseries {\cellcolor{mygreen}} constructed & \bfseries {\cellcolor{myoat}} categorical \\
 \midrule
\bfseries  & \bfseries  & \bfseries  & \bfseries N Missing: & \bfseries 0 \\
\bfseries  & \bfseries  & \bfseries  & \bfseries N Unique: & \bfseries 7 \\
 &  & \multicolumn{2}{r}{White } & 969 \\
 &  & \multicolumn{2}{r}{Black } & 122 \\
 &  & \multicolumn{2}{r}{Hispanic } & 121 \\
 &  & \multicolumn{2}{r}{Asian } & 95 \\
 &  & \multicolumn{2}{r}{Prefer not to say } & 86 \\
 &  & \multicolumn{2}{r}{Mixed } & 68 \\
 &  & \multicolumn{2}{r}{Other } & 39 \\
 & \multicolumn{4}{p{0.9\textwidth}}{\textit{Notes: Simplified version of ethnicity\_categorised for more aggregate analysis. }}  \\
 & \cline{1-4}
 \midrule
\bfseries 29 & \bfseries location & \bfseries Dictionary of location information. & \bfseries NA & \bfseries {\cellcolor{myoat}} dict \\
 \midrule
 & \multicolumn{4}{p{0.9\textwidth}}{\textit{Notes: Keys explained below. }}  \\
 & \cline{1-4}
 \midrule
\bfseries 30 & \bfseries location\_birth\_country & \bfseries Participant country of birth & \bfseries {\cellcolor{myyellow}} direct & \bfseries {\cellcolor{myoat}} categorical \\
 \midrule
 & \multicolumn{4}{p{0.9\textwidth}}{\textit{Question text:} In which country were you born? }  \\
 & \cline{1-4}
\bfseries  & \bfseries  & \bfseries  & \bfseries N Missing: & \bfseries 0 \\
\bfseries  & \bfseries  & \bfseries  & \bfseries N Unique: & \bfseries 75 \\
 &  & \multicolumn{2}{r}{Too many values to show } & - \\
 & \multicolumn{4}{p{0.9\textwidth}}{\textit{Notes: Selected from standardised dropdown country list. }}  \\
 & \cline{1-4}
 \midrule
\bfseries 31 & \bfseries location\_birth\_countryISO & \bfseries ISO 3166-1 alpha-3 code for the country of birth & \bfseries {\cellcolor{mygreen}} constructed & \bfseries {\cellcolor{myoat}} categorical \\
 \midrule
\bfseries  & \bfseries  & \bfseries  & \bfseries N Missing: & \bfseries 0 \\
\bfseries  & \bfseries  & \bfseries  & \bfseries N Unique: & \bfseries 75 \\
 &  & \multicolumn{2}{r}{Too many values to show } & - \\
 \midrule
\bfseries 32 & \bfseries location\_birth\_subregion & \bfseries Participant sub-region of birth & \bfseries {\cellcolor{mygreen}} constructed & \bfseries {\cellcolor{myoat}} categorical \\
 \midrule
\bfseries  & \bfseries  & \bfseries  & \bfseries N Missing: & \bfseries 0 \\
\bfseries  & \bfseries  & \bfseries  & \bfseries N Unique: & \bfseries 16 \\
 &  & \multicolumn{2}{r}{Too many values to show } & - \\
 & \multicolumn{4}{p{0.9\textwidth}}{\textit{Notes: Mapped from country of birth, based on United Nations defined subregions. }}  \\
 & \cline{1-4}
 \midrule
\bfseries 33 & \bfseries location\_reside\_country & \bfseries Participant country of residence & \bfseries {\cellcolor{myyellow}} direct & \bfseries {\cellcolor{myoat}} categorical \\
 \midrule
 & \multicolumn{4}{p{0.9\textwidth}}{\textit{Question text:} In which country do you currently reside? }  \\
 & \cline{1-4}
\bfseries  & \bfseries  & \bfseries  & \bfseries N Missing: & \bfseries 0 \\
\bfseries  & \bfseries  & \bfseries  & \bfseries N Unique: & \bfseries 38 \\
 &  & \multicolumn{2}{r}{Too many values to show } & - \\
 & \multicolumn{4}{p{0.9\textwidth}}{\textit{Notes: Selected from standardised dropdown country list. }}  \\
 & \cline{1-4}
 \midrule
\bfseries 34 & \bfseries location\_reside\_countryISO & \bfseries ISO 3166-1 alpha-3 code for the country of residence & \bfseries {\cellcolor{mygreen}} constructed & \bfseries {\cellcolor{myoat}} categorical \\
 \midrule
\bfseries  & \bfseries  & \bfseries  & \bfseries N Missing: & \bfseries 0 \\
\bfseries  & \bfseries  & \bfseries  & \bfseries N Unique: & \bfseries 38 \\
 &  & \multicolumn{2}{r}{Too many values to show } & - \\
 \midrule
\bfseries 35 & \bfseries location\_reside\_subregion & \bfseries Participant sub-region of residence & \bfseries {\cellcolor{mygreen}} constructed & \bfseries {\cellcolor{myoat}} categorical \\
 \midrule
\bfseries  & \bfseries  & \bfseries  & \bfseries N Missing: & \bfseries 0 \\
\bfseries  & \bfseries  & \bfseries  & \bfseries N Unique: & \bfseries 11 \\
 &  & \multicolumn{2}{r}{Too many values to show } & - \\
 & \multicolumn{4}{p{0.9\textwidth}}{\textit{Notes: Mapped from country of residence, based on United Nations defined subregions. }}  \\
 & \cline{1-4}
 \midrule
\bfseries 36 & \bfseries location\_same\_birth\_reside\_country & \bfseries Whether the participant was born and resides in the same country & \bfseries {\cellcolor{mygreen}} constructed & \bfseries {\cellcolor{myoat}} binary \\
 \midrule
\bfseries  & \bfseries  & \bfseries  & \bfseries N Missing: & \bfseries 0 \\
\bfseries  & \bfseries  & \bfseries  & \bfseries N Unique: & \bfseries 3 \\
 &  & \multicolumn{2}{r}{Yes } & 1320 \\
 &  & \multicolumn{2}{r}{No } & 177 \\
 &  & \multicolumn{2}{r}{Prefer not to say } & 3 \\
 \midrule
\bfseries 37 & \bfseries location\_special\_region & \bfseries Adjusted regional categories for unique sample properties & \bfseries {\cellcolor{mygreen}} constructed & \bfseries {\cellcolor{myoat}} categorical \\
 \midrule
\bfseries  & \bfseries  & \bfseries  & \bfseries N Missing: & \bfseries 0 \\
\bfseries  & \bfseries  & \bfseries  & \bfseries N Unique: & \bfseries 11 \\
 &  & \multicolumn{2}{r}{US } & 338 \\
 &  & \multicolumn{2}{r}{Europe } & 313 \\
 &  & \multicolumn{2}{r}{UK } & 292 \\
 &  & \multicolumn{2}{r}{Latin America and the Caribbean } & 146 \\
 &  & \multicolumn{2}{r}{Australia and New Zealand } & 129 \\
 &  & \multicolumn{2}{r}{Africa } & 118 \\
 &  & \multicolumn{2}{r}{Asia } & 60 \\
 &  & \multicolumn{2}{r}{Northern America } & 50 \\
 &  & \multicolumn{2}{r}{Middle East } & 50 \\
 &  & \multicolumn{2}{r}{Prefer not to say } & 3 \\
 &  & \multicolumn{2}{r}{Oceania } & 1 \\
 & \multicolumn{4}{p{0.9\textwidth}}{\textit{Notes: Within regions and sub-regions, some countries are split out to better represent sample density (e.g., treating UK and US samples seperately from Europe and North America). }}  \\
 & \cline{1-4}
 \midrule
\bfseries 38 & \bfseries study\_id & \bfseries Unique study idenfitier on Prolific & \bfseries {\cellcolor{myred}} meta & \bfseries {\cellcolor{myoat}} string id \\
 \midrule
\bfseries  & \bfseries  & \bfseries  & \bfseries N Missing: & \bfseries 0 \\
\bfseries  & \bfseries  & \bfseries  & \bfseries N Unique: & \bfseries 51 \\
 \midrule
\bfseries 39 & \bfseries study\_locale & \bfseries Recruitment country of Prolific study & \bfseries {\cellcolor{myred}} meta & \bfseries {\cellcolor{myoat}} categorical \\
 \midrule
\bfseries  & \bfseries  & \bfseries  & \bfseries N Missing: & \bfseries 0 \\
\bfseries  & \bfseries  & \bfseries  & \bfseries N Unique: & \bfseries 33 \\
 &  & \multicolumn{2}{r}{Too many values to show } & - \\
 \midrule
\bfseries 40 & \bfseries generated\_datetime & \bfseries Recorded date of the survey completion & \bfseries {\cellcolor{myred}} meta & \bfseries {\cellcolor{myoat}} datetime \\
 \midrule
\bfseries  & \bfseries  & \bfseries  & \bfseries N Missing: & \bfseries 0 \\
\bfseries  & \bfseries  & \bfseries  & \bfseries N Unique: & \bfseries 1492 \\
 &  & \multicolumn{2}{r}{earliest date } & 2023-11-22 15:48:46 \\
 &  & \multicolumn{2}{r}{latest\_date } & 2023-12-22 06:56:27 \\
 & \multicolumn{4}{p{0.9\textwidth}}{\textit{Notes: End time, not start time }}  \\
 & \cline{1-4}
 \midrule
\bfseries 41 & \bfseries timing\_duration\_s & \bfseries Duration of the survey session (in seconds) & \bfseries {\cellcolor{myred}} meta & \bfseries {\cellcolor{myoat}} float \\
 \midrule
\bfseries  & \bfseries  & \bfseries  & \bfseries N Missing: & \bfseries 0 \\
\bfseries  & \bfseries  & \bfseries  & \bfseries N Unique: & \bfseries 977 \\
 &  & \multicolumn{2}{r}{mean } & 2154.2 \\
 &  & \multicolumn{2}{r}{std } & 20557.1 \\
 &  & \multicolumn{2}{r}{min } & 160.0 \\
 &  & \multicolumn{2}{r}{max } & 529927.0 \\
 & \multicolumn{4}{p{0.9\textwidth}}{\textit{Notes: Extreme values are caused by participants completing task in multiple sessions. }}  \\
 & \cline{1-4}
 \midrule
\bfseries 42 & \bfseries timing\_duration\_mins & \bfseries Duration of the survey session (in minutes) & \bfseries {\cellcolor{mygreen}} constructed & \bfseries {\cellcolor{myoat}} float \\
 \midrule
\bfseries  & \bfseries  & \bfseries  & \bfseries N Missing: & \bfseries 0 \\
\bfseries  & \bfseries  & \bfseries  & \bfseries N Unique: & \bfseries 977 \\
 &  & \multicolumn{2}{r}{mean } & 35.9 \\
 &  & \multicolumn{2}{r}{std } & 342.6 \\
 &  & \multicolumn{2}{r}{min } & 2.7 \\
 &  & \multicolumn{2}{r}{max } & 8832.1 \\
 & \multicolumn{4}{p{0.9\textwidth}}{\textit{Notes: timing\_duration\_s / 60. Extreme values are caused by participants completing task in multiple sessions. }}  \\
 & \cline{1-4}
 \midrule
\bfseries 43 & \bfseries included\_in\_UK\_REP & \bfseries Indicator if participant was included in the rebalanced UK representative sample & \bfseries {\cellcolor{mygreen}} constructed & \bfseries {\cellcolor{myoat}} binary \\
 \midrule
\bfseries  & \bfseries  & \bfseries  & \bfseries N Missing: & \bfseries 0 \\
\bfseries  & \bfseries  & \bfseries  & \bfseries N Unique: & \bfseries 2 \\
 &  & \multicolumn{2}{r}{False } & 1257 \\
 &  & \multicolumn{2}{r}{True } & 243 \\
 & \multicolumn{4}{p{0.9\textwidth}}{\textit{Notes: Census-representative samples were rebalanced to mitigate sampling issues. See paper for details. }}  \\
 & \cline{1-4}
 \midrule
\bfseries 44 & \bfseries included\_in\_US\_REP & \bfseries Indicator if participant was included in the rebalanced US representative sample & \bfseries {\cellcolor{mygreen}} constructed & \bfseries {\cellcolor{myoat}} binary \\
 \midrule
\bfseries  & \bfseries  & \bfseries  & \bfseries N Missing: & \bfseries 0 \\
\bfseries  & \bfseries  & \bfseries  & \bfseries N Unique: & \bfseries 2 \\
 &  & \multicolumn{2}{r}{False } & 1270 \\
 &  & \multicolumn{2}{r}{True } & 230 \\
 & \multicolumn{4}{p{0.9\textwidth}}{\textit{Notes: Census-representative samples were rebalanced to mitigate sampling issues. See paper for details. }}  \\
 & \cline{1-4}
 \midrule
\bfseries 45 & \bfseries included\_in\_balanced\_subset & \bfseries Indicator if participant's conversations are included in the balanced subset & \bfseries {\cellcolor{mygreen}} constructed & \bfseries {\cellcolor{myoat}} binary \\
 \midrule
\bfseries  & \bfseries  & \bfseries  & \bfseries N Missing: & \bfseries 0 \\
\bfseries  & \bfseries  & \bfseries  & \bfseries N Unique: & \bfseries 2 \\
 &  & \multicolumn{2}{r}{True } & 1246 \\
 &  & \multicolumn{2}{r}{False } & 254 \\
 & \multicolumn{4}{p{0.9\textwidth}}{\textit{Notes: Balanced subset was created to equally sample conversations of three types (unguided, values, controversy).
We only include participants who have at least one of each conversation type, and then ensure equal numbers of each type are retained.
See paper for details. }}  \\
 & \cline{1-4}
\end{longtable}

\clearpage
\subsection{Conversations Codebook}
\label{sec:convos_codebook}
\fontsize{6pt}{6pt}\selectfont
\begin{longtable}{p{0.005\textwidth}p{0.20\textwidth}p{0.40\textwidth}p{0.1\textwidth}p{0.1\textwidth}}
\toprule
 & VARIABLE & LABEL & CATEGORY & TYPE \\
\midrule
\endfirsthead
\toprule
 & VARIABLE & LABEL & CATEGORY & TYPE \\
\midrule
\endhead
\midrule
\multicolumn{5}{r}{Continued on next page} \\
\midrule
\endfoot
\bottomrule
\endlastfoot
\bfseries 0 & \bfseries user\_id & \bfseries Unique participant identifier & \bfseries {\cellcolor{myred}} meta & \bfseries {\cellcolor{myoat}} string id \\
 \midrule
\bfseries  & \bfseries  & \bfseries  & \bfseries N Missing: & \bfseries 0 \\
\bfseries  & \bfseries  & \bfseries  & \bfseries N Unique: & \bfseries 1396 \\
 & \multicolumn{4}{p{0.9\textwidth}}{\textit{Notes: Pseudonymized from Prolific worker ID. Used to link conversation data to survey data. }}  \\
 & \cline{1-4}
 \midrule
\bfseries 1 & \bfseries conversation\_id & \bfseries Unique conversation identifier & \bfseries {\cellcolor{myred}} meta & \bfseries {\cellcolor{myoat}} string id \\
 \midrule
\bfseries  & \bfseries  & \bfseries  & \bfseries N Missing: & \bfseries 0 \\
\bfseries  & \bfseries  & \bfseries  & \bfseries N Unique: & \bfseries 8011 \\
 \midrule
\bfseries 2 & \bfseries opening\_prompt & \bfseries Opening human-written prompt of the conversation & \bfseries {\cellcolor{myyellow}} direct & \bfseries {\cellcolor{myoat}} string \\
 \midrule
 & \multicolumn{4}{p{0.9\textwidth}}{\textit{Question text:} Now start the conversation with your question, request or statement. }  \\
 & \cline{1-4}
\bfseries  & \bfseries  & \bfseries  & \bfseries N Missing: & \bfseries 0 \\
\bfseries  & \bfseries  & \bfseries  & \bfseries N Unique: & \bfseries 7811 \\
 &  & \multicolumn{2}{r}{mean chars } & 65.7 \\
 &  & \multicolumn{2}{r}{std chars } & 59.2 \\
 &  & \multicolumn{2}{r}{min chars } & 2.0 \\
 &  & \multicolumn{2}{r}{max chars } & 1195.0 \\
 & \multicolumn{4}{p{0.9\textwidth}}{\textit{Notes: We provide the following soft guidance:
Need some inspiration? You can request help with a task (like writing a recipe, organising an activity or event, completing an assignment)... You can chitchat, have casual conversation or seek personal advice. You can ask questions about the world, current events or your viewpoints. }}  \\
 & \cline{1-4}
 \midrule
\bfseries 3 & \bfseries open\_feedback & \bfseries Participant written feedback on the conversation as a whole. & \bfseries {\cellcolor{myyellow}} direct & \bfseries {\cellcolor{myoat}} string \\
 \midrule
 & \multicolumn{4}{p{0.9\textwidth}}{\textit{Question text:} Give the model some feedback on the conversation as whole. Hypothetically, what would an ideal interaction for you look like here? What was good and what was bad? What (if anything) was missing? What would you change to make the conversation better?
Please write 2-5 sentences in your own words. }  \\
 & \cline{1-4}
\bfseries  & \bfseries  & \bfseries  & \bfseries N Missing: & \bfseries 0 \\
\bfseries  & \bfseries  & \bfseries  & \bfseries N Unique: & \bfseries 7953 \\
 &  & \multicolumn{2}{r}{mean chars } & 160.1 \\
 &  & \multicolumn{2}{r}{std chars } & 106.4 \\
 &  & \multicolumn{2}{r}{min chars } & 2.0 \\
 &  & \multicolumn{2}{r}{max chars } & 1581.0 \\
 & \multicolumn{4}{p{0.9\textwidth}}{\textit{Notes: Entry box reads: Enter text here. Do not copy and paste. }}  \\
 & \cline{1-4}
 \midrule
\bfseries 4 & \bfseries conversation\_type & \bfseries Type of conversation (from pre-defined categories) & \bfseries {\cellcolor{myyellow}} direct & \bfseries {\cellcolor{myoat}} categorical \\
 \midrule
 & \multicolumn{4}{p{0.9\textwidth}}{\textit{Question text:} Choose what type of conversation you want to have. }  \\
 & \cline{1-4}
\bfseries  & \bfseries  & \bfseries  & \bfseries N Missing: & \bfseries 0 \\
\bfseries  & \bfseries  & \bfseries  & \bfseries N Unique: & \bfseries 3 \\
 &  & \multicolumn{2}{r}{unguided } & 3113 \\
 &  & \multicolumn{2}{r}{values guided } & 2460 \\
 &  & \multicolumn{2}{r}{controversy guided } & 2438 \\
 & \multicolumn{4}{p{0.9\textwidth}}{\textit{Notes: Participants pick from the following radio buttons:
Unguided. Ask, request or talk to the model about anything . It is up to you!
Values guided. Ask, request or talk to the model about something important to you or that represents your values. This could be related to work, religion, family and relationship, politics or culture.
Controversy guided. Ask, request or talk to the model about something controversial or where people would disagree in your community, culture or country.
We also provide the additional instruction: Remember if you are here as a paid study participant, you need to do two of each type. If you are here as a volunteer, then take your pick!
 }}  \\
 & \cline{1-4}
 \midrule
\bfseries 5 & \bfseries conversation\_turns & \bfseries Number of human-model turns (back-and-forths) in the conversation. & \bfseries {\cellcolor{myred}} meta & \bfseries {\cellcolor{myoat}} int \\
 \midrule
\bfseries  & \bfseries  & \bfseries  & \bfseries N Missing: & \bfseries 0 \\
\bfseries  & \bfseries  & \bfseries  & \bfseries N Unique: & \bfseries 13 \\
 &  & \multicolumn{2}{r}{mean } & 3.4 \\
 &  & \multicolumn{2}{r}{std } & 1.6 \\
 &  & \multicolumn{2}{r}{min } & 2.0 \\
 &  & \multicolumn{2}{r}{max } & 22.0 \\
 & \multicolumn{4}{p{0.9\textwidth}}{\textit{Notes: We force 2 turns as the minimum. After the opening turn, we give the instruction:
Now continue the conversation. Conversations can be between 2 and 10 turns. Try to vary the length. When you're done, click Finish. }}  \\
 & \cline{1-4}
 \midrule
\bfseries 6 & \bfseries conversation\_history & \bfseries Full conversation history (human and model messages, with scores and model metadata) & \bfseries {\cellcolor{myyellow}} direct & \bfseries {\cellcolor{myoat}} dict \\
 \midrule
 &  & \multicolumn{2}{r}{Too many values to show } & - \\
 & \multicolumn{4}{p{0.9\textwidth}}{\textit{Notes: We provide an example of what this nested conversation history looks like below. }}  \\
 & \cline{1-4}
 \midrule
\bfseries 7 & \bfseries performance\_attributes & \bfseries How well the top-rated model response performed across different attributes & \bfseries nested & \bfseries {\cellcolor{myoat}} dict \\
 \midrule
 & \multicolumn{4}{p{0.9\textwidth}}{\textit{Question text:} Tell us how the model performed. Consider your first message and the top-rated response.
Rate the following statements about the performance across different attributes.
This response... }  \\
 & \cline{1-4}
\bfseries  & \bfseries  & \bfseries  & \bfseries N Missing: & \bfseries 1824 \\
\bfseries  & \bfseries  & \bfseries  & \bfseries N Unique: & \bfseries 7532 \\
 & \cline{1-4}
 & values & \multicolumn{3}{p{0.6\textwidth}}{...reflected my values or cultural perspective } \\
 &  & \multicolumn{2}{r}{mean } & 74.1 \\
 &  & \multicolumn{2}{r}{std } & 22.2 \\
 &  & \multicolumn{2}{r}{min } & 1.0 \\
 &  & \multicolumn{2}{r}{max } & 100.0 \\
 & \cline{1-4}
 & fluency & \multicolumn{3}{p{0.6\textwidth}}{...was well-written and coherent } \\
 &  & \multicolumn{2}{r}{mean } & 84.3 \\
 &  & \multicolumn{2}{r}{std } & 18.3 \\
 &  & \multicolumn{2}{r}{min } & 1.0 \\
 &  & \multicolumn{2}{r}{max } & 100.0 \\
 & \cline{1-4}
 & factuality & \multicolumn{3}{p{0.6\textwidth}}{...was factual and informative } \\
 &  & \multicolumn{2}{r}{mean } & 79.2 \\
 &  & \multicolumn{2}{r}{std } & 21.5 \\
 &  & \multicolumn{2}{r}{min } & 1.0 \\
 &  & \multicolumn{2}{r}{max } & 100.0 \\
 & \cline{1-4}
 & safety & \multicolumn{3}{p{0.6\textwidth}}{...was safe and doesn't risk harm to myself and others } \\
 &  & \multicolumn{2}{r}{mean } & 85.1 \\
 &  & \multicolumn{2}{r}{std } & 19.3 \\
 &  & \multicolumn{2}{r}{min } & 1.0 \\
 &  & \multicolumn{2}{r}{max } & 100.0 \\
 & \cline{1-4}
 & diversity & \multicolumn{3}{p{0.6\textwidth}}{...summarised multiple viewpoints or different worldviews } \\
 &  & \multicolumn{2}{r}{mean } & 68.7 \\
 &  & \multicolumn{2}{r}{std } & 25.3 \\
 &  & \multicolumn{2}{r}{min } & 1.0 \\
 &  & \multicolumn{2}{r}{max } & 100.0 \\
 & \cline{1-4}
 & creativity & \multicolumn{3}{p{0.6\textwidth}}{...was creative and inspiring } \\
 &  & \multicolumn{2}{r}{mean } & 63.7 \\
 &  & \multicolumn{2}{r}{std } & 26.1 \\
 &  & \multicolumn{2}{r}{min } & 1.0 \\
 &  & \multicolumn{2}{r}{max } & 100.0 \\
 & \cline{1-4}
 & helpfulness & \multicolumn{3}{p{0.6\textwidth}}{...was helpful and relevant to my request } \\
 &  & \multicolumn{2}{r}{mean } & 81.5 \\
 &  & \multicolumn{2}{r}{std } & 21.9 \\
 &  & \multicolumn{2}{r}{min } & 1.0 \\
 &  & \multicolumn{2}{r}{max } & 100.0 \\
 & \multicolumn{4}{p{0.9\textwidth}}{\textit{Notes: Sliders from [Performed very poorly] to [Performed very well] are recorded on a 1-100 scale. Participant does not see numeric value.
Note that the attributes align choice\_attributes, as well as with the stated preference ratings from The Survey.
Participants had option to select N/A, which is recorded as Null.
N Missing indicates the number of participants who have at least one missing value in the nested columns.
N Unique indicates the unique combinations of use cases selected by participants.
There was no option for 'other'.
Note, these sliders run from 1-100 (on Dynabench). The sliders for stated\_prefs (in Survey on Qualtrics) run 0-100. }}  \\
 & \cline{1-4}
 \midrule
\bfseries 8 & \bfseries choice\_attributes & \bfseries How different attributes influenced the participant's choice of the top-rated model response & \bfseries {\cellcolor{myyellow}} direct & \bfseries {\cellcolor{myoat}} dict \\
 \midrule
 & \multicolumn{4}{p{0.9\textwidth}}{\textit{Question text:} Tell us why you chose this response over others. Consider your first message and top-rated response compared to other responses.
Rate the following statements about the importance of different attributes in your decision.
I chose this response... }  \\
 & \cline{1-4}
\bfseries  & \bfseries  & \bfseries  & \bfseries N Missing: & \bfseries 1740 \\
\bfseries  & \bfseries  & \bfseries  & \bfseries N Unique: & \bfseries 7526 \\
 & \cline{1-4}
 & values & \multicolumn{3}{p{0.6\textwidth}}{...reflected my values or cultural perspective } \\
 &  & \multicolumn{2}{r}{mean } & 66.9 \\
 &  & \multicolumn{2}{r}{std } & 27.2 \\
 &  & \multicolumn{2}{r}{min } & 1.0 \\
 &  & \multicolumn{2}{r}{max } & 100.0 \\
 & \cline{1-4}
 & fluency & \multicolumn{3}{p{0.6\textwidth}}{...was well-written and coherent } \\
 &  & \multicolumn{2}{r}{mean } & 82.5 \\
 &  & \multicolumn{2}{r}{std } & 18.5 \\
 &  & \multicolumn{2}{r}{min } & 1.0 \\
 &  & \multicolumn{2}{r}{max } & 100.0 \\
 & \cline{1-4}
 & factuality & \multicolumn{3}{p{0.6\textwidth}}{...was factual and informative } \\
 &  & \multicolumn{2}{r}{mean } & 79.3 \\
 &  & \multicolumn{2}{r}{std } & 21.0 \\
 &  & \multicolumn{2}{r}{min } & 1.0 \\
 &  & \multicolumn{2}{r}{max } & 100.0 \\
 & \cline{1-4}
 & safety & \multicolumn{3}{p{0.6\textwidth}}{...was safe and doesn't risk harm to myself and others } \\
 &  & \multicolumn{2}{r}{mean } & 72.1 \\
 &  & \multicolumn{2}{r}{std } & 27.8 \\
 &  & \multicolumn{2}{r}{min } & 1.0 \\
 &  & \multicolumn{2}{r}{max } & 100.0 \\
 & \cline{1-4}
 & diversity & \multicolumn{3}{p{0.6\textwidth}}{...summarised multiple viewpoints or different worldviews } \\
 &  & \multicolumn{2}{r}{mean } & 66.0 \\
 &  & \multicolumn{2}{r}{std } & 26.5 \\
 &  & \multicolumn{2}{r}{min } & 1.0 \\
 &  & \multicolumn{2}{r}{max } & 100.0 \\
 & \cline{1-4}
 & creativity & \multicolumn{3}{p{0.6\textwidth}}{...was creative and inspiring } \\
 &  & \multicolumn{2}{r}{mean } & 62.1 \\
 &  & \multicolumn{2}{r}{std } & 27.1 \\
 &  & \multicolumn{2}{r}{min } & 1.0 \\
 &  & \multicolumn{2}{r}{max } & 100.0 \\
 & \cline{1-4}
 & helpfulness & \multicolumn{3}{p{0.6\textwidth}}{...was helpful and relevant to my request } \\
 &  & \multicolumn{2}{r}{mean } & 82.5 \\
 &  & \multicolumn{2}{r}{std } & 20.0 \\
 &  & \multicolumn{2}{r}{min } & 1.0 \\
 &  & \multicolumn{2}{r}{max } & 100.0 \\
 & \multicolumn{4}{p{0.9\textwidth}}{\textit{Notes: Sliders from [Very unimportant] to [Very important] are recorded on a 1-100 scale. Participant does not see numeric value.
Note that the attributes align with performance\_attributes, as well as the stated preference ratings from The Survey.
Participants had option to select N/A, which is recorded as Null.
num\_missing indicates the number of participants who have at least one missing value in the nested columns.
num\_unique indicates the unique combinations of use cases selected by participants.
There was no option for 'other'.
Note, these sliders run from 1-100 (on Dynabench). The sliders for stated\_prefs (in Survey on Qualtrics) run 0-100. }}  \\
 & \cline{1-4}
 \midrule
\bfseries 9 & \bfseries generated\_datetime & \bfseries Recorded date of the conversation completion & \bfseries {\cellcolor{myred}} meta & \bfseries {\cellcolor{myoat}} datetime \\
 \midrule
\bfseries  & \bfseries  & \bfseries  & \bfseries N Missing: & \bfseries 0 \\
\bfseries  & \bfseries  & \bfseries  & \bfseries N Unique: & \bfseries 7820 \\
 &  & \multicolumn{2}{r}{earliest date } & 2023-11-22 15:55:46 \\
 &  & \multicolumn{2}{r}{latest\_date } & 2023-12-22 08:04:46 \\
 & \multicolumn{4}{p{0.9\textwidth}}{\textit{Notes: Recorded at end of conversation, before fine-grained feedback page shown. }}  \\
 & \cline{1-4}
 \midrule
\bfseries 10 & \bfseries timing\_duration\_s & \bfseries Duration of the conversation (in seconds) & \bfseries {\cellcolor{myred}} meta & \bfseries {\cellcolor{myoat}} float \\
 \midrule
\bfseries  & \bfseries  & \bfseries  & \bfseries N Missing: & \bfseries 0 \\
\bfseries  & \bfseries  & \bfseries  & \bfseries N Unique: & \bfseries 7656 \\
 &  & \multicolumn{2}{r}{mean } & 555.9 \\
 &  & \multicolumn{2}{r}{std } & 422.1 \\
 &  & \multicolumn{2}{r}{min } & 73.5 \\
 &  & \multicolumn{2}{r}{max } & 17145.8 \\
 & \multicolumn{4}{p{0.9\textwidth}}{\textit{Notes: Extreme values are caused by participants completing task in multiple sessions. }}  \\
 & \cline{1-4}
 \midrule
\bfseries 11 & \bfseries timing\_duration\_mins & \bfseries Duration of the conversation (in minutes) & \bfseries {\cellcolor{mygreen}} constructed & \bfseries {\cellcolor{myoat}} float \\
 \midrule
\bfseries  & \bfseries  & \bfseries  & \bfseries N Missing: & \bfseries 0 \\
\bfseries  & \bfseries  & \bfseries  & \bfseries N Unique: & \bfseries 1948 \\
 &  & \multicolumn{2}{r}{mean } & 9.3 \\
 &  & \multicolumn{2}{r}{std } & 7.0 \\
 &  & \multicolumn{2}{r}{min } & 1.2 \\
 &  & \multicolumn{2}{r}{max } & 285.8 \\
 & \multicolumn{4}{p{0.9\textwidth}}{\textit{Notes: timing\_duration\_s / 60. Extreme values are caused by participants completing task in multiple sessions. }}  \\
 & \cline{1-4}
 \midrule
\bfseries 12 & \bfseries included\_in\_balanced\_subset & \bfseries Indicator if participant's conversations are included in the balanced subset & \bfseries {\cellcolor{mygreen}} constructed & \bfseries {\cellcolor{myoat}} binary \\
 \midrule
\bfseries  & \bfseries  & \bfseries  & \bfseries N Missing: & \bfseries 0 \\
\bfseries  & \bfseries  & \bfseries  & \bfseries N Unique: & \bfseries 2 \\
 &  & \multicolumn{2}{r}{True } & 6696 \\
 &  & \multicolumn{2}{r}{False } & 1315 \\
 & \multicolumn{4}{p{0.9\textwidth}}{\textit{Notes: Balanced subset was created to equally sample conversations of three types (unguided, values, controversy).
We only include participants who have at least one of each conversation type, and then ensure equal numbers of each type are retained.
See paper for details. }}  \\
 & \cline{1-4}
\end{longtable}

\clearpage
\subsection{Utterances Codebook}
\label{sec:utterances_codebook}
\fontsize{6pt}{6pt}\selectfont
\begin{longtable}{p{0.005\textwidth}p{0.20\textwidth}p{0.40\textwidth}p{0.1\textwidth}p{0.1\textwidth}}
\toprule
 & VARIABLE & LABEL & CATEGORY & TYPE \\
\midrule
\endfirsthead
\toprule
 & VARIABLE & LABEL & CATEGORY & TYPE \\
\midrule
\endhead
\midrule
\multicolumn{5}{r}{Continued on next page} \\
\midrule
\endfoot
\bottomrule
\endlastfoot
\bfseries 0 & \bfseries user\_id & \bfseries Unique participant identifier & \bfseries {\cellcolor{myred}} meta & \bfseries {\cellcolor{myoat}} string id \\
 \midrule
\bfseries  & \bfseries  & \bfseries  & \bfseries N Missing: & \bfseries 0 \\
\bfseries  & \bfseries  & \bfseries  & \bfseries N Unique: & \bfseries 1396 \\
 & \multicolumn{4}{p{0.9\textwidth}}{\textit{Notes: Pseudonymized from Prolific worker ID. Used to link utterance data to survey data. }}  \\
 & \cline{1-4}
 \midrule
\bfseries 1 & \bfseries conversation\_id & \bfseries Unique conversation identifier & \bfseries {\cellcolor{myred}} meta & \bfseries {\cellcolor{myoat}} string id \\
 \midrule
\bfseries  & \bfseries  & \bfseries  & \bfseries N Missing: & \bfseries 0 \\
\bfseries  & \bfseries  & \bfseries  & \bfseries N Unique: & \bfseries 8011 \\
 & \multicolumn{4}{p{0.9\textwidth}}{\textit{Notes: Used to link utterance data to conversation data. }}  \\
 & \cline{1-4}
 \midrule
\bfseries 2 & \bfseries interaction\_id & \bfseries Unique interaction identifier, where an interaction is a turn within a conversation (single human message with multiple model responses) & \bfseries {\cellcolor{myred}} meta & \bfseries {\cellcolor{myoat}} string id \\
 \midrule
\bfseries  & \bfseries  & \bfseries  & \bfseries N Missing: & \bfseries 0 \\
\bfseries  & \bfseries  & \bfseries  & \bfseries N Unique: & \bfseries 27172 \\
 \midrule
\bfseries 3 & \bfseries utterance\_id & \bfseries Unique utterance identifier, where an utterance is a single human message - single model response pair & \bfseries {\cellcolor{myred}} meta & \bfseries {\cellcolor{myoat}} string id \\
 \midrule
\bfseries  & \bfseries  & \bfseries  & \bfseries N Missing: & \bfseries 0 \\
\bfseries  & \bfseries  & \bfseries  & \bfseries N Unique: & \bfseries 68371 \\
 \midrule
\bfseries 4 & \bfseries within\_turn\_id & \bfseries Within turn identifier of up to four model responses to a single human message & \bfseries {\cellcolor{myred}} meta & \bfseries {\cellcolor{myoat}} string id \\
 \midrule
\bfseries  & \bfseries  & \bfseries  & \bfseries N Missing: & \bfseries 0 \\
\bfseries  & \bfseries  & \bfseries  & \bfseries N Unique: & \bfseries 4 \\
 & \multicolumn{4}{p{0.9\textwidth}}{\textit{Notes: Order is random, not based on score or presentation in interface }}  \\
 & \cline{1-4}
 \midrule
\bfseries 5 & \bfseries conversation\_type & \bfseries Type of conversation (from pre-defined categories) & \bfseries {\cellcolor{myyellow}} direct & \bfseries {\cellcolor{myoat}} categorical \\
 \midrule
 & \multicolumn{4}{p{0.9\textwidth}}{\textit{Question text:} Choose what type of conversation you want to have. }  \\
 & \cline{1-4}
\bfseries  & \bfseries  & \bfseries  & \bfseries N Missing: & \bfseries 0 \\
\bfseries  & \bfseries  & \bfseries  & \bfseries N Unique: & \bfseries 3 \\
 &  & \multicolumn{2}{r}{unguided } & 3113 \\
 &  & \multicolumn{2}{r}{values guided } & 2460 \\
 &  & \multicolumn{2}{r}{controversy guided } & 2438 \\
 & \multicolumn{4}{p{0.9\textwidth}}{\textit{Notes: Participants pick from the following radio buttons:
Unguided. Ask, request or talk to the model about anything . It is up to you!
Values guided. Ask, request or talk to the model about something important to you or that represents your values . This could be related to work, religion, family and relationship, politics or culture.
Controversy guided. Ask, request or talk to the model about something controversial or where people would disagree in your community, culture or country.
We also provide the additional instruction: Remember if you are here as a paid study participant, you need to do two of each type. If you are here as a volunteer, then take your pick!
 }}  \\
 & \cline{1-4}
 \midrule
\bfseries 6 & \bfseries turn & \bfseries Turn of conversation when prompt was entered & \bfseries {\cellcolor{myred}} meta & \bfseries {\cellcolor{myoat}} int \\
 \midrule
\bfseries  & \bfseries  & \bfseries  & \bfseries N Missing: & \bfseries 0 \\
\bfseries  & \bfseries  & \bfseries  & \bfseries N Unique: & \bfseries 22 \\
 &  & \multicolumn{2}{r}{mean } & 1.2 \\
 &  & \multicolumn{2}{r}{std } & 1.6 \\
 &  & \multicolumn{2}{r}{min } & 0.0 \\
 &  & \multicolumn{2}{r}{max } & 21.0 \\
 & \multicolumn{4}{p{0.9\textwidth}}{\textit{Notes: In the paper, we refer to the first turn as T=1. Here, we index the first turn as 0. }}  \\
 & \cline{1-4}
 \midrule
\bfseries 7 & \bfseries model\_name & \bfseries Name of LLM & \bfseries {\cellcolor{myred}} meta & \bfseries {\cellcolor{myoat}} categorical \\
 \midrule
\bfseries  & \bfseries  & \bfseries  & \bfseries N Missing: & \bfseries 0 \\
\bfseries  & \bfseries  & \bfseries  & \bfseries N Unique: & \bfseries 21 \\
 &  & \multicolumn{2}{r}{command } & 4812 \\
 &  & \multicolumn{2}{r}{claude-instant-1 } & 4292 \\
 &  & \multicolumn{2}{r}{models/chat-bison-001 } & 4168 \\
 &  & \multicolumn{2}{r}{HuggingFaceH4/zephyr-7b-beta } & 4133 \\
 &  & \multicolumn{2}{r}{meta-llama/Llama-2-7b-chat-hf } & 3995 \\
 &  & \multicolumn{2}{r}{command-light } & 3929 \\
 &  & \multicolumn{2}{r}{command-nightly } & 3816 \\
 &  & \multicolumn{2}{r}{gpt-4-1106-preview } & 3735 \\
 &  & \multicolumn{2}{r}{gpt-4 } & 3515 \\
 &  & \multicolumn{2}{r}{meta-llama/Llama-2-70b-chat-hf } & 3493 \\
 &  & \multicolumn{2}{r}{gpt-3.5-turbo } & 3471 \\
 &  & \multicolumn{2}{r}{timdettmers/guanaco-33b-merged } & 3468 \\
 &  & \multicolumn{2}{r}{claude-2.1 } & 3338 \\
 &  & \multicolumn{2}{r}{mistralai/Mistral-7B-Instruct-v0.1 } & 3261 \\
 &  & \multicolumn{2}{r}{claude-2 } & 3209 \\
 &  & \multicolumn{2}{r}{tiiuae/falcon-7b-instruct } & 2608 \\
 &  & \multicolumn{2}{r}{OpenAssistant/oasst-sft-4-pythia-12b-epoch-3.5 } & 2314 \\
 &  & \multicolumn{2}{r}{meta-llama/Llama-2-13b-chat-hf } & 1744 \\
 &  & \multicolumn{2}{r}{luminous-supreme-control } & 1722 \\
 &  & \multicolumn{2}{r}{google/flan-t5-xxl } & 1715 \\
 &  & \multicolumn{2}{r}{luminous-extended-control } & 1633 \\
 & \multicolumn{4}{p{0.9\textwidth}}{\textit{Notes: We provide the long name as it appeared on our backend.
We provide a mapping of long names to shorter more familiar names on our Github or in the paper. }}  \\
 & \cline{1-4}
 \midrule
\bfseries 8 & \bfseries model\_provider & \bfseries Provider of the LLM & \bfseries {\cellcolor{myred}} meta & \bfseries {\cellcolor{myoat}} categorical \\
 \midrule
\bfseries  & \bfseries  & \bfseries  & \bfseries N Missing: & \bfseries 0 \\
\bfseries  & \bfseries  & \bfseries  & \bfseries N Unique: & \bfseries 6 \\
 &  & \multicolumn{2}{r}{huggingface\_api } & 26731 \\
 &  & \multicolumn{2}{r}{cohere } & 12557 \\
 &  & \multicolumn{2}{r}{anthropic } & 10839 \\
 &  & \multicolumn{2}{r}{openai } & 10721 \\
 &  & \multicolumn{2}{r}{google } & 4168 \\
 &  & \multicolumn{2}{r}{aleph } & 3355 \\
 & \multicolumn{4}{p{0.9\textwidth}}{\textit{Notes: Note for open-access LLMs, HuggingFace API is always listed as the source and does not imply they built the model. }}  \\
 & \cline{1-4}
 \midrule
\bfseries 9 & \bfseries user\_prompt & \bfseries Human-written message. & \bfseries {\cellcolor{myyellow}} direct & \bfseries {\cellcolor{myoat}} string \\
 \midrule
\bfseries  & \bfseries  & \bfseries  & \bfseries N Missing: & \bfseries 0 \\
\bfseries  & \bfseries  & \bfseries  & \bfseries N Unique: & \bfseries 26673 \\
 &  & \multicolumn{2}{r}{mean chars } & 69.9 \\
 &  & \multicolumn{2}{r}{std chars } & 62.0 \\
 &  & \multicolumn{2}{r}{min chars } & 1.0 \\
 &  & \multicolumn{2}{r}{max chars } & 1311.0 \\
 \midrule
\bfseries 10 & \bfseries model\_response & \bfseries Model-generated response & \bfseries {\cellcolor{myyellow}} direct & \bfseries {\cellcolor{myoat}} string \\
 \midrule
\bfseries  & \bfseries  & \bfseries  & \bfseries N Missing: & \bfseries 0 \\
\bfseries  & \bfseries  & \bfseries  & \bfseries N Unique: & \bfseries 66614 \\
 &  & \multicolumn{2}{r}{mean chars } & 565.3 \\
 &  & \multicolumn{2}{r}{std chars } & 387.9 \\
 &  & \multicolumn{2}{r}{min chars } & 1.0 \\
 &  & \multicolumn{2}{r}{max chars } & 4630.0 \\
 & \multicolumn{4}{p{0.9\textwidth}}{\textit{Notes: An empty string is stored as `EMPTY STRING'. }}  \\
 & \cline{1-4}
 \midrule
\bfseries 11 & \bfseries score & \bfseries Score of the model response & \bfseries {\cellcolor{myyellow}} direct & \bfseries {\cellcolor{myoat}} int \\
 \midrule
 & \multicolumn{4}{p{0.9\textwidth}}{\textit{Question text:} Rate the model responses. There are no right or wrong answers. Use your subjective judgement. }  \\
 & \cline{1-4}
\bfseries  & \bfseries  & \bfseries  & \bfseries N Missing: & \bfseries 0 \\
\bfseries  & \bfseries  & \bfseries  & \bfseries N Unique: & \bfseries 100 \\
 &  & \multicolumn{2}{r}{mean } & 65.1 \\
 &  & \multicolumn{2}{r}{std } & 29.3 \\
 &  & \multicolumn{2}{r}{min } & 1.0 \\
 &  & \multicolumn{2}{r}{max } & 100.0 \\
 & \multicolumn{4}{p{0.9\textwidth}}{\textit{Notes: Sliders from [Terrible] to [Perfect] are recorded on a 1-100 scale. Participant does not see numeric value. }}  \\
 & \cline{1-4}
 \midrule
\bfseries 12 & \bfseries if\_chosen & \bfseries Whether model response was highest-rated by participant & \bfseries {\cellcolor{mygreen}} constructed & \bfseries {\cellcolor{myoat}} binary \\
 \midrule
\bfseries  & \bfseries  & \bfseries  & \bfseries N Missing: & \bfseries 0 \\
\bfseries  & \bfseries  & \bfseries  & \bfseries N Unique: & \bfseries 2 \\
 &  & \multicolumn{2}{r}{False } & 40934 \\
 &  & \multicolumn{2}{r}{True } & 27437 \\
 & \multicolumn{4}{p{0.9\textwidth}}{\textit{Notes: In case of a tie, a random response is chosen. }}  \\
 & \cline{1-4}
 \midrule
\bfseries 13 & \bfseries included\_in\_balanced\_subset & \bfseries Indicator if participant's conversations are included in the balanced subset & \bfseries {\cellcolor{mygreen}} constructed & \bfseries {\cellcolor{myoat}} binary \\
 \midrule
\bfseries  & \bfseries  & \bfseries  & \bfseries N Missing: & \bfseries 0 \\
\bfseries  & \bfseries  & \bfseries  & \bfseries N Unique: & \bfseries 2 \\
 &  & \multicolumn{2}{r}{True } & 57401 \\
 &  & \multicolumn{2}{r}{False } & 10970 \\
 & \multicolumn{4}{p{0.9\textwidth}}{\textit{Notes: Balanced subset was created to equally sample conversations of three types (unguided, values, controversy).
We only include participants who have at least one of each conversation type, and then ensure equal numbers of each type are retained.
See paper for details. }}  \\
 & \cline{1-4}
\end{longtable}

\clearpage
\subsection{Metadata Codebook}
\label{sec:metadata_codebook}
\fontsize{6pt}{6pt}\selectfont
\begin{longtable}{p{0.005\textwidth}p{0.20\textwidth}p{0.40\textwidth}p{0.1\textwidth}p{0.1\textwidth}}
\toprule
 & VARIABLE & LABEL & CATEGORY & TYPE \\
\midrule
\endfirsthead
\toprule
 & VARIABLE & LABEL & CATEGORY & TYPE \\
\midrule
\endhead
\midrule
\multicolumn{5}{r}{Continued on next page} \\
\midrule
\endfoot
\bottomrule
\endlastfoot
\bfseries  & \bfseries & \bfseries  & \bfseries  & \bfseries  \\
\bfseries 0 & \bfseries column\_id & \bfseries Source of text utterance & \bfseries {\cellcolor{myred}} meta & \bfseries {\cellcolor{myoat}} categorical \\
 \midrule
\bfseries  & \bfseries  & \bfseries  & \bfseries N Missing: & \bfseries 0 \\
\bfseries  & \bfseries  & \bfseries  & \bfseries N Unique: & \bfseries 5 \\
 &  & \multicolumn{2}{r}{model\_response } & 68371 \\
 &  & \multicolumn{2}{r}{user\_prompt } & 27172 \\
 &  & \multicolumn{2}{r}{open\_feedback } & 8011 \\
 &  & \multicolumn{2}{r}{self\_description } & 1500 \\
 &  & \multicolumn{2}{r}{system\_string } & 1500 \\
 \midrule
\bfseries 1 & \bfseries user\_id & \bfseries Unique participant identifier & \bfseries {\cellcolor{myred}} meta & \bfseries {\cellcolor{myoat}} string id \\
 \midrule
\bfseries  & \bfseries  & \bfseries  & \bfseries N Missing: & \bfseries 0 \\
\bfseries  & \bfseries  & \bfseries  & \bfseries N Unique: & \bfseries 1500 \\
 & \multicolumn{4}{p{0.9\textwidth}}{\textit{Notes: Pseudonymized from Prolific worker ID. Used to link metadata to main data. }}  \\
 & \cline{1-4}
 \midrule
\bfseries 2 & \bfseries conversation\_id & \bfseries Unique conversation identifier & \bfseries {\cellcolor{myred}} meta & \bfseries {\cellcolor{myoat}} string id \\
 \midrule
\bfseries  & \bfseries  & \bfseries  & \bfseries N Missing: & \bfseries 3000 \\
\bfseries  & \bfseries  & \bfseries  & \bfseries N Unique: & \bfseries 8011 \\
 & \multicolumn{4}{p{0.9\textwidth}}{\textit{Notes: Used to link metadata to main data. }}  \\
 & \cline{1-4}
 \midrule
\bfseries 3 & \bfseries interaction\_id & \bfseries Unique interaction identifier, where an interaction is a turn within a conversation (single human message with multiple model responses) & \bfseries {\cellcolor{myred}} meta & \bfseries {\cellcolor{myoat}} string id \\
 \midrule
\bfseries  & \bfseries  & \bfseries  & \bfseries N Missing: & \bfseries 11011 \\
\bfseries  & \bfseries  & \bfseries  & \bfseries N Unique: & \bfseries 27172 \\
 & \multicolumn{4}{p{0.9\textwidth}}{\textit{Notes: Used to link metadata to main data. }}  \\
 & \cline{1-4}
 \midrule
\bfseries 4 & \bfseries utterance\_id & \bfseries Unique utterance identifier, where an utterance is a single human message - single model response pair & \bfseries {\cellcolor{myred}} meta & \bfseries {\cellcolor{myoat}} string id \\
 \midrule
\bfseries  & \bfseries  & \bfseries  & \bfseries N Missing: & \bfseries 38183 \\
\bfseries  & \bfseries  & \bfseries  & \bfseries N Unique: & \bfseries 68371 \\
 & \multicolumn{4}{p{0.9\textwidth}}{\textit{Notes: Used to link metadata to main data. }}  \\
 & \cline{1-4}
 \midrule
\bfseries 5 & \bfseries pii\_flag & \bfseries Automated flag for personally identifiable information & \bfseries {\cellcolor{myred}} meta & \bfseries {\cellcolor{myoat}} binary \\
 \midrule
\bfseries  & \bfseries  & \bfseries  & \bfseries N Missing: & \bfseries 0 \\
\bfseries  & \bfseries  & \bfseries  & \bfseries N Unique: & \bfseries 2 \\
 &  & \multicolumn{2}{r}{False } & 105443 \\
 &  & \multicolumn{2}{r}{True } & 1111 \\
 & \multicolumn{4}{p{0.9\textwidth}}{\textit{Notes: Uses scrubadub https://scrubadub.readthedocs.io/en/stable/ to find PII. There may be some misclassifications.
Many of the inspected positives were false positives. All positive human-written texts checked.
See pii\_manual\_flag. }}  \\
 & \cline{1-4}
 \midrule
\bfseries 6 & \bfseries pii\_manual\_flag & \bfseries Manual verification of personally identifiable information in human-written texts & \bfseries {\cellcolor{myred}} meta & \bfseries {\cellcolor{myoat}} binary \\
 \midrule
\bfseries  & \bfseries  & \bfseries  & \bfseries N Missing: & \bfseries 106387 \\
\bfseries  & \bfseries  & \bfseries  & \bfseries N Unique: & \bfseries 1 \\
 &  & \multicolumn{2}{r}{nan } & 106387 \\
 &  & \multicolumn{2}{r}{0.0 } & 167 \\
 & \multicolumn{4}{p{0.9\textwidth}}{\textit{Notes: For any automated PII flags, we manually checked the human-written text for PII.
All were false positives so this flag overules the automated flag. We did not check model-generated text for PII.
NaN indicates entry was not manually checked. }}  \\
 & \cline{1-4}
 \midrule
\bfseries 7 & \bfseries language\_flag & \bfseries Automated language detection & \bfseries {\cellcolor{myred}} meta & \bfseries {\cellcolor{myoat}} categorical \\
 \midrule
\bfseries  & \bfseries  & \bfseries  & \bfseries N Missing: & \bfseries 0 \\
\bfseries  & \bfseries  & \bfseries  & \bfseries N Unique: & \bfseries 59 \\
 &  & \multicolumn{2}{r}{Too many values to show } & - \\
 & \multicolumn{4}{p{0.9\textwidth}}{\textit{Notes: Uses langid. There may be some misclassifications. }}  \\
 & \cline{1-4}
 \midrule
\bfseries 8 & \bfseries en\_flag & \bfseries Whether detected language is English & \bfseries {\cellcolor{myred}} meta & \bfseries {\cellcolor{myoat}} binary \\
 \midrule
\bfseries  & \bfseries  & \bfseries  & \bfseries N Missing: & \bfseries 0 \\
\bfseries  & \bfseries  & \bfseries  & \bfseries N Unique: & \bfseries 2 \\
 &  & \multicolumn{2}{r}{Too many values to show } & - \\
 & \multicolumn{4}{p{0.9\textwidth}}{\textit{Notes: Constructed based on automated language detection. }}  \\
 & \cline{1-4}
 \midrule
\bfseries 9 & \bfseries moderation\_flag & \bfseries Automated flag for moderation & \bfseries {\cellcolor{myred}} meta & \bfseries nested dict \\
 \midrule
 & \multicolumn{4}{p{0.9\textwidth}}{\textit{Notes: Uses OpenAI moderation API. There may be some misclassifications.
Nested dictionary with binary flags and probabilities for sub-categories of harm. }}  \\
 & \cline{1-4}
\end{longtable}

\end{document}